%% file: thesis.tex
\documentclass{ucbthesis}
\usepackage{biblatex}
\usepackage{subcaption}
\usepackage{graphicx}
\usepackage{pifont}
\usepackage{threeparttable}
\usepackage{amsmath} 
\usepackage{amssymb}  
\usepackage{dsfont}
\usepackage{multirow}
\usepackage[ruled,vlined]{algorithm2e}

\usepackage{mathtools}
\usepackage{mathrsfs}
\usepackage{listings}

\usepackage{caption} 
\usepackage{makecell}
\usepackage{pgfplots}
\pgfplotsset{compat = 1.10}
\usepgfplotslibrary{external}

\input{math_commands.tex}

\newcommand{\xmark}{\ding{55}}%

\bibliography{references}

\begin{document}


\title{\normalfont{} Efficient Deep Neural Networks}
\author{Bichen Wu}
\degreesemester{Summer}
\degreeyear{2019}
\degree{Doctor of Philosophy}
\chair{Professor Kurt Keutzer}
\othermembers{Professor Joseph Gonzalez \\
  Professor Sara McMains}
\numberofmembers{3}

\field{Engineering - Electrical Engineering and Computer Sciences}
\campus{Berkeley}

\maketitle
\approvalpage
\copyrightpage

\include{abstract}

\begin{frontmatter}

\begin{dedication}
\null\vfil
\begin{center}
To my wife, Zeyu Yang, my mother, Professor Aihua Yang, my father, Weifan Wu, and my sister, Virginia Wu, for their unconditional love and support. \\
\end{center}
\vfil\null
\end{dedication}


\tableofcontents
\clearpage
\listoffigures
\clearpage
\listoftables

\begin{acknowledgements}
First of all, I want to thank my advisor, Professor Kurt Keutzer. Kurt is probably the most visionary, inspiring, and supportive advisor a student can ever expect. On research, his guidance is not just on how to solve specific problems, but more importantly, how to discover and define impactful problems. I am very fortunate to have worked with Kurt in the past four years, and what I have learned from him is beyond count.

I want to thank Professor Joseph Gonzalez for years of collaboration and his guidance on my dissertation. Every time I talk to Joey, I need to pay 120\% of attention so that I can keep up with the speed he sparks new ideas. He is not only a role model of how to conduct the highest quality research but also a role model of how to organize team effort and support others. I also want to thank Professor Sara McMains for her feedback and encouragement on my dissertation. I had the privilege of working with Professor Jaijeet Roychowdhury during the first two years at Berkeley. I want to thank him for his guidance.

It has been an honor to work with many great collaborators during my time at UC Berkeley. I want to thank Forrest Iandola, Peter Jin, Matthew Moskewicz, and Khalid Ashraf for helping me start the journey of searching for efficient neural networks. Thanks to Song Han for years of collaboration starting from Tsinghua University. Our discussion sparked a lot of new ideas that later advanced the state-of-the-art of efficient neural network research.  Thanks to Alvin Wan, Xiangyu Yue, Sicheng Zhao, Xuanyu Zhou, Amir Gholami, Noah Golmant, and Kiseok Kwon for their collaboration that leads to the ``Squeeze-AI'' series of work, including SqueezeDet, SqueezeSeg, SqueezeSegV2, SqueezeNext, and ShiftNet. I want to thank Yifan Yang, Qijing Huang, Tianjun Zhang, Liang Ma, Giulio Gambardella, Michaela Blott, Professor Luciano Lavagno, Kees Vissers, and Professor John Wawrzynek for their collaboration on co-designing efficient neural networks and hardware accelerators. Thanks to Virginia Wu and Bernie Wang for their collaboration on the LATTE project. Thanks to my collaborators from Facebook: Xiaoliang Dai, Peizhao Zhang, Yanghan Wang, Fei Sun, Yiming Wu, Yuandong Tian, Peter Vajda, and Yangqing Jia for their help with the work on neural architecture search. Thanks to fellow researchers from Google: Andrew Howard, Mark Sandler, Bo Chen, Quoc Le, Bo Chen, Minxing Tan, and Hanxiao Liu for their inspiring prior work and their discussion that helped my research. I want to thank my other collaborators, including (in no particular order): Xin Wang, Fisher Yu, Tianshi Wang, Aadithya Karthik, Palak Bushan, Zhen Dong, Tianyuan Zhang, Zheng Liang, Flora Xue, Tianren Gao, Bohan Zhai, Suresh Krishna, Ravi Krishna, Rob Roddick, Professor Sanjit Seshia, and Professor Alberto Sangiovanni-Vincentelli. 

Over the last few years, our research has been partially funded by DARPA PERFECT program, Award HR0011-12-2-0016, together with ASPIRE Lab sponsors, Berkeley AI Research Lab sponsors, Berkeley Deep Drive (BDD) Industry Consortium, and individual gifts from BMW, Intel, and the Samsung Global Research Organization. I want to thank them for their generous support. 

I want to thank my mother, Professor Aihua Yang, my father, Weifan Wu, and my younger sister, Virginia Wu. They selflessly supported me to explore the destiny of my life and to create an impact to make this world a better place. Finally and most importantly, I would like to thank my wife, Zeyu Yang, for her love and encouragements, for all the late nights and early mornings, and for faithfully supporting me throughout the peaks and valleys. I want to dedicate this milestone of my life to them for their unconditional love and support. 

\end{acknowledgements}

\end{frontmatter}

\pagestyle{headings}

\include{chap1}

\include{chap2}
\include{chap3}
\include{chap4}
\include{chap5}
\include{chap6}
\include{chap7}
\include{chap8}
\include{conclusion}

\printbibliography

\end{document}

%% file: math_commands.tex
\usepackage{amsmath,amsfonts,bm}

\def\eqref#1{equation~\ref{#1}}

\def\1{\bm{1}}

\def\ra{{\textnormal{a}}}

\def\rg{{\textnormal{g}}}

\def\rM{{\textnormal{m}}}

\def\rvg{{\mathbf{g}}}

\def\rvm{{\mathbf{m}}}

\def\vtheta{{\bm{\theta}}}

\def\vm{{\bm{m}}}

\def\vw{{\bm{w}}}

\DeclareMathAlphabet{\mathsfit}{\encodingdefault}{\sfdefault}{m}{sl}
\SetMathAlphabet{\mathsfit}{bold}{\encodingdefault}{\sfdefault}{bx}{n}

\DeclareMathOperator*{\argmin}{arg\,min}

%% file: abstract.tex
\begin{abstract}
The success of deep neural networks (DNNs) is attributable to three factors: increased compute capacity, more complex models, and more data. These factors, however, are not always present, especially for edge applications such as autonomous driving, augmented reality, and internet-of-things. Training DNNs requires a large amount of data, which is difficult to obtain. Edge devices such as mobile phones have limited compute capacity, and therefore, require specialized and efficient DNNs. However, due to the enormous design space and prohibitive training costs, designing efficient DNNs for different target devices is challenging. So the question is, with limited data, compute capacity, and model complexity, can we still successfully apply deep neural networks?

This dissertation focuses on the above problems and improving the efficiency of deep neural networks at four levels. \textbf{Model efficiency}: we designed neural networks for various computer vision tasks and achieved more than 10x faster speed and lower energy. \textbf{Data efficiency}: we developed an advanced tool that enables 6.2x faster annotation of a LiDAR point cloud. We also leveraged domain adaptation to utilize simulated data, bypassing the need for real data. \textbf{Hardware efficiency}: we co-designed neural networks and hardware accelerators and achieved 11.6x faster inference. \textbf{Design efficiency}: the process of finding the optimal neural networks is time-consuming. Our automated neural architecture search algorithms discovered, using 421x lower computational cost than previous search methods, models with state-of-the-art accuracy and efficiency.
\end{abstract}

%% file: chap1.tex
\chapter{Introduction}
\section{Three factors for the success of deep learning}
In recent years, research on deep neural networks has achieved tremendous progress in a wide range of artificial intelligence problems, including but not limited to computer vision, natural language processing, and reinforcement learning. In 2015, ResNet \cite{resnet} surpassed the human-level accuracy in the ImageNet classification task. In 2016, an automated agent named AlphaGo \cite{silver2017mastering} beat the world champion Lee Sedol in the game of Go. In 2018, Hassan et al. \cite{hassan2018achieving} achieved human-level parity in Chinese-to-English translation. 

\begin{figure}[h]
\centering
  \includegraphics[width=\textwidth]{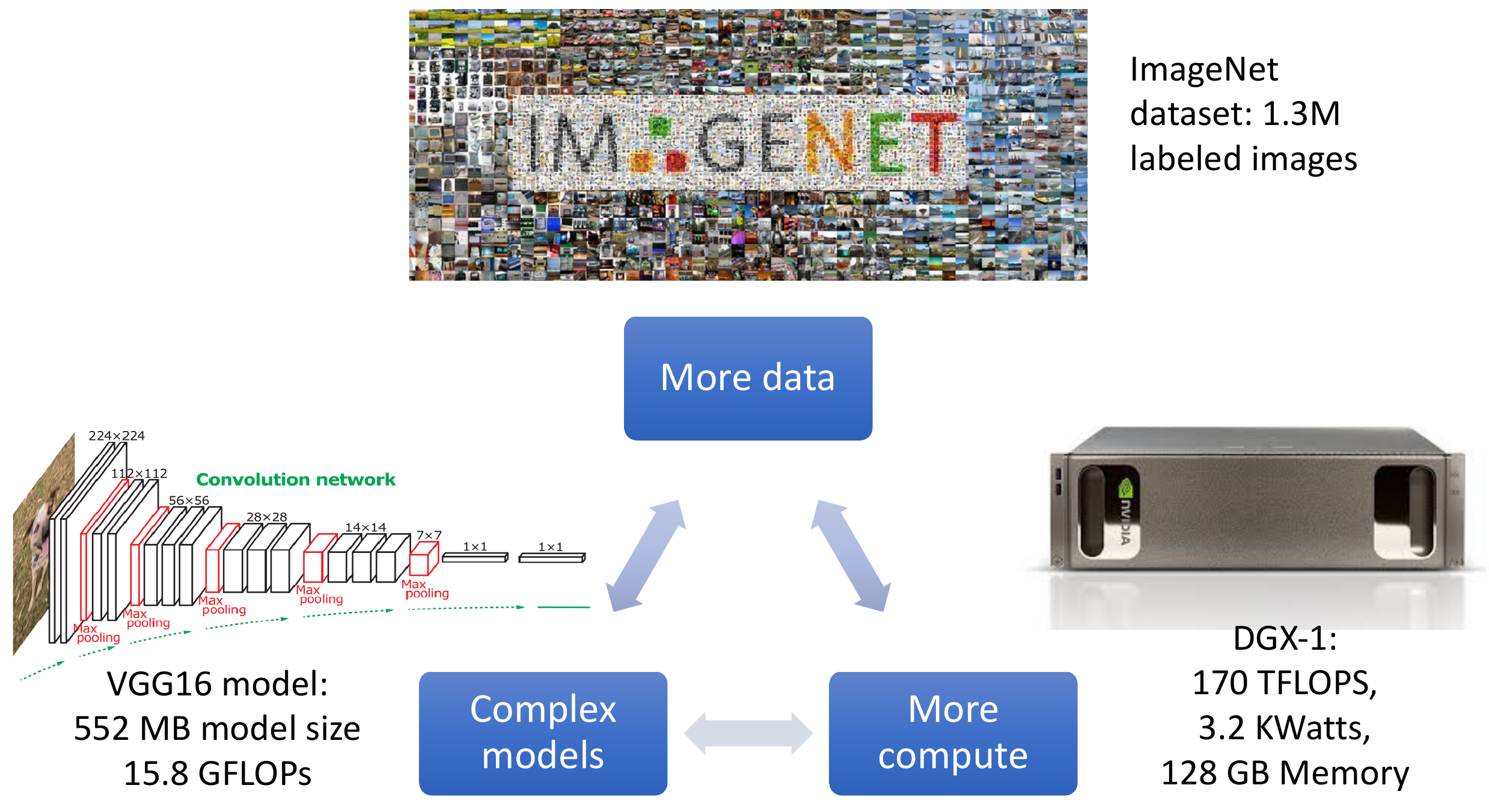}
  \caption{Three factors for the success of deep neural networks.
}
\label{fig:DL-success-factors} 
\end{figure}

The success of deep neural networks is typically attributable to three factors: more complex models, more data, and increased compute capacity, as illustrated in Figure \ref{fig:DL-success-factors}. 

\textbf{More complex models}: In many previous works \cite{resnet, zoph2017learning, howard2017mobilenets}, it is commonly observed that the performance of deep neural networks is highly dependent on the model complexity, which is measured by a neural network's parameter size or the number of floating-point operations (FLOPs). A general trend is that the larger the model is, the higher accuracy it can achieve in a given task. For examples, AlexNet \cite{krizhevsky2012imagenet} achieves 58\% top-1 accuracy on ImageNet and the network contains about 60 million parameters. VGG16 \cite{VGG} contains 138 million parameters and significantly improves the accuracy to 71\%. As a result,  in order to achieve better performance, people tend to use the largest model that can still fit in the computational constraints.  

\textbf{More data}: Before deep learning, people realized that with the same learning models, more data can effectively improve their performance \cite{halevy2009unreasonable}. In the deep learning era, this rule is confirmed by many works. For example, Sun et al. \cite{sun2017revisiting} observed that the performance of an object detection model ``increases logarithmically based on the volume of training data''.  For image classification, ImageNet \cite{deng2009imagenet} contains 1.3 million labeled images. For natural language processing,  GPT-2 \cite{radford2019language} was pre-trained on a dataset consisting of 8 million documents for a total of 40 GB of text. For reinforcement learning, AlphaGo \cite{silver2017mastering} was trained on 29 million self-played games. As a result, in order to get better performance, people have devoted significant efforts into creating large-scale datasets for training deep neural networks. 

\textbf{More compute}: As datasets become larger and models become more complex, correspondingly, deep neural networks rely increasingly more on powerful hardware processors for training and inference. One forward pass of the ResNet50 model requires 4 GFLOPs of compute and the training of it requires $10^{18}$ FLOPs \cite{you2018imagenet}, which takes 14 days on an NVIDIA M40 GPU. To make the training and inference faster, people spend significant efforts in building more powerful processors. In 2014, Nvidia's K80 GPU was able to deliver a compute capacity of 5.6 TFLOPs per second while in 2018, Nvidia's Tesla V GPU reaches a compute capacity 125 TFLOPs per second. 

This trend is best summarized by Richard Sutton\footnote{\url{http://www.incompleteideas.net/IncIdeas/BitterLesson.html}}:
\begin{quote}
    The biggest lesson that can be read from 70 years of AI research is that general methods that leverage computation are ultimately the most effective, and by a large margin. 
\end{quote}

\section{Deep learning on the edge}
Powered by the rapid progress of deep neural networks, people have begun to deploy deep neural networks to solve practical problems. Among many applications, a wide range of them require deploying neural networks on the ``edge,'' i.e., on light-weight devices that perform computation locally. For example, users of applications such as biometric identification (face or finger ID) may have privacy concerns and therefore wish to avoid sending data to the cloud. Applications such as autonomous driving and augmented reality (AR), on the other hand, depend on the real-time perception of the environment in order to interact in real time with the real-world. They cannot tolerate the latency of sending data to the cloud, processing, and then receiving the results. For internet-of-things (IoT) applications, sensors are usually deployed at large scale, and they cannot afford the cost of transmitting data to a centralized server for processing. In such applications, the execution of neural networks has to be completed locally on light-weight devices.

\begin{figure}[h]
\centering
  \includegraphics[width=\textwidth]{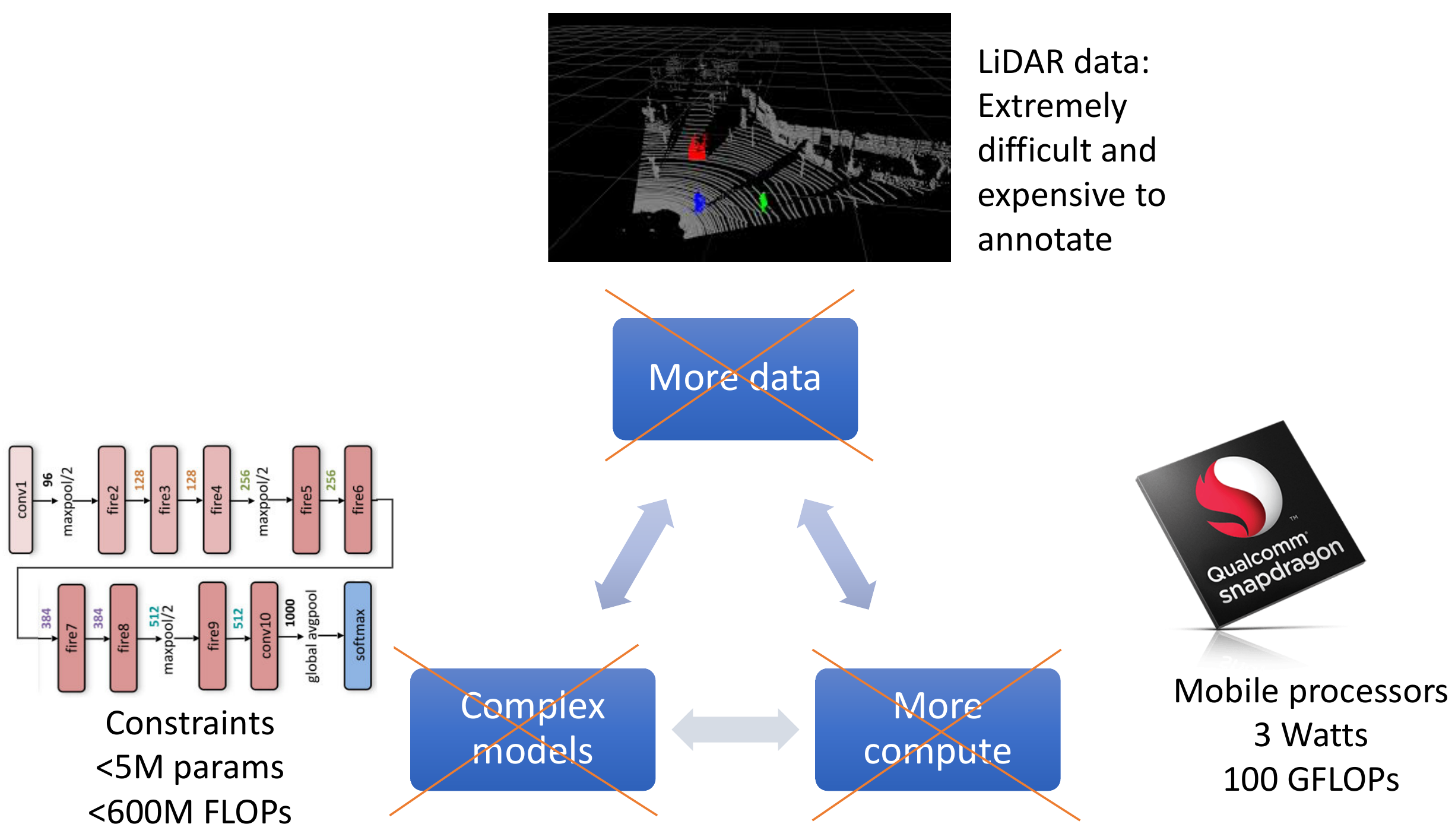}
  \caption[Deep learning on the edge.]{For edge-based applications, the three factors for the success of deep learning are usually not present.
 }
\label{fig:DL-on-the-edge} 
\end{figure}

However, for edge-based applications, the three factors for the success of deep learning are usually not present. This is summarized in Figure \ref{fig:DL-on-the-edge}. 

\textbf{Limited compute capacity}: The compute capacity of edge-based processors is much lower than desktop or server GPUs. For example, the Titan X\footnote{\url{https://www.nvidia.com/en-us/geforce/products/10series/titan-x-pascal/}}, a common desktop GPU used for most computer vision experiments in 2017, can deliver 11 TFLOPs of compute and consumes 223 watts of power. In comparison, the thermal design point (maximum power) of mobile processors is typically under 5 watts. One of the latest mobile SoC, Helio P90\footnote{\url{https://www.mediatek.com/products/smartphones/mediatek-helio-p90}}  contains two ARM Cortex A75\footnote{\url{https://www.arm.com/products/silicon-ip-cpu/cortex-a/cortex-a75}} CPUs and each of them can deliver around 2.2G half-precision floating point operations. The compute capacity on the edge is significantly lower. 

\textbf{Limited model complexity}: In order to fit in the compute constraint of edge processors and meet the real-time speed requirement at the same time, neural network models for the edge need to be significantly smaller. While the VGG16 model requires 15.8 GFLOPs of compute per image and 552MB of model size,  mobile models typically need to constrain the FLOPs to under 600 million and model size to less than 5 MB \cite{howard2017mobilenets, zoph2017learning}.  For always-on applications such as visual wake up, models need to be constrained to have less than 250 KB of model size and 60M FLOPs \cite{chowdhery2019visual}. 

\textbf{Limited data}: In many edge applications, collecting and annotating data is prohibitively expensive. Take LiDAR-based perception for autonomous driving as an example: collecting LiDAR point cloud data requires using LiDAR sensors, which typically cost thousands of dollars. Moreover, annotating LiDAR point clouds is significantly harder than annotating images for human due to the much lower resolution of the sensor. For such applications, creating a large-scale dataset is very difficult. 

\section{Organization of this thesis}
\begin{figure}[h]
\centering
  \includegraphics[width=.9\textwidth]{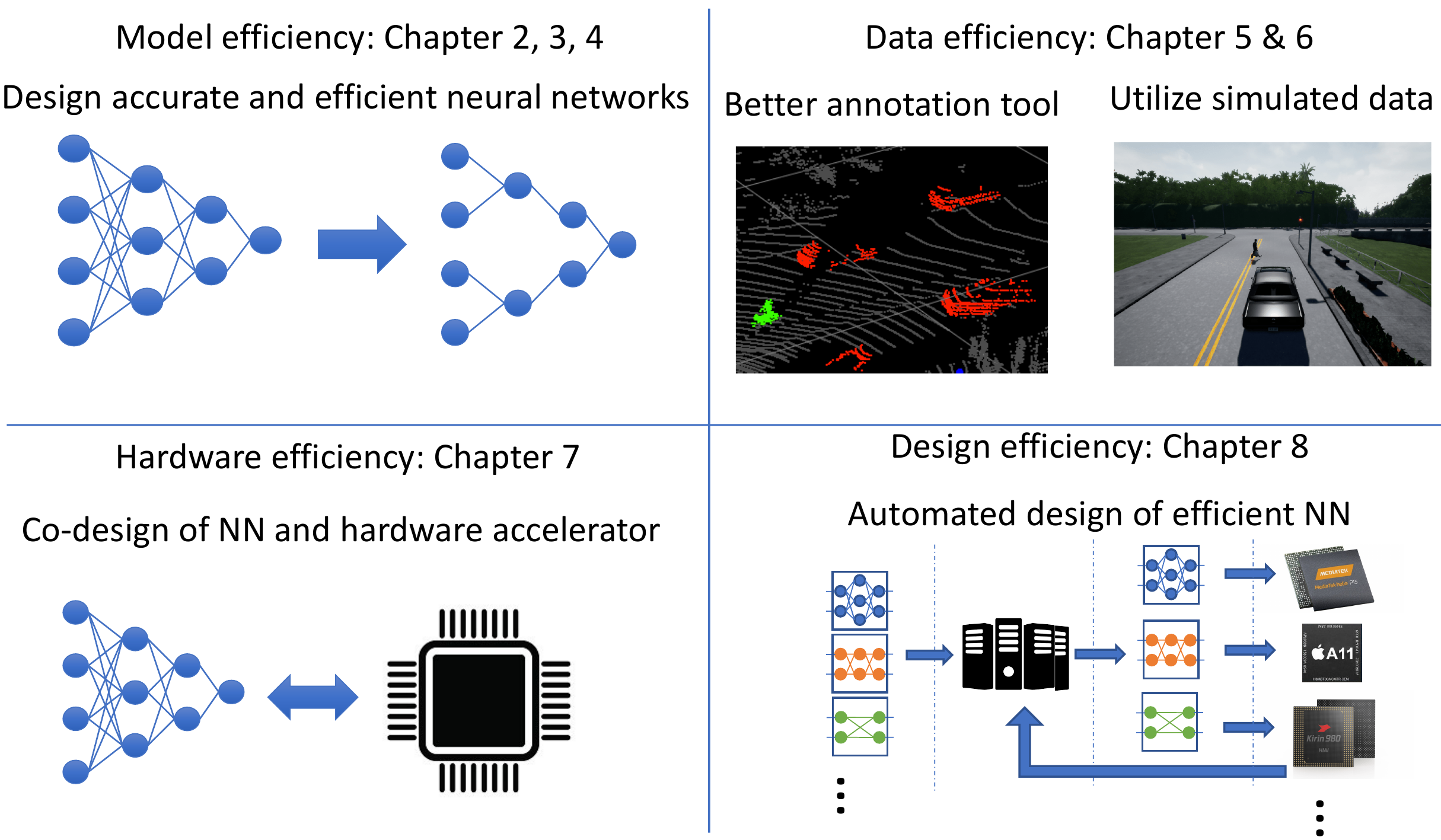}
  \caption{An overview of this thesis.}
\label{fig:thesis-overview} 
\end{figure}

In order to address these problems, in this thesis, we focus on improving the efficiency of deep neural networks at four different levels, as illustrated in Figure \ref{fig:thesis-overview}. 

First, we challenge the popular view that more complex models are necessary for improved performance. We show that by carefully designing deep neural networks, we can find much more compact models that achieve the same level of accuracy with significantly lower complexity. We demonstrate this on two diverse computer vision problems: image-based object detection and LiDAR-based segmentation. In Chapter \ref{chap:metrics}, we first discuss metrics for evaluating the efficiency (complexity) of neural networks. In Chapter \ref{chap:sqdt}, we introduce SqueezeDet, a highly efficient neural network model designed for object detection for autonomous driving. Compared with previous baselines, SqueezeDet achieved the same accuracy, with 30x fewer parameters, 20x speedup, and 35x better energy efficiency. In Chapter \ref{chap:sqsg}, we introduce SqueezeSeg, a neural network-based pipeline for LiDAR point-cloud segmentation. By carefully designing the problem formulation, the network architecture, and data representation, we successfully adapted 2D convolution neural networks to process 3D point cloud data and were able to achieve high accuracy with a speed of more than 100 frames per second, significantly faster than previous traditional methods. 

Second, we discuss strategies for improving data efficiency of deep neural networks via better annotation and domain adaptation. We use LiDAR-based detection as a motivating example, since collecting and annotating a LiDAR point cloud dataset is significantly more difficult than an image dataset. To address these problems, in Chapter \ref{chap:latte}, we systematically analyze the problems in LiDAR point cloud annotation: low resolution, complex annotation operation, and temporal correlation. To solve these problems, we built a new annotation tool that improves annotation efficiency by 6.2x. In Chapter \ref{chap:sqsgv2}, we discuss a more radical strategy to leverage simulated data to train neural networks and adapt the model to the real world. By improving the model structure to make it less sensitive to domain shift and applying a data-adaptation training pipeline, we are able to train a model on simulated data and reach the accuracy of the same model trained on real-world data. This allows us to bypass the need to collect and annotate real data.

Third, to fully optimize the efficiency of deep neural networks, we not only need to improve neural network models, but also the hardware processor. In Chapter \ref{chap:shift}, we discuss the gap between neural network design and hardware accelerator design. To close this gap, we perform co-design of neural networks and hardware accelerators. By proposing a novel operator named \textit{shift}, we not only significantly reduce the FLOPs and parameter size of a ConvNet but also simplify the operators and enable building a ConvNet consisting of only 1x1 convolutions. This, in turn, simplifies the design of hardware, allowing us to build a dedicated compute unit for 1x1 convolutions and achieve 11x speedup over the previous state-of-the-art hardware accelerators. 

Finally, in chpater \ref{chap:dnas}, we discuss the design efficiency of deep neural networks. Designing optimal neural networks for given hardware processors and applications is a difficult task due to the challenges of intractable design space, conditional optimality, and inaccurate efficiency metrics. Traditional iterative and manual design usually has a very long design cycle, and the results are suboptimal due to insufficient design space explorations. Recently, people have begun to use automated neural architecture search (NAS) to design neural networks. While the networks searched automatically outperform manual designed counterparts, most of the NAS methods are computationally expensive, costing tens of thousands of GPU-hours to find the optimal model. To solve this problem, we present differentiable neural architecture search (DNAS), an algorithm that automatically searches for neural network models that surpass the previous state-of-the-art while the search cost is 421x lower.

With this thesis, we show that we are able to significantly improve the efficiency of deep neural networks at four different levels, which facilitates the broad adoption of deep neural networks in many practical problems.

%% file: chap2.tex
\chapter{Model Efficiency: Metrics of model efficiency}
\label{chap:metrics}

In previous works, it has been shown that increasing the model complexity is an effective way to improve the performance of neural networks. For example, on the ImageNet classification task,  AlexNet contains 60 million parameters and achieved a top-1 accuracy of 58\%.  In comparison, VGG16 \cite{simonyan2014very} contains 138 million parameters and significantly boosted the top-1 accuracy to 71\%. 

Despite the performance improvement, the increased model complexity leads to a higher computational burden and therefore requires higher compute capacity. As a result, while we were able to run neural networks on powerful GPUs in data centers (in the cloud), it was infeasible to deploy neural networks on the edge where the compute capacity and power budget are limited. To solve this problem, we focus on the following \textit{key question}: 
\begin{quote}
    Is it possible to design neural networks to achieve the same performance with lower model complexity?
\end{quote}

We discuss this topic in Chapter \ref{chap:metrics}, \ref{chap:sqdt}, and \ref{chap:sqsg}. In this chapter, we first discuss a prerequisite question: 
\begin{quote}
    How should we evaluate the efficiency (complexity) of a neural network?
\end{quote}

\section{Background: memory hierarchy}
Before we discuss the efficiency (complexity) of neural networks, we first introduce some background of computer architecture briefly. This can help us understand how the computation of neural networks is executed on a hardware processor, and what characteristics of a neural network we should consider when we measure its efficiency. 

\begin{figure}[h]
  \centering
 \includegraphics [width=\textwidth]{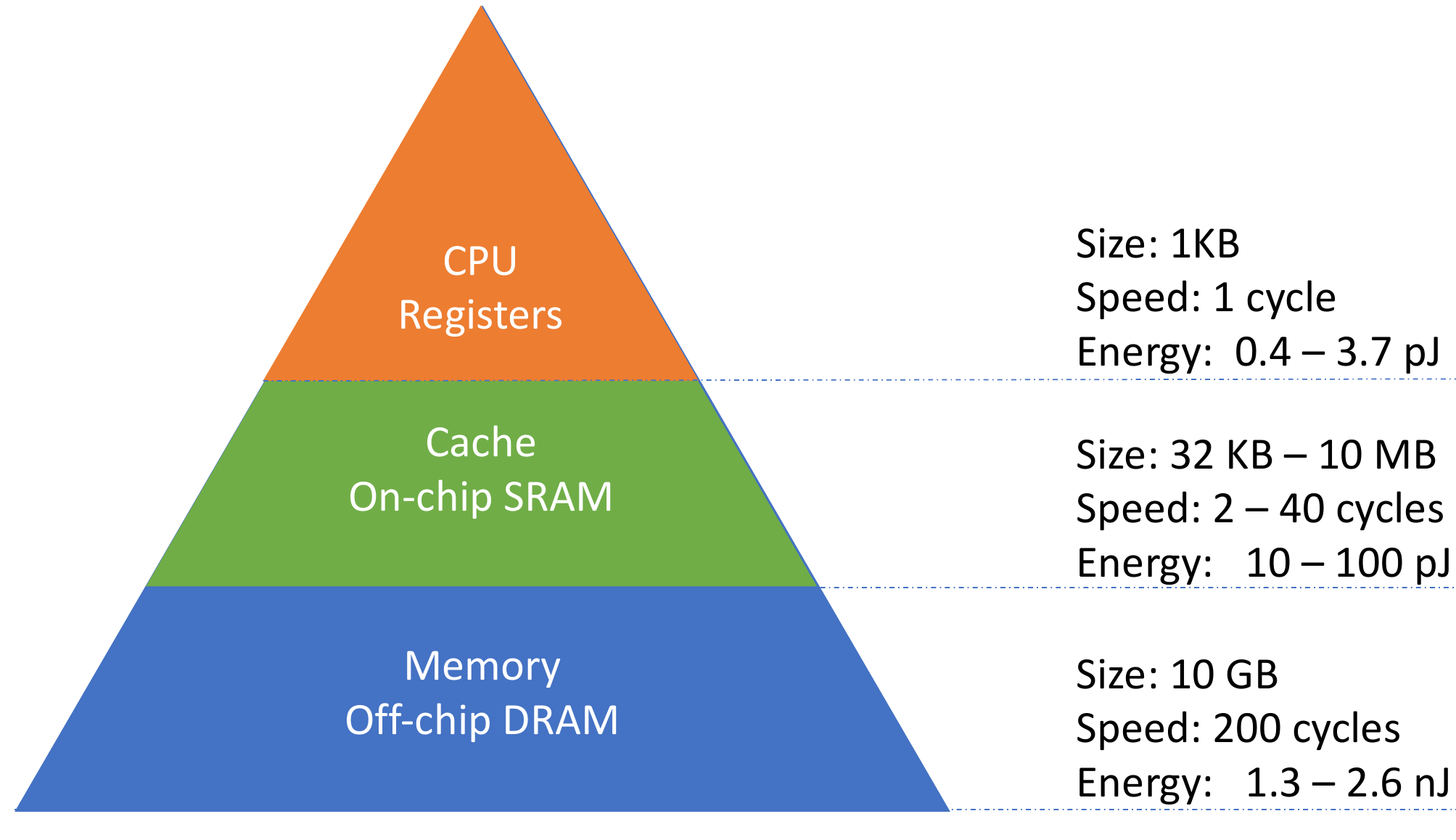}
 \caption[Memory Hierarchy.]{The memory hierarchy of computer architecture. The top-level is CPU and its register files. Data stored in the register file can be accessed in 1 clock cycle, and the computation consumes a tiny amount of energy (0.4 pJ for a 16-bit floating-point add with 45nm technology \cite{pedram2016dark}). However, the size of the register file is limited to a typical size of 1 KB. The next level is a larger cache memory made of on-chip SRAMs (static random-access memory). The memory size is larger (typically, 32KB, 256KB, 10MB for L1, L2, and L3 cache respectively) and the access speed is slower (3, 10, 40 cycles for L1, L2, and L3 cache respectively). The next stage is the main memory made of off-chip DRAMs (dynamic random-access memory). Accessing or storing data to and from off-chip DRAMs is significantly slower and consumes 3,556x more energy than a 16bit floating point add operation \cite{pedram2016dark}. 
 }
\label{fig:memory-hierarchy}
\end{figure}

Modern computer processors organize memories in a hierarchical structure to create an illusion that CPUs can access a large amount of fast memory.  Figure \ref{fig:memory-hierarchy} illustrates the typical memory hierarchy of computer architecture. At the top of the hierarchy are the CPU and its register files. In order to perform a compute operation, such as adding two numbers, the CPU first fetches data from register files in as fast as 1 clock cycle. The compute operation itself consumes little energy, typically 0.4 pJ for a 16bit floating point addition with 45nm technology \cite{pedram2016dark}. If the data is not available in the register file, the processor goes to the next level of cache memory that is made of on-chip SRAMs. Cache memories are larger (typically 32KB, 256KB, 10MB for L1, L2, and L3 cache respectively) and slower (typically 3, 10, 40 cycles for L1, L2, and L3 cache respectively) than register files. Depending on the level of cache, accessing a piece (64bit) of data from cache can consume 10 to 100 pJ of energy \cite{pedram2016dark}. If the data needed is still not available, the processor needs to go to the main memory, which is made of cheaper off-chip DRAMs. Main memories are much larger than on-chip SRAMs, but also much slower (typically 200 cycles) and consumes significantly more energy (3,556x more than an add operation \cite{pedram2016dark}). 

Compared with compute operations such as addition and multiplication, memory accesses, especially DRAM accesses, require orders-of-magnitude higher energy and latency. As a result, the efficiency of most modern computer programs is bounded by memory, instead of compute, so the most effective way to improve efficiency is to reduce the memory accesses. 

\section{Compute characteristics and metrics of neural networks}
\subsection{Compute characteristics}
Deep neural networks (DNNs) consist of layers of transformation functions parameterized by learnable weights. Despite the tremendous diversity of neural network types, the core computation of neural networks are essentially variations of matrix-multiplications. 

\begin{figure}[h]
  \centering
 \includegraphics [width=.8\textwidth]{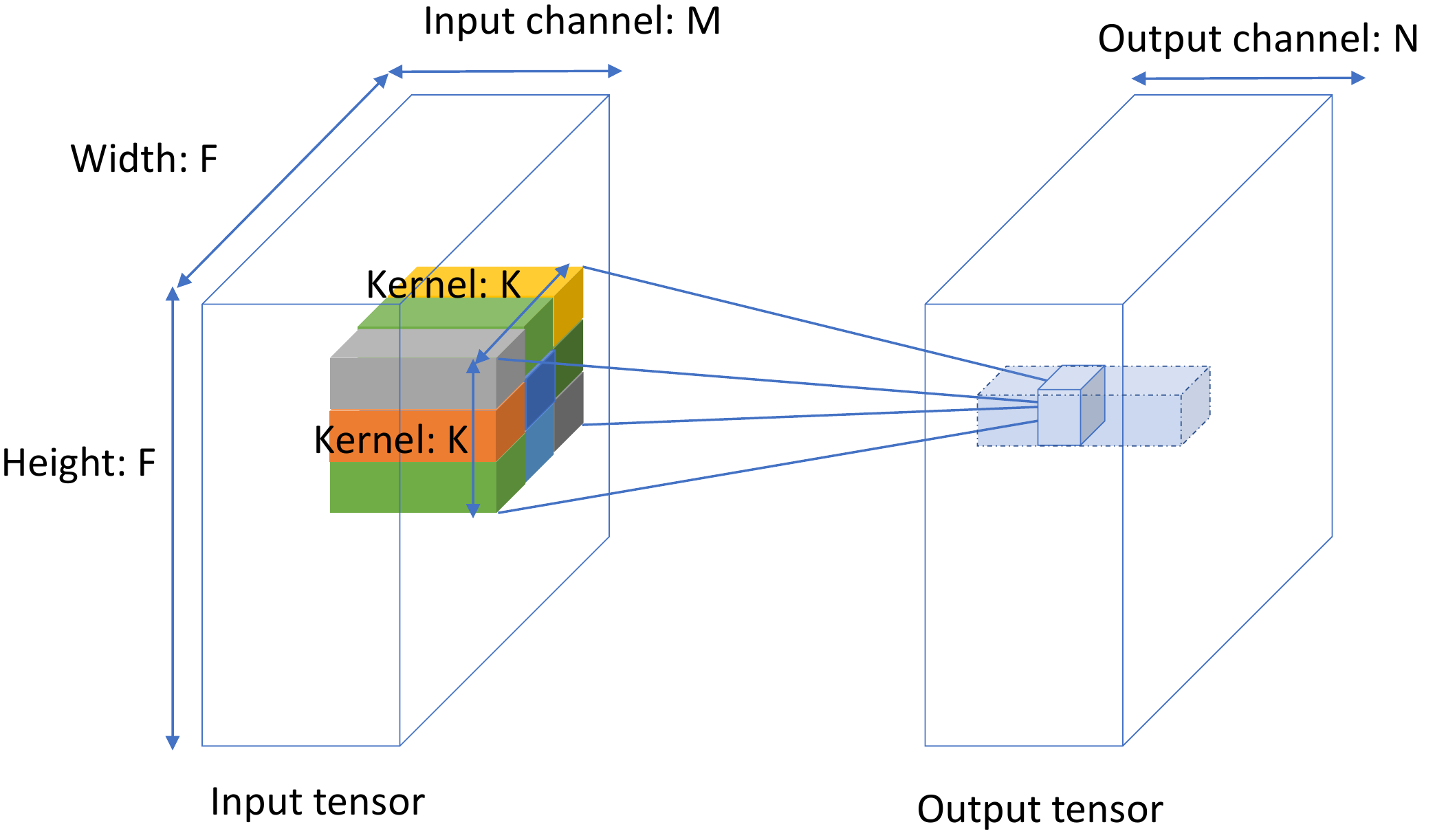}
 \caption[An illustration of a convolutional layer.]{An illustration of a convolutional layer. The computation of this layer is illustrated in Listing \ref{lst:conv-code}. 
 }
\label{fig:conv}
\end{figure}

We use a convolutional layer as an example to study the compute characteristics of neural networks. An illustration of a convolutional layer is in Figure \ref{fig:conv}, and the computation is described in Listing \ref{lst:conv-code}.  For simplicity, we assume the input tensor \texttt{x} has the same spatial height and width \texttt{F}, and channel size \texttt{M}. We stack \texttt{B} input tensors together to process them in a batch. We multiply each input tensor with \texttt{kernel}, a weight tensor whose horizontal and vertical sizes are both \texttt{K}, and the filter size is \texttt{N}. For simplicity, we assume stride of the convolution is 1, and we use a padding strategy such that the output tensor \texttt{y} has the same spatial dimensions \texttt{F} as input \texttt{x}. The channel size of the output is \texttt{N}. We ignore the computation of adding a bias to the output and applying the nonlinear activation functions since their computational cost is negligible. This representation can also be used for other layer types. For example, by setting \texttt{F=1, K=1}, Listing \ref{lst:conv-code} can represent a fully connected layer. 

\definecolor{codegreen}{rgb}{0,0.6,0}
\definecolor{codegray}{rgb}{0.5,0.5,0.5}
\definecolor{codepurple}{rgb}{0.58,0,0.82}
\definecolor{backcolour}{rgb}{0.95,0.95,0.92}
\begin{lstlisting}[
    language=Python,
    caption=Pseudo code to compute a convolutional layer,
    label={lst:conv-code},
    commentstyle=\color{codegreen},
    backgroundcolor=\color{backcolour},
    keywordstyle=\color{magenta},
    numberstyle=\tiny\color{codegray},
    stringstyle=\color{codepurple},
    tabsize=2, 
    numbers=left]
for b in range(0, B): # batch size
  for i in range(0, F): # image height
    for j in range(0, F): # image width
      for n in range(0, N): # filter
        for m in range(0, M): # channel 
          for p in range(-K/2,K/2): # vertical kernel
            for q in range(-K/2,K/2): # horizontal kernel
              y[b,n,i,j] += x[b,m,i+p,j+q]*kernel[n,m,p+K/2,q+K/2]
\end{lstlisting}

\subsection{Theoretical metrics}

Using this nested for-loop representation, we can easily calculate several theoretical metrics to evaluate the complexity of a neural network layer. 

\textbf{MACs (FLOPs)}: The inner-most operation in Listing \ref{lst:conv-code} is a multiply-and-accumulate (MAC) operation: it multiplies a number \texttt{x[b,m,i+p,j+q]} from the input tensor with \texttt{kernel[n,m,p+K/2,q+K/2]} from the weight tensor , and add the result to \texttt{y[b,n,i,j]} in the output tensor. The number of MACs directly reflects the number of compute operations of a layer, therefore its complexity. For a convolutional layer with the above configurations, the number of MACs is computed as 
\begin{equation}
    B \times F \times F \times M \times N \times K \times K.
\end{equation}
When comparing different neural networks, the batch size is by default set to 1. Since neural network weights and activations are typically represented in floating-point numbers, sometimes people also use the number of floating-point operations (FLOPs) to refer to the same concept. Strictly speaking, a MAC operation consists of a multiplication and an addition so that it can be counted as two separate floating-point operations. In addition to matrix multiplication, neural network layers can also add a bias term to the output. Such operations can also be counted as FLOPs, but not MACs. However, since the number of bias addition operations is many fewer than MACs, we can ignore bias additions, and only consider MACs. Following the convention, in this thesis, we use FLOPs and MACs interchangeably to denote the same concept unless otherwise noted. 

\textbf{Parameter size}: To perform a MAC operation, we need to load both the input and weight of a layer into CPUs. As explained above, data loading, especially from off-chip DRAMs, can consume a huge amount of energy and time, so it is important to figure out how many memory accesses is needed. First, we consider the parameters. The parameter size of a convolutional layer is calculated as 
\begin{equation}
     M \times N \times K \times K.
\end{equation}
In streaming tasks such as video recognition, new inputs are continuously fed into the network, but the network weights can be reused, so it does not grow with the batch size \texttt{B}. In an ideal case, if the parameter size of the neural network is small enough and can fit in the on-chip cache memory, we can load the weights once and avoid the repeated weight accesses from off-chip DRAMs. This can save a huge amount of time and energy. In practice, even if the weights of a neural network cannot fit into the on-chip cache memory, reducing the parameter size can also reduce memory accesses and therefore improve the efficiency.

\textbf{Activation size}: The input and output of a neural network layer are also referred to as activations. During the computation, input needs to be loaded to the CPU while the output needs to be written back to the memory. Different from neural network weights, during inference, the input of a layer can be discarded after the output is computed. This is called the ``double-buffer'' strategy. For a convolutional layer, memory accesses needed by the input and output activations can be computed as 
\begin{equation}
      B \times (M + N) \times F \times F.
\end{equation}

\textbf{Arithmetic intensity}: Since memory accesses consume significantly more energy than compute operations, for a given neural network layer, we would like to perform as many compute operations as possible while reducing memory accesses. To evaluate the potential of data reuse of a neural network layer, we compute its arithmetic intensity by dividing the number of MACs by the number of memory accesses as 
\begin{equation}
 \frac{B \times F \times F \times M \times N \times K \times K}{M \times N \times K \times K + B \times (M + N) \times F \times F}.  
\label{eqn:arithmetic-intensity-batch}
\end{equation}
The number of memory accesses includes both the weights and activations. The arithmetic intensity provides an estimation of how much we can reuse the fetched data during the computation of a neural network layer. 

\subsection{Example: variants of convolutional layers}
We use the theoretical metrics above to analyze some popular variations of convolutional layers for computer vision. The equations to calculate the parameters, MACs, and arithmetic intensity are summarized in Table \ref{tab:conv-metrics}. To provide more intuitive examples, we compute those metrics with typical layer configurations. The results are summarized in Table \ref{tab:metrics-comparison}.  

\begin{table}[h]
\centering
\begin{tabular}{|c|c|c|c|}
\hline
                                                                     & Params  & MACs         & \begin{tabular}[c]{@{}c@{}}Arithmetic\\ intensity\end{tabular} \\ \hline
\begin{tabular}[c]{@{}c@{}}Spatial\\ conv\end{tabular}               & $MNK^2$   & $BMNK^2F^2$   & $\frac{BMNK^2F^2}{BF^2(M+N) + K^2MN}$                            \\ \hline
\begin{tabular}[c]{@{}c@{}}Spatially\\ separable \\ conv\end{tabular} & $MNK$     & $BMNKF^2$     & $\frac{BMNKF^2}{BF^2(M+N) + KMN}$                                \\ \hline
\begin{tabular}[c]{@{}c@{}}Pointwise\\ conv\end{tabular}             & $MN$      & $BMNF^2$      & $\frac{BMNF^2}{BF^2(M+N) + MN}$                                  \\ \hline
\begin{tabular}[c]{@{}c@{}}Group\\ conv\end{tabular}                 & $MNK^2/G$ & $BMNK^2F^2/G$ & $\frac{BMNK^2F^2/G}{BF^2(M+N) + K^2MN/G}$                        \\ \hline
\begin{tabular}[c]{@{}c@{}}Depthwise\\ conv\end{tabular}             & $MK^2$    & $BMK^2F^2$    & $\frac{BMK^2F^2}{2BMF^2 + K^2M}$                                 \\ \hline
\end{tabular}
\caption{Theoretical metrics of convolutional layers.}
\label{tab:conv-metrics}
\end{table}

\begin{table}[h]
\centering
\begin{tabular}{|c|c|c|c|c|c|c|}
\hline
\multirow{2}{*}{}                                                    & \multicolumn{2}{c|}{Params} & \multicolumn{2}{c|}{MACs}             & \multicolumn{2}{c|}{\begin{tabular}[c]{@{}c@{}}Arithmetic\\ intensity\end{tabular}} \\ \cline{2-7} 
                                                                     & Early   & Late              & Early             & Late              & Early                                     & Late                                    \\ \hline
\begin{tabular}[c]{@{}c@{}}Spatial\\ conv\end{tabular}               & 9,216   & $2.36\times10^6$  & $1.16\times10^8$  & $1.16\times10^8$  & 143                                       & 48.1                                    \\ \hline
\begin{tabular}[c]{@{}c@{}}Spatially\\ separable\\ conv\end{tabular} & 3072    & $7.86\times10^5$  & $3.85\times 10^7$ & $3.85\times 10^7$ & 47.8                                      & 46.1                                    \\ \hline
\begin{tabular}[c]{@{}c@{}}Pointwise\\ conv\end{tabular}             & 1024    & $2.62\times10^5$  & $1.28\times 10^7$ & $1.28\times 10^7$ & 16.0                                      & 41.1                                    \\ \hline
\begin{tabular}[c]{@{}c@{}}Group\\ conv\end{tabular}                 & 2304    & $5.89\times10^5$  & $2.89\times 10^7$ & $2.89\times 10^7$ & 35.9                                      & 45.2                                    \\ \hline
\begin{tabular}[c]{@{}c@{}}Depthwise\\ conv\end{tabular}             & 288     & 4,608             & $3.61\times10^6$  & $2.26\times10^5$  & 4.50                                      & 4.12                                    \\ \hline
\end{tabular}
\caption[Theoretical metrics of convolutional layers with typical layer configurations.]{Comparison of convolutional layers with typical layer configurations. ``Early'' denotes a layer next to the input of a convolutional neural network. ``Late'' denotes a layer close to the output of a network. In all calculations we assume the batch size $B=1$. For the early layer, we assume $M=N=32$, $F=112$, and $K=3$. For the late layer, we assume  $M=N=512$, $F=7$ and $K=3$. For the group convolution, we assume $G=4$.}
\label{tab:metrics-comparison}
\end{table}

\textbf{Spatial convolution}: a vanilla spatial convolution is characterized by the following hyper-parameters: the number of input channels M, the number of output filters N and kernel size K. Its parameter size, number of MACs and arithmetic intensity are summarized in Table \ref{tab:conv-metrics}. Spatial convolutions are expensive in parameter size and FLOPs as they grow quadratically with the kernel size. Therefore, many works \cite{SqueezeNet, howard2017mobilenets, wu2017shift} aim to reduce the parameter size and MACs by replacing spatial convolutions. We will talk about this in more detail in Chapter \ref{chap:shift}. On the other hand, however, spatial convolutions have a higher arithmetic intensity and therefore have higher potential for data reuse, as shown in Table \ref{tab:metrics-comparison}. 

\textbf{Spatially separable convolution}: Using tensor factorization techniques, a spatial convolution of size $K \times K$  can be factorized to two convolutions of with kernel sizes of $K \times 1$ and $1 \times K$. Compared with spatial convolution, a spatially separable convolution has $K$ times smaller parameter size and MACs. This technique is adopted in works such as \cite{gholami2018squeezenext} to reduce the neural network complexity. 

\textbf{Pointwise convolution} is a special case of the spatial convolution with kernel size $K=1$. Compared with spatial convolutions where $K > 1$, pointwise convolutions have smaller parameter size and MACs, but the arithmetic intensity is lower because there is less opportunity for data reuse.

\textbf{Group convolution} divides the input and output channels into G groups, and output in one group is only dependent on the corresponding input channels. As a result, for a given input channel size M and number of filters N, the parameter size and FLOPs of a group convolution is G times smaller than spatial convolutions. However, in practice, this radical reduction in parameter size and FLOPs usually leads to degraded accuracy. A common compensation strategy to mitigate the accuracy loss is to increase the channel size (M) or the number of filters (N), for example, increase N by G times such that the parameter size and FLOPs are the same as the original spatial convolution. This strategy usually leads to better performance \cite{Xie2016resnext}. However, such a strategy results in lower arithmetic intensity and therefore requires higher memory bandwidth from the hardware.

\textbf{Depth-wise convolution} is an extreme case of group convolution where the group number G equals the input channel size M, and also generates M output channels. As discussed above, depth-wise convolution has a very low arithmetic intensity, as shown in Table \ref{tab:metrics-comparison}. As a result, even though its FLOPs and parameter size is trivial, if not carefully handled, it can be very slow, as reported in ShuffleNet \cite{ShuffleNet}. 

\section{Practical efficiency metrics and how to optimize them}
\subsection{Practical efficiency metrics}
All the theoretical metrics mentioned above are simple to compute and hardware-agnostic. Therefore, they are widely adopted to measure and compare the efficiency of neural networks. However, in practical applications, what people care about more are metrics such as are speed (latency or throughput), power, and energy. 

\textbf{Latency}: Latency means the interval between the start and end of the computation of a neural network. It is critical for applications such as autonomous driving and augmented reality, where real-time interaction with the environment is needed. As a first-order approximation, on a given processor, a neural network with more FLOPs, more parameters, and activation size will need more time to finish the computation and therefore will have higher latency.

\textbf{Throughput}: Throughput means the number of input processed per unit time. This is different from latency, and it is a critical metric, especially for cloud-based, high-volume applications. On parallel architectures, an effective way to improve throughput is to stack input data in a batch and process them together. This not necessarily improves, sometimes even hurts the latency, but it can process more inputs in the same amount of time. Adding a new batch dimension in Listing \ref{lst:conv-code} increases more degrees of freedom to optimize the dataflow, therefore, leads to better utilization. In addition, batching can leads to higher arithmetic intensity. From Equation \ref{eqn:arithmetic-intensity-batch}, we can see that while MACs and activation size grows linearly with the batch size \texttt{B}, the parameter size stay the same, so increasing the batch size can improve the arithmetic intensity. As a concrete example, for VGG16 \cite{VGG}, the $fc1$ layer is a fully connected layer that transforms an unrolled $7 \times 7 \times 512$ input tensor to a vector with $4096$ dimensions. The arithmetic intensity of this layer is
\[
\frac{B \times 7 \times 7 \times 512 \times 4096}{7 \times 7 \times 512 \times 4096 + B \times (7 \times 7 \times 512 + 4096)}.
\]
In this case, when the batch size is small, the memory accesses for parameters dominate, so the arithmetic intensity grows almost linearly with the batch size, until it eventually converges when the batch size \texttt{B} $\to \infty$.

\textbf{Power}: Power efficiency is the energy consumed per unit time. For a deep learning system, the constraint on power can come from several factors, including power supply, heat dissipation, or overtime, energy constraint. The power efficiency of a deep learning system depends on both the thermal design point of the hardware processor and the neural network. To improve the power efficiency, an effective way is to optimize the neural network to reduce its MACs, and more importantly, parameter and activation size for fewer memory accesses. Moreover, a more compact neural network design enables us to deploy the neural network on low-power processors with weaker compute capacity. 

\textbf{Energy}: The energy efficiency of a DNN is defined as the energy consumed per data point (such as image, voice, sentence). In many battery-based embedded applications, the energy consumption of a neural network is a primary metric for efficiency. As explained before, the energy consumption is dominated by memory operations, which are highly dependent on the parameter and activation size. Optimizing those metrics lead to lower energy consumption. 

\subsection{Guidelines for optimizing practical efficiency}
The practical efficiency of neural networks not only depends on the neural network architecture itself but also on the underlying hardware processor and how the computations are mapped to the hardware. Given the complexity of modern computer architectures, it is difficult, if not impossible, to establish a simple relationship to map the theoretical metrics such as MACs and parameter size to practical metrics such as latency and energy. In this section, we try to provide some high-level guidelines on how to optimize neural network efficiency. Details of how to apply these guidelines will be discussed in later chapters of this thesis. 

\textbf{Reducing MACs, parameter sizes, and activation sizes}. As a first-order approximation, hardware-agnostic metrics are very useful to estimate the practical efficiency of a neural network. Taking parameter size as an example, for streaming applications such as video recognition, if a neural network is small enough that it can fit into the cache memory entirely, we can load the model once and reuse its parameters for all the inputs. This leads to a significant energy reduction. Even if parameters cannot fit into the cache memory, reducing the parameter size can still lead to fewer memory accesses, therefore improve the efficiency. We will discuss this strategy in more detail in Chapter \ref{chap:sqdt} and \ref{chap:sqsg}.

\textbf{Model-hardware co-design}. Given the complexity of modern computer architectures, theoretical metrics such as FLOPs and parameter sizes do not always align with practical efficiency metrics. A neural network with fewer FLOPs can have much higher latency due to its complicated network structure that cannot be efficiently computed on hardware processors \cite{zoph2017learning, howard2019searching, sandler2018mobilenetv2}. Therefore, it is important that we understand how the computation of a neural network is executed on the underlying hardware and how customized hardware can boost the efficiency of neural networks. In Chapter \ref{chap:shift}, we will discuss how we co-design a neural network and hardware accelerator to achieve significant efficiency improvements.

\textbf{Neural architecture search}. Designing efficient neural networks is intrinsically a difficult problem since both the accuracy and practical efficiency of neural networks are difficult, if not impossible, to predict. Also, for a deep neural network with many layers, each layer can choose a different layer configuration. This leads to an intractable design space. In Chapter \ref{chap:dnas}, we will discuss how to formulate the design of neural networks as an optimization problem and use efficient algorithms to automatically search for neural networks with high accuracy and practical efficiency. 

%% file: chap3.tex
\chapter{Model Efficiency: SqueezeDet}
\label{chap:sqdt} 
In this chapter, we continue to discuss the model efficiency of deep neural networks. It has been shown in previous works that increasing the model complexity is an effective way to improve the performance of neural networks. However, the increased model complexity also leads to higher computational burdens, making it more difficult to deploy neural networks to mobile and embedded devices. In Chapter \ref{chap:metrics}, we raised a key question: 
\begin{quote}
    Is it possible to design neural networks to achieve the same performance with lower model complexity?
\end{quote}
 This question is first addressed by SqueezeNet \cite{iandola2016squeezenet} in 2015. SqueezeNet is a neural network that achieved the same accuracy as AlexNet, but with only 1.2 million parameters, or 50x fewer than AlexNet. However, SqueezeNet is designed only for the image classification problem. Starting from SqueezeNet, we want to explore two \textit{key questions}: 
\begin{quote}
    Can we design efficient neural networks to solve more general computer vision problems, such as object detection?
\end{quote}
Further more:
\begin{quote}
    Can we design efficient neural networks to process other visual modalities, such as depth measurements from LiDAR sensors? 
\end{quote}

We answer these two questions in Chapters \ref{chap:sqdt} and \ref{chap:sqsg}. In this chapter, we will introduce SqueezeDet, an efficient network designed for image object detection. In Chapter \ref{chap:sqsg}, we discuss SqueezeSeg, an efficient network designed for LiDAR point cloud segmentation. We show that by carefully formulating the problem, designing the training protocol and the neural network model, we are able to achieve over more than 10x efficiency improvement over baseline solutions.

\section{Object detection for autonomous driving}
Object detection is a fundamental problem in computer vision, and it is a crucial task in many applications, such as autonomous driving. In this chapter, we will use autonomous driving as a motivating application. 

A safe and robust autonomous driving system relies on accurate perception of the environment. To be more specific, an autonomous vehicle needs to accurately detect cars, pedestrians, cyclists, road signs, and other objects in real-time in order to make right control decisions that ensure safety. Moreover, to be economical and widely deployable, this object detector must operate on embedded processors that dissipate far less power than powerful GPUs used for benchmarking in typical computer vision experiments.

While recent research has been primarily focused on improving accuracy, for actual deployment in an autonomous vehicle, there are other issues of image object detection that are equally critical. For autonomous driving, some basic requirements for image object detectors include the following: a) Accuracy. More specifically, the detector ideally should achieve $100\%$ recall with high precision on objects of interest. b) Speed. The detector should have real-time (typically 30 frames per second) or faster inference speed to reduce the latency of the vehicle control loop. c) Small model size. As discussed in~\cite{iandola2016squeezenet}, smaller model size brings benefits of more efficient distributed training, less communication overhead to export new models to clients through wireless update, less energy consumption, and more feasible embedded system deployment. d) Power and energy efficiency. Desktop and rack systems may have the luxury of burning 250W of power for neural network computation, but embedded processors targeting the automotive market must fit within a much smaller power and energy envelope due to both energy and heat dissipation constraints. While precise figures vary, the new Xavier\footnote{https://blogs.nvidia.com/blog/2016/09/28/xavier/} processor from Nvidia, for example, is targeting a 20W thermal design point. Processors targeting battery-based applications have an even smaller power and energy budget and must fit in the 3W--10W range to prevent overheating and ensure battery lives. Without addressing the problems of a) accuracy, b) speed, c) small model size, and d) energy and power efficiency, we will not be able to truly leverage the power of deep neural networks for autonomous driving.  

\section{Related work}
\subsection{Neural Networks for object detection}
From 2005 to 2013, various techniques were applied to advance the accuracy of object detection on datasets such as PASCAL~\cite{PASCAL}.
In most of these years, versions of HOG (histogram of oriented gradients) + SVM (support vector machine)~\cite{HOG} or DPM (deformable parts model)~\cite{DPM-journal} led the state-of-art accuracy on these datasets.
However, in 2013, Girshick et al. proposed Region-based Convolutional Neural Networks (R-CNN)~\cite{R-CNN}, which led to substantial gains in object detection accuracy.
The R-CNN approach begins by identifying region proposals (i.e. regions of interest that are likely to contain objects) and then classifying these regions using a CNN.
One disadvantage of R-CNN is that it computes the CNN independently on each region proposal, leading to time-consuming ($\leq$ 1 fps) and energy-inefficient ($\geq$ 200 J/frame) computation. 
To remedy this, Girshick et al. experimented with a number of strategies to amortize computation across the region proposals~\cite{DPMareCNN,DenseNet,Fast-R-CNN}, culminating in {\em Faster R-CNN}~\cite{Faster-R-CNN}. Another model, R-FCN, is fully-convolutional and delivers accuracy that is competitive with R-CNN, but R-FCN is fully-convolutional, which allows it to amortize more computation across the region proposals.

There have been a number of works that have adapted the R-CNN approach to address object detection for autonomous driving. Almost all the top-ranked published methods on the KITTI \cite{KITTI} leader board are based on Faster R-CNN. Ashraf et al. \cite{ShallowNetworks} modified the CNN architecture to use shallower networks to improve accuracy. Multi-scale CNN \cite{mscnn} and Sub-CNN \cite{subcnn} on the other hand focused on generating better region proposals. Most of these methods focused on better accuracy, while few previous methods have reported real-time inference speeds on the KITTI dataset \cite{KITTI}. 

{\em Region proposals} are a cornerstone in all of the object detection methods that we have discussed so far. However, in YOLO (You Only Look Once)~\cite{YOLO}, region proposition and classification are integrated into one single stage. Compared with R-CNN and Faster R-CNN based methods, YOLO's single-stage detection pipeline is extremely fast, making YOLO the first CNN based general-purpose object detection model that achieved real-time speed.

\subsection{Small CNN models}
For any particular accuracy level on a computer vision benchmark, it is usually feasible to develop multiple CNN architectures that are able to achieve that level of accuracy.
Given the same level of accuracy, it is often beneficial to develop smaller CNNs (i.e. CNNs with fewer model parameters), as discussed in~\cite{SqueezeNet}. AlexNet~\cite{alexnet} and VGG-19~\cite{VGG-19} are CNN model architectures that were designed for image classification and have since been modified to address other computer vision tasks.
The AlexNet model contains 240MB of parameters, and it delivers approximately 80\% top-5 accuracy on ImageNet~\cite{deng2009imagenet} image classification.
The VGG-19 model contains 575MB of parameters and delivers about 87\% top-5 accuracy on ImageNet.
However, models with fewer parameters can deliver similar levels of accuracy.
The SqueezeNet~\cite{SqueezeNet} model has only 4.8MB of parameters (50x smaller than AlexNet), and it matches or exceeds AlexNet-level accuracy on ImageNet.
The GoogLeNet-v1~\cite{googlenet} model only has 53MB of parameters, and it matches VGG-19-level accuracy on ImageNet.

\section{Method}
\subsection{Detection Pipeline}
\begin{figure}[h]
  \centering
 \includegraphics [width=.8\textwidth]{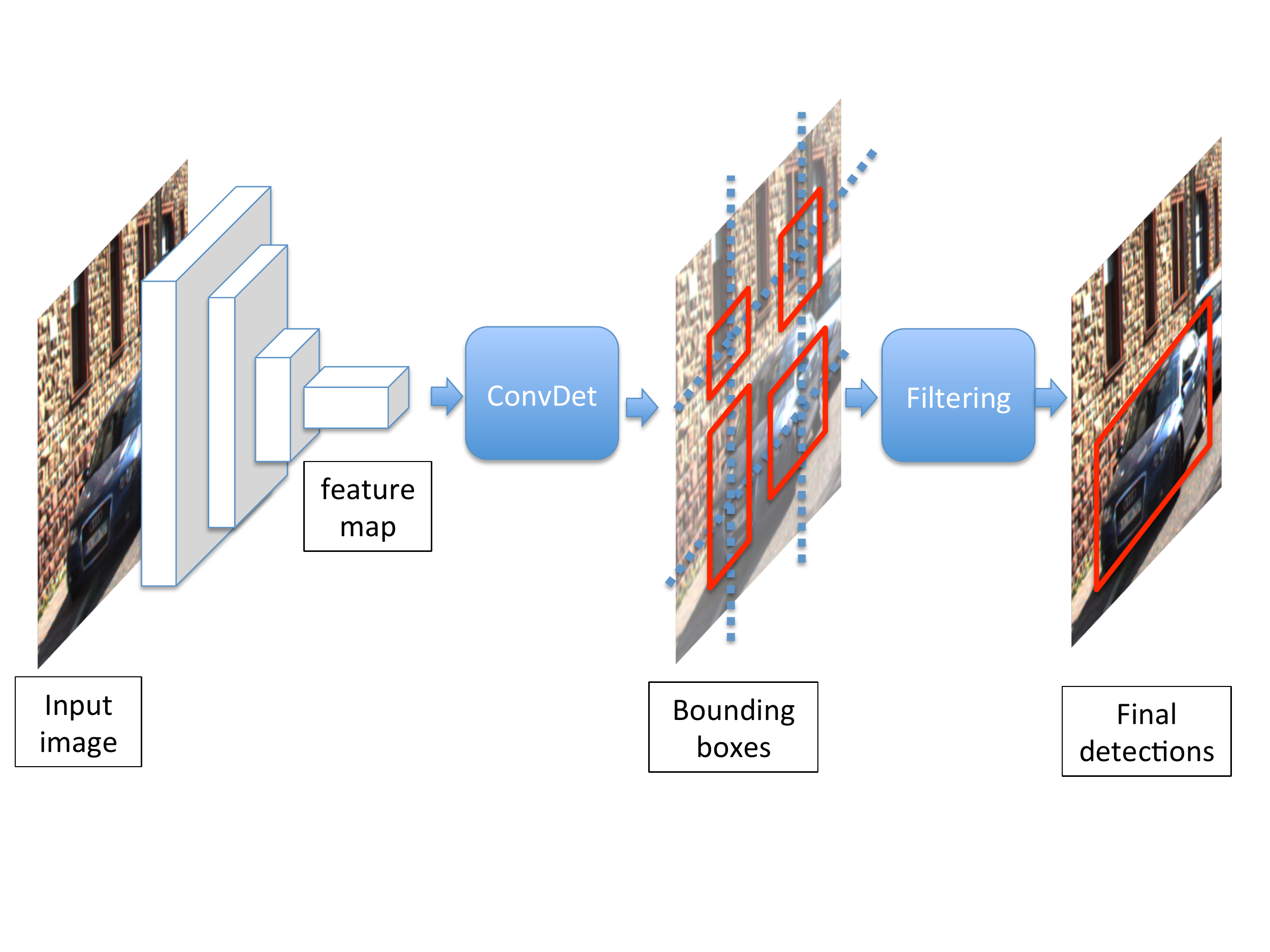}
 \caption[SqueezeDet detection pipeline.]{SqueezeDet detection pipeline. A convolutional neural network extracts a feature map from the input image and feeds it into the \textit{ConvDet} layer. The \textit{ConvDet} layer then computes bounding boxes centered around $W\times H$ uniformly distributed grid centers. Each bounding box is associated with $1$ confidence score and $C$ conditional class probabilities. Then, we keep the top $N$ bounding boxes with the highest confidence and use NMS to filter them to get the final detections.}
\label{fig:DetPipeline}
\end{figure}

Inspired by YOLO~\cite{YOLO}, we adopt a single-stage detection pipeline: region proposition and classification is performed by one single network simultaneously. As shown in Fig.\ref{fig:DetPipeline}, a convolutional neural network first takes an image as input and extract a low-resolution, high dimensional feature map from the image. Then, the feature map is fed into the \textit{ConvDet} layer to compute bounding boxes centered around $W\times H$ uniformly distributed spatial grids. Here, $W$ and $H$ are numbers of grid centers along horizontal and vertical directions. 

Each bounding box is associated with $C+1$ values, where $C$ is the number of classes to distinguish, and the extra $1$ is for the confidence score, which indicates how likely does the bounding box contain an object. Similar to YOLO~\cite{YOLO}, we define the confidence score as $Pr(\text{Object})*\text{IOU}_{truth}^{pred}$. A high confidence score implies a high probability that an object of interest does exist and that the overlap between the predicted bounding box and the ground truth is high. The other $C$ scalars represent the conditional class probability distribution given that the object exists within the bounding box. More formally, we denote the conditional probabilities as $Pr(\text{class}_c|\text{Object}), c \in [1,C].$ We assign the label with the highest conditional probability to this bounding box, and we use 
\[
	\max_c Pr(\text{class}_c|\text{Object}) * Pr(\text{Object})*\text{IOU}_{truth}^{pred}
\]
as the metric to estimate the confidence of the bounding box prediction. 

Finally, we keep the top $N$ bounding boxes with the highest confidence and use Non-Maximum Suppression (NMS) to filter redundant bounding boxes to obtain the final detections. During inference, the entire detection pipeline consists of only one forward pass of one neural network with minimal post-processing.

\subsection{ConvDet}
\label{sec:convdet}

The \textit{SqueezeDet} detection pipeline is inspired by YOLO~\cite{YOLO}. However, as we will describe in this section, the design of the \textit{ConvDet} layer enables SqueezeDet to generate tens-of-thousands of region proposals with fewer model parameters compared to YOLO. 

\textit{ConvDet} is essentially a convolutional layer that is trained to output bounding box coordinates and class probabilities. It works as a sliding window that moves through each spatial position on the feature map. At each position, it computes \(K\times(4+1+C)\) values that encode the bounding box predictions. Here, $K$ is the number of reference bounding boxes with pre-selected shapes. Following the notation from~\cite{Faster-R-CNN}, we call these reference bounding boxes anchors. Each position on the feature map corresponds to a grid center in the original image, so each anchor can be described by 4 scalars \( (\hat{x}_i, \hat{y}_j, \hat{w}_k, \hat{h}_k), i \in [1, W], j\in [1, H], k \in [1, K]. \) Here $\hat{x}_i, \hat{y}_i$ are spatial coordinates of the reference grid center $(i, j)$. $\hat{w}_k, \hat{h}_k$ are the width and height of the $k$-th reference bounding box. We use the method described by Ashraf et al.~\cite{ShallowNetworks} to select reference bounding box shapes to match the data distribution. 

\begin{figure}[h]
  \centering
 \includegraphics [width=.8\linewidth]{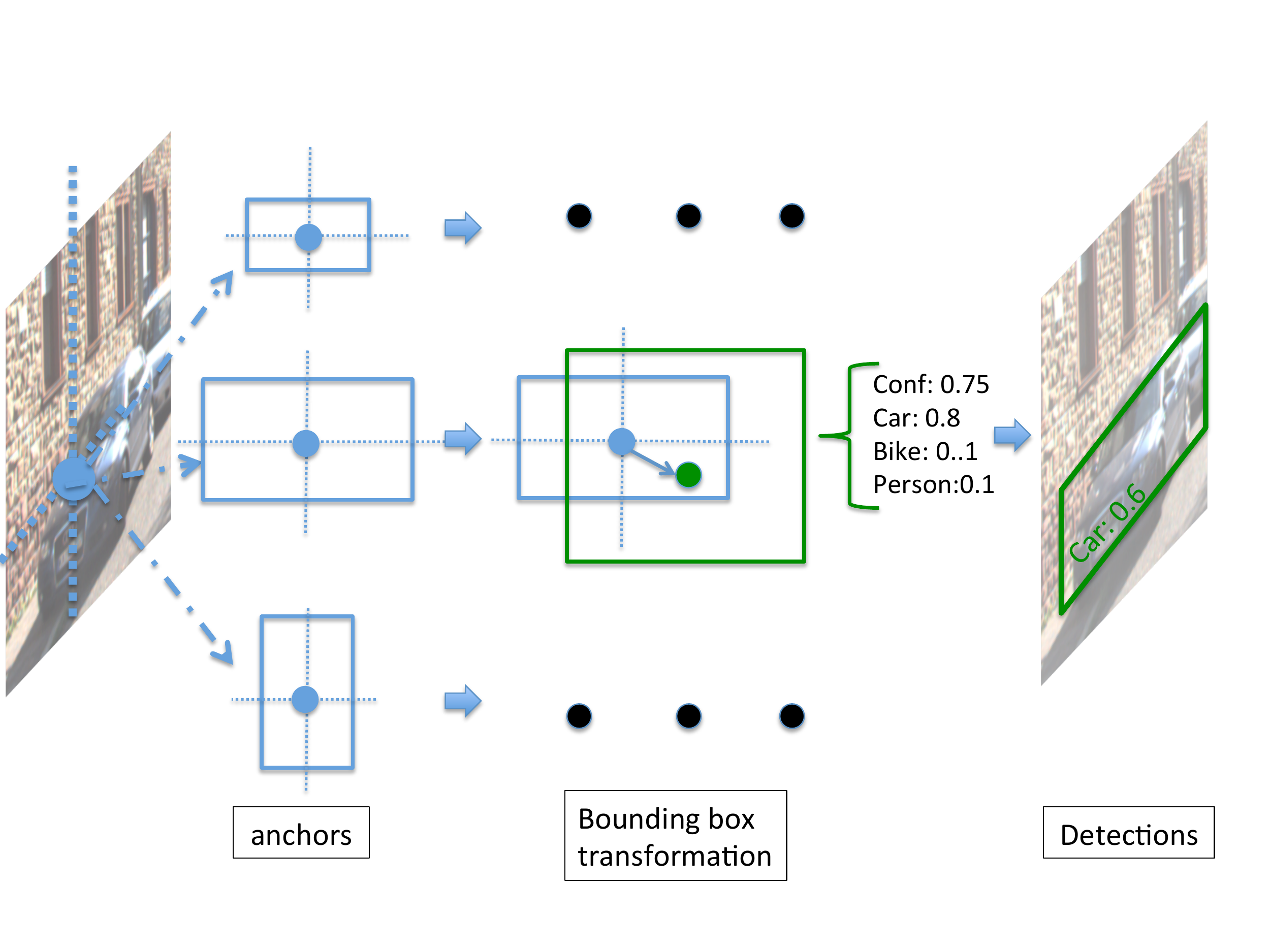}
 \caption[Bounding box transformation.]{Bounding box transformation. Each grid center has $K$ anchors with pre-determined shapes. Each anchor is transformed into its new position and shape using the relative coordinates computed by the \textit{ConvDet} layer. Each anchor is associated with a confidence score and class probabilities of the object within the bounding box.}
\label{fig:bbox_transform}
\end{figure}

For each anchor $(i, j, k)$, we compute $4$ relative coordinates \((\delta x_{ijk},  \delta y_{ijk}, \delta w_{ijk}, \delta h_{ijk})\) to transform the anchor into a predicted bounding box, as shown in Fig. \ref{fig:bbox_transform}. Following Faster R-CNN~\cite{R-CNN-supp}, the transformation is described as 
\begin{equation}
\label{eq:bbox_trans}
\begin{gathered}
	x_i^p = \hat{x}_i + \hat{w}_k\delta x_{ijk}, ~
    y_j^p = \hat{y}_j + \hat{h}_k\delta y_{ijk}, \\
    w_k^p = \hat{w}_k \exp(\delta w_{ijk}), ~
    h_k^p = \hat{h}_k \exp(\delta h_{ijk}),
\end{gathered}
\end{equation}
where $x_i^p, y_j^p, w_k^p, h_k^p$ are predicted bounding box coordinates. As explained in the previous section, the other $C+1$ outputs for each anchor encode the confidence score for this prediction and conditional class probabilities.

\textit{ConvDet} is similar to the last layer of RPN in Faster R-CNN~\cite{Faster-R-CNN}. The main difference is that RPN is regarded as a ``weak'' detector that is only responsible for detecting whether an object exists and generating bounding box proposals for the object. The classification is handed over to fully connected layers, which are regarded as a ``strong'' classifier. However,  convolutional layers are in fact ``strong'' enough to detect, localize, and classify objects at the same time.

\begin{figure}[ht]
\centering
  \begin{subfigure}[t]{.49\linewidth}
    \centering\includegraphics[clip, trim=1cm 3.5cm 2.5cm 6cm, width=\linewidth]{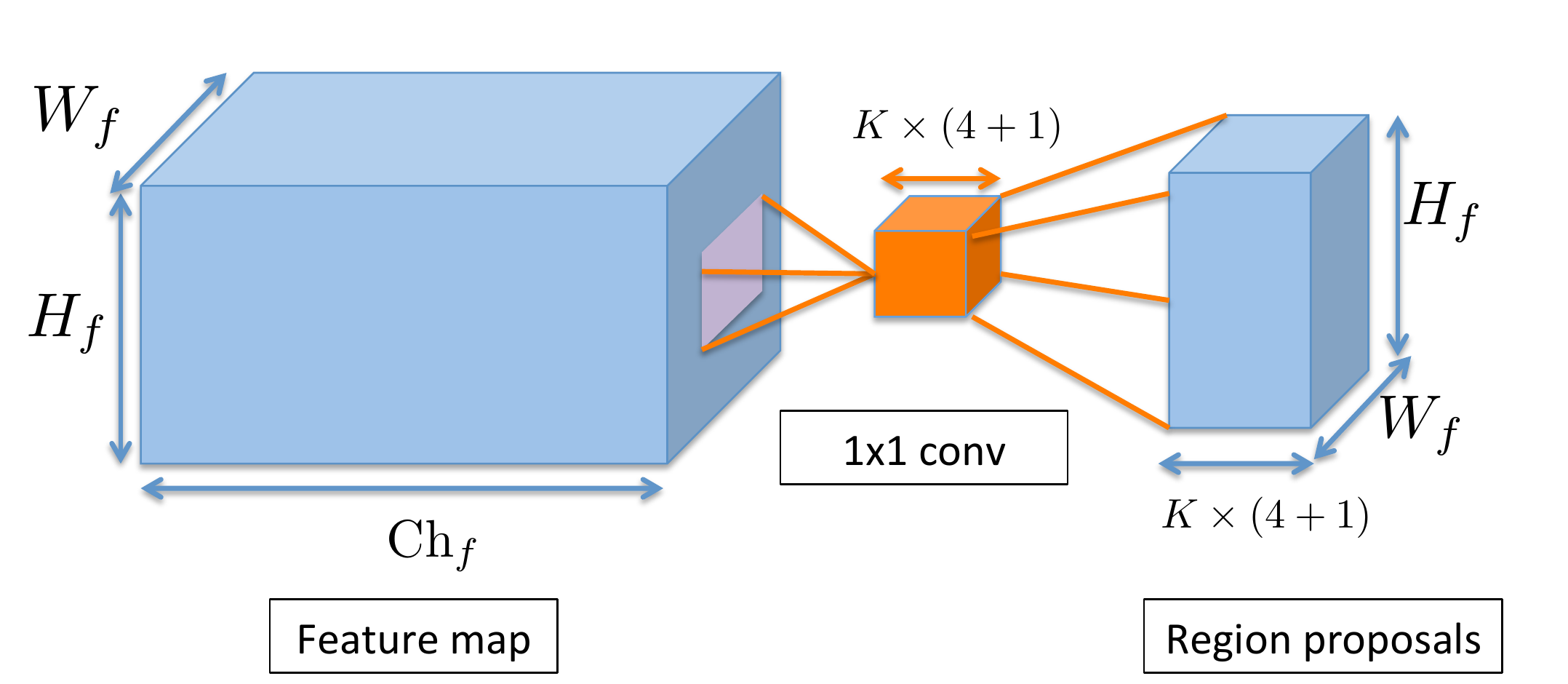}
    \caption{Last layer of Region Proposal Network (RPN) is a 1x1 convolution with $K\times (4+1)$ outputs. $4$ is the number of relative coordinates, and $1$ is the confidence score. It's only responsible for generating region proposals. The parameter size for this layer is $\text{Ch}_f \times K \times 5$.}
  \end{subfigure} \hfill
  \begin{subfigure}[t]{.49\linewidth}
    \centering\includegraphics[clip, trim=1cm 3.5cm 1cm 5cm, width=\linewidth]{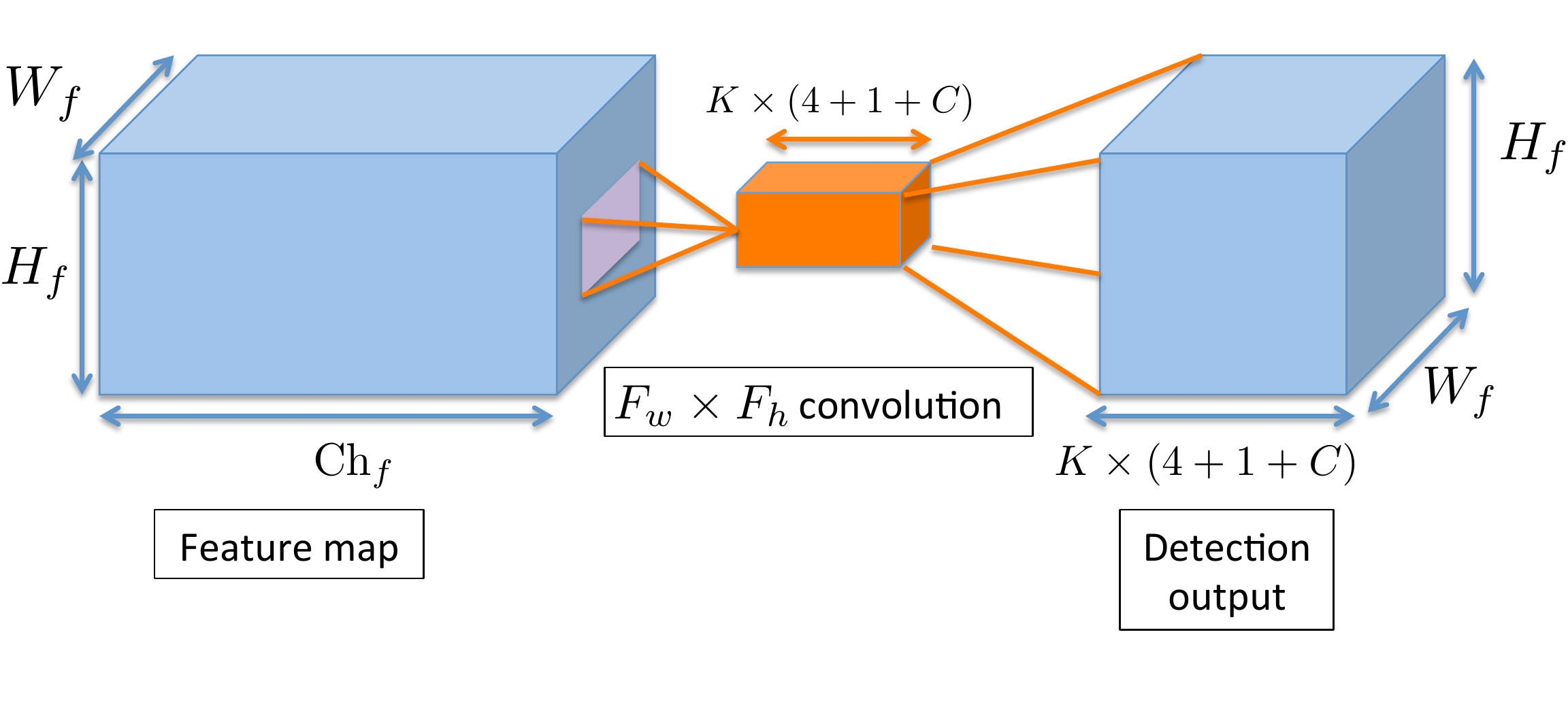}
    \caption{The \textit{ConvDet} layer is a $F_w \times F_h$ convolution with output size of $K\times (5+C)$. It's responsible for both computing bounding boxes and classifying the object within. The parameter size for this layer is $F_w F_h \text{Ch}_fK(5+C)$. }
  \end{subfigure} \hfill
  \begin{subfigure}[t]{.49\linewidth}
    \centering\includegraphics[clip, trim=0cm 5cm 0cm 1cm, width=\linewidth]{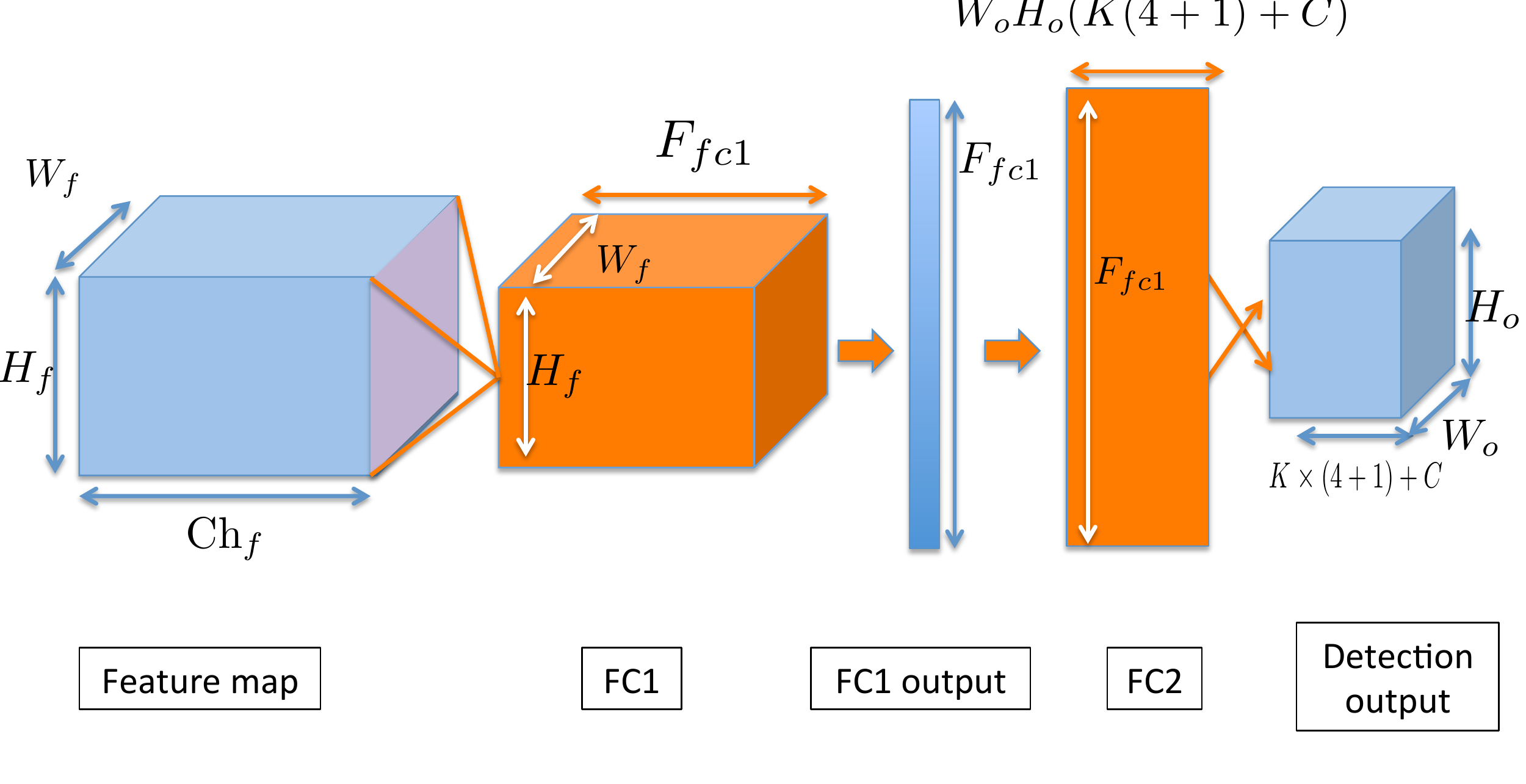}
    \caption{The detection layer of YOLO~\cite{YOLO} contains 2 fully connected layers. The first one is of size $W_fH_f\text{Ch}_fF_{fc1}$. The second one is of size $F_{fc1} W_oH_oK(5+C)$. }
  \end{subfigure}
\caption[RPN vs. ConvDet vs. FcDet.]{Comparing RPN, \textit{ConvDet} and the detection layer of YOLO~\cite{YOLO}. Activations are represented as blue cubes and layers (and their parameters) are represented as orange ones. Activation and parameter dimensions are also annotated.}
\label{fig:RPN_ConvDet_YOLO}
\end{figure}

For simplicity, we denote the detection layers of YOLO~\cite{YOLO} as \textit{FcDet} (only counting the last two fully connected layers). Compared with \textit{FcDet}, the \textit{ConvDet} layer has orders of magnitude fewer parameters and is still able to generate more region proposals with higher spatial resolution. The comparison between \textit{ConvDet} and \textit{FcDet} is illustrated in Fig.~\ref{fig:RPN_ConvDet_YOLO}. 

Assume that the input feature map is of size $(W_f, H_f, \text{Ch}_f)$, $W_f$ is the width of the feature map, $H_f$ is the height, and $\text{Ch}_f$ is the number of input channels to the detection layer. Denote \textit{ConvDet}'s filter width as $F_w$ and height as $F_h$. With proper padding/striding strategy, the output of \textit{ConvDet} keeps the same spatial dimension as the feature map. To compute $K\times(4+1+C)$ outputs for each reference grid, the number of parameters required by the \textit{ConvDet} layer is $F_wF_h\text{Ch}_fK(5+C)$. 

The \textit{FcDet} layer described in YOLO \cite{YOLO} is comprised of two fully connected layers. Using the same notation for the input feature map and assuming the number of outputs of the $fc1$ layer is $F_{fc1}$, then the number of parameters in the $fc1$ layer is $W_fH_f\text{Ch}_fF_{fc1}$. The second fully connected layer in YOLO \cite{YOLO} generates $C$ class probabilities as well as $K\times(4+1)$ bounding box coordinates and confidence scores for each of the $W_o \times H_o$ grids. Thus, the number of parameters in the $fc2$ layer is $F_{fc1}W_oH_o(5K+C)$. The total number of parameters in these two fully connected layers is $F_{fc1}(W_fH_f\text{Ch}_f + W_oH_o(5K+C))$. 

In \cite{YOLO}, the input feature map is of size 7x7x1024. $F_{fc1} = 4096$, $K=2$, $C=20$, $W_o = H_o = 7$, thus the total number of parameters required by the two fully connected layers is approximately $212\times 10^6$. If we keep the feature map sizes, number of output grid centers, classes, and anchors the same, and use 3x3 \textit{ConvDet}, it would only require $3\times3\times 1024\times 2 \times 25 \approx 0.46\times 10^6$ parameters, which is 460X smaller then \textit{FcDet}. The comparison of RPN, \textit{ConvDet} and \textit{FcDet} is illustrated in Fig.~\ref{fig:RPN_ConvDet_YOLO} and summarized in Table~\ref{table:RPN_CONVDET_YOLO}.

\begin{table}
\begin{center}
\begin{tabular}{c|ccc}

 & RP & $cls$ & \#Parameter\\
\hline
RPN & \checkmark & \xmark & 5$\text{Ch}_fK$ \\
\textit{ConvDet} & \checkmark & \checkmark & $F_wF_h\text{Ch}_fK(5+C)$ \\
\textit{FcDet} & \checkmark & \checkmark & $F_{fc1}(W_fH_f\text{Ch}_f + W_oH_o(5K+C))$\\
\hline
\end{tabular}
\end{center}
\caption[Comparison between RPN, \textit{ConvDet} and \textit{FcDet}.]{Comparison between RPN, \textit{ConvDet} and \textit{FcDet}. RP stands for region proposition. $cls$ stands for classification. }
\label{table:RPN_CONVDET_YOLO}
\end{table}

\subsection{Training protocol}

Unlike Faster R-CNN~\cite{Faster-R-CNN}, which deploys a (4-step) alternating training strategy to train RPN and detector network, our SqueezeDet detection network can be trained end-to-end.

To train the \textit{ConvDet} layer to learn detection, localization, and classification, we define a multi-task loss function:   
\begin{equation}
\label{eq:loss}
\begin{gathered}
	\frac{\lambda_{bbox}}{N_{obj}} \sum_{i=1}^W \sum_{j=1}^H \sum_{k=1}^K 
    I_{ijk} [(\delta x_{ijk} - \delta x_{ijk}^G)^2
    +(\delta y_{ijk} - \delta y_{ijk}^G)^2 \\ 
    + (\delta w_{ijk} - \delta w_{ijk}^G)^2 + (\delta h_{ijk} - \delta h_{ijk}^G)^2]\\
    +  \sum_{i=1}^W \sum_{j=1}^H \sum_{k=1}^K 
   \frac{\lambda_{conf}^+}{N_{obj}} I_{ijk} (\gamma_{ijk} - \gamma_{ijk}^G)^2 
   + \frac{\lambda_{conf}^-}{WHK - N_{obj}}\bar{I}_{ijk} \gamma_{ijk}^2 \\
   + \frac{1}{N_{obj}} \sum_{i=1}^W \sum_{j=1}^H \sum_{k=1}^K \sum_{c=1}^C
   I_{ijk} l_c^G \log(p_c).
\end{gathered}
\end{equation}

The first part of the loss function is the bounding box regression. \((\delta x_{ijk}, \delta y_{ijk}, \delta w_{ijk}, \delta h_{ijk})\) corresponds to the relative coordinates of anchor-$k$ located at grid center-$(i,j)$. They are outputs of the \textit{ConvDet} layer. The ground truth bounding box $\delta_{ijk}^G$, or \((\delta x_{ijk}^G, \delta y_{ijk}^G, \delta w_{ijk}^G, \delta h_{ijk}^G)\), is computed as: 
\begin{equation}
\label{eq:bbox_trans_inv}
\begin{gathered}
	\delta x_{ijk}^G = (x^G - \hat{x}_i)/\hat{w}_k, ~
    \delta y_{ijk}^G = (y^G - \hat{y}_j)/\hat{h}_k, \\
    \delta w_{ijk}^G = \log(w^G/\hat{w}_k), ~
    \delta h_{ijk}^G = \log(h^G/\hat{h}_k). \\
\end{gathered}
\end{equation}
Note that Equation~\ref{eq:bbox_trans_inv} is essentially the inverse transformation of Equation~\ref{eq:bbox_trans}. \((x^G, y^G, w^G, h^G)\) are coordinates of a ground truth bounding box. During training, we compare ground truth bounding boxes with all anchors and assign them to the anchors that have the largest overlap (Intersection-over-Union) with each of them. The reason is that we want to select the ``closest'' anchor to match the ground truth box such that the transformation needed is reduced to the minimum. $I_{ijk}$ evaluates to 1 if the $k$-th anchor at position-$(i, j)$ has the largest overlap with a ground truth box, and to 0 if no ground truth is assigned to it. This way, we only include the loss generated by the ``responsible'' anchors. As there can be multiple objects per image, we normalize the loss by dividing it by the number of objects.

The second part of the loss function is confidence score regression. \(\gamma_{ijk}\) is the output from the \textit{ConvDet} layer, representing the predicted confidence score for anchor-$k$ at position-$(i,j)$.  \(\gamma_{ijk}^G\) is obtained by computing the IOU of the predicted bounding box with the ground truth bounding box. As above, we only include the loss generated by the anchor box with the largest overlap with the ground truth. For anchors that are not ``responsible'' for the detection, we penalize their confidence scores with the $\bar{I}_{ijk}\gamma_{ijk}^2$ term, where $\bar{I}_{ijk} = 1 - I_{ijk}$. Usually, there are much more anchors that are not assigned to any object. In order to balance their influence, we use $\lambda_{conf}^+$ and $\lambda_{conf}^-$ to adjust the weight of these two loss components. By definition, the confidence score's range is [0, 1]. To guarantee that $\gamma_{ijk}$ falls into that range, we feed the corresponding \textit{ConvDet} output into a \textit{sigmoid} function to normalize it.

The last part of the loss function is just cross-entropy loss for classification. $l_c^G \in \{0, 1\}$ is the ground truth label and $p_c \in [0, 1], c\in[1, C]$ is the probability distribution predicted by the neural net. We used \textit{softmax} to normalize the corresponding \textit{ConvDet} output to make sure that $p_c$ is ranged between $[0, 1]$.

The hyper-parameters in Equation~\ref{eq:loss} are selected empirically. In our experiments, we set $\lambda_{bbox}=5, \lambda_{conf}^+=75,  \lambda_{conf}^-=100$. This loss function can be optimized directly using back-propagation. 

\subsection{Neural Network Design}

So far in this section, we described the single-stage detection pipeline, the \textit{ConvDet} layer, and the end-to-end training protocol. These parts are universal and can work with various CNN architectures, including VGG16\cite{VGG}, ResNet\cite{resnet}, and so on. When choosing the ``backbone'' CNN structure, our focus is mainly on model size and energy \& power efficiency, and SqueezeNet\cite{SqueezeNet} is our top candidate. 

\textbf{Model size.} SqueezeNet is built upon the \textit{Fire Module}, which is comprised of a \textit{squeeze} layer as input, and two parallel \textit{expand} layers as output. The \textit{squeeze} layer is a 1x1 convolutional layer that compresses an input tensor with large channel size to one with the same batch and spatial dimension, but smaller channel size. The \textit{expand} layer is a mixture of 1x1 and 3x3 convolution filters that take the compressed tensor as input, retrieve the rich features and output an activation tensor with large channel size. The alternating \textit{squeeze} and \textit{expand} layers effectively reduce parameter size without losing too much accuracy.

\textbf{Energy \& power efficiency.} Different operations involved in neural network inference have varying energy needs. The most expensive operation is the off-chip DRAM access, which uses 3,556x more energy than an addition operation~\cite{pedram2016dark}. Thus, we want to reduce off-chip DRAM accesses as many as possible. 

The most straightforward strategy to reduce off-chip DRAM access is to use small models, which reduces memory access for parameters. An effective way to reduce parameter size is to use convolutional layers instead of fully connected layers when possible. Convolution parameters can be accessed once and reused across all neighborhoods of all data items (if batch$>$1) of the input data. However, the fully-connected layer only exposes parameter reuse opportunities in the ``batch" dimension, and each parameter is only used on one neighborhood of the input data. Besides model size, another important aspect is to control the size of intermediate activations. Assume the on-chip SRAM size of the computing hardware is 16MB, the SqueezeNet model size is 5MB. If the total size of activation output of any two consecutive layers is less than 11MB, then all the memory accesses can be completed in the on-chip SRAM, no off-chip DRAM accesses are needed. 

In this chapter, we adopted two versions of the SqueezeNet architecture. The first one is the SqueezeNet v1.1 model\footnote{\url{https://github.com/DeepScale/SqueezeNet/}}
with a model size$4.72$MB and $>80.3\%$ ImageNet top-5 accuracy. The second one is a more powerful SqueezeNet variation with a squeeze ratio of $0.75$, $86.0\%$ of ImageNet accuracy and a model size of $19$MB ~\cite{SqueezeNet}. In this chapter, we denote the first model as SqueezeDet and the second one as SqueezeDet+. We pre-train these two models for ImageNet classification, then we add two fire modules with randomly initialized weight on top of the pretrained model, and connect to the \textit{ConvDet} layer. 

\section{Experiments}

We evaluated the model on the KITTI~\cite{KITTI} object detection dataset, which is designed with autonomous driving in mind. We analyzed our model's accuracy measured by average precision (AP), recall, speed, and model size, and then compare with our previous work \cite{ShallowNetworks}, a faster-RCNN-based object detector trained on the KITTI dataset under the same experimental setting. Next, we analyzed the trade-off between accuracy and cost in terms of model size, FLOPS, and activation size by tuning several key hyperparameters. We implemented training, evaluation, error analysis, and visualization pipeline using Tensorflow~\cite{TensorFlow}, compiled with the cuDNN~\cite{cuDNN} computational kernels. The source code is released at \url{https://github.com/BichenWuUCB/squeezeDet}. 

\begin{table}[h!]
\centering
\begin{tabular}{c|cccccc}
Method                                           & \begin{tabular}[c]{@{}c@{}}Car\\ mAP \end{tabular} & \begin{tabular}[c]{@{}c@{}}Cyclist\\ mAP \end{tabular} & \begin{tabular}[c]{@{}c@{}}Pedestrian\\ mAP \end{tabular} & \begin{tabular}[c]{@{}c@{}}All\\ mAP \end{tabular} & \begin{tabular}[c]{@{}c@{}}Model size\\ (MB)\end{tabular} & \begin{tabular}[c]{@{}c@{}}Speed\\ (FPS)\end{tabular} \\ \hline
FRCN+Alex\cite{ShallowNetworks} & 82.6                                                   & -                                                          & -                                                             & -                                                      & 240                                                       & 2.9                                                   \\ 
FRCN+VGG\cite{ShallowNetworks}  & 86.0                                                   & -                                                          & -                                                             & -                                                      & 485                                                       & 1.7                                                   \\ \hline
SqueezeDet                                       & 82.9                                                   & 76.8                                                       & 70.4                                                          & 76.7                                                   & \textbf{7.9}                                              & \textbf{57.2}                                         \\
SqueezeDet+                                      & 85.5                                                   & \textbf{82.0}                                              & \textbf{73.7}                                                 & \textbf{80.4}                                          & 26.8                                                      & 32.1                                                  \\
VGG16-Det                                        & \textbf{86.9}                                          & 79.6                                                       & 70.7                                                          & 79.1                                                   & 57.4                                                      & 16.6                                                  \\
ResNet50-Det                                     & 86.7                                                   & 80.0                                                       & 61.5                                                          & 76.1                                                   & 35.1                                                      & 22.5                                                  \\ \hline
\end{tabular}
\caption[Summary of detection accuracy, model size, and inference speed.]{Summary of detection accuracy, model size, and inference speed. The mAP (mean-average precision) for each category are averaged across three difficulty levels. The mAP for All is averaged across all categories and difficulty levels. }
\label{table:AP}
\end{table}

\begin{table}[h!]
\footnotesize
\begin{center}
\begin{tabular}{c|ccc|ccc|ccc|c}
& \multicolumn{3}{c|}{car} & \multicolumn{3}{c|}{cyclist} &\multicolumn{3}{c|}{pedestrian} & mAP \\
method & E & M & H & E & M & H & E & M & H & \\
\hline
FRCN+VGG\cite{ShallowNetworks} & 92.9 & 87.9 & 77.3 & - & - & - & - & - & - & - \\
FRCN+Alex\cite{ShallowNetworks} & \textbf{94.7} & 84.8 & 68.3 & - & - & - & - & - & - & - \\
\hline
SqueezeDet & 90.2 & 84.7 & 73.9 & 82.9 & 75.4 & 72.1 & 77.1 & 68.3 & 65.8 &76.7 \\
SqueezeDet+ & 90.4 & 87.1 & 78.9 & \textbf{87.6} & \textbf{80.3} & \textbf{78.1} & \textbf{81.4} & \textbf{71.3} & \textbf{68.5} & \textbf{80.4} \\
VGG16-Det & 93.5 & 88.1 & 79.2 & 85.2 & 78.4 & 75.2 & 77.9 & 69.1 & 65.1 & 79.1 \\
ResNet50-Det & 92.9 & \textbf{87.9} & \textbf{79.4} & 85.0 & 78.5 & 76.6 & 67.3 & 61.6 & 55.6 & 76.1 \\
\hline
\end{tabular}
\end{center}
\caption[Detailed average precision results.]{Detailed average precision results for each difficulty level and each category.}
\label{tab:detailed-AP}
\end{table}

\subsection{KITTI object detection}

\textbf{Experimental setup.} 
In our experiments, unless otherwise specified, we scaled all the input images to 1242x375. We randomly split the $7381$ training images in half into a training set and a validation set. SqueezeDet, including variations, and the baseline model \cite{ShallowNetworks} are trained and evaluated on the same training-validation dataset. Our average precision (AP) results are from the validation set. We used Stochastic Gradient Descent with momentum to optimize the loss function. We set the initial learning rate to $0.01$, learning rate decay factor to $0.5$ and decay step size to $10000$. Instead of using a fixed number of steps, we trained our model all the way until the mean average precision (mAP)\footnote{Mean of average precision in 3 difficulty levels (easy, medium, hard) of 3 categories (car, cyclist, pedestrian).} on the training set converges, and then evaluated the model on the validation set. Unless otherwise specified, we used a batch size of 20. We adopted data augmentation techniques such as random cropping and flipping to reduce overfitting.  We trained our model to detect three categories of object, car, cyclist, pedestrian. We used nine anchors for each grid in our model. At the inference stage, we only kept the top 64 detections with the highest confidence, and use NMS to filter the bounding boxes. We used NVIDIA TITAN X GPUs for our experiments.  

\begin{figure*}[h]
\centering
  \begin{subfigure}[t]{.48\linewidth}
    \centering\includegraphics[clip, trim=0cm 6cm 0cm 6cm, width=\linewidth]{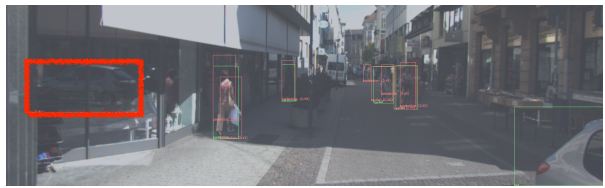}
    \caption{Example of a background error. The detector is confused by a car mirrored in the window.}
  \end{subfigure} \hfill
  \begin{subfigure}[t]{.48\linewidth}
    \centering\includegraphics[clip, trim=0cm 6cm 0cm 6cm, width=\linewidth]{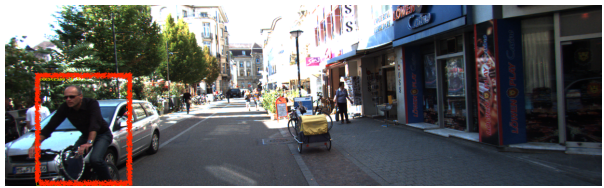}
    \caption{Classification error. The detector predict a cyclist to be a pedestrian.}
  \end{subfigure}
  \begin{subfigure}[t]{.48\linewidth}
    \centering\includegraphics[clip, trim=0cm 6cm 0cm 6cm, width=\linewidth]{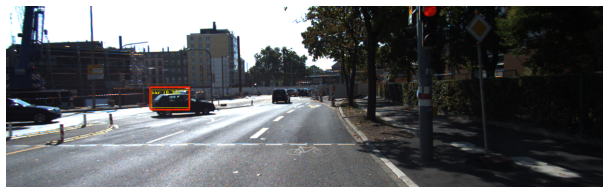}
    \caption{Localization error. The predicted bounding box doesn't have an IOU $>0.7$ with the ground truth.}
  \end{subfigure}\hfill
  \begin{subfigure}[t]{.48\linewidth}
    \centering\includegraphics[clip, trim=0cm 6cm 0cm 6cm, width=\linewidth]{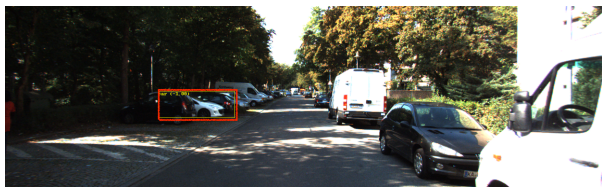}
    \caption{Missed object. The missed car is highly truncated and overlapped with other cars.}
  \end{subfigure}  
\caption{Example of detection errors.}
\label{fig:det_samples}
\end{figure*}

\textbf{Average Precision.} The detection accuracy, measured by average precision, is shown in Table~\ref{table:AP}. Compared with the baseline \cite{ShallowNetworks}, SqueezeDet+ is on-par with the Faster-RCNN + VGG16 model in terms of car detection accuracy. SqueezeDet is slightly better than Faster-RCNN + AlexNet. In terms of overall accuracy, our proposed SqueezeDet+ model achieved the highest mean average precision among all classes and difficulty levels.  To evaluate whether \textit{ConvDet} can be applied to other backbone CNNs, we appended \textit{ConvDet} to the convolution layers of the VGG16 and ResNet50 models. Both variations achieved competitive AP scores. Example of error detections of SqueezeDet by types are visualized in Fig.~\ref{fig:det_samples}. More detailed accuracy results are reported in Table~\ref{tab:detailed-AP}. 

\textbf{Recall.} Recall is an essential metric for the safety of autonomous vehicles, so we now analyze the recall of our proposed models. For each image with a resolution of 1242x375, SqueezeDet generates in total 15048 bounding box predictions. It is intractable to perform non-maximum suppression (NMS) on this many bounding boxes because of the quadratic time complexity of NMS with respect to the number of bounding boxes. Thus we only kept the top 64 predictions to feed into NMS. 

\begin{figure}[h]
  \centering
  \includegraphics[clip, trim=3cm 5cm 6.5cm 5cm, width=.8\textwidth]{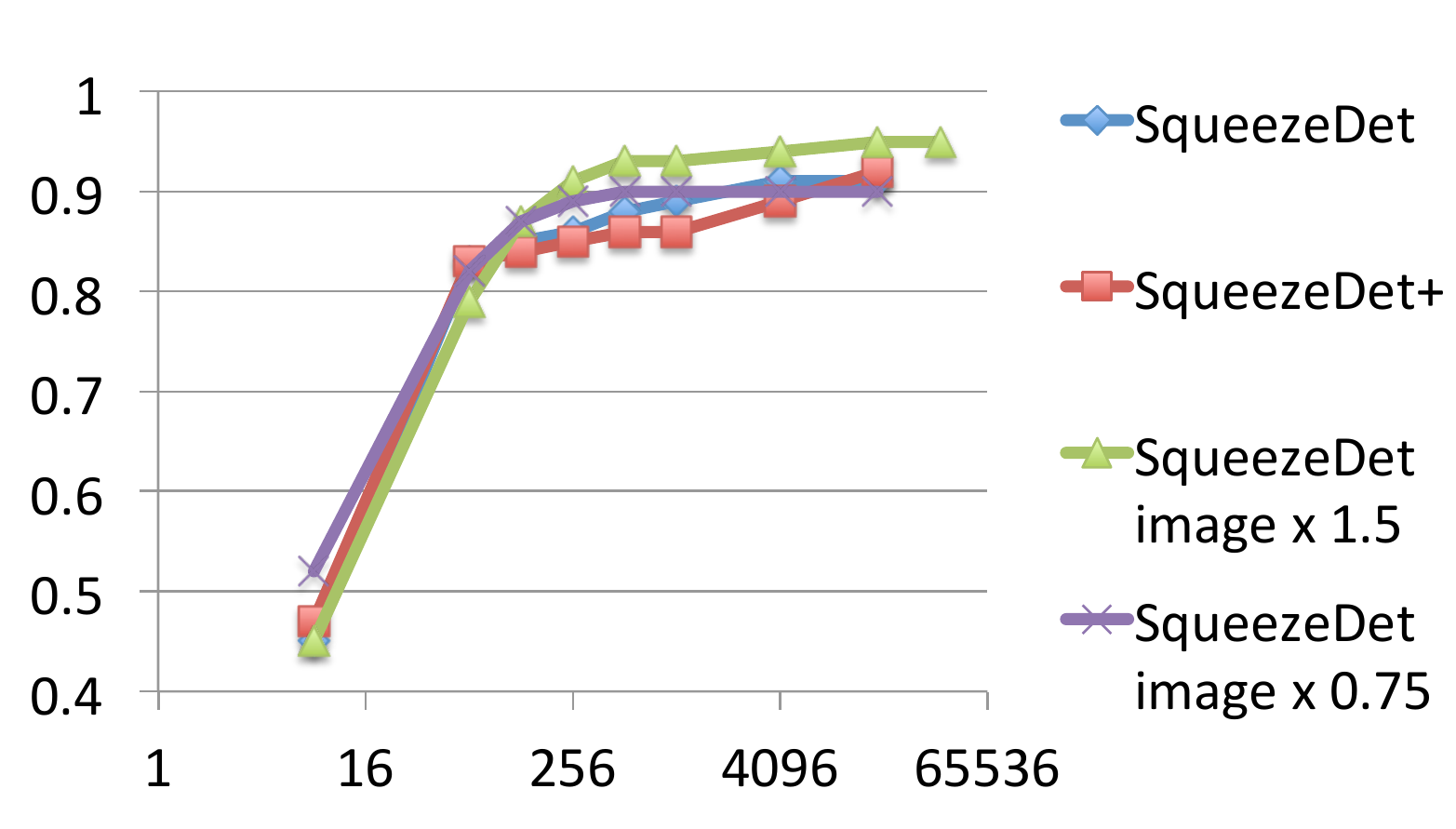}
 \caption[Recall vs. $N_{obj}$ for SqueezeDet and SqueezeDet+.]{Overall recall vs $N_{obj}$ for SqueezeDet and SqueezeDet+ models. We also tried to re-scale the input image by 1.5X and 0.75X. The squeezeDet and SqueezeDet+ models achieved the best recall of 0.91 and 0.92 respectively with all bounding boxes. SqueezeDet with 1.5X image resolution achieved 0.95. SqueezeDet with 0.75X image resolution achieved 0.90.}
\label{fig:recall}
\end{figure}

An interesting question to ask is, how does the number of bounding boxes kept affect recall? We tested this with the following experiment: First, we collect all the bounding box predictions and sort them by their confidence. Next, for each image, we choose the top $N_{box}$ bounding box predictions, and sweep $N_{box}$, the number of bounding boxes to keep, from 8 to 15,048. Then, we evaluate the overall recall for all difficulty levels of all categories. The Recall-$N_{box}$ curve is plotted in Fig.~\ref{fig:recall}. As we could see, for SqueezeDet and its strengthened model, the top 64 bounding boxes' overall recall is already larger than 80\%. If using all the bounding boxes, the SqueezeDet models can achieve $91\%$ and $92\%$ overall recall. Increasing the image size by 1.5X, the total number of bounding boxes increased to 35,190, and the maximum recall using all bounding boxes increases to 95\%.

\textbf{Speed}. We benchmark the inference speed of SqueezeDet and baselines on a TITAN X GPU with a batch size of 1. Our models are the first to achieve real-time inference speed on the KITTI dataset. Compared with the baseline \cite{ShallowNetworks}, SqueezeDet+ model achieved almost the same accuracy as Faster-RCNN+VGG16 on the same validation set, but the inference speed is 19x faster. The smaller SqueezeDet achieved a speed of 57.2 frames per second, which is almost twice the standard of real-time speed (30 FPS).

\begin{figure}[h]
\centering
    \centering\includegraphics[clip, trim=1cm 3cm 0cm 3cm, width=.8\linewidth]{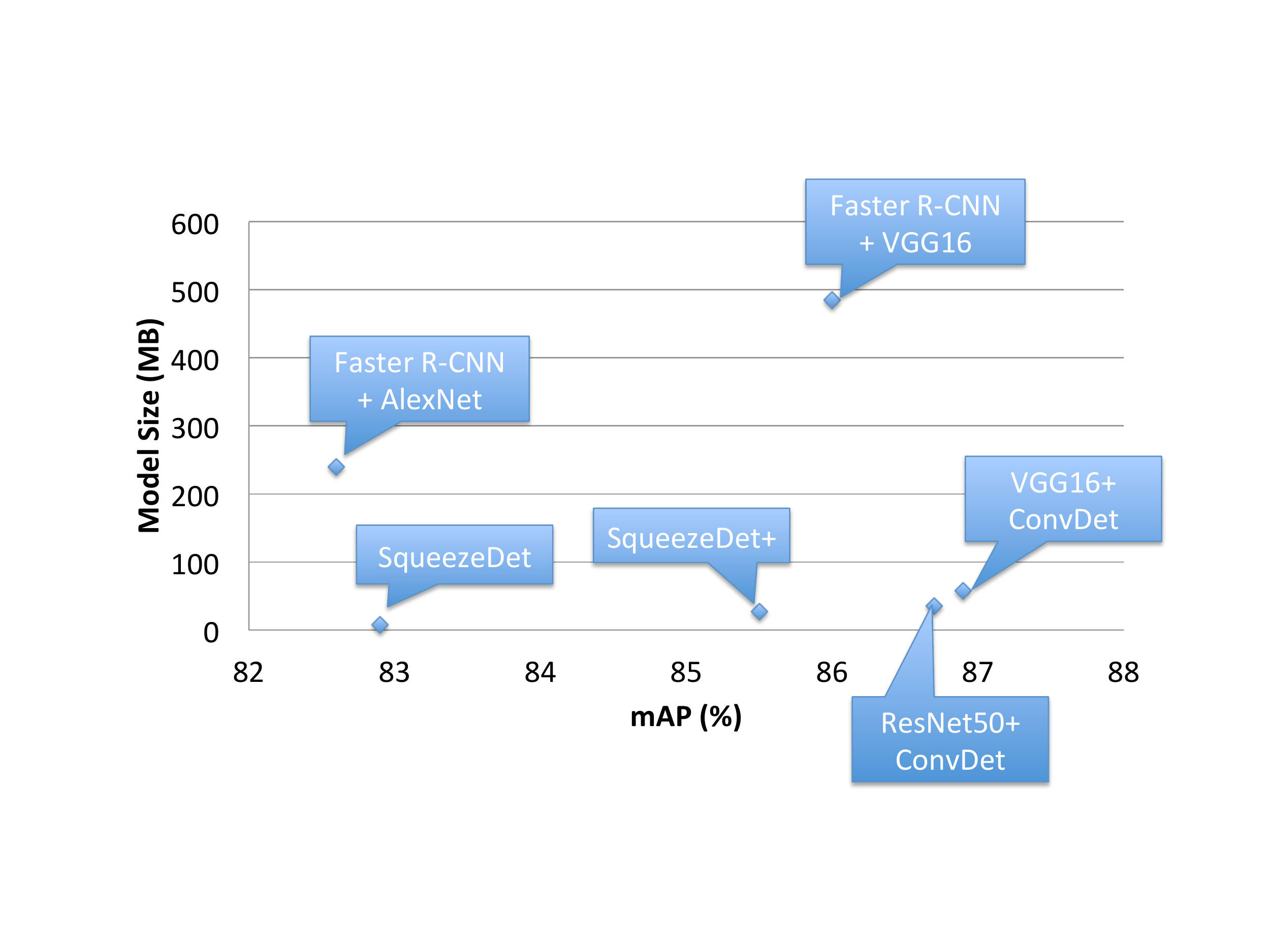}
    \caption[Model size vs. mean average precision for car detection.]{Model size vs. mean average precision for car detection. Each point on this plane represents a method's model size and accuracy tradeoff.}
    \label{fig:model-size-vs-mAP}
\end{figure}

\textbf{Model size.} We compare our proposed models with Faster-RCNN based models from~\cite{ShallowNetworks}. We plot the model sizes and their mean average precisions for three difficulty levels of the car category in Fig.~\ref{fig:model-size-vs-mAP} and summarize them in Table~\ref{table:AP}. As can be seen in Table~\ref{table:AP}, the SqueezeDet model is 61X smaller than the \textit{Faster R-CNN + VGG16} model, and it is 30X smaller than the \textit{Faster R-CNN + AlexNet} model. Almost $80\%$ of the parameters of the VGG16 model are from the fully connected layers. Thus, after we replace the fully connected layers and RPN layer with \textit{ConvDet}, the model size is only $57.4$MB. YOLO~\cite{YOLO} is comprised of 24 convolutional layers and two fully connected layers, and the model size of 753MB. SqueezeDet, without any compression, is 95X smaller than YOLO. 

\subsection{Design space exploration} 

\begin{table}
\begin{center}
\footnotesize
\begin{tabular}{c|ccccccc}
& &  & &   & Activation  &  \\
 & &  & & Model  & Memory  &  \\
 & mAP & Speed & FLOPs & Size & Footprint \\
DSE &  (\%) & (FPS) & $\times 10^9$  & (MB) & (MB) \\ 
\hline
SqueezeDet & 76.7 & 57.2  & 9.7 & 7.9 & 117.0 \\
scale-up & 72.4 & 31.3 & 22.5 & 7.9 & 263.3 \\
scale-down & 73.2 & 92.5 &5.3  & 7.9 & 65.8 \\
16 anchors & 66.9 & 51.4 & 11.0 & 9.4 & 117.4  \\
SqueezeDet+ & 80.4 & 32.1 & 77.2 & 26.8 & 252.7   \\
\hline
\end{tabular}
\end{center}
\caption[Design space exploration for SqueezeDet.]{Design space exploration for SqueezeDet. Different approaches with their accuracy, FLOPs per image, inference speed, model size, and activation memory footprint. The speed, FLOPS, and activation memory footprint are measured for a batch size of 1. We used mean average precision (mAP) to evaluate the overall accuracy of the KITTI object detection task. }
\label{table:DSE}
\end{table}

We conducted design space exploration to evaluate some key hyper-parameters' influence on our model's overall detection accuracy (measured in mAP). Meanwhile, we also investigated the ``cost'' of these variations in terms of FLOPs, inference speed, model size, and memory footprint. The results are summarized in Table~\ref{table:DSE}, where the first row is our SqueezeDet architecture, subsequent rows are modifications to SqueezeDet, and the final row is SqueezeDet+.

\textbf{Image resolution.} For object detection, increasing image resolution is often an effective approach to improve detection accuracy~\cite{ShallowNetworks}. However, larger images lead to larger activations, more FLOPs, longer training time, and so on. We now evaluate some of these tradeoffs. In our experiments, we scaled the image resolution by $1.5$X and $0.75$X receptively. With larger input images, the training becomes much slower, so we reduced the batch size to 10. 
As we can see in Table~\ref{table:DSE}, scaling up the input image decreases the mAP and also leads to more FLOPs, lower speed, and larger memory footprint. We also experiment with decreasing the image size. Scaling down the image leads to an astonishing 92.5 FPS of inference speed and a smaller memory footprint, though it suffers from a 3 percentage point drop in mean-average precision.

\textbf{Number of anchors.}
Another hyper-parameter to tune is the number of anchors. Intuitively, the more anchors to use, the more bounding box proposals are to be generated, thus should result in better accuracy. However, in our experiment in Table~\ref{table:DSE}, using more anchors actually leads to lower accuracy. However, it also shows that for models that use \textit{ConvDet}, increasing the number of anchors only modestly increases the model size, FLOPs, and memory footprint. 

\textbf{Model architecture.}
As we discussed before, by using a more powerful backbone model with more parameters significantly improved accuracy (See Table~\ref{table:DSE}). However, this modification also costs substantially more in terms of FLOPs, model size and memory footprint.

\subsection{Energy \& power efficiency}
\begin{figure}[h]
\centering
  \includegraphics[clip, trim=2cm 0cm 0cm 0cm, width=.8\textwidth]{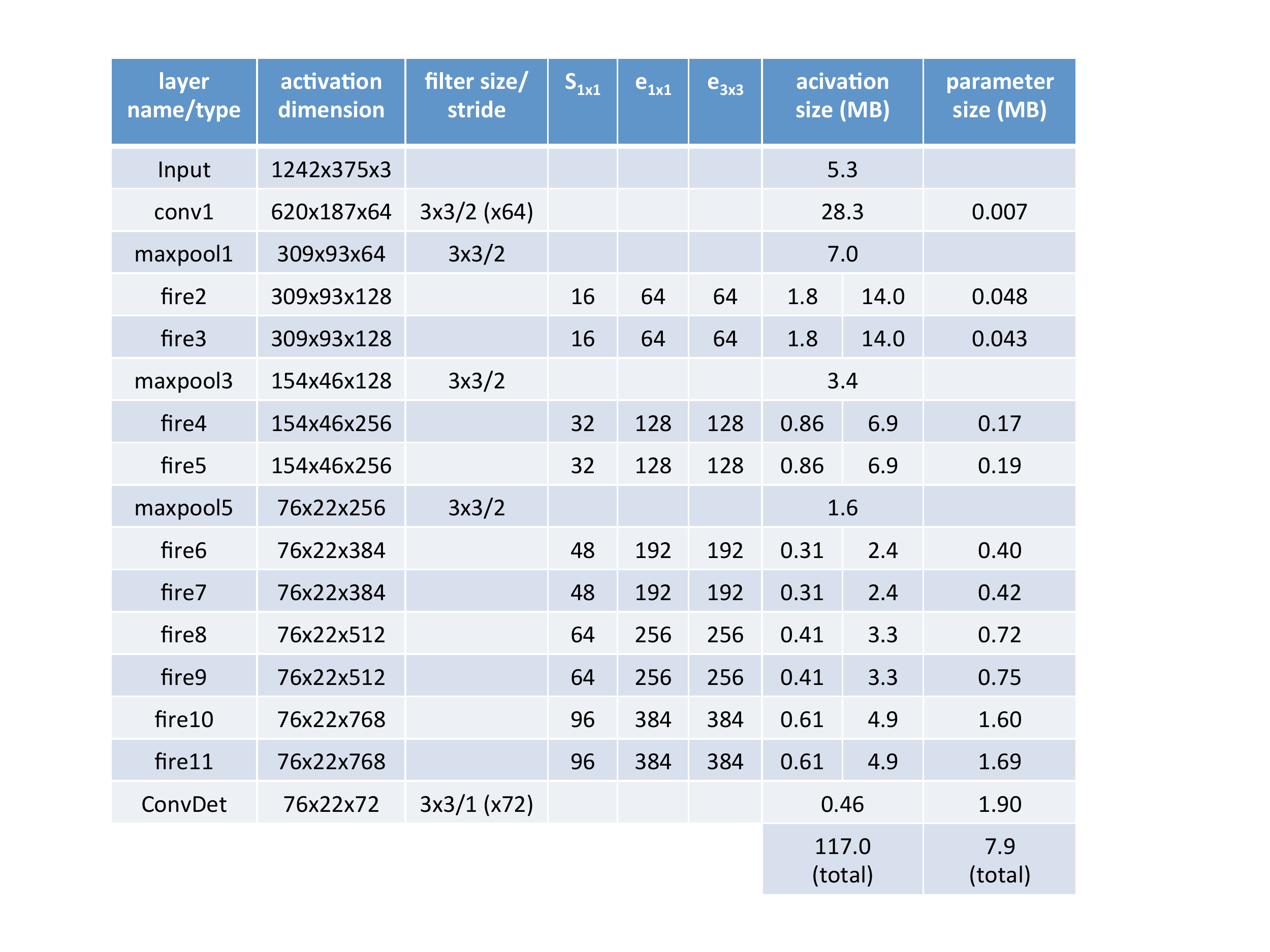}
  \captionof{table}[Layer specification of SqueezeDet.]{Layer specification of SqueezeDet. $s_{1x1}$ represents the number of 1x1 output filters in the \textit{squeeze} layer, $e_{1x1}$ is number of 1x1 filters in the \textit{expand} layer and $e_{3x3}$ is number of 3x3 filters in the \textit{expand} layer. 
 }
\label{fig:SqueezeDet} 
\end{figure}

Different operations involved in the computation of a neural network consume different amounts of energy. According to Pedram et al.~\cite{pedram2016dark}, an off-chip DRAM access consumes 3,556x more energy than an addition operation, so our primary focus is on reducing memory accesses, which can be realized by reducing the model parameters and intermediate layer activations. We analyze the memory footprint of SqueezeDet layer by layer. Details of the SqueezeDet model are shown in Table~\ref{fig:SqueezeDet}.

We counted the activation memory footprint for several models, including SqueezeDet, variations thereof, and others. Our results are summarized in Table~\ref{table:energy}. As we can see, SqueezeDet has a much lower memory footprint and performs fewer FLOPs compared to other models, leading to better energy and power efficiency.

\begin{table*}
\begin{center}
\footnotesize
\begin{tabular}{c|ccccccc}
& & & Activation &  &  &  \\
& Model & & Memory & Average & Inference & Energy \\
& Size & FLOPs& Footprint &  GPU Power  &  Speed & Efficiency &mAP \\ 
model & (MB) & $\times 10^9$ & (MB) &  (W)  &  (FPS) & (J/frame) &(\%) \\ 
\hline
SQDT & 7.9 & 9.7 & 117.0 & 80.9 & 57.2 & 1.4 & 76.7 \\
SQDT: scale-up & 7.9 & 22.5 & 263.3 & 89.9 & 31.3 & 2.9 & 72.4 \\
SQDT: scale-down & 7.9 & 5.3 & 65.8 & 77.8 & 92.5 & 0.84 & 73.2 \\
SQDT: 16 anchors & 9.4 & 11.0 & 117.4 & 82.9 & 51.4 & 1.6& 66.9\\
SQDT+ & 26.8 & 77.2 & 252.7 & 128.3 & 32.1 & 4.0 & 80.4 \\
\hline
VGG-Det & 57.4 & 288.4 & 540.4 & 153.9 &16.6 & 9.3 & 79.1 \\
ResNet50-Det & 35.1 & 61.3 & 369.0  & 95.4 &22.5 & 4.2 & 76.1 \\
\hline
FRCN+VGG16~\cite{ShallowNetworks} & 485 &- & - & 200.1 & 1.7 & 117.7 & -\\
FRCN-Alex~\cite{ShallowNetworks} & 240 & - & - & 143.1 & 2.9& 49.3 & -\\
YOLO${}^{\star}$ & 753 & - & - & 187.3 & 25.8 & 7.3 & -\\
\end{tabular}
\end{center}
\caption[Energy efficiency of SqueezeDet and baselines. ]{Comparing SqueezeDet and other models in terms of energy efficiency and other aspects. The default image resolution is 1242x375, but the ``SQDT: scale-up'' variation up-sampled input image's height and width by 1.5X. The ``scale-down'' variation scaled image resolution by 0.75X. The default SqueezeDet model contains 9 anchors. However, the 16-anchor variation contains 16 anchors for each grid. ${}^{\star}$ We launched YOLO to detect  $4,952$ VOC 2007 test images, and it took 192 seconds to finish. We then compute the inference speed as $4,952/192 \approx$ $25.8$FPS, which is slower than the speed reported in~\cite{YOLO}. The input image to YOLO is scaled to 448x448.}
\label{table:energy}
\end{table*}

\begin{figure}[h]
\centering
  \includegraphics[clip, trim=0cm 0cm 0cm 0cm, width=.8\linewidth]{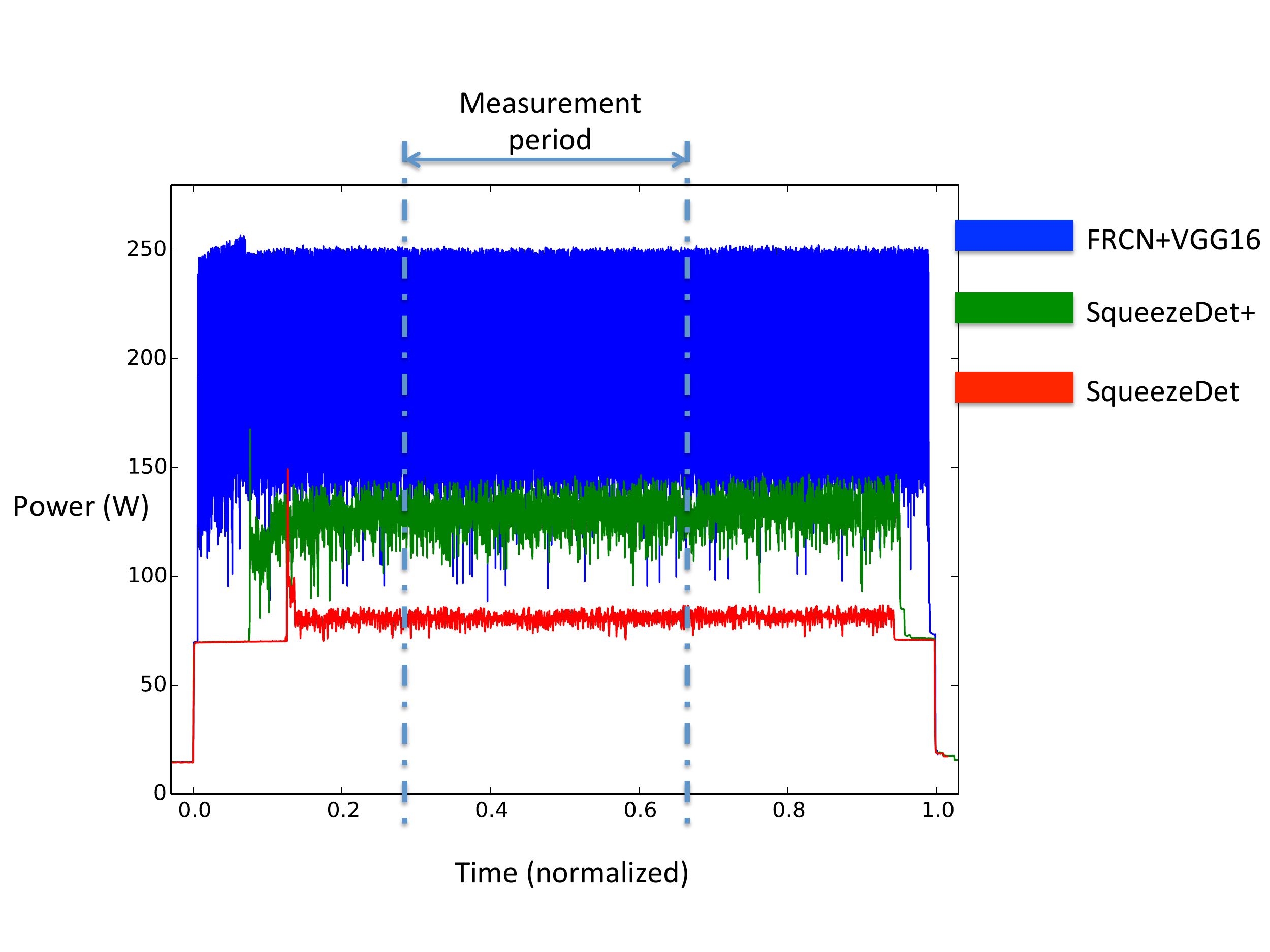}
  \caption[GPU power measurement of SqueezeDet.]{GPU power measured by \texttt{nvidia-smi}. Here we plot the power measurement curve of 3 models, SqueezeDet, SqueezeDet+, and Faster R-CNN + VGG16 model. We normalize the working period of 3 models to the same range of [0, 1]. We divide the working period evenly into 3 parts and use the middle part to compute the average GPU power for each model.}
\label{fig:power-curve}
\end{figure}

We measured the energy consumption of SqueezeDet and the other models during the object detection evaluation of $3741$ images from the KITTI dataset~\cite{KITTI}. The default input image resolution is 1242x375, and the batch size is set to 1. Meanwhile, we measured the GPU power usage with Nvidia's system monitor interface (\texttt{nvidia-smi}).  We sampled the power reading with a fixed interval of 0.1 seconds. Then, we obtained the power-vs-time curve as shown in Fig~\ref{fig:power-curve}.  When the GPU is idle, it consumes about $15$W of power. Through the evaluation process, the GPU went through several stages from idle to working and then to idle again. We denote the period with power measurement $\ge20$W as the working period. Then, we divide the working period evenly into 3 parts, and we average the measurements from the middle part to compute the average GPU power. The energy consumption per image is then computed as \[\text{Average Energy [Joule / frame]} = \frac{\text{Average Power [Joule / Second]}}{\text{Inference Speed [Frame / Second]}}.\]

We measured the energy consumption of SqueezeDet and several other models using the above approach, and our experimental results are listed in Table~\ref{table:energy}.  SqueezeDet consumes only $1.4$J per image, which is 84$\times$ less than the Faster R-CNN + VGG16 model.  Scaling the image resolution down by 0.75$\times$, the mAP drops by 3 percentage points, but the inference speed is 1.6$\times$ faster, and the energy consumption is less than $1$J per image. With much better accuracy, SqueezeDet+  only consumes $4$J per image, which is $>$10X more efficient than Faster R-CNN based methods. We combined the convolutional layers of VGG16 and ResNet50 with ConvDet, and both models achieved much better energy efficiency compared with Faster R-CNN based models, as shown in Table \ref{table:energy}.

We also compared our models with YOLO. We use YOLO to detect 4,952 images from the VOC 2007~\cite{PASCAL} test set. The input images are scaled to 448x448, and the batch size is 1. It took YOLO $192$ seconds to finish the evaluation. Using the same approach to measure the GPU power of YOLO, we compute the energy per frame of YOLO as 7.3J. Using the reported frame rate of 45FPS \cite{YOLO}, YOLO's energy consumption per frame is 4.2J, which is comparable with SqueezeDet+. However, note that the input image (with the size of 1242x375) to SqueezeDet+ in our experiment contains 2X more pixels than the input image (448x448) to YOLO.  

Our experiments show that SqueezeDet and its variations are very energy efficient compared with previous neural network-based object detectors.

\section{Conclusion}
\label{sec:conclusion}

We present SqueezeDet, a fully convolutional neural network for real-time object detection. We integrate the region proposition and classification into \textit{ConvDet}, which is orders of magnitude smaller than its fully-connected counterpart. With the constraints of autonomous driving in mind, our proposed SqueezeDet and SqueezeDet+ models are designed to be small, fast, energy-efficient, and accurate. Compared with previous baselines, we achieve the same accuracy with 30.4x smaller model size, 19.7x faster inference speed, and 35.2x lower energy.

%% file: chap4.tex
\chapter{Model Efficiency: SqueezeSeg}
\label{chap:sqsg} 

In this chapter, we continue the discussion of model efficiency and focus on the following \textit{key question}:
\begin{quote}
    Can we design efficient neural networks to process 3D visual modalities such as depth measurements from LiDAR? 
\end{quote}
To answer this question, we discuss SqueezeSeg, an efficient neural network for LiDAR point cloud segmentation. 

\section{Introduction to LiDAR-based perception}
In this chapter, we continue to use autonomous driving as a motivating application, since it motivates us to not only consider RGB images, but also other visual modalities such as depth measurements. 

Autonomous driving systems rely on accurate, real-time, and robust perception of the environment. An autonomous vehicle needs to accurately categorize and locate ``road-objects'', which we define to be driving-related objects such as cars, pedestrians, cyclists, and other obstacles. Different autonomous driving solutions may have different combinations of sensors, but a 3D LiDAR scanner is one of the most prevalent components. LiDAR scanners directly produce distance measurements of the environment, which are then used by vehicle controllers and planners. Moreover, LiDAR scanners are robust under almost all lighting conditions, whether it be day or night, with or without glare and shadows. As a result, LiDAR-based perception tasks have attracted significant research attention. 

\begin{figure}[t]
    \centering
    \includegraphics[width=.7\textwidth,trim={0cm 0cm 0cm 0cm},clip]{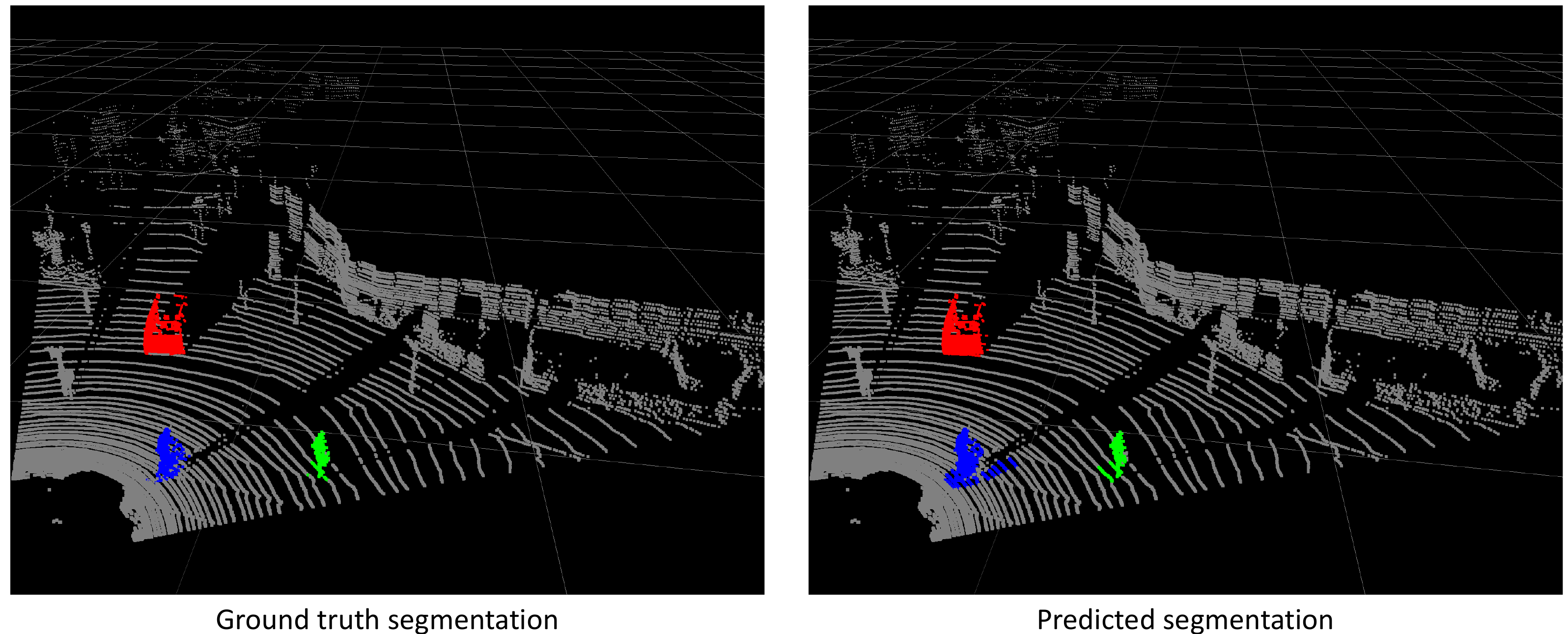}
    \caption[An example of SqueezeSeg segmentation results.]{An example of SqueezeSeg segmentation results. Our predicted result is on the right and the ground truth is on the left. Cars are annotated in red, pedestrians in green and cyclists in blue.}
    \label{fig:LiDAR}
\end{figure}

In this work, we focus on road-object segmentation using (Velodyne style) 3D LiDAR point clouds. Given the point cloud output from a LiDAR scanner, the task aims to isolate objects of interest and predict their categories, as shown in Fig. \ref{fig:LiDAR}. Previous approaches comprise or use parts of the following stages: Remove the ground, cluster the remaining points into instances, extract (hand-crafted) features from each cluster, and classify each cluster based on its features. This paradigm, despite its popularity~\cite{LiDARSegICRA2012,himmelsbach2008lidar,wang2012could,zermas2017fast}, has several disadvantages: a) Ground segmentation in the above pipeline usually relies on hand-crafted features or decision rules -- some approaches rely on a scalar threshold~\cite{thrun2006stanley} and others require more complicated features such as surface normals~\cite{LocalConvexitySeg} or invariant descriptors~\cite{wang2012could}. None of these rules are general enough to deal with the real-world complexity. b) Multi-stage pipelines see aggregate effects of compounded errors, and classification or clustering algorithms in the pipeline above are unable to leverage context, most importantly the immediate surroundings of an object. c) Many approaches for ground removal rely on iterative algorithms such as RANSAC (random sample consensus)~\cite{zermas2017fast}, GP-INSAC (Gaussian Process Incremental Sample Consensus)~\cite{LiDARSegICRA2012}, or agglomerative clustering~\cite{LiDARSegICRA2012}. 
The runtime and accuracy of these algorithmic components depend on the quality of random initialization and, therefore, can be unstable. This instability is not acceptable for many embedded applications such as autonomous driving. We take an alternative approach: use deep learning to extract features, develop a single-stage pipeline and thus sidestep iterative algorithms.

\begin{figure*}[h]
    \centering
    \includegraphics[width=\linewidth]{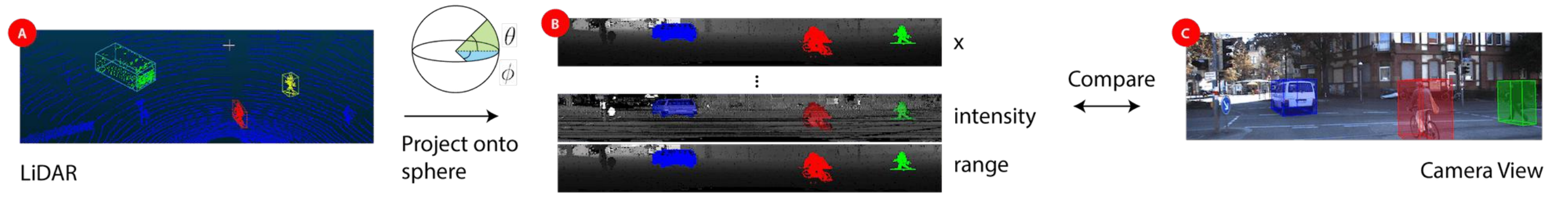}
    \caption[LiDAR projections.]{LiDAR Projections. Note that each channel reflects structural information in the camera-view image.}
    \label{fig:ProjectedData}
\end{figure*}

In this chapter, we propose an end-to-end pipeline based on convolutional neural networks (CNNs) and conditional random fields (CRFs). CNNs and CRFs have been successfully applied to segmentation tasks on 2D images ~\cite{FCN,CRF, DeepLab, CRFasRNN}. To apply CNNs to 3D LiDAR point clouds, we designed a CNN that accepts transformed LiDAR point clouds and outputs a point-wise map of labels, which is further refined by a CRF model. Instance-level labels are then obtained by applying conventional clustering algorithms (such as DBSCAN) on points within a category. To feed 3D point clouds to a 2D CNN, we adopt a spherical projection to transform sparse, irregularly distributed 3D point clouds to dense, 2D grid representations. The proposed CNN model draws inspiration from SqueezeNet~\cite{SqueezeNet} and is carefully designed to reduce parameter size and computational complexity and achieve real-time inference speed for our target embedded applications. The CRF model is reformulated as a recurrent neural network (RNN) module as~\cite{CRFasRNN} and can be trained end-to-end together with the CNN model. Our model is trained on LiDAR point clouds from the KITTI dataset~\cite{KITTI} and point-wise segmentation labels are converted from 3D bounding boxes in KITTI. To obtain even more training data, we leveraged \textit{Grand Theft Auto V (GTA-V)} as a simulator to retrieve LiDAR point clouds and point-wise labels.

Experiments show that SqueezeSeg achieves high accuracy and is extremely fast and stable, making it suitable for autonomous driving applications. We additionally find that supplanting our dataset with artificial, noise-injected simulation data further boosts validation accuracy on real-world data.

\section{Related work}
\subsection{Semantic segmentation for 3D LiDAR point clouds}
Previous work saw a wide range of granularity in LiDAR segmentation, handling anything from specific components to the whole pipeline.
\cite{LocalConvexitySeg} proposed mesh-based ground and object segmentation relying on local surface convexity conditions.  
\cite{LiDARSegICRA2012} summarized several approaches based on iterative algorithms such as RANSAC and GP-INSAC for ground removal. 
Recent work also focused on algorithmic efficiency. \cite{zermas2017fast} proposed efficient algorithms for ground segmentation and clustering while \cite{shin2017real} bypassed ground segmentation to directly extract foreground objects. 
\cite{wang2012could} expanded its focus to the whole pipeline, including segmentation, clustering, and classification. It proposed to directly classify point patches into background and foreground objects of different categories then use EMST-RANSAC ~\cite{zermas2017fast} to further cluster instances.  

\subsection{CNN for 3D point clouds}
CNN approaches consider LiDAR point clouds in either two or three dimensions. Work with two-dimensional data considers raw images with projections of LiDAR point clouds top-down~\cite{CNNRoadSeg} or from a number of other views~\cite{MultiView}. Other work considers three-dimensional data itself, discretizing the space into voxels and engineering features such as disparity, mean, and saturation~\cite{LiDARFusion}. Regardless of data preparation, deep learning methods consider end-to-end models that leverage 2D convolutional~\cite{LiDARDet} or 3D convolutional~\cite{3DCNN} neural networks.

\subsection{Semantic Segmentation for Images}
Both CNNs and CRFs have been applied to semantic segmentation tasks for images. \cite{FCN} proposed transforming CNN models, trained for classification, to fully convolutional networks to predict pixel-wise labels. \cite{CRF} proposed a CRF formulation for image segmentation and solved it approximately with the mean-field iteration algorithm. CNNs and CRFs are combined in~\cite{DeepLab}, where the CNN is used to produce an initial probability map and the CRF is used to refine and restore details. In \cite{CRFasRNN}, mean-field iteration is re-formulated as a recurrent neural network (RNN) module.  

\subsection{Data Collection through Simulation}
Obtaining annotations, especially point-wise or pixel-wise annotations for computer vision tasks is usually very difficult. As a consequence, synthetic datasets have seen growing interest. In the autonomous driving community, the video game Grand Theft Auto has been used to retrieve data for object detection and segmentation~\cite{eccv_playing_for_data,drive_in_matrix}. 

\section{Method}
\subsection{Point cloud transformation}
\label{sec:point_cloud}
Conventional CNN models operate on images, which can be represented by 3-dimensional tensors of size $H\times W \times 3$. The first two dimensions encode spatial position, where $H$ and $W$ are the image height and width, respectively. The last dimension encodes features, most commonly RGB values. However, a 3D LiDAR point cloud is usually represented as a set of cartesian coordinates, $(x, y, z)$. Extra features can also be included, such as intensity or RGB values. Unlike image pixels, LiDAR point clouds are sparse and irregular. It is not possible to directly use conventional CNNs to process such data. A naive alternative is to discretize the 3D space into voxels and extract features based on the point distribution in each voxel. However, this method is problematic. Discretizing the space with high resolution will result in high computational complexity. Since most of the voxels are empty, most of the computation is wasted. On the contrary, if using low resolution to discretize the space, the accuracy will drop significantly. 

In this work, we present a novel way to represent LiDAR point cloud data to bypass this dilemma. CNNs are effective and efficient at processing 2D grid data. To obtain such a compact representation, we project the LiDAR point cloud onto a sphere for a dense, grid-based representation as
\begin{gather}
\label{eqn:projection}
\begin{split}
\theta = \arcsin{\frac{z}{\sqrt{x^2+y^2+z^2}}}, \ 
\tilde{\theta} = \lfloor\theta / \triangle \theta\rfloor, \\
\phi = \arcsin{\frac{y}{\sqrt{x^2+y^2}}}, \ 
\tilde{\phi} = \lfloor\phi / \triangle \phi\rfloor.
\end{split}
\end{gather}
$\phi$ and $\theta$ are \textit{azimuth} and \textit{zenith} angles, as shown in Fig. \ref{fig:ProjectedData} (a). $(\tilde{\theta}, \tilde{\phi})$ denotes the position of a point on a 2D spherical grid  and $\triangle \theta$ and $\triangle \phi$ are resolutions for discretization. Applying Equation (\ref{eqn:projection}) to each point in the cloud, we can obtain a 3D tensor of size $H\times W\times C$. In this chapter, we consider data collected from a Velodyne HDL-64E LiDAR with 64 vertical channels, so $H=64$. Limited by data annotations from the KITTI dataset, we only consider the front view area of $90^{\circ}$ and divide it into 512 grids so $W=512$.  $C$ is the number of features for each point. In our experiments, we used 5 features for each point: 3 cartesian coordinates $(x, y, z)$, an intensity measurement and range $r=\sqrt{x^2+y^2+z^2}$. An example of a projected point cloud can be found at Fig. \ref{fig:ProjectedData} (b). As can be seen, such representation is dense and regularly distributed, resembling an ordinary image Fig. \ref{fig:ProjectedData} (c). This featurization allows us to avoid hand-crafted features, bettering the odds that our representation generalizes.

\subsection{Network structure}
\begin{figure*}[h]
    \centering
    \includegraphics[width=\textwidth]{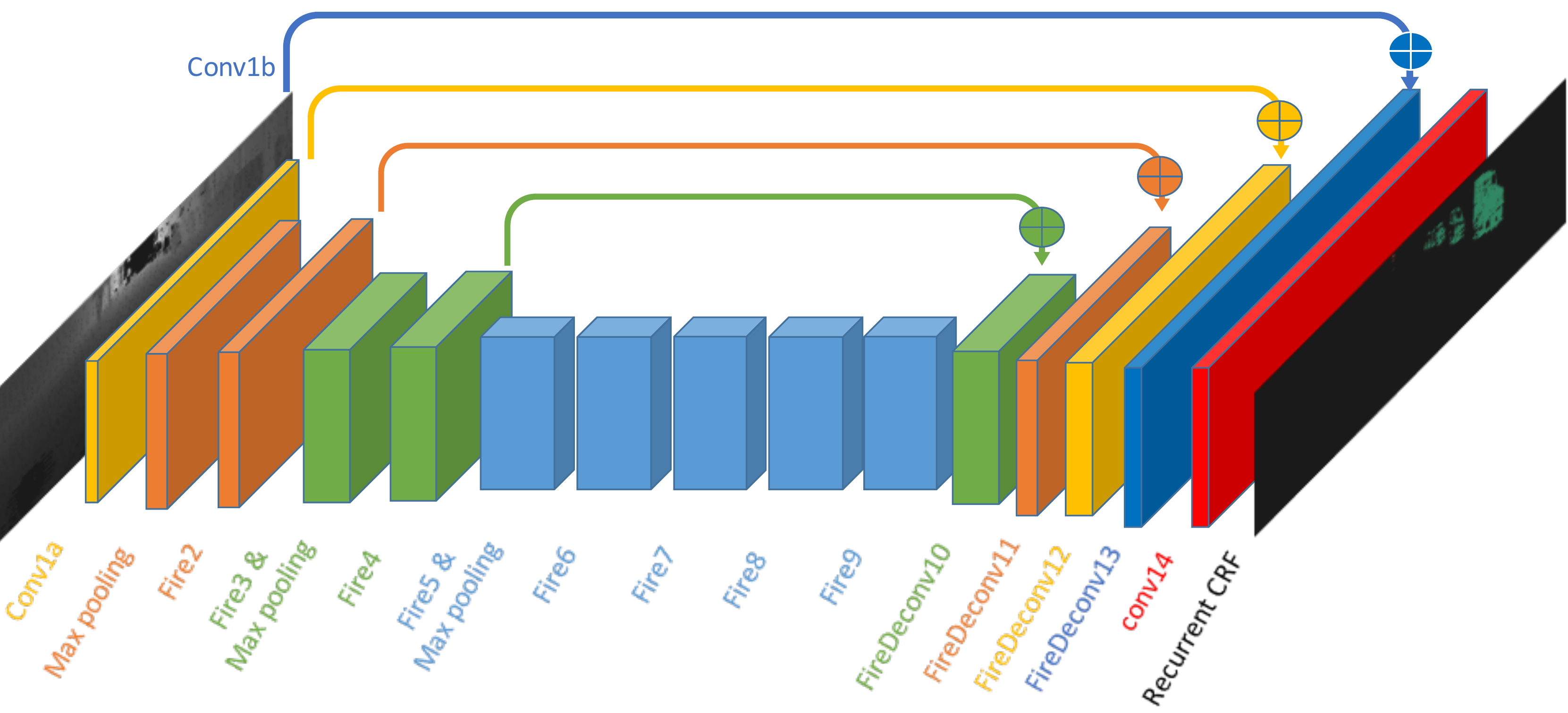}
    \caption{Network structure of SqueezeSeg.}
    \label{fig:NNstructure}
\end{figure*}

Our convolutional neural network structure is shown in Fig. \ref{fig:NNstructure}. SqueezeSeg is derived from SqueezeNet~\cite{SqueezeNet}, a light-weight CNN that achieved AlexNet level accuracy with 50X fewer parameters. SqueezeNet has been successfully adopted for image based object detection and achieved the state-of-the-art efficiency \cite{SqueezeDet}.

The input to \textit{SqueezeSeg} is a $64\times 512 \times 5$ tensor as described in the previous section. We ported layers (\textit{conv1a} to \textit{fire9}) from SqueezeNet for feature extraction. SqueezeNet used \textit{max-pooling} to down-sample intermediate feature maps in both width and height dimensions, but since our input tensor's height is much smaller than its width, we only down-sample the width. The output of \textit{fire9} is a down-sampled feature map that encodes the semantics of the point cloud. 

To obtain full resolution label predictions for each point, we used deconvolution  modules (more precisely, ``transposed convolutions'') to up-sample feature maps in the width dimension. We used skip-connections to add up-sampled feature maps to lower-level feature maps of the same size, as shown in Fig. \ref{fig:NNstructure}. The output probability map is generated by a convolutional layer (\textit{conv14}) with \textit{softmax} activation. The probability map is further refined by a recurrent CRF layer, which will be discussed in the next section. 

\begin{figure}[h]
    \centering
    \includegraphics[width=.8\textwidth]{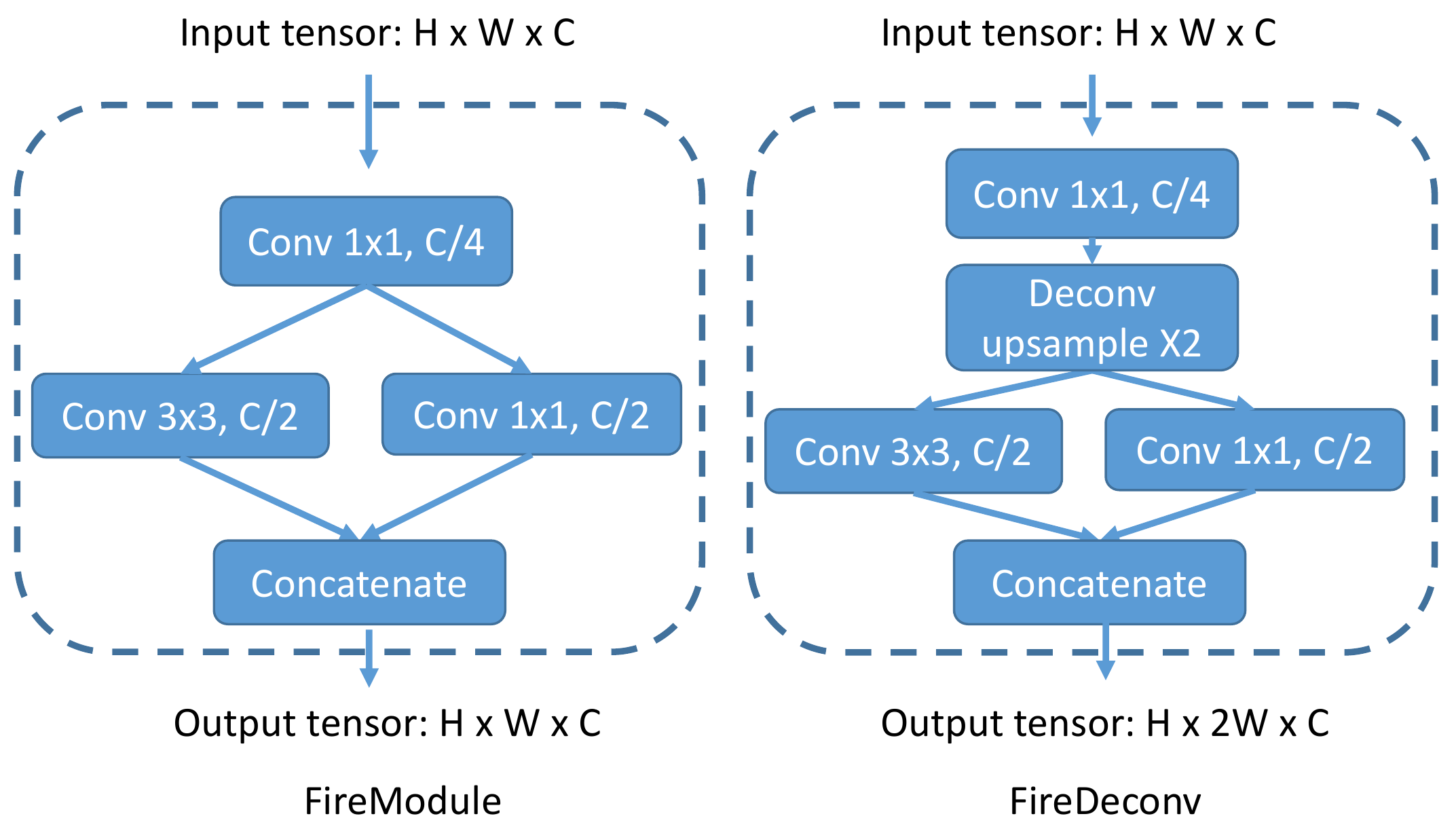}
    \caption{Structure of a \textit{FireModule} (left) and a \textit{fireDeconv} (right).}
    \label{fig:FireDeconv}
\end{figure}

In order to reduce the number of model parameters and computation, we replaced convolution and deconvolution layers with \textit{fireModule}s~\cite{SqueezeNet} and \textit{fireDeconv}s. The architecture of both modules are shown in Fig. \ref{fig:FireDeconv}. In a \textit{fireModule}, the input tensor of size $H\times W \times C$ is first fed into a 1x1 convolution to reduce the channel size to $C/4$. Next, a 3x3 convolution is used to fuse spatial information. Together with a parallel 1x1 convolution, they recover the channel size of $C$. The input 1x1 convolution is called the \textit{squeeze} layer and the parallel 1x1 and 3x3 convolution together is called the \textit{expand} layer. Given matching input and output size, a 3x3 convolutional layer requires $9C^2$ parameters and $9HWC^2$ computations, while the \textit{fireModule} only requires $\frac{3}{2}C^2$ parameters and $\frac{3}{2}HWC^2$ computations. In a \textit{fireDeconv} module, the deconvolution layer used to up-sample the feature map is placed between \textit{squeeze} and \textit{expand} layers. To up-sample the width dimension by 2, a regular 1x4 deconvolution layer must contain $4C^2$ parameters and $4HWC^2$ computations. With the \textit{fireDeconv} however, we only need $\frac{7}{4}C^2$ parameters and $\frac{7}{4}HWC^2$ computations.

\subsection{Conditional Random Field}
With image segmentation, label maps predicted by CNN models tend to have blurry boundaries. This is due to the loss of low-level details in down-sampling operations such as max-pooling. Similar phenomena are also observed with SqueezeSeg.

Accurate point-wise label prediction requires understanding not only the high-level semantics of the object and scene but also low-level details. The latter is crucial for the consistency of label assignments. For example, if two points in the cloud are next to each other and have similar intensity measurements, it is likely that they belong to the same object and thus have the same label. Following~\cite{DeepLab}, we used a conditional random field (CRF) to refine the label map generated by the CNN. For a given point cloud and a label prediction $\mathbf{c}$ where $c_i$ denotes the predicted label of the $i$-th point, a CRF model employs the energy function
\begin{equation}
E(\mathbf{c}) = \sum_i u_i(c_i) + \sum_{i, j} b_{i, j}(c_i, c_j).
\label{eqn:CRF}
\end{equation}
The unary potential term $u_i(c_i) = -\log{P(c_i)}$ considers the predicted probability $P(c_i)$ from the CNN classifier. The binary potential terms define the ``penalty'' for assigning different labels to a pair of similar points and is defined as $b_{i,j}(c_i, c_j) = \mu(c_i, c_j) \sum_{m=1}^M w_m k^m(\mathbf{f_i}, \mathbf{f_j})$ where $\mu(c_i, c_j) = 1$ if $c_i \neq c_j$ and 0 otherwise, $k^m$ is the $m$-th Gaussian kernel that depends on features $\mathbf{f}$ of point $i$ and $j$ and $w_m$ is the corresponding coefficient. In our work, we used two Gaussian kernels
\begin{equation}
\begin{gathered}
w_1\exp (-\frac{\|\mathbf{p_i} - \mathbf{p_j}\|^2}{2\sigma_\alpha^2} 
       -\frac{\|\mathbf{x_i} - \mathbf{x_j}\|^2}{2\sigma_\beta^2}) \\
+ w_2\exp (-\frac{\|\mathbf{p_i} - \mathbf{p_j}\|^2}{2\sigma_\gamma^2}).
\end{gathered}
\label{eqn:kernels}
\end{equation}
The first term depends on both angular position $\mathbf{p} (\tilde{\theta}, \tilde{\phi})$ and cartesian coordinates $\mathbf{x}(x, y, z)$ of two points. The second term only depends on angular positions. $\sigma_\alpha$, $\sigma_\beta$ and $\sigma_\gamma$ are three hyper parameters chosen empirically. Extra features such as intensity and RGB values can also be included. 

\begin{figure}[h]
    \centering
    \includegraphics[width=\textwidth]{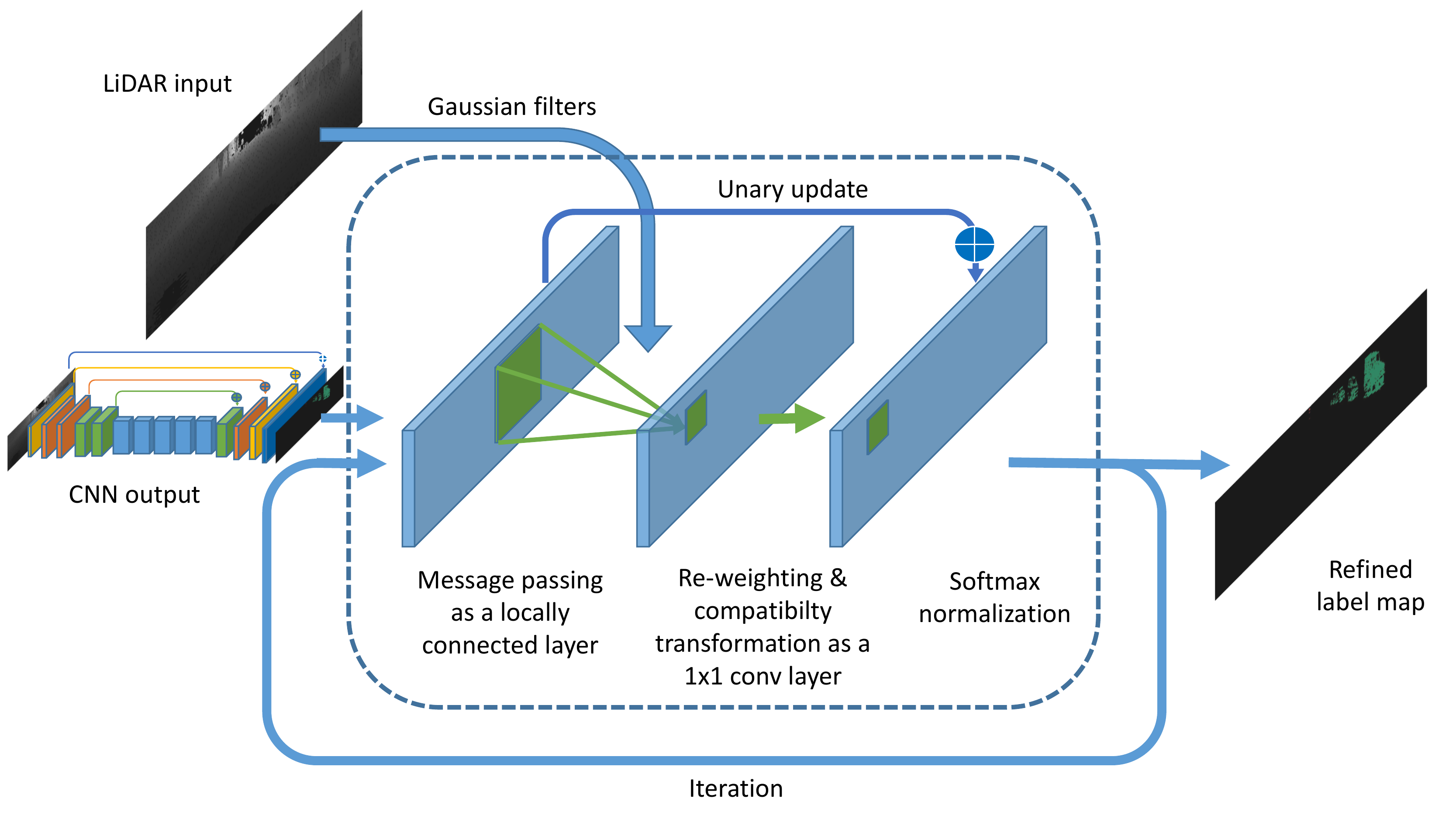}
    \caption{Conditional Random Field (CRF) as an RNN layer.}
    \label{fig:CRF}
\end{figure}

Minimizing the above CRF energy function yields a refined label assignment. Exact minimization of Equation (\ref{eqn:CRF}) is intractable, but~\cite{CRF} proposed a mean-field iteration algorithm to solve it approximately and efficiently. ~\cite{CRFasRNN} reformulated the mean-field iteration as a recurrent neural network (RNN). We refer readers to \cite{CRF} and \cite{CRFasRNN} for a detailed derivation of the mean-field iteration algorithm and its formulation as an RNN. Here, we provide just a brief description of our implementation of the mean-field iteration as an RNN module as shown in Fig.~\ref{fig:CRF}. 

The output of the CNN model is fed into the CRF module as the initial probability map. Next, we compute Gaussian kernels based on the input feature as Equation (\ref{eqn:kernels}). The value of above Gaussian kernels drops very fast as the distance (in the 3D cartesian space and the 2D angular space) between two points increases. Therefore, for each point, we limit the kernel size to a small region of $3\times 5$ on the input tensor. We then filter the initial probability map using the above Gaussian kernels. This step is also called message passing in \cite{CRFasRNN} since it essentially aggregates probabilities of neighboring points. This step can be implemented as a locally connected layer with the above Gaussian kernels as parameters. Next, we re-weight the aggregated probability and use a ``compatibility transformation'' to decide how much it changes each point's distribution. This step can be implemented as a 1x1 convolution whose parameters are learned during training. We update the initial probability by adding it to the output of the 1x1 convolution and use \textit{softmax} to normalize it. The output of the module is a refined probability map, which can be further refined by applying this procedure iteratively. In our experiment, we used 3 iterations to achieve an accurate label map. This recurrent CRF module together with the CNN model can be trained together end-to-end. With a single-stage pipeline, we sidestep the thread of propagated errors present in multi-stage workflows and leverage contextual information accordingly.

\subsection{Data collection}
Our initial data is from the KITTI raw dataset, which provides images, LiDAR scans and 3D bounding boxes organized in sequences. Point-wise annotations are converted from 3D bounding boxes. All points within an object's 3D bounding box are considered part of the target object. We then assign the corresponding label to each point. An example of such a conversion can be found in Fig. \ref{fig:ProjectedData} (a, b).
Using this approach, we collected 10,848 images with point-wise labels. 

In order to obtain more training samples (both point clouds and point-wise labels), we built a LiDAR simulator in GTA-V. The framework of the simulator is based on DeepGTAV\footnote{\url{https://github.com/aitorzip/DeepGTAV}}, which uses Script Hook V\footnote{\url{http://www.dev-c.com/gtav/scripthookv/}} as a plugin. 

We mounted a virtual LiDAR scanner atop an in-game car, which is then set to drive autonomously. The system collects both LiDAR point clouds and the game screen. In our setup, the virtual LiDAR and game camera are placed at the same position, which offers two advantages: First, we can easily run sanity checks on the collected data, since the points and images need to be consistent. Second, the points and images can be exploited for other research projects, e.g. sensor fusion, etc. 

We use ray casting to simulate each laser ray that LiDAR emits. The direction of each laser ray is based on several parameters of the LiDAR setup: vertical field of view (FOV), vertical resolution, pitch angle, and the index of the ray in the point cloud scan. Through a series of APIs, the following data associated with each ray can be obtained: a) the coordinates of the first point the ray hits, b) the class of the object hit, c) the instance ID of the object hit (which is useful for instance-wise segmentation, etc.), d) the center and bounding box of the object hit. 

\begin{figure}[h]
    \centering
    \includegraphics[width=.8\textwidth]{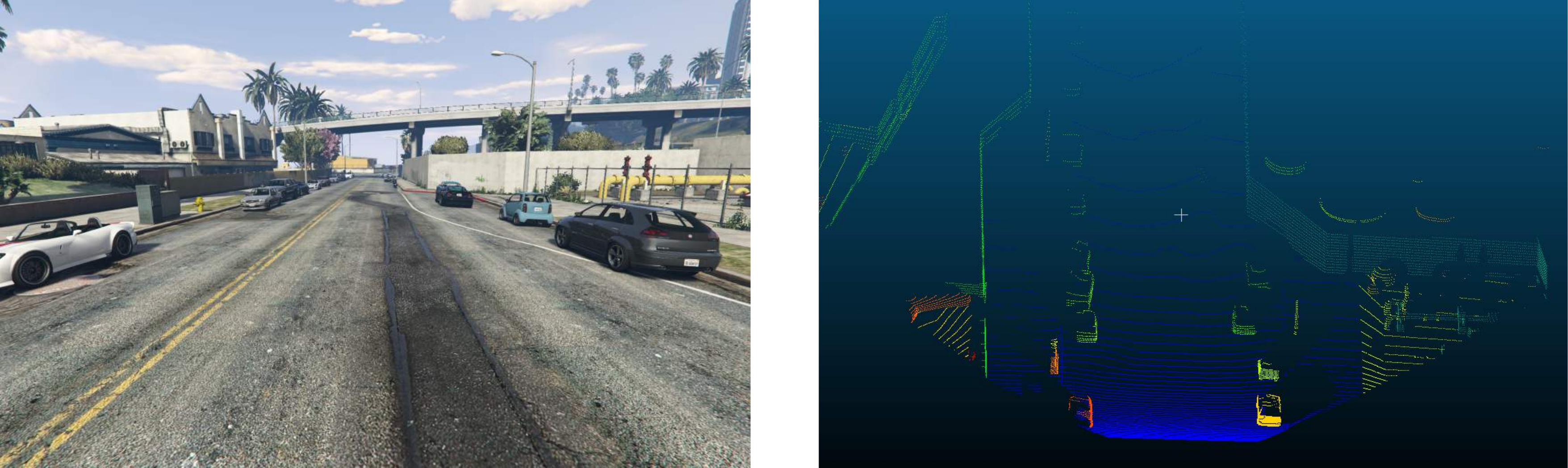}
    \caption[A GTA-V synthetic image vs. a real image.]{Left: Image of a game scene from GTA-V. Right: A LiDAR point cloud corresponding to the game scene.}
    \label{fig:GTA}
\end{figure}

\begin{figure}[h]
    \centering
    \includegraphics[width=0.9\linewidth]{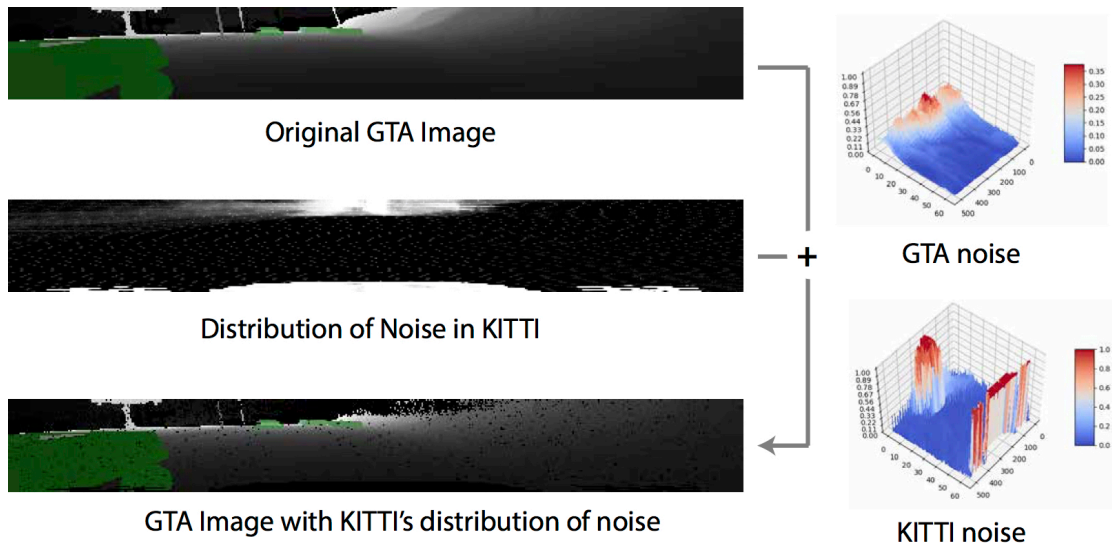}
    \caption{Fixing distribution of noise in synthesized data}
    \label{fig:noise}
\end{figure}

Using this simulator, we built a synthesized dataset with 8,585 samples, roughly doubling our training set size. To make the data more realistic, we further analyzed the distribution of noise across KITTI point clouds (Fig.~\ref{fig:noise}). We took empirical frequencies of noise at each radial coordinate and normalized to obtain a valid probability distribution. First, let $P_i$ be a 3D tensor in the format described earlier in Section \ref{sec:point_cloud} denoting the spherically projected ``pixel values'' of the $i$-th KITTI point cloud. For each of the $n$ KITTI point clouds, consider whether or not the pixel at the $(\tilde{\theta}, \tilde{\phi})$ coordinate contains ``noise.'' For simplicity, we consider ``noise'' to be missing data, where all pixel channels are zero. Then, the empirical frequency of noise at the $(\tilde{\theta}, \tilde{\phi})$ coordinate is
$$
\epsilon(\tilde{\theta}, \tilde{\phi}) = \frac{1}{n}\sum_{i=1}^n \mathds{1}_{\{P_i[\tilde{\theta}, \tilde{\phi}] = 0\}}.
$$
Then, we can then augment the synthesized data using the distribution of noise in the KITTI data. For any point cloud in the synthetic dataset, at each $(\tilde{\theta}, \tilde{\phi})$ coordinate of the point cloud, we randomly add noise by setting all feature values to $0$ with probability $\epsilon(\tilde{\theta}, \tilde{\phi})$.

It is worth noting that GTA-V used very simple physical models for pedestrians, often reducing people to cylinders.  In addition, GTA-V does not encode a separate category for cyclists, instead of marking people and vehicles separately on all accounts. For these reasons, we decided to focus on the ``car'' class for KITTI evaluation when training with synthetic data.

\section{Experiments}
\subsection{Evaluation metrics}
We evaluate our model's performance on both class-level and instance-level segmentation tasks. For class-level segmentation, we compare predicted with ground-truth labels, point-wise, and evaluate precision, recall, and IoU (intersection-over-union) scores, which are defined as follows:
\begin{gather*}
Pr_c = \frac{|\mathcal{P}_c \cap \mathcal{G}_c|}{|\mathcal{P}_c|}, 
recall_c = \frac{|\mathcal{P}_c \cap \mathcal{G}_c|}{|\mathcal{G}_c|}, 
IoU_c = \frac{|\mathcal{P}_c \cap \mathcal{G}_c|}{|\mathcal{P}_c \cup \mathcal{G}_c|},
\end{gather*}
where $\mathcal{P}_c$ and $\mathcal{G}_c$ respectively denote the predicted and ground-truth point sets that belong to class-$c$. $|\cdot|$ denotes the cardinality of a set. IoU score is used as the primary accuracy metric in our experiments.

For instance-level segmentation, we first match each predicted instance-$i$ with a ground truth instance. This index-matching procedure can be denoted as $\mathcal{M}(i) = j$ where $i \in \{1, \cdots, N\}$ is the predicted instance index and $j \in \{\emptyset, 1, \cdots, M\}$ is the ground truth index. If no ground truth is matched to instance-$i$, then we set $\mathcal{M}(i)$ to $\emptyset$. The matching procedure $\mathcal{M}(\cdot)$ 1) sorts ground-truth instances by the number of points and 2) for each ground-truth instance, finds the predicted instance with the largest IoU. The evaluation script can be found in our code release.

For each class-$c$, we compute  instance-level precision, recall, and IoU scores as
\begin{gather*}
Pr_c = \frac{\sum_i|\mathcal{P}_{i,c} \cap \mathcal{G}_{\mathcal{M}(i),c}|}{|\mathcal{P}_c|}, \\
recall_c = \frac{\sum_i |\mathcal{P}_{i,c} \cap \mathcal{G}_{\mathcal{M}(i),c}|} 
{|\mathcal{G}_{c}|}, \\
IoU_c = \frac{\sum_i|\mathcal{P}_{i,c} \cap \mathcal{G}_{\mathcal{M}(i),c}|}{|\mathcal{P}_c \cup \mathcal{G}_c|}.
\end{gather*}
$\mathcal{P}_{i,c}$ denotes the $i$-th predicted instance that belongs to class-$c$. Different instance sets are mutually exclusive, thus $\sum_i|\mathcal{P}_{i,c}| = |\mathcal{P}_c|$. Likewise for $\mathcal{G}_{\mathcal{M}(i),c}$. If no ground truth instance is matched with prediction-$i$, then $\mathcal{G}_{\mathcal{M}(i),c}$ is an empty set. 

\subsection{Experimental setup}
Our primary dataset is the converted KITTI dataset described above. We split the publicly available raw dataset into a training set with 8,057 frames and a validation set with 2,791 frames. Note that KITTI LiDAR scans can be temporally correlated if they are from the same sequence. In our split, we ensured that frames in the training set do not appear in validation sequences. Our training/validation split will be released as well. We developed our model in Tensorflow and used NVIDIA TITAN X GPUs, Drive PX2 \textit{AutoCruise} and \textit{AutoChauffeur} systems for our experiments. Since the KITTI dataset only provides reliable 3D bounding boxes for front-view LiDAR scans, we limit our horizontal field of view to the forward-facing $90^\circ$. We used DBSCAN to further process the output of SqueezeSeg to obtain instance-level segmentation, with a proximity radius of $0.3m$ and \textit{minPts} of 20. Details of our model training protocols and experiment parameters can be found in our code release: \url{https://github.com/BichenWuUCB/SqueezeSeg}.

\subsection{Experimental results}
Segmentation accuracy of SqueezeSeg is summarized in Table \ref{tab:main}. We compared two variations of SqueezeSeg, one with the recurrent CRF layer and one without. Although our proposed metric is very challenging -- as a high IoU requires point-wise correctness--SqueezeSeg still achieved high IoU scores, especially for the car category. Note that both class-level and instance-level recalls for the car category are higher than 90\%, which is desirable for autonomous driving, as false negatives are more likely to lead to accidents than false positives. We attribute lower performance on pedestrian and cyclist categories to two reasons: 1) there are many fewer instances of pedestrian and cyclist in the dataset. 2) Pedestrian and cyclist instances are much smaller in size and have much finer details, making it more difficult to segment. 

By combining our CNN with a CRF, we increased accuracy (IoU) for the car category significantly. The performance boost mainly comes from improvement in precision since CRF better filters misclassified points on the borders. At the same time, we also noticed that the CRF resulted in slightly worse performance for pedestrian and cyclist segmentation tasks. This may be due to lack of CRF hyperparameter tuning for pedestrians and cyclists. 

\begin{figure}[h]
    \centering
     \includegraphics[width=.9\linewidth,clip]{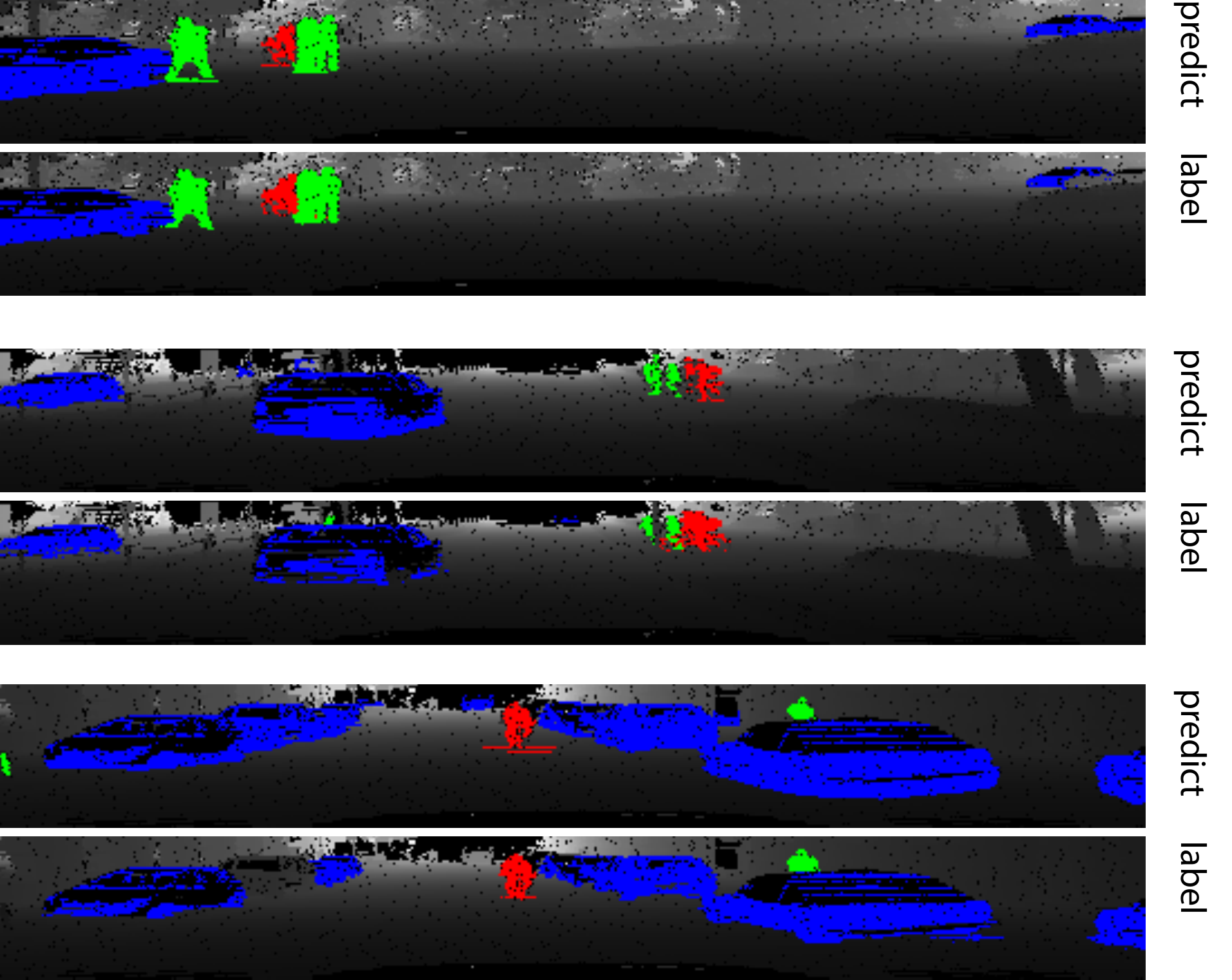}
    \caption[Visualization of SqueezeSeg's prediction.]{Visualization of SqueezeSeg's prediction on a projected LiDAR depth map. For comparison, visualization of the ground-truth labels is plotted below the predicted ones. Notice that SqueezeSeg additionally and accurately segments objects that are unlabeled in the ground truth.}
    \label{fig:viz}
\end{figure}

\begin{table}[h]
\centering
\vspace{0.4em}
\caption{Segmentation Performance of SqueezeSeg}
\label{tab:main}
\begin{tabular}{cc|ccc|ccc}
                                                 &         & \multicolumn{3}{c|}{Class-level} & \multicolumn{3}{c}{Instance-level} \\
                                                 &         & P         & R         & \textbf{IoU}      & P          & R         & \textbf{IoU}       \\ \hline
\multicolumn{1}{c|}{\multirow{2}{*}{car}}        & w/ CRF  & 66.7      & 95.4      & 64.6     & 63.4       & 90.7      & 59.5      \\
\multicolumn{1}{c|}{}                            & w/o CRF & 62.7      & 95.5      & 60.9     & 60.0       & 91.3      & 56.7      \\ \hline
\multicolumn{1}{c|}{\multirow{2}{*}{pedestrian}} & w/ CRF  & 45.2      & 29.7      & 21.8     & 43.5       & 28.6      & 20.8       \\
\multicolumn{1}{c|}{}                            & w/o CRF & 52.9      & 28.6      & 22.8     & 50.8       & 27.5      & 21.7      \\ \hline
\multicolumn{1}{c|}{\multirow{2}{*}{cyclist}}    & w/ CRF  & 35.7      & 45.8      & 25.1     & 30.4       & 39.0      & 20.6      \\
\multicolumn{1}{c|}{}                            & w/o CRF & 35.2      & 51.1      & 26.4     & 30.1       & 43.7      & 21.7      \\ \hline
\end{tabular}
\begin{tablenotes}
\small \item Summary of SqueezeSeg's segmentation performance. \textit{P, R, IoU} in the header row respectively represent precision, recall and intersection-over-union. IoU is used as the primary accuracy metric. All the values in this table are in percentages. 
\end{tablenotes}
\end{table}

Runtime of two SqueezeSeg models are summarized in Table~\ref{tab:runtime}. On a TITAN X GPU, SqueezeSeg without CRF only takes $8.7$ ms to process one LiDAR point cloud frame. Combined with a CRF layer, the model takes $13.5$ ms each frame. This is much faster than the sampling rate of most LiDAR scanners today. The maximum and typical rotation rate for Velodyne HDL-64E LiDAR, for example, is 20Hz and 10Hz respectively. On vehicle embedded processors such as Nividia Drive PX2 AutoCruise and AutoChauffeur,  where computational resources are more limited, SqueezeSeg can still achieve a frame rate of 13.5Hz to 40.0Hz. Also, note that the standard deviation of runtime for both SqueezeSeg models is very small, which is crucial for the stability of the entire autonomous driving system. However, our instance-wise segmentation currently relies on conventional clustering algorithms such as DBSCAN\footnote{We used the implementation from 
\url{http://scikit-learn.org/0.15/modules/generated/sklearn.cluster.DBSCAN.html}}, which in comparison takes much longer and has much larger variance. A more efficient and stable clustering implementation is necessary, but it is out of the scope of this chapter.

\begin{table}[h]
\footnotesize
\centering
\caption{Average runtime and standard deviation of SqueezeSeg}
\label{tab:runtime}
\begin{tabular}{c|c|c|c|c}
                  \footnotesize unit: ms                                    & \footnotesize Titan X  & \begin{tabular}[c]{@{}c@{}} \footnotesize Drive PX2 \\ \scriptsize AutoCruise\end{tabular} & \begin{tabular}[c]{@{}c@{}} \footnotesize Drive PX2\\ \scriptsize AutoChauffeur\end{tabular} & \begin{tabular}[c]{@{}c@{}} \footnotesize Xeon E5\\ \footnotesize CPU\end{tabular} \\ \hline
\footnotesize SqueezeSeg                                                   & $13.6/0.8$ & $74.0/0.8$                                                        & $37.8/1.7$                                                          & -                                                     \\
\begin{tabular}[c]{@{}c@{}} \footnotesize SqueezeSeg\\ \footnotesize w/o CRF\end{tabular} & $8.7/0.5$  & $52.0/1.3$                                                        & $25.1/0.8$                                                         & -                                                     \\
\footnotesize DBSCAN                                                       & -        & -                                                               & -                                                                 & $27.3/45.8$                                             \\ \hline
\end{tabular}
\begin{tablenotes}
\small \item Average runtime and standard deviation of SqueezeSeg, SqueezeSeg without CRF and DBSCAN on different processors. The unit for above values is millisecond. The first number in each cell is the average runtime and the second is the standard deviation. 
\end{tablenotes}
\end{table}

We tested our model's accuracy on KITTI data when trained on GTA simulated data. The results are summarized in Table~\ref{tab:gta}. Our GTA simulator is currently still limited in its ability to provide realistic labels for pedestrians and cyclists, so we consider only segmentation performance for cars. Additionally, our simulated point cloud does not contain intensity measurements; we, therefore, excluded intensity as an input feature to the network. To quantify the effects of training on synthesized data, we trained a SqueezeSeg model on the KITTI training set, without using intensity measurements, and validated on the KITTI validation set. The model's performance is shown in the first row of the table. Compared with Table~\ref{tab:main}, the IoU score is worse, due to the loss of the intensity channel. If we train the model completely on GTA simulated data, we see a significantly worse performance. However, combining the KITTI training set with our GTA-simulated dataset, we see significantly increased accuracy that is even better than Table~\ref{tab:main}. 

A visualization of the segmentation result by SqueezeSeg \textit{vs.} ground truth labels can be found in Fig.\ref{fig:viz}. For most of the objects, the predicted result is almost identical to the ground-truth, save for the ground beneath target objects. Also, notice SqueezeSeg additionally and accurately segments objects that are unlabeled in the ground truth. These objects may be obscured or too small, therefore placed in the ``Don't Care" category for the KITTI benchmark.

\begin{table}[h]
\centering
\caption{Segmentation Performance on the Car Category with Simulated Data}
\label{tab:gta}
\begin{tabular}{c|ccc|ccc}
\multicolumn{1}{l|}{} & \multicolumn{3}{c|}{Class-level}                                                  & \multicolumn{3}{c}{Instance-level}                                               \\
\multicolumn{1}{l|}{} & \multicolumn{1}{l}{P} & \multicolumn{1}{l}{R} & \multicolumn{1}{l|}{\textbf{IoU}} & \multicolumn{1}{l}{P} & \multicolumn{1}{l}{R} & \multicolumn{1}{l}{\textbf{IoU}} \\ \hline
KITTI                 & 58.9                  & 95.0                  & 57.1                              & 56.1                  & 90.5                  & 53.0                             \\
GTA                   & 30.4                  & 86.6                  & 29.0                              & 29.7                  & 84.6                  & 28.2                             \\
KITTI + GTA           & 69.6                  & 92.8                  & 66.0                              & 66.6                  & 88.8                  & 61.4                             \\ \hline
\end{tabular}
\end{table}

\section{Conclusion}

We present SqueezeSeg, an accurate, fast, and stable end-to-end approach for road-object segmentation from LiDAR point clouds. Addressing the deficiencies of previous approaches that were discussed in the Introduction, our deep learning approach 1) does not rely on hand-crafted features, but utilizes convolutional filters learned through training; 2) uses a deep neural network and therefore has no reliance on iterative algorithms such as RANSAC, GP-INSAC, and agglomerative clustering; and 3) reduces the pipeline to a single-stage, sidestepping the issue of propagated errors and allowing the model to leverage object context fully. 

LiDAR points sparse and irregularly distributed in a 3D space. Such data are challenging to process efficiently. To bypass this problem, we propose a novel spherical transformation to turn a 3D point cloud into a compact 2D representation that can be directly fed into a CNN.  In order to obtain training data, we propose an idea to convert bounding box annotations into point-wise labels. To obtain more data, we leverage simulation engine to synthesize large amounts of training data. Finally, we design the model structure of SqueezeSeg to be compact and high efficiency. As a result, SqueezeSeg accomplishes very high accuracy at faster-than-real-time inference speeds with small variance, as required for applications such as autonomous driving.

%% file: chap5.tex
\chapter{Data Efficiency: LATTE}
\label{chap:latte}

Deep neural networks require a large amount of data for training and evaluation. In many applications, however, obtaining such large amounts of data is not feasible due to the cost and difficulty of data collection and annotation. A typical example of this is LiDAR-based perception. Collecting LiDAR point cloud data requires installing costly LiDAR sensors, and more importantly, annotating LiDAR point clouds is significantly more difficult than annotating images, due to low resolution, complex operation, and temporal correlation. Therefore, getting enough LiDAR point cloud data to train neural networks is very challenging and expensive. It is important for us to address the following \textit{key question}:
\begin{quote}
    Can we improve the data efficiency of deep neural networks? 
\end{quote}
In the next two chapters, we discuss how to improve the \textit{data efficiency} of deep neural networks from two perspectives. In Chapter \ref{chap:latte}, we discuss how to build efficient annotation tools to accelerate the process of LiDAR annotation. In Chapter \ref{chap:sqsgv2}, we adopt a more aggressive approach to utilize simulated data to train neural networks, bypassing the need for collecting and annotating real data. 

\section{Introduction}
LiDAR (Light detection and ranging) is an essential and widely adopted sensor for autonomous vehicles. This is particularly true for applications such as Robo-Taxis that require higher levels (L4-L5) of autonomy. Compared with cameras, LiDARs are more robust to ambient light condition changes. They can also provide very accurate distance measurements (error $<2$cm $^1$) to nearby obstacles, which is essential for the planning and control of autonomous vehicles. 
\begin{figure}[h]
    \centering
    \includegraphics[width=0.8\linewidth]{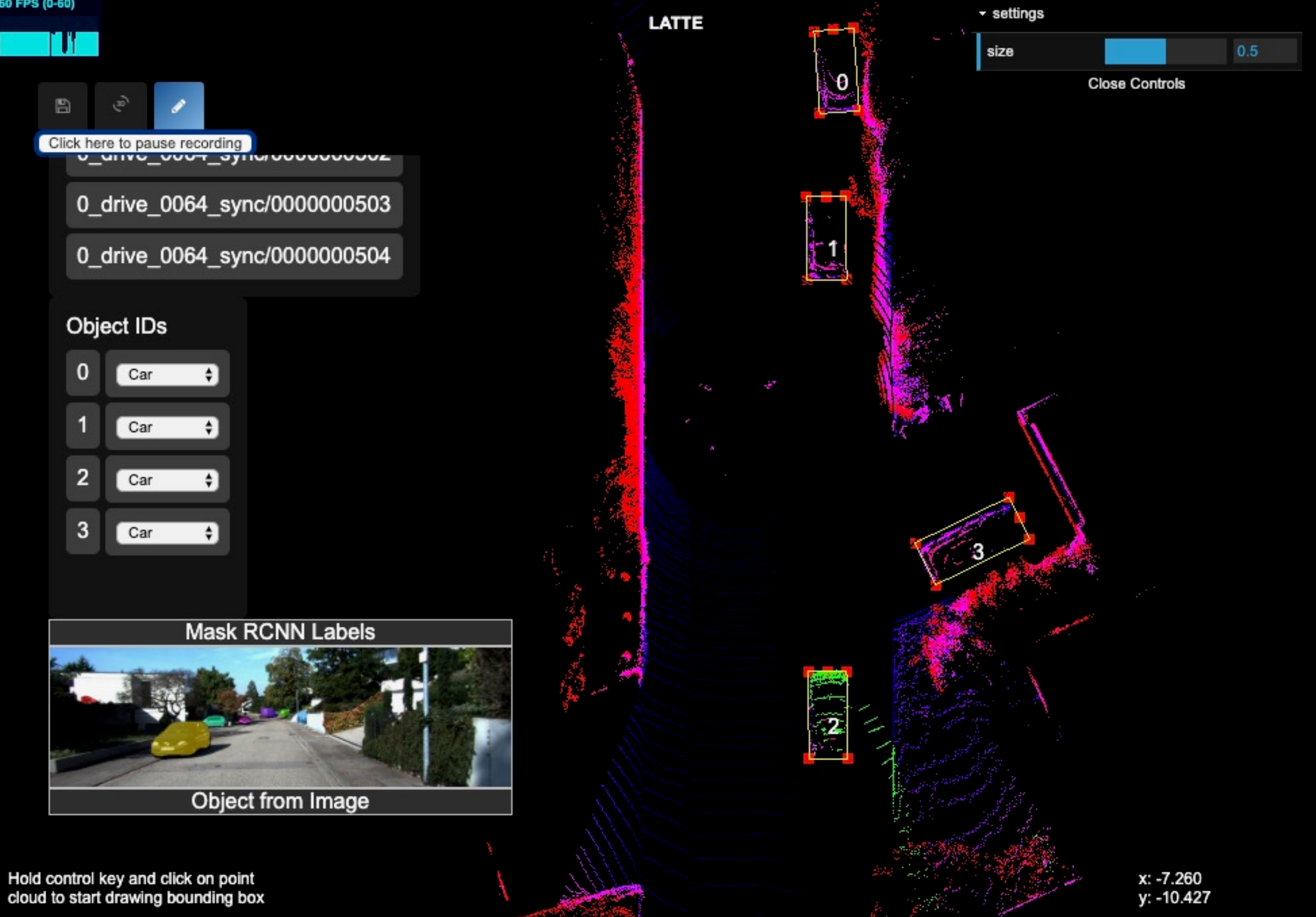}
    \caption[A screenshot of LATTE.]{A screenshot of LATTE. Best viewed in color.}
    \label{fig:abstract}
\end{figure}

To understand the environment through LiDAR, autonomous vehicles need to extract semantic meaning from the point cloud and accurately identify and locate objects such as cars, pedestrians, cyclists, and so on. Such problems are called LiDAR-based detection, and they have long been studied by the research community. An increasing number of works \cite{wu2018squeezeseg, wu2018squeezesegv2,qi2017frustum,LiDARDet} have demonstrated the promise of using Deep Learning to solve this problem. Compared with previous approaches, deep learning solutions obtain superior accuracy and faster speed, but they are also extremely data-hungry, requiring large amounts of data for training.  

\begin{figure*}[h]
    \centering
    \includegraphics[width=\textwidth]{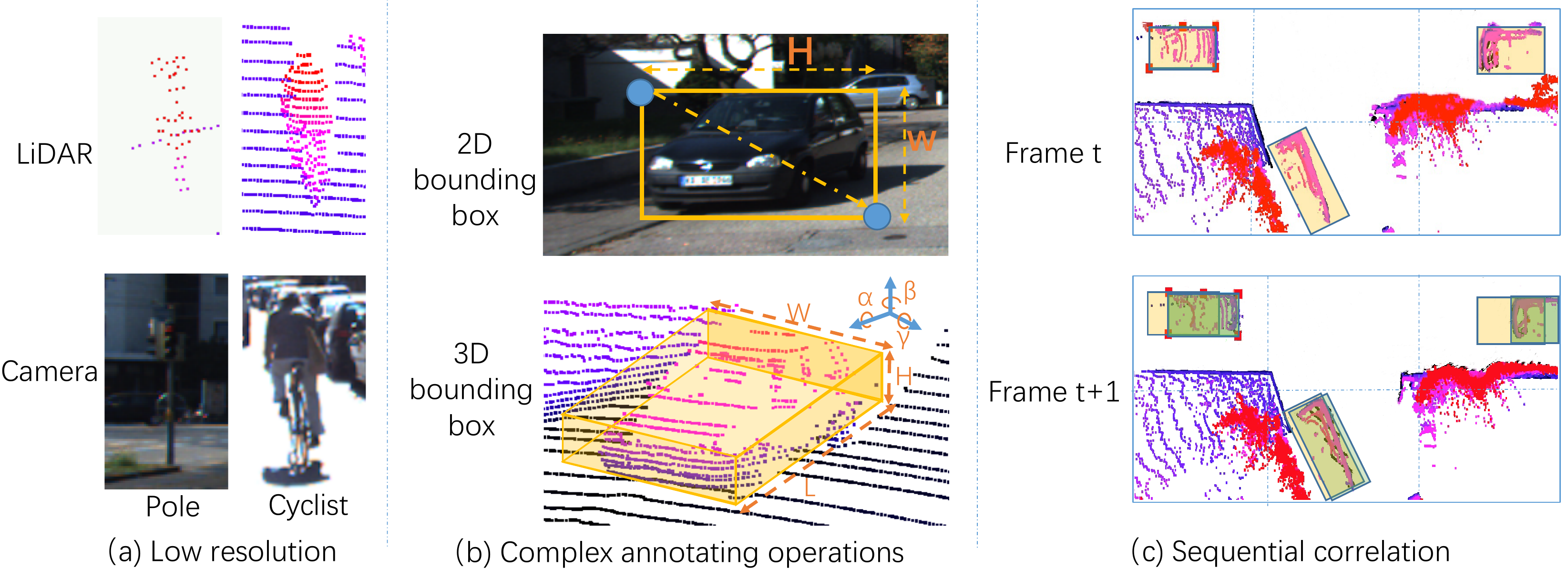}
    \caption[Challenges of annotating LiDAR point clouds.]{Challenges of annotating LiDAR point clouds. (a) LiDAR point clouds have low resolution and therefore objects are difficult for humans to recognize. The upper two figures are point clouds of a traffic pole and a cyclist, but both are difficult to recognize. The lower two are the corresponding images. (b) Annotating 2D bounding boxes on an image vs. 3D bounding boxes on a point cloud. Annotating 3D bounding boxes is more complicated due to more degrees of freedom of 3D scaling and rotation. (c) Point clouds of two consecutive frames are shown here. Even though the two frames are highly similar, target objects are moving and have different speeds. As a result, new bounding boxes are needed on new frames.}
    \label{fig:challenges}
\end{figure*}

Compared to annotating camera images, annotating LiDAR point clouds is much more difficult. The challenges can be summarized in the following three aspects and are illustrated in Fig. \ref{fig:challenges}: 1) \textbf{Low resolution}: Cameras can easily capture high-resolution images, with widespread support for 4K resolution (3840 x 2160 pixels). In comparison, LiDAR sensors have much more limited resolution. For example, typical vertical resolutions for Velodyne LiDARs are 32 or 64 lines (pixels). For the 64-line LiDAR\footnote{\url{https://velodynelidar.com/hdl-64e.html}} with a vertical angular resolution of 0.41$^\circ$, the spatial resolution at 50 meters is only 0.36 meters. As a result, LiDAR point clouds are very sparse, making it difficult for human annotators to identify objects, as shown in Fig. \ref{fig:challenges}(a). 2) \textbf{Complex annotating operations}: LiDAR-based detection problems have different formulations, and the most popular two are 3D bounding box detection and point-wise segmentation. The former requires predicting a 3D bounding box that tightly covers a target object, and the latter requires finding all the points that belong to a target object. In both scenarios, annotating 3D point cloud is significantly more complex than annotating a 2D image. In bounding box detection, for example, a 2D bounding box can be determined by drawing two corners. For 3D bounding boxes, however, annotators have to determine not only the center position, the length, width, and height of the target, but also 3D rotations. As a result, correctly annotating a 3D bounding box is much more complex for human annotators, as shown in Fig. \ref{fig:challenges}(b). 3) \textbf{Sequential correlation}: Many LiDAR point cloud data are collected in sequences, so consecutive frames are different but highly correlated. If we were to annotate LiDAR point cloud frame by frame naively, most of the annotations would be repeated, as shown in Fig. \ref{fig:challenges}(c).

Without addressing these challenges, it is difficult to annotate LiDAR data efficiently over a large dataset. This limits the progress of research for LiDAR-based detection. Furthermore, although several efforts are trying to create more open-sourced datasets (\cite{KITTI,huang2018apolloscape}) for LiDAR-based detection, the annotation tools behind these datasets are not publicly accessible. Moreover, there are obvious advantages in enabling groups to efficiently annotate LiDAR datasets for their own LiDAR sensors, configurations, and so on.

To address these problems, we propose \textit{LATTE}, an open-sourced LiDAR annotation tool, as shown in Fig. \ref{fig:abstract}. We address the challenges above with the following solutions: 1) \textbf{Sensor fusion}: Cameras have much higher resolution than LiDAR sensors, and image-based detection algorithms are much more mature than LiDAR-based.  LiDAR sensors are usually paired with cameras, and the two sensors are calibrated such that each point from the cloud can be projected to a corresponding pixel in the image. Therefore, we can apply camera-based image detection algorithms and transfer labels from an image to a 3D point cloud. The algorithm-generated labels are not perfect and are limited by the algorithm accuracy, projection and synchronization errors, but we can use them as pre-labels to help human annotators recognize objects and ``fine-tune'' the labels. 2) \textbf{One-click annotation}: We simplify the operational complexity for LiDAR annotation from drawing point-wise labels to drawing 3D bounding boxes, then to top-view 2D bounding boxes, and eventually to one-click annotations. For a target object, a human annotator need only click on one point on it, and we utilize clustering algorithms to expand the point-annotation to the entire object, and automatically estimate a top-view 2D bounding box around the object. From the top-view of a 2D bounding box, we can infer point-wise labels for segmentation problems, simply by treating each point inside the bounding box as part of the target object. 3) \textbf{Tracking}: To reduce repeated annotations on consecutive frames in a sequence, we utilize tracking algorithms to transfer annotations from one frame to subsequent ones. By integrating all these solutions, our annotation tool enables a 6.2x reduction in annotation time while delivering better label quality, as measured by 23.6\% and 2.2\% higher instance-level precision and recall, and 2.0\% higher bounding box IoU. Furthermore, we open-source our annotation tool and build it in a modular way such that each component can be replaced and improved easily if more advanced algorithms for each part become available. 

\section{Related work}
\label{sec:related work}
\textbf{LiDAR-based detection and datasets:} LiDAR-based detection aims to identify and locate objects of interest from a LiDAR point cloud. Two main problem formulations for these point clouds are bounding box object detection \cite{LiDARDet} and semantic segmentation\cite{wu2018squeezeseg}. Object detection aims to draw a tight bounding box that covers the target object, and semantic segmentation aims to predict labels for each point in the cloud and therefore find the cluster corresponding to target objects. Earlier works mainly rely on handcrafted geometrical features for segmentation and classification \cite{LiDARSegICRA2012,himmelsbach2008lidar,wang2012could}. More recent works adopt deep learning to solve this problem \cite{wu2018squeezeseg,wu2018squeezesegv2,qi2017pointnet,qi2017pointnet++,qi2017frustum, LiDARDet} and achieve significant improvements in accuracy and efficiency.

Deep learning methods require much more data for training, therefore many efforts have focused on creating public datasets for LiDAR-based detection. The KITTI dataset \cite{KITTI} contains about 15,000 frames of 3D bounding box annotations for road-objects. The Apolloscape dataset \cite{huang2018apolloscape} contains 140,000 frames of point-wise background annotation. Public datasets serve as benchmarks to facilitate research, but they are not enough to support product adoption since different configurations of LiDAR sensors, locations, and so on requires creating different datasets. Therefore, it is equally important to provide annotation tools to enable more groups to create their own datasets.

\textbf{Annotation tools:} Many efforts focus on improving annotation tools to generate more data for deep learning training, but most of the annotation tools are focusing on images. VIA \cite{dutta2016via} provides a simple yet powerful webpage-based tool for drawing bounding boxes and polygons on images. Later works such as PolygonRNN \cite{castrejon2017annotating, acuna2018efficient} seek to utilize more advanced algorithms to facilitate and accelerate human annotations. For video annotation, VATIC \cite{vatic} integrates tracking (mainly linear interpolation) to reduce annotating repeated entities in consecutive frames. For autonomous driving applications, Yu et al. propose Scalabel \cite{yu2018bdd100k}, a package of tools that support annotating bounding boxes and semantic masks on images. Few works have focused on building LiDAR annotation tools. The Apolloscape \cite{huang2018apolloscape} dataset's annotation pipeline uses sensor fusion and image-based detection to generate labels through images. But their 3D annotations are for static backgrounds instead of moving objects. \cite{piewak2018boosting} adopts a sensor-fusion strategy to generate LiDAR point cloud labels using image-based detectors, and directly use them to train LiDAR-based detectors. However, the correctness of such labels is limited by the accuracy of the image detector, projection and synchronization error. Moreover, neither \cite{huang2018apolloscape} nor \cite{piewak2018boosting} have open-sourced their annotation tools. 

\textbf{Data collection through simulation:} To sidestep the difficulty of data collection and annotation, many research efforts aim at using simulation to generate LiDAR point cloud data to train neural networks. Yue et al. \cite{yue2018lidar} built a LiDAR simulator on top of the video game GTA-V. The simulated data are then used to train, evaluate, and verify deep learning models \cite{wu2018squeezeseg,wu2018squeezesegv2,yue2018lidar}. Carla \cite{Dosovitskiy17} is an open-source simulator for autonomous driving, and it supports image and LiDAR data generation. However, due to the distribution shift between the simulated data and the real-world data, deep learning models trained on simulated data perform poorly on the real-world data. Many works aim to close the gap between simulation and the real world by domain adaptation \cite{wu2018squeezesegv2}. Despite some promising progress, domain adaptation remains a challenging problem and the gap has been reduced but not closed. Therefore, collecting and annotating real-world data is still critical.

\section{Method}
\label{sec:method}
In this section, we discuss in detail three features of LATTE that aim to accelerate LiDAR point cloud annotation: sensor fusion, one-click annotation, and tracking.

\subsection{Sensor Fusion}
\begin{figure}[h]
    \centering
    \includegraphics[width=\linewidth]{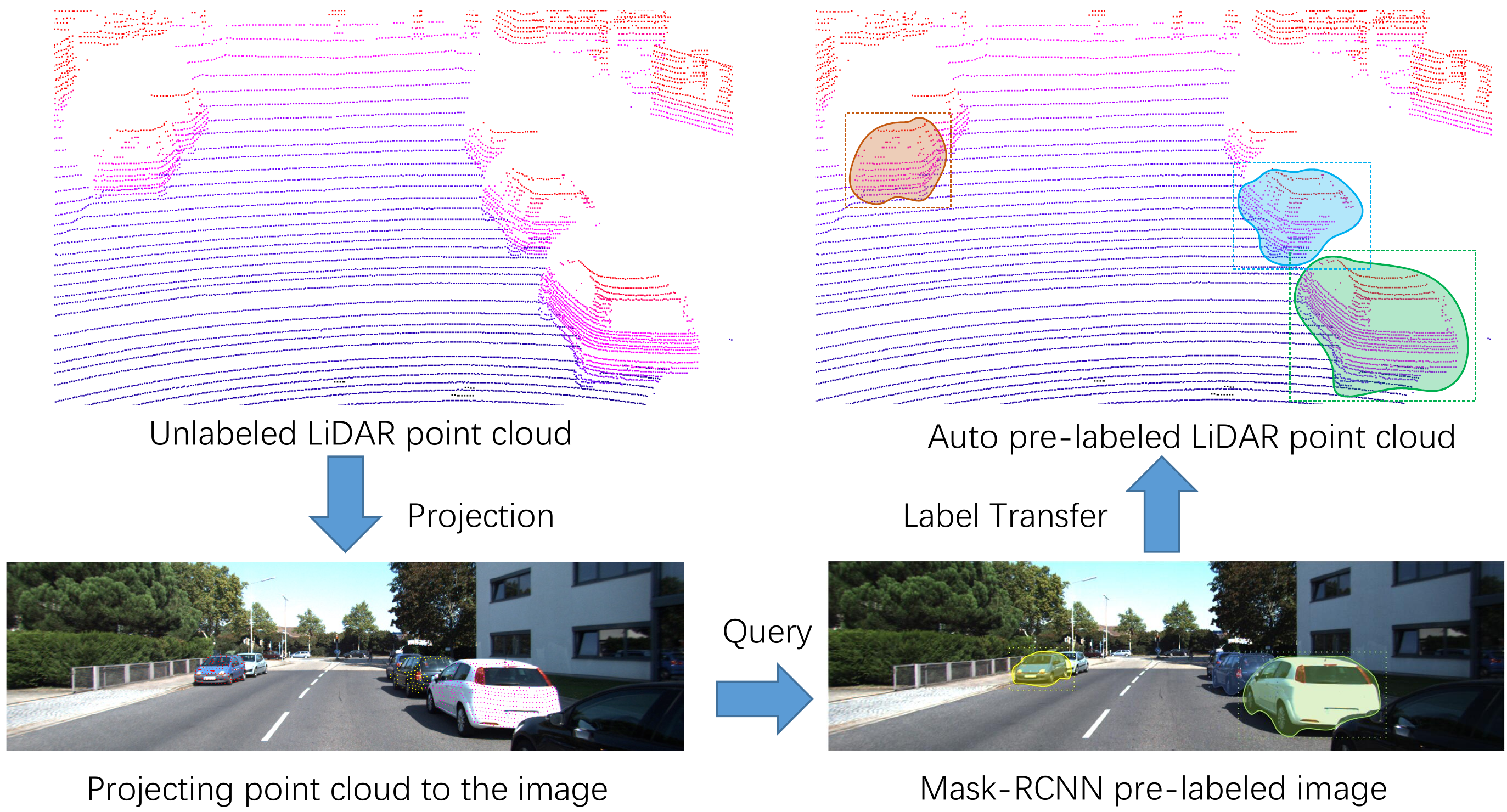}
    \caption[The sensor-fusion pipeline of LATTE.]{The sensor-fusion pipeline of LATTE. A LiDAR point cloud is projected onto its corresponding image. Next, we use Mask-RCNN to predict semantic labels on the image. The labels are then transferred back to the LiDAR point cloud.}
    \label{fig:sensor-fusion}
\end{figure}

As shown in Fig.\ref{fig:challenges}(a), LiDAR sensors are low resolution and are difficult for human annotators to recognize. In comparison, cameras have higher resolution, and image-based detection algorithms are more mature than LiDAR-based. Therefore, we use image-based detection to help us annotate LiDAR point cloud. Our pipeline is illustrated in Fig. \ref{fig:sensor-fusion}.

LiDAR sensors are paired with cameras, and the two sensors are calibrated such that for each point $\mathbf{p}_i$ with a 3D coordinate $(x_i, y_i, z_i)$ in the point cloud, we can project it to a pixel $\mathbf{q}_i$ with a 2D coordinate $(u_i, v_i)$ in the corresponding image. The projection can be mathematically described as
\[
\mathbf{q}_i = P\mathbf{p}_i + \mathbf{t},
\]
where $P\in \mathcal{R}^{2\times 3}$ is the projection matrix and $\mathbf{t} \in \mathcal{R}^2$ is the translation vector.  

We then apply semantic segmentation on the image. In our annotator, we use Mask R-CNN \cite{he2017mask} to get semantic labels for each pixel in the image. The semantic labels can be regarded as a mask $\mathcal{M}$ such that for each pixel $\mathbf{q}_i$ with coordinates $(u_i, v_i)$, we can find its label $l_i$ as $l_i = \mathcal{M}(u_i, v_i)$. Finally, this label can be transferred to its corresponding point in the 3D space. This way, we can automatically generate pre-labels for the point cloud. 

\begin{figure}[h]
    \centering
    \includegraphics[width=0.9\linewidth]{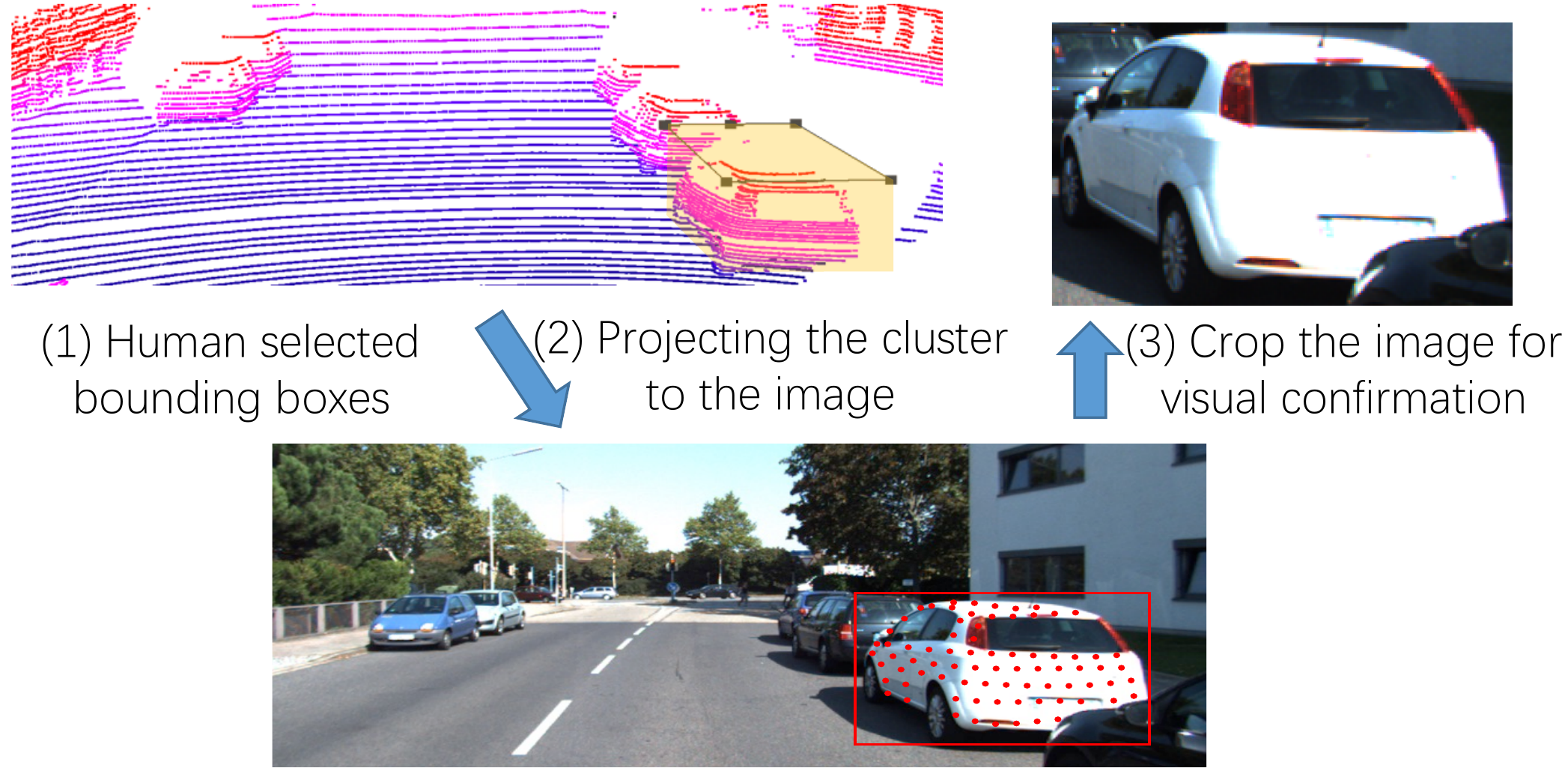}
    \caption[The visual confirmation pipline of LATTE.]{We use sensor fusion to help annotators confirm the category of a selected object. Once a 3D bounding box is chosen, we project all the points within the bounding box to the image and show the corresponding crop of the image to human annotators for visual confirmation.}
    \label{fig:visual-conf}
\end{figure}

We highlight the pre-labeled points in the original point cloud such that human annotators can quickly identify objects of interest. After an annotator draws a bounding box over the target object, we again project all the points in the cluster back to the image, find the patch of the image that contains the target object, crop the patch and show it to human annotators for confirmation, as illustrated in Fig. \ref{fig:visual-conf}. 

\subsection{One-click Annotation}
\label{sec:one-click}
\begin{figure}[h]
    \centering
    \includegraphics[width=\linewidth]{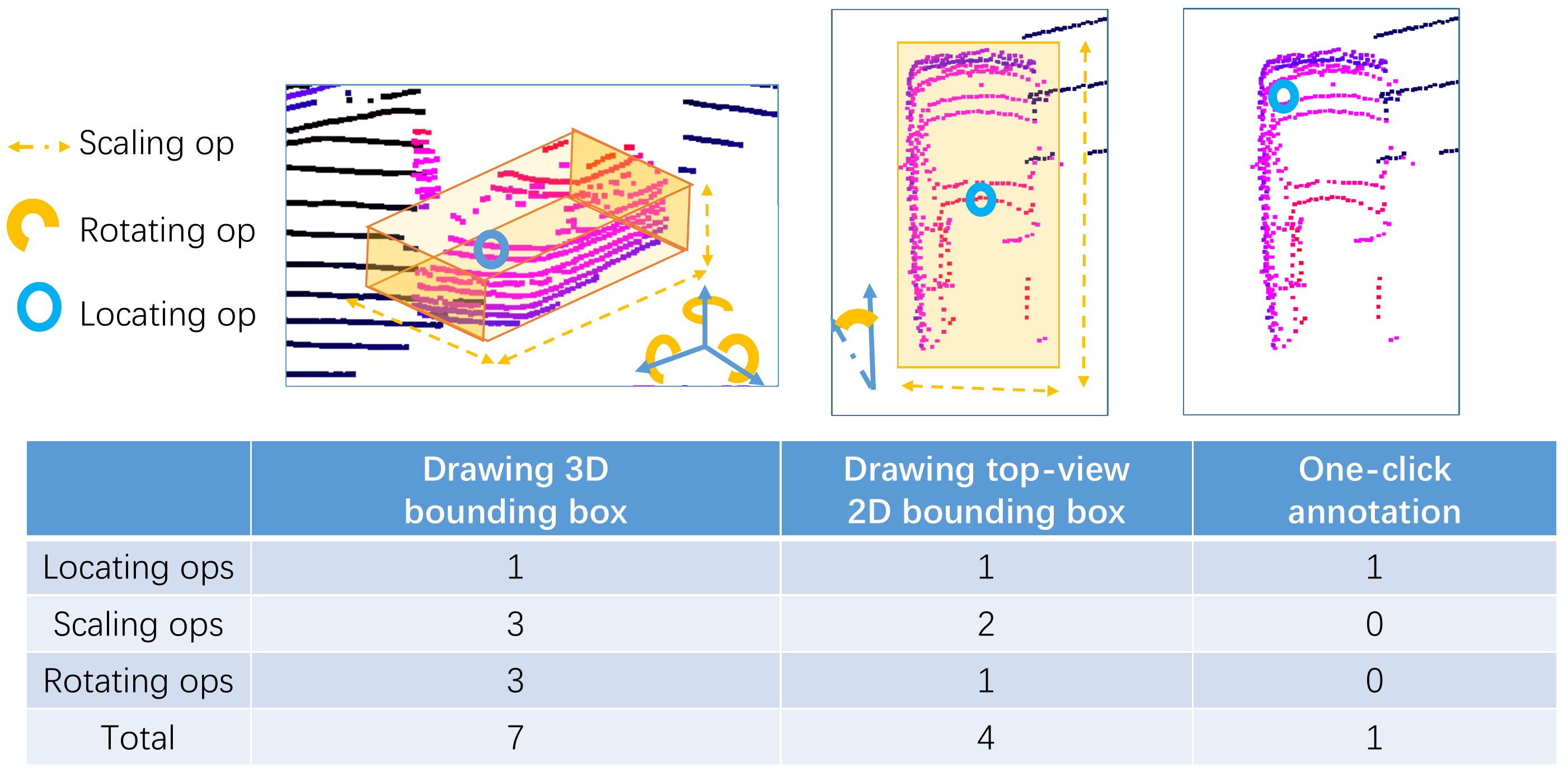}
    \caption[A comparison of annotation operations. ]{A comparison of drawing a 3D bounding box, a top-view 2D bounding box, and one-click annotation.}
    \label{fig:ops-comparison}
\end{figure}

In this section, we discuss how we simplify the annotation operation from drawing point-wise labels to drawing 3D bounding box, then to top-view 2D bounding boxes, and eventually to simply one-click annotation. A comparison of drawing a 3D bounding box, a top-view 2D bounding box, and one-click annotation is illustrated in Fig.\ref{fig:ops-comparison}.

LiDAR-based perception can be formulated as a bounding box detection problem or a semantic segmentation problem. The former requires annotating 3D bounding boxes as shown in Fig.\ref{fig:challenges}(b), while the latter requires annotating point-wise labels. Naively annotating each point to obtain point-wise categorical labels is not feasible. Fortunately for 3D point cloud, point-wise labels can be obtained from 3D bounding boxes, as explained in \cite{wu2018squeezeseg}. For most of the road-objects we care about, their bounding boxes do not overlap in 3D space. As a result, point-wise labels can be converted from 3D bounding boxes, simply by treating each point inside a bounding box as part of the target object and therefore with the same categorical label. 

However, drawing 3D bounding boxes is still operationally complex. As illustrated in Fig. \ref{fig:ops-comparison}, ideally, drawing a 3D bounding box requires 1 operation to locate the object, 3 operations to scale the sides of the bounding box, and 3 rotations to adjust the orientation. For autonomous driving applications, what is more important is to locate the object from the top-view. Therefore, we can simplify a 3D bounding box to a top-view 2D bounding box, which can be determined by its 2D center position, 2D sizes of width and length, and its yaw angle. The operations needed to draw such a bounding box include 1 locating operation, 2 scaling operations, and 1 rotation, as illustrated in Fig. \ref{fig:ops-comparison}. 

\begin{figure}[h]
    \centering
    \includegraphics[width=\linewidth]{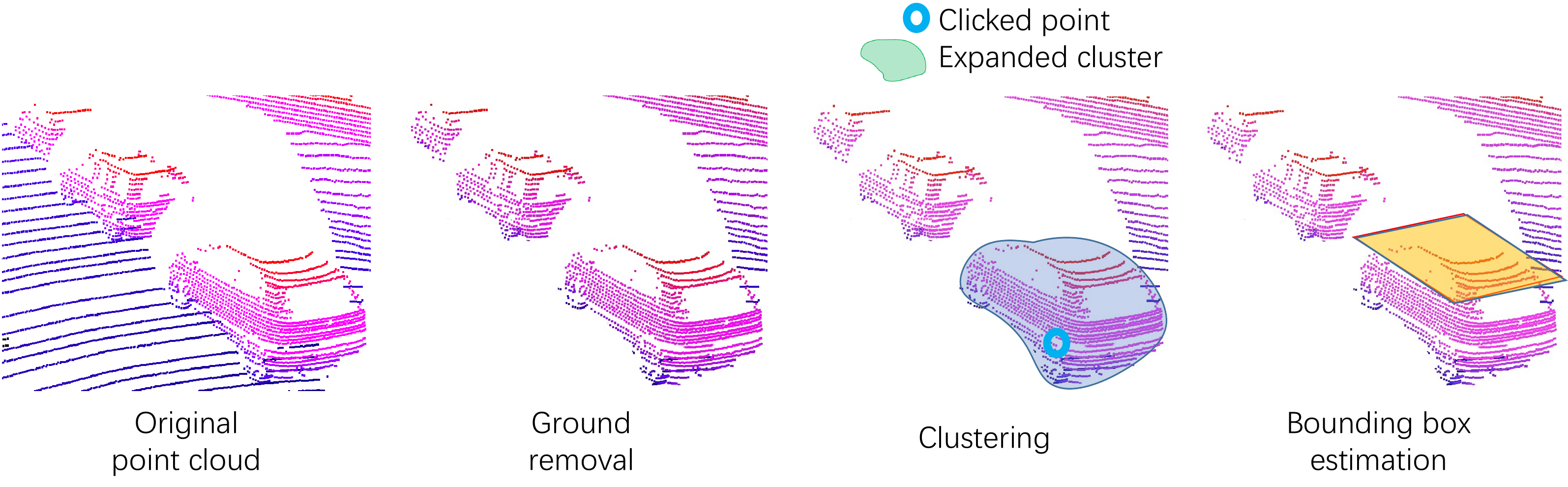}
    \caption[The one-click annotation pipeline of LATTE.]{The one-click annotation pipeline of LATTE. For a given LiDAR point cloud, we first remove the ground. After an annotator clicks on one point on a target object, we use clustering algorithms to expand from the clicked point to the entire object. Finally, we estimate a top-view 2D bounding box for the object.}
    \label{fig:one-click}
\end{figure}

To further reduce the annotation complexity, we built one-click annotation -- human annotators only need to click on one point on the target object, as illustrated in Fig. \ref{fig:ops-comparison}. After the locating operations by human annotators, we automatically apply clustering algorithms to find all the points for the target object. From the point cluster, we estimate a top-view 2D bounding box for the target object. Then, human annotators only need to adjust the automatically generated bounding box if it does not fit the object perfectly. Our one-click annotation is summarized in Fig. \ref{fig:one-click}. It mainly contains three steps: ground removal, clustering, and bounding box estimation.

\textbf{Ground removal:} We model the ground as a segment of planes where each plane is characterized by a linear model: 
\[
 \mathbf{n}^{T}\mathbf{p} = d,
\]
where $\mathbf{n} = [a, b, c]^T$ is the normal vector,  $\mathbf{p} = [x, y, z]^T$ is a point at the plane, and $d$ is the distance to the ground. To determine the ground, we need to estimate the normal vector $n$ from the noisy LiDAR data. We initially estimate the normal vector by sampling a set of lowest points in the vertical direction (z-direction). We denote the set as $G_0$ and compute the covariance matrix $C_0 \in \mathcal{R}^{3 \times 3}$:
\begin{equation*}
\begin{gathered}
   \bar{\mathbf{p}} = \frac{1}{|G_0|}\sum_{i=1}^{|G_0|}{\mathbf{p}_i}, \\
    C_0 = \sum_{i=1}^{|G_0|}{(\mathbf{p}_i - \bar{\mathbf{p}})(\mathbf{p}_i - \bar{\mathbf{p}})^T}.
\end{gathered}
\end{equation*}

where $\bar{\mathbf{p}}$ is the mean of the points in $G_0$ and the covariance matrix $C_0$ represents how spread-out the points in $G_0$ are. We analyze the direction of the dispersion by computing its singular value decomposition (SVD). The first two singular vectors corresponding to the two largest singular values represent the span of the plane. The singular vector corresponding to the smallest singular value is a good approximation for the normal because the variance in the direction of the normal is the smallest among all directions. 

After computing the normal vector, we now have an updated estimate of the ground plane. With our estimated plane we resample the ground points by their distance and iteratively update the normal vector:
\[
G_{k} = \{\mathbf{p} : |\mathbf{n}_{k-1}^T \cdot \mathbf{p}| < thresh\}
\]
where $|\mathbf{n}_{k-1}^T \cdot \mathbf{p}|$ is the distance between point $p$ and the plane whose normal vector at iteration $k-1$ is $\mathbf{n}_{k-1}$. 
The normal vector approximation and ground sampling are repeated until the segmentation converges or for a fixed number of iterations. 

\textbf{Clustering:}
After removing the ground we find the nearest cluster to the point where the human annotator clicks on. The clustering algorithm is based on density-based spatial clustering of applications with noise (DBSCAN) \cite{Ester:1996:DAD:3001460.3001507} and is described in Algorithm \ref{alg:cluster}, where FindNeighbor(p, X, $\epsilon$) finds the  neighbors in X that are $\epsilon$-close to p.

\begin{algorithm}
\SetAlgoLined
\KwIn{seed $s \in \mathcal{R}^{3}$, point cloud $P \in \mathcal{R}^{n \times 3}$, distance threshold $\epsilon$}
\KwOut{cluster $C \in \mathcal{R}^{m \times 3}$}
 seen = $\emptyset$, initialize Q\;
 Q.push(seed)\;
 \While{Q not empty}{
  neighbors = FindNeighbors(Q.pop(), P, $\epsilon$)\; 
  \For{neighbor in neighbors}{
   \If{neighbor not in seen}{
    seen.add(neighbor)\;
    Q.push(neighbor)\;
   }
  }
  }
  return seen\;
 \caption{Clustering Algorithm}
 \label{alg:cluster}
\end{algorithm}

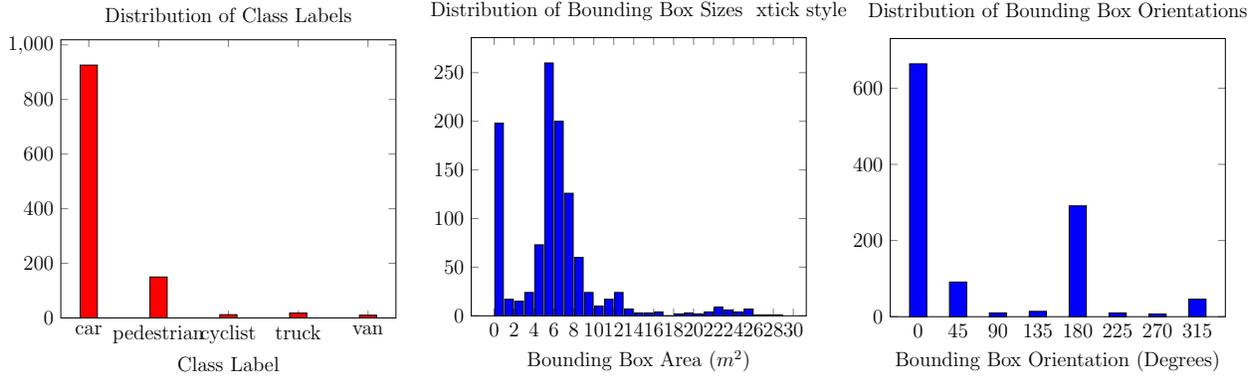
\begin{figure*}
\centering
    \begin{subfigure}[b]{0.32\textwidth}
    \begin{tikzpicture}[scale=0.65]
        \begin{axis}[
            title={Distribution of Class Labels},
            symbolic x coords={car, pedestrian, cyclist, truck, van},
            ymin=0,
            xlabel=Class Label]
        ]
        \addplot[ybar, fill=red]
	        coordinates {(car, 926) (pedestrian, 150) (cyclist, 12) (truck, 18) (van, 10)};
        \end{axis}
        \end{tikzpicture}
    \end{subfigure} \hfill
    \begin{subfigure}[b]{0.32\textwidth}
    \begin{tikzpicture}[scale=0.65]
        \begin{axis}[
            title={Distribution of Bounding Box Sizes }
            xtick style={draw=none},
            xtick={0,2,...,30},
            ymin=0,
            bar width=5pt,
            xlabel=Bounding Box Area $(m^2)$]
        ]
        \addplot[ybar, fill=blue]
	        coordinates {(.5, 198) (1.5, 17) (2.5, 15) (3.5, 24) (4.5, 73) (5.5, 260) (6.5, 200) (7.5, 126) (8.5, 60) (9.5, 24) (10.5, 10) (11.5, 17) (12.5, 24) (13.5, 7) (14.5, 3) (15.5, 3) (16.5, 4) (17.5, 0) (18.5, 2) (19.5, 3) (20.5, 2) (21.5, 4) (22.5, 9) (23.5, 6) (24.5, 4) (25.5, 7) (26.5, 1) (27.5, 1) (28.5, 1)};
        \end{axis}
        \end{tikzpicture}
    \end{subfigure} \hfill
        \begin{subfigure}[b]{0.32\textwidth}
        \begin{tikzpicture}[scale=0.65]
        \begin{axis}[
            title={Distribution of Bounding Box Orientations},
            xtick style={draw=none},
            xtick={0,45,...,360},
            ymin=0,
            xlabel=Bounding Box Orientation (Degrees)]
        ]
        \addplot[ybar, fill=blue]
	        coordinates {(0, 664) (45, 91) (90, 10) (135, 14) (180, 291) (225, 10) (270, 7) (315, 46)};
        \end{axis}
        \end{tikzpicture}
    \end{subfigure}
\caption[Distribution of bounding boxes in test data.]{Distribution of bounding boxes by class, box area, and orientation in our test benchmark.}
\label{fig:stats}
\end{figure*}

Since LiDAR point clouds can contain a large number of points (approximately 100,000 points per frame for Velodyne LiDAR), we perform pruning and downsampling in order to make one-click annotation efficient. Fig. \ref{fig:stats} shows that the distribution of bounding box sizes is concentrated around 6$m^2$, the size of a typical car. Therefore we can assume an upper bound on the dimensions of an object and appropriately prune the point cloud.  

\textbf{Bounding box estimation:} After we find the cluster, we use a search-based rectangle fitting \cite{Zhang-2017-26536} to estimate bounding boxes. Other methods, such as PCA based ones, can also be plugged into LATTE. To have the optimal rectangle fitting for a cluster, we need to know the appropriate heading of the rectangle. Ideally, the rectangle can be found by solving the following optimization problem:
\begin{equation*}
 \begin{aligned}
  & \argmin_{\theta, U, V, c_1, c_2 } & &\sum_{i \in |U|}{(x_i \cos\theta + y_i \sin\theta - c_1)^2} +  \\
  & & &\sum_{j \in |V|}{(-x_j \sin\theta + y_j \cos\theta - c_2)^2}, \\ 
  & \text{subject to } & & U \cup V = G, U \cap V = \emptyset. \\ 
 \end{aligned}
\end{equation*}
It aims to partition observed points in $G$ into two mutually exclusive groups $U$ and $V$ depending on which edges they are closer to. Points in $U$ are closer to the edge $x\cos\theta + y\sin\theta - c_1 $, and points in $V$ are closer to $-x\sin\theta + y\cos\theta - c_2$. 

Due to the combinatorial nature of the problem, it is infeasible to solve it exactly. To solve this problem approximately and efficiently, we use search-based rectangle fitting algorithm \cite{Zhang-2017-26536} that searches headings and projects the points in the cluster to two perpendicular edges. It searches the optimal heading to minimize a loss function as:
\begin{equation*}
    \theta^{*} = \argmin_{\theta \in [0, \pi]} L(G\mathbf{e}_{\theta,1}, G \mathbf{e}_{\theta,2}),
\end{equation*}
where $G \in \mathcal{R}^{n\times 2}$ denotes a matrix where each row contains the $(x, y)$-coordinate of a point. $\mathbf{e}_{\theta,1} = [\cos\theta, \sin\theta]^T, \mathbf{e}_{\theta,2} = [-\sin\theta, \cos\theta]^T$ are orthogonal unit vectors representing the directions of two perpendicular edges. The loss function $L(\cdot, \cdot)$ is defined as the following. We denote $\mathbf{c}_1 = G\mathbf{e}_{\theta,1}, \mathbf{c}_2 = G\mathbf{e}_{\theta,2}$, which represent projection of points to $\mathbf{e}_{\theta,1}, \mathbf{e}_{\theta,2}$. Then the distances from points in $G$ to the closer edge is computed as 
\begin{equation*}
    \begin{gathered}
        \mathbf{d}_1 = \argmin_{\mathbf{v} \in \{\mathbf{c}_1 - \min \{\mathbf{c}_1\}, \mathbf{c}_1 - \max \{\mathbf{c}_1\}\}} \|\mathbf{v}\|_2, \\
        \mathbf{d}_2 = \argmin_{\mathbf{v} \in \{\mathbf{c}_2 - \min \{\mathbf{c}_2\}, \mathbf{c}_2 - \max \{\mathbf{c}_2\}\}} \|\mathbf{v}\|_2.
    \end{gathered}
\end{equation*}
We can then divide all the points to two groups according to above distances and compute the loss function as
\begin{equation*}
 \begin{aligned}
  L(G\mathbf{e}_{\theta, 1}, G\mathbf{e}_{\theta, 2}) = & - \mathrm{Var}(\{\mathbf{d}_{1,i} : \mathbf{d}_{1,i} < \mathbf{d}_{2,i}\}) \\ & - \mathrm{Var}(\{\mathbf{d}_{2,i} : \mathbf{d}_{2,i} < \mathbf{d}_{1,i}\})
\end{aligned}
\end{equation*}
where $\mathbf{d}_{1, i}$ is the $i$-th element of $\mathbf{d}_{1}$, and similar for $\mathbf{d}_{2, i}$. $\mathrm{Var}(\cdot)$ computes the variance of a set of values. After solving for the optimal heading $\theta^*$, the following equations are used to compute the four edges of the rectangle: 
\begin{equation*}
 \begin{aligned}
\text{Edge 1:} & x\cos\theta^* + y\sin\theta^* - \min\{\mathbf{c}_{1}^*\} = 0 \\
\text{Edge 2:} & x\cos\theta^* + y\sin\theta^* - \max\{\mathbf{c}_{1}^*\} = 0 \\
\text{Edge 3:} & -x\sin\theta^* + y\cos\theta^* - \min\{\mathbf{c}_{2}^*\} = 0 \\
\text{Edge 4:} & -x\sin\theta^* + y\cos\theta^* - \max\{\mathbf{c}_{2}^*\} = 0
\end{aligned}
\end{equation*}
where  $\mathbf{c}_1^* = G\mathbf{e}_{\theta^{*},1},
\mathbf{c}_2^* = G\mathbf{e}_{\theta^{*},2}$.

\subsection{Tracking}
\begin{figure}[h]
    \centering
    \includegraphics[width=0.75\linewidth]{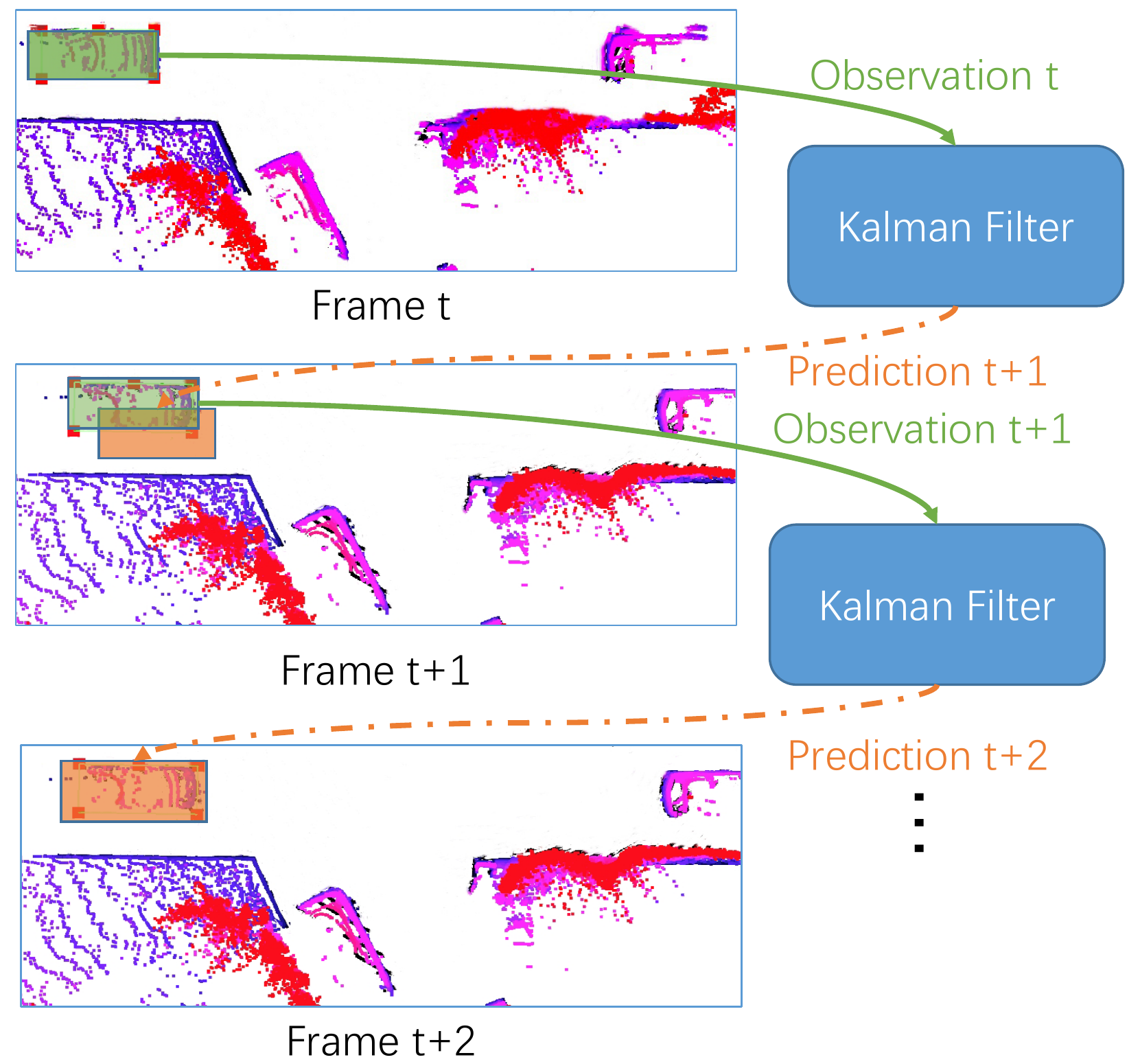}
    \caption[The tracking pipeline of LATTE.]{The tracking pipeline of LATTE. Annotators label a bounding box in the initial frame. Next, we use Kalman filtering to predict the center position of the bounding box at the next frame. Human annotators then adjust the bounding box, and we use the new center position as a new observation to update the Kalman filter.}
    \label{fig:tracking}
\end{figure}

To accelerate annotation on sequences, we integrate tracking to LATTE such that annotations from one frame can be transferred to subsequent ones, as illustrated in Fig. \ref{fig:tracking}.

LATTE is constructed in a modular way such that it can support different tracking algorithms, but we adopt Kalman filtering \cite{Welch:1995:IKF:897831} in our implementation. We use Kalman filtering to track the bounding box center of a target object. Human annotators need to label the first frame of a sequence. Next, our algorithm predicts the centers of bounding boxes for the next frame. For non-rigid objects (such as pedestrians), their bounding boxes do not have fixed shapes. Therefore, we re-estimate the bounding boxes using a similar algorithm as the one-click annotation, as described in section \ref{sec:one-click}. For rigid objects such as cars, their bounding boxes should not change over time, so we only estimate the yaw angle. The predicted bounding box is displayed at the next frame, and the human annotator can simply make adjustments to the bounding boxes. The adjusted bounding boxes then serve as observations in the Kalman update step. 

Formally, we define the state vector of a bounding box at frame $k$ as $x_k = [p_x, p_y, v_x, v_y, a_x, a_y]^T$, where $p_x$ and $p_y$ are the coordinates for the center of the bounding box at frame $k$. $v_x$, $v_y$ and $a_x$, $a_y$ represent the velocity and acceleration of the center, respectively. A human annotates the first frame in the sequence, and we can obtain the initial values for the center position. The initial velocity and acceleration values are left to be zero. Next, we predict the center coordinates at the next frame as 
\begin{equation*}
\begin{gathered}
    \mathbf{\hat{x}}_{k|k-1} = F\mathbf{\hat{x}}_{k-1|k-1}, \\
    P_{k|k-1}=F P_{k-1|k-1} F^{T}+Q,
\end{gathered}
\end{equation*}
where $F \in \mathcal{R}^{6 \times 6}$ represents the state transition model. We assume a constant acceleration model and define $F$ as 
\[
\mathbf{F} = \begin{bmatrix} 
1 & 0 & \Delta t & 0 & \frac{1}{2} \Delta t^2 & 0\\
0 & 1 & 0 & \Delta t & 0 & \frac{1}{2} \Delta t^2\\
0 & 0 & 1 & 0 & \Delta t & 0\\ 
0 & 0 & 0 & 1 & 0 & \Delta t\\ 
0 & 0 & 0 & 0 & 1 & 0\\
0 & 0 & 0 & 0 & 0 & 1\\
\end{bmatrix},
\]
where $\Delta t$ is the sampling interval of the sensor. $Q \in \mathcal{R}^{6\times 6}$ represents the process noise covariance matrix and is modeled as $Q = \text{diag}(n_x, n_y, n_{v_x}, n_{v_y}, n_{a_x}, n_{a_y})$, where the coefficients are tuned by trial-and-error. The values are based their on the units of the state vector ($m$, $m/s$, $m/s^2$) and their uncertainty level. Since we can directly observe center coordinates, we have higher certainty for $n_x$ and $n_y$ than others. $\hat{x}_{k-1|k-1}$ represents the a posteriori state estimate at time $k-1$ given observations up to and including at time k. $P_{k-1|k-1} \in \mathcal{R}^{6\times 6}$ represents the a posteriori error covariance matrix, and the initial value $\mathbf{P}_{0|0}$ is estimated empirically by computing the error covariance matrix on a sample of 100 tracking objects. 

Based on the center position prediction, we then estimate the bounding box at frame $k$ and ask the human annotator to adjust it. The adjusted bounding box provides us an observation of the new center coordinates $\mathbf{z}_{k} = [p_{x,k}, p_{y,k}]^T$, which we use to update the Kalman filter as 
\begin{equation*}
    \begin{gathered}
    \mathbf{\tilde{y}}_{k} =  \mathbf{z}_{k} - H\mathbf{\hat{x}}_{k|k-1}, \\
    K_k = P_{k|k-1}H^{T} (R + H P_{k|k-1}H^T)^{-1}, \\
    \mathbf{\hat{x}}_{k|k} = \mathbf{\hat{x}}_{k|k-1} + K_k \mathbf{\tilde{y}}_{k},  \\
    P_{k|k} = (I - K_k H)P_{k|k-1}(I - K_k H)^T + K_k R K_k^T,
    \end{gathered}
\end{equation*}
where $H \in \mathcal{R}^{2\times 6}$ is the observation model. Since we only observe the center position through, we define
\[
H = \begin{bmatrix} 
1 & 0 & 0 & 0 & 0 & 0\\
0 & 1 & 0 & 0 & 0 & 0\\
\end{bmatrix}.
\]
$R \in \mathcal{R}^{2\times 2}$ is the covariance of the observation noise and is defined as $R = \text{diag}(\Delta x, \Delta y)$ where $\Delta x$ and $\Delta y$ are the resolution of the LiDAR scanner in the x and y direction. We iteratively apply Kalman filtering from the first frame to the end, as shown in Fig. \ref{fig:tracking}.

\section{Experiments}
\label{sec:experiment}
\subsection{Experiment Setup}
Providing precise quantitative measurements of productivity improvements using human subjects is very time consuming and complicated \cite{lazar2017research}. Nevertheless, while we feel that the productivity advantages of features such as "one-click annotation" are obvious, we wanted to provide some estimate of the productivity improvements that LATTE provided. So, to estimate productivity we simply measured the time and operation count used by volunteers using LATTE to annotate LiDAR point cloud data from the KITTI dataset \cite{KITTI}. The KITTI dataset involves eight object categories and includes 3D Velodyne point cloud data, accompanying color images, GPS/IMU data, 3D object tracklet labels, and camera-to-Velodyne calibration data. We randomly selected 30 sequences of LiDAR data, where each sequence contains five frames. This test benchmark contains a total amount of 1,116 instances. More detailed analysis of object statistics can be found in Fig. \ref{fig:stats}. 

We asked nine volunteers to annotate these frames using LATTE with all three features (sensor fusion, one-click annotation, and tracking). We divide the data and volunteers such that each feature is evaluated on the entire dataset. In other words, each frame is annotated using each feature so that every feature is evaluated on the same frames. This is to ensure that we are testing each feature on the same data. 

The annotators were asked to draw bounding boxes for instances that they feel confident, for example, instances of vehicles where at least two complete edges are visible. Instances that far away tend to be sparse or occluded and are therefore not annotated. We only evaluate objects whose bounding box intersects with a ground truth bounding box. To form a baseline for comparison, we asked volunteers to draw top-view 2D bounding boxes on the test LiDAR point cloud without using LATTE's advanced features. To further evaluate the efficacy of each component, we also asked volunteers to use only one of the three features for annotation. Each volunteer annotates a sequence of 5 frames with one feature a time. We also asked the volunteers to annotate with a fully-featured version of the tool. In order to eliminate the case where annotation efficiency is improved solely due to the fact that the annotator has seen the frame before, each frame is seen only once by each volunteer. Therefore we do not falsely attribute an improvement in efficiency to the feature we are testing.

\subsection{Metrics}
To evaluate the accuracy of annotations, we used our own ground truth instead of using the bounding boxes from KITTI dataset. This is because bounding boxes provided by KITTI do not include all instances in a scene, particularly the ones that are behind the car. Therefore, we asked an experienced human annotator to provide high-quality annotations as ground truth. We measure the \textbf{intersection-over-union (IoU)} between an annotated bounding box and the ground truth as the accuracy metric per instance, and we report the IoU averaged over all the instances annotated by all of the volunteers. Note that comparing our ground truth with KITTI's bounding boxes, we see 86.0\% average IoU. 

\begin{figure}[h]
    \centering
    \includegraphics[width=\linewidth]{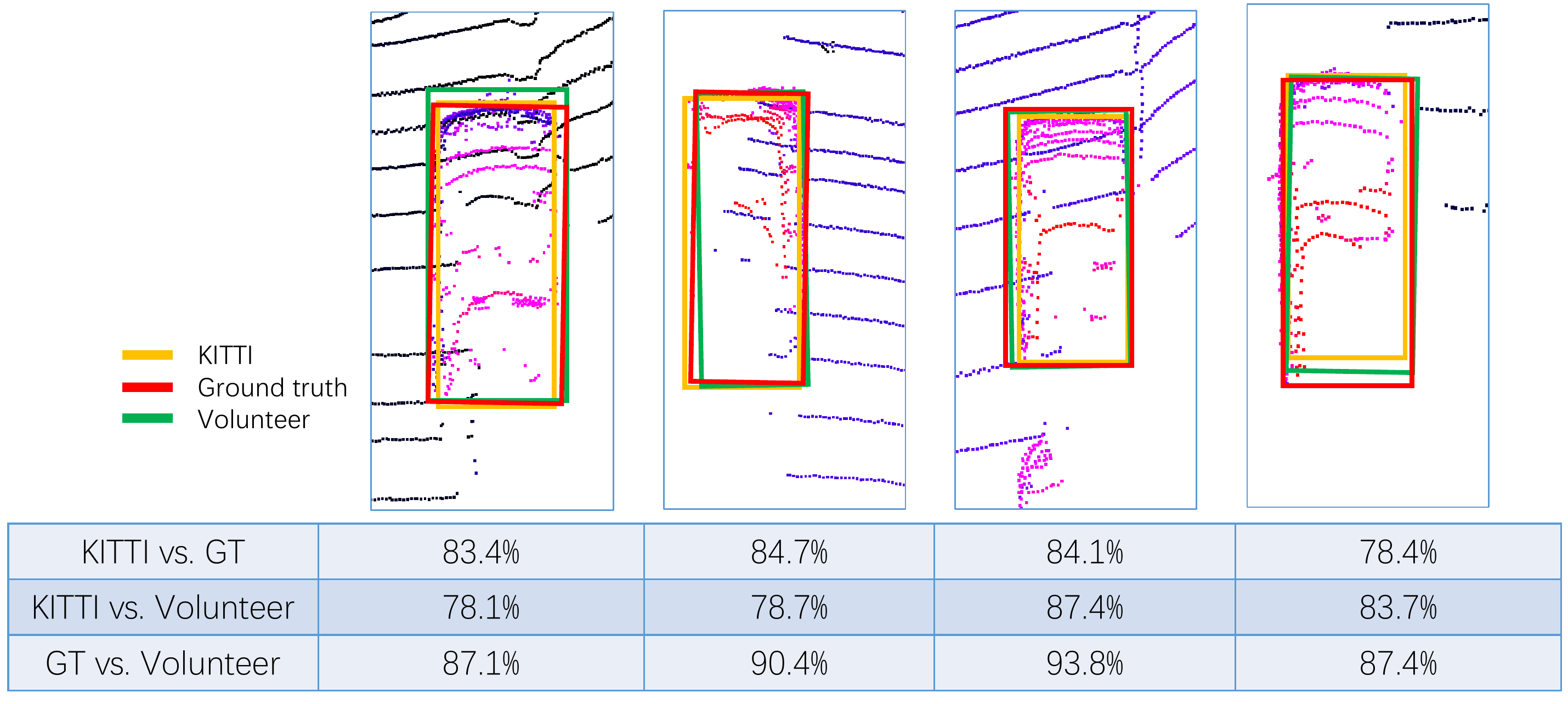}
    \caption[Visualization of bounding box annotations.]{Visualization of bounding box annotations from our volunteers, our ground truth, and KITTI. The IoU between each pair is listed. }
    \label{fig:iou-viz}
\end{figure}

To better understand the typical agreement between two human annotators, we ask different volunteers to use LATTE to label the same frames and instances and compute the pair-wise IoU agreement per instance. Among 452 pairs of annotations on 132 instances, the average IoU is 84.5\% with a standard deviation of 8.74\%. This serves as a reference for considering other IoU results. In addition, we select a few samples of bounding box annotations and compare them with our ground truth and KITTI's annotation in Fig. \ref{fig:iou-viz}. As we can see, the IoU between each pair ranges from 78.1\% to 93.8\%, but the bounding boxes are very similar to each other.  

Another factor of accuracy is the correct identification of objects of interest. To evaluate this, we measure the \textbf{instance-level precision and recall}. We compare all the annotated bounding boxes and ground truth bounding boxes. We pair each ground truth box to an annotated box with the highest overlap. Next, if a ground truth box has over 50\% of overlap with an annotated box, then we the annotated box as a true positive, or TP. If the ground truth is not matched with any other annotation, it is a false negative, or FN. If an annotated box is not matched with any ground truth boxes, it is a false positive, or FP. Then, the instance level precision is defined as 
\begin{equation*}
    \frac{TP}{FP + TP}
\end{equation*}
while the recall is defined as 
\begin{equation*}
    \frac{TP}{TP + FN}.
\end{equation*}
This is different from the IoU above since IoU is counted on objects that are both annotated by a volunteer and the ground truth. We report the IoU result in Table \ref{table:efficiency_results} and the precision and recall in Table \ref{tab:prec-recall}. 

To evaluate efficiency, we record the \textbf{time} spent on annotation per instance. In addition, we also measure the \textbf{number of annotation operations}, which is defined as bounding box adjustments (resizing, rotation, translation) and assigning classes. The efficiency result is reported in Table \ref{table:efficiency_results}. 

\subsection{Results}
\begin{table}[h]
\centering
\begin{threeparttable}
\caption{Accuracy and efficiency of LATTE}
\label{table:efficiency_results}
\begin{tabular}{c|c|c|c|c}
\hline
Method               & IoU (\%) & \begin{tabular}[c]{@{}c@{}}IoU(\%)\\ w/ KITTI\end{tabular} & Time (s) & \#ops \\ \hline
Ground Truth         & 100.0    & 86.0                                                       & -        & -     \\ \hline
Baseline             & 85.5     & 82.3                                                       & 9.51      & 3.76   \\
Sensor fusion        & 86.3     & 82.5                                                       & 3.88      & 2.88   \\
One-click annotation & 86.2     & 82.9                                                       & 2.55      & 1.29   \\
Tracking             & 86.4     & 83.5                                                       & 2.41      & 1.53   \\
Full features        & 87.5     & 84.7                                                       & 1.53      & 1.02   \\ \hline
\end{tabular}
\begin{tablenotes}
\item The accuracy and efficiency of LATTE are listed. The  ``IoU'' column shows the average IoU of annotation vs. our ground truth. The ``IoU w/ KITTI'' column shows the average IoU of annotations vs. the KITTI ground truth. The ``Time'' column shows the average time spent on annotating one instance. The ``\#ops'' column shows the average number of operations spent on annotating one instance. As a reference for the IoU result, the average pair-wise IoU from different annotators is 84.5\% with a standard deviation of 8.74\%.  
\end{tablenotes}
\end{threeparttable}
\end{table}

\begin{table}[h]
\centering
\caption{Instance-level precision \& recall of LATTE}
\label{tab:prec-recall}
\begin{tabular}{c|c|c}
\hline
Method & Precision (\%) & Recall (\%) \\ \hline
Baseline & 69.9 & 82.9 \\
Sensor fusion & 83.9 & 85.0 \\
One-click annotation & 78.8 & 84.8 \\
Tracking & 91.8 & 85.2 \\
Full features & 93.5 & 85.1 \\ 
\hline
\end{tabular}
\end{table}

\textbf{Baseline}: Since there are no other open-source tools with similar functionality, we compare LATTE to a stripped-down version with all of the new features removed. The volunteers are asked to manually annotate all instances by drawing top-view bounding boxes, as illustrated in Fig. \ref{fig:ops-comparison}. From Table \ref{table:efficiency_results}, we can see that it takes an average of 9.51s, and 3.76 operations to annotate one instance. Despite the longer time, the label quality is the poorest among all variations. The IoU with the ground truth is just 85.5\%, the precision is 69.9\%, and the recall is 82.9\%.

\textbf{Full features:} To test LATTE with all of the features, we ask the volunteers to use the following workflow. For the first frame of a sequence, we use sensor fusion to highlight objects of interest and ask human annotators to use one-click annotation to draw annotations and make necessary adjustments. After volunteers are confident with the first frame, they can move on to the next frame, where tracking will generate bounding box proposals for volunteers to verify and adjust. The full features show a 6.2 times speed-up in annotation time and 3.7 times reduction in the number of operations while achieving higher IoU at 87.5\% over the baseline's 85.5\%, higher recall (85.1\% vs. 82.9\%) and significantly higher precision (93.5\% vs. 69.9\%). We visualize some samples of annotations and compare them with the ground truth and KITTI dataset's annotations in Fig. \ref{fig:iou-viz}. 

\textbf{Ablation study}: To study the effectiveness of each of the proposed features, we ask volunteers to use a variation of LATTE with only one feature enabled. When volunteers are only allowed to use the \textbf{sensor fusion} feature, we are able to achieve a 2.4 times speed-up in annotation time compared to the baseline while achieving higher IoU (+0.8\%), recall (+2.1\%), and significantly higher precision (+14.0\%). In addition, the number of operations per instance on average reduces by 0.9, nearly an entire operation. This supports our claim that sensor fusion helps human annotators to better recognize objects from point clouds. With only \textbf{one-click annotation}, our method shows a 3.8 times speed-up in annotation time and 2.9x reduction in the number of annotation operations while achieving higher IoU, precision, and recall. With only \textbf{tracking},  we show a speed-up of 4.74x while achieving better annotation agreement than the baseline. This is largely due to the fact that tracking saved redundant annotations on sequential frames.

\section{Conclusion}
Efficiently annotating LiDAR point clouds at scale is crucial for the development of LiDAR-based detection and autonomous driving. However, annotating LiDAR point clouds is difficult due to the challenges of low resolution, complex annotating operations, and sequential correlation. To solve these problems, we propose LATTE, an open-sourced LiDAR annotation tool that features sensor-fusion, one-click annotation, and tracking. Based on our experiments we estimate that LATTE achieves a 6.2x speedup compared with baseline annotation tools and delivers better label quality with 23.6\% and 2.2\% higher instance-level precision and recall, and 2.0\% higher bounding box IoU.  

%% file: chap6.tex
\chapter{Data Efficiency: SqueezeSegV2}
\label{chap:sqsgv2}

As discussed in Chapter \ref{chap:latte}, collecting and annotating large amounts of data to train deep neural networks is very difficult, especially for LiDAR point cloud data. Advanced tools such as LATTE significantly reduce the cost of data annotation by 6.2x, from about 10 seconds per frame to 1.5 seconds. However, it is still costly to annotate large scale datasets that typically consist of millions of frames. This motivates us to explore more aggressive approaches to improve the data efficiency of deep learning. 

A promising approach is to leverage simulated data to train deep neural networks. Today, many simulators and video games such as CARLA\cite{Dosovitskiy17} and GTA-V \cite{yue2018lidar, philip2018free} can synthesize photo-realistic images and depth maps. The advantage of leveraging simulated data to train deep neural networks include: 1) we can easily obtain large amounts of, or even unlimited data through simulation; 2) labels for simulated data are readily available, therefore bypassing the need for manual annotation. However, models trained on simulated data usually do not generalize to the real world, due to the problem of domain shift -- the data distribution of the target domain (real world) is different from that of the source domain (simulation). In this chapter, we discuss our approaches to solve this problem and address the following \textit{key question}:
\begin{quote}
    Can we use simulated data to train deep neural networks and transfer the trained model to the real world?  
\end{quote}

\section{Introduction}
\begin{figure}[!t]
\begin{center}
\centering \includegraphics[width=0.7\linewidth]{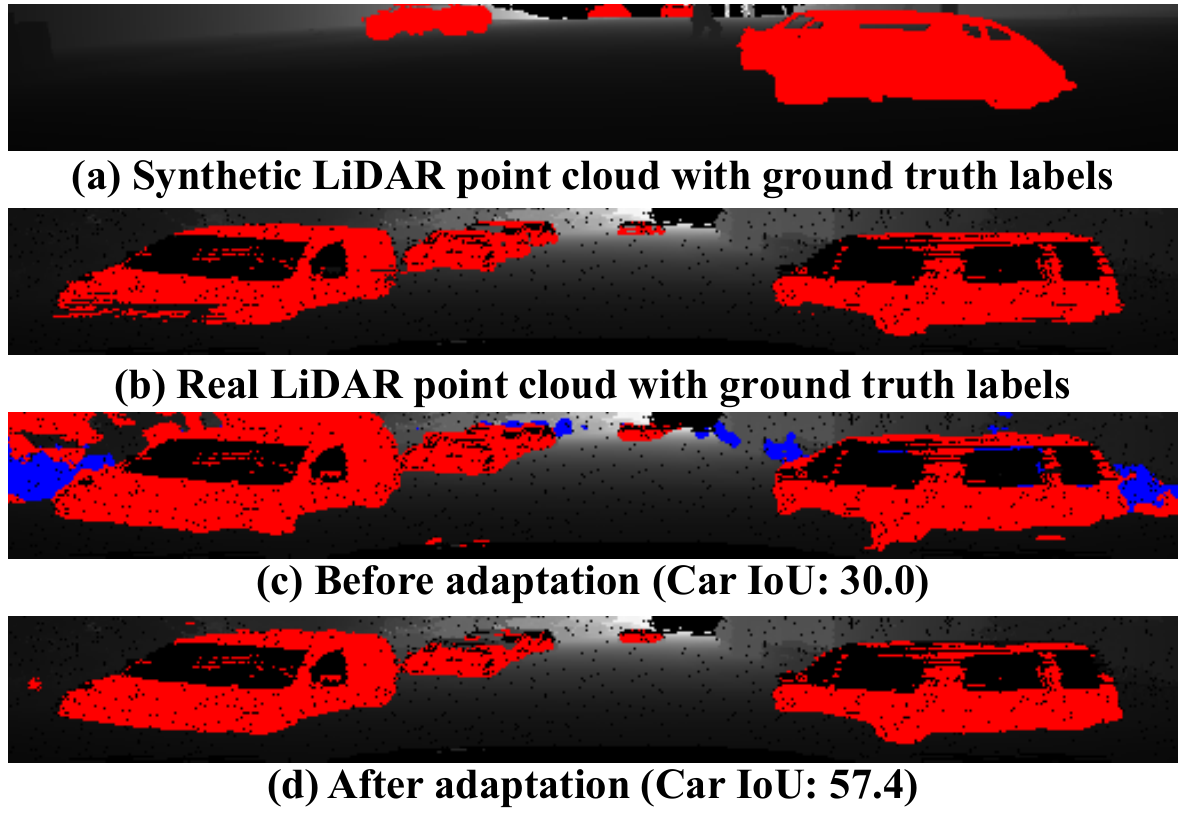}
\caption[An example of \emph{domain shift}]{An example of \emph{domain shift}. The point clouds are projected onto a spherical surface for visualization (cars in red, pedestrians in blue). Our domain adaptation pipeline improves the segmentation from (c) to (d) while trained on synthetic data.}
\label{fig:DomainShift}
\end{center}
\end{figure}

In Chapter \ref{chap:sqsg}, we introduced SqueezeSeg for LiDAR point cloud segmentation. SqueezeSeg is extremely efficient -- the fastest version achieves an inference speed of over 100 frames per second. However, SqueezeSeg still has several limitations: first, its accuracy still needs to be improved to be practically useful. A critical reason for accuracy degradation is \textit{dropout noise} -- missing points from the sensed point cloud caused by limited sensing range, mirror diffusion of the sensing laser, or jitter in incident angles. A visualization of the dropout noise is provided in Figure \ref{fig:DomainShift} (b). Such dropout noise can corrupt the output of convolution filters and therefore reduces overall accuracy. More importantly, training deep learning models requires tens of thousands of labeled point clouds; however, collecting and annotating such data is even more time consuming and expensive than collecting comparable data from cameras. GTA-V is used to synthesize LiDAR point cloud as an extra source of training data~\cite{wu2017squeezeseg}; however, this approach suffers from the domain shift problem \cite{torralba2011unbiased} --  models trained on synthetic data fail catastrophically on the real data, as shown in Fig.~\ref{fig:DomainShift}. Domain shift comes from different sources, but the absence of dropout noise and intensity signals in GTA-V are two important factors. Simulating realistic dropout noise and intensity is very difficult, as it requires sophisticated modeling of both the LiDAR device and the environment, while both contain a lot of non-deterministic factors. As such, the LiDAR point clouds generated by GTA-V do not contain dropout noise and intensity signals. The comparison of simulated data and real data is shown in Fig.~\ref{fig:DomainShift} (a), (b).

In this chapter, we focus on addressing the challenges above. First, to improve the accuracy, we mitigate the impact of dropout noise by proposing the Context Aggregation Module (CAM), a novel CNN module that aggregates contextual information from a larger receptive field and improves the robustness of the network to dropout noise. Adding CAM to the early layers of SqueezeSegV2 not only significantly improves its performance when trained on real data, but also effectively reduces the domain gap, boosting the network's real-world test accuracy when trained on synthetic data. In addition to CAM, we adopt several improvements to SqueezeSeg, including using focal loss \cite{lin2018focal}, batch normalization \cite{ioffe2015batch}, and LiDAR mask as an input channel. As shown in section \ref{ssec:Results_SqueezeSegV2}, these improvements together boosted the accuracy of SqueezeSegV2 by 6.0\% - 8.6\% in all categories on the converted KITTI dataset \cite{wu2017squeezeseg}. A qualitative visualization of such improvement is provided in Figure \ref{fig:SqueezeSegV1V2}. 

Second, to better utilize synthetic data for training the model, we propose a domain adaptation training pipeline that contains the following steps: first, before training, we render intensity channels in synthetic data through \textit{learned intensity rendering}. We train a neural network that takes point coordinates as input and predicts intensity values. This rendering network can be trained in a "self-supervised" fashion on unlabeled real data. After training the network, we feed the synthetic data into the network and render the intensity channel. Second, we use the synthetic data augmented with rendered intensity to train the network. Meanwhile, we follow \cite{morerio2018minimal} and use \textit{geodesic correlation alignment} to align the batch statistics between real data and synthetic data. 3) After training, we propose \textit{progressive domain calibration} to further reduce the gap between the target domain and the trained network. Experiments show that the above domain-adaptation training pipeline doubles the test accuracy on real data from 29.0\% to 57.4\% while training on synthetic data. A qualitative visualization of this improvement is provided in Figure \ref{fig:DomainShift}, where the adapted model's performance is obviously better than the original model's. 

\section{Related work}
\label{sec:RelatedWork}
\textbf{3D LiDAR Point Cloud Segmentation} aims to recognize objects from point clouds by predicting point-wise labels. Non-deep-learning methods~\cite{moosmann2009segmentation, douillard2011segmentation, zermas2017fast} usually involve several stages such as ground removal, clustering, and classification. SqueezeSeg~\cite{wu2017squeezeseg} is one early work that applies deep learning to this problem. Piewak et al.~\cite{piewak2018boosting} adopted a similar problem formulation and pipeline to SqueezeSeg and proposed a new network architecture called LiLaNet. They created a dataset by utilizing image-based semantic segmentation to generate labels for the LiDAR point cloud. However, the dataset was not released, so we were not able to conduct a direct comparison to their work. Another category of methods is based on PointNet~\cite{qi2017pointnet, qi2017pointnet++}, which treats a point cloud as an unordered set of 3D points. This is effective with 3D perception problems such as classification and segmentation. Limited by its computational complexity; however, PointNet is mainly used to process indoor scenes where the number of points is limited. Frustum-PointNet~\cite{qi2017frustum} is proposed for out-door object detection, but it relies on image object detection to first locate object clusters and feeds the cluster, instead of the whole point cloud, to the PointNet.

\textbf{Unsupervised Domain Adaptation (UDA)} aims to adapt the models from one labeled source domain to another unlabeled target domain. Recent UDA methods have focused on transferring deep neural network representations~\cite{patel2015visual,csurka2017domain}. Typically, deep UDA methods employ a conjoined architecture with two streams to represent the models for the source and target domains, respectively. In addition to the task-related loss computed from the labeled source data, deep UDA models are usually trained jointly with another loss, such as a discrepancy loss~\cite{long2015learning,sun2017correlation,zhuo2017deep,zhang2017curriculum,morerio2018minimal}, adversarial loss~\cite{liu2016coupled,ganin2016domain,tzeng2017adversarial,shrivastava2017learning,bousmalis2017unsupervised,hoffman2018cycada}, label distribution loss~\cite{zhang2017curriculum} or reconstruction loss~\cite{ghifary2015domain,ghifary2016deep}.

The most relevant work is the exploration of synthetic data~\cite{shrivastava2017learning,zhang2017curriculum,hoffman2018cycada}. By enforcing a self-regularization loss, Shrivastava et al.~\cite{shrivastava2017learning} proposed SimGAN to improve the realism of synthetic data using unlabeled real data. Another category of relevant work employs a discrepancy loss~\cite{long2015learning,sun2017correlation,zhuo2017deep,morerio2018minimal}, which explicitly measures the discrepancy between the source and target domains on corresponding activation layers of the two network streams. Instead of working on 2D images, we try to adapt synthetic 3D LiDAR point clouds by a novel adaptation pipeline.

\textbf{Simulation} has recently been used for creating large-scale ground truth data for training purposes. Richter et al.~\cite{playingfordata} provided a method to extract semantic segmentation for the synthesized in-game images. In \cite{drivinginthematrix}, the same game engine is used to extract ground truth 2D bounding boxes for objects in the image. Yue et al.~\cite{yue2018lidar} proposed a framework to generate synthetic LiDAR point clouds. Richter et al.~\cite{playingforbenchmarks} and Kr\"ahenb\"uhl~\cite{philip2018free} extracted more types of information from video games.

\section{Improving the model structure}
\label{sec:Better}

\begin{figure*}[!t]
\begin{center}
\centering \includegraphics[width=\linewidth]{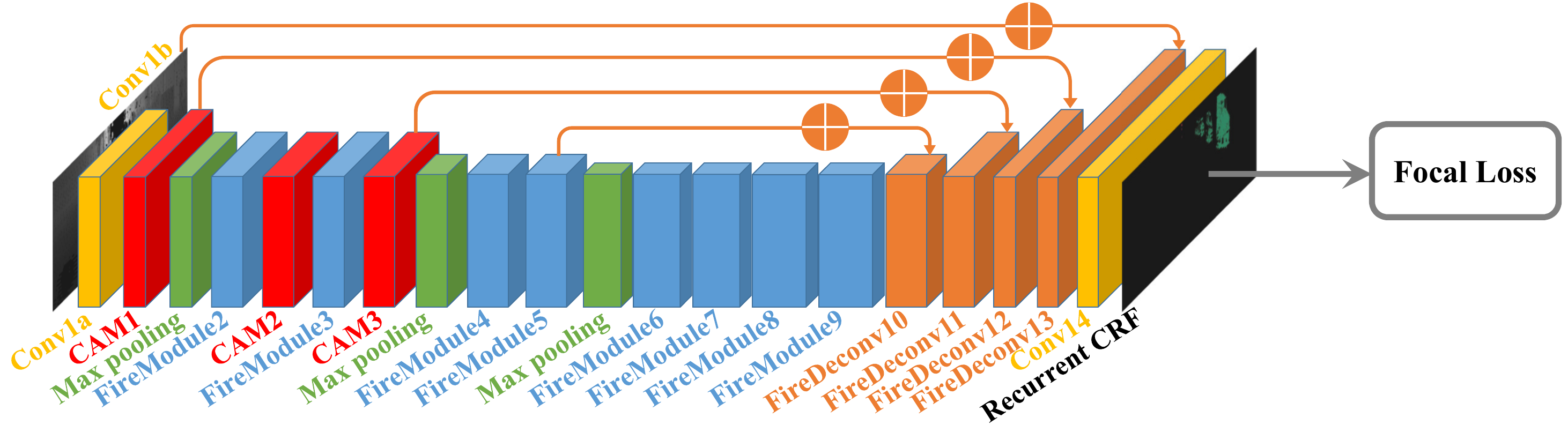}
\caption[The network structure of the proposed SqueezeSegV2.]{The network structure of the proposed SqueezeSegV2 model for road-object segmentation from 3D LiDAR point clouds.}
\label{fig:SqueezeSegV2Framework}
\end{center}
\end{figure*}

We propose SqueezeSegV2, by improving upon the base SqueezeSeg model, adding Context Aggregation Module (CAM), adding LiDAR mask as an input channel, using batch normalization \cite{ioffe2015batch}, and employing the focal loss \cite{lin2018focal}. The network structure of SqueezeSegV2 is shown in Fig.~\ref{fig:SqueezeSegV2Framework}.

\subsection{Context Aggregation Module}
\label{ssec:CAM}
\begin{figure}[!t]
\begin{center}
\centering \includegraphics[width=0.8\linewidth]{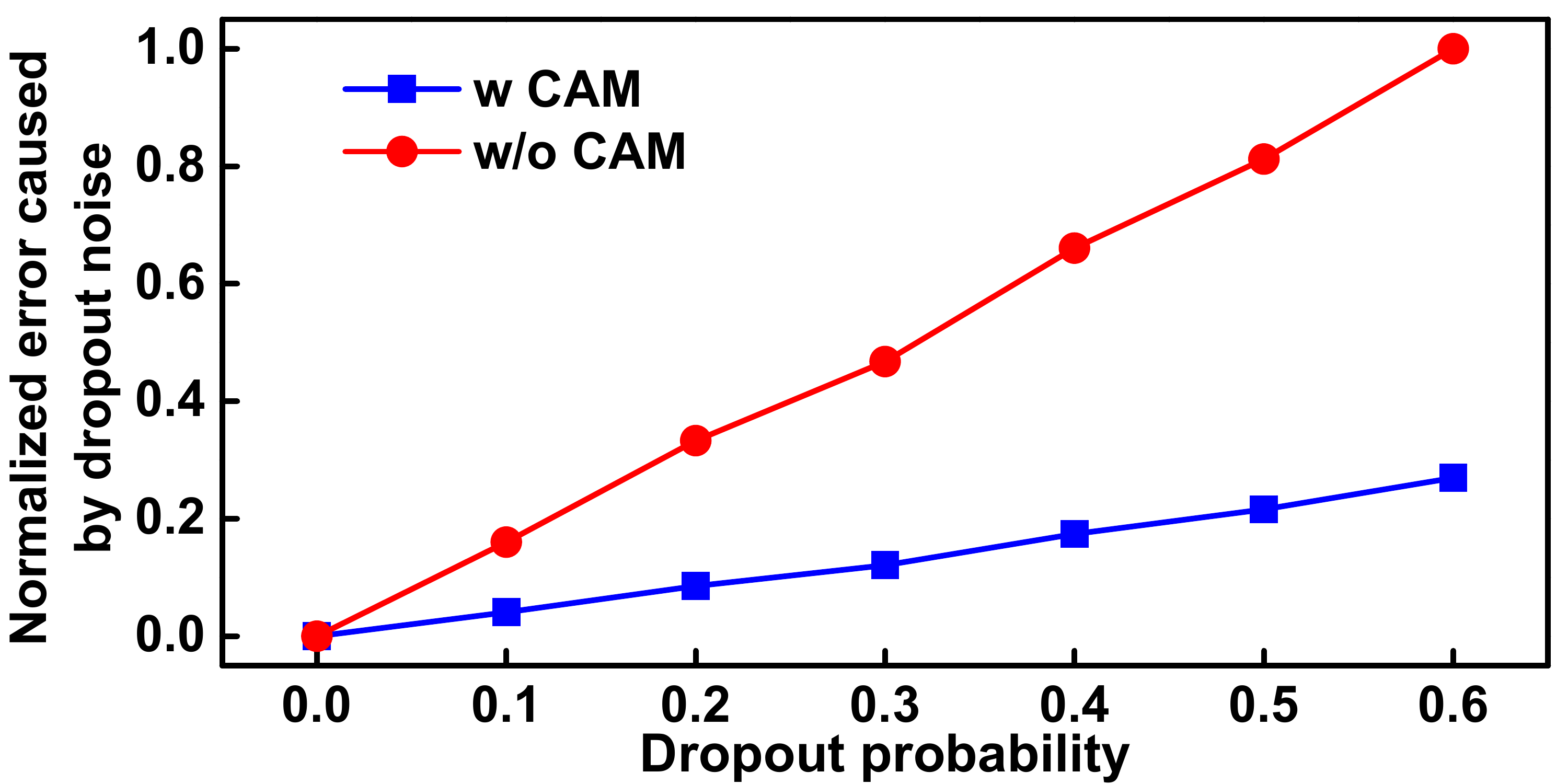}
\caption[A numerical experiment for CAM.]{We feed a random tensor to a convolutional filter, one with CAM before a $3\times3$ convolution filter and the other one without CAM. We randomly add dropout noise to the input and measure the output errors. As we increase the dropout probability, the error also increases. For all dropout probabilities, adding CAM improve the robustness towards the dropout noise and therefore, the error is always smaller.}
\label{fig:CAM_Comparison}
\end{center}
\end{figure}

LiDAR point cloud data contains many missing points, which we refer to as dropout noise, as shown in Fig.~\ref{fig:DomainShift}(b). As can be seen, in many pixels in the projected LiDAR depth map, the LiDAR measurements are missing. This usually happen on cars' window areas, end of the roads, cars that are far away, and so on. Dropout noise is mainly caused by 1) limited sensor range, 2) mirror reflection (instead of diffusion reflection) of sensing lasers on smooth surfaces, and 3) jitter of the incident angle. Dropout noise has a significant impact on SqueezeSeg, especially in the early layers of a network. At early layers where the receptive field of the convolution filter is very small, missing points in a small neighborhood can corrupt the output of the filter significantly. To illustrate this, we conduct a simple numerical experiment, where we randomly sample an input tensor and feed it into a $3\times3$ convolution filter. We randomly drop out some pixels from the input tensor, and as shown in Fig.~\ref{fig:CAM_Comparison}, the errors of the corrupted output increases with the dropout probability. 

\begin{figure}[!t]
\begin{center}
\centering \includegraphics[width=1.0\linewidth]{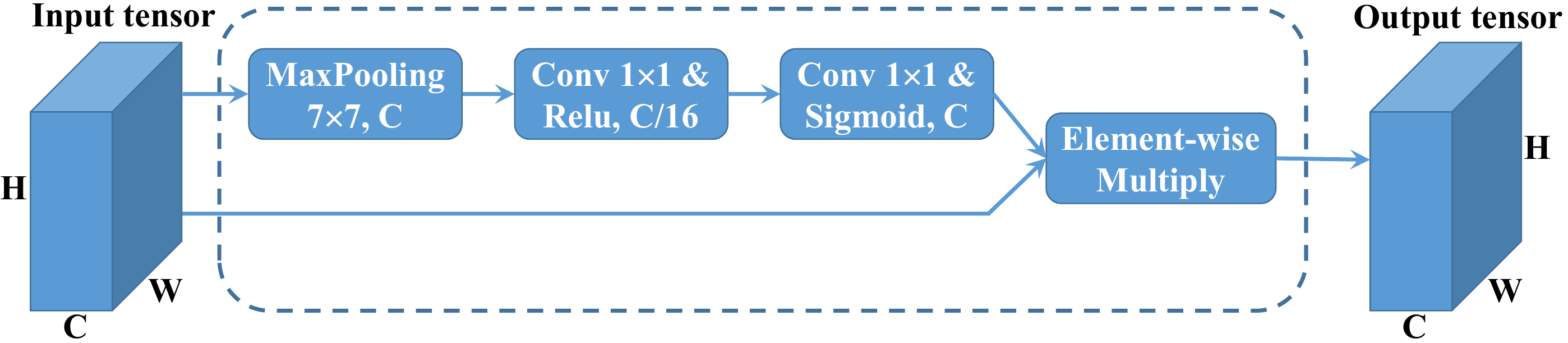}
\caption[Structures of the Context Aggregation Module.]{Structures of the Context Aggregation Module. The module starts with a max-layer with a kernel size of 7. Next, two 1x1 convolution layers are used, followed by a sigmoid activation at the output. Following \cite{hu2018squeeze}, we perform an element-wise multiplication to combine the output of CAM and its input.}
\label{fig:CAM}
\end{center}
\end{figure}

This problem not only impacts SqueezeSeg when trained on real data but also leads to a serious domain gap between synthetic data and real data, since simulating realistic dropout noise from the same distribution is very difficult.

To solve this problem, we propose a novel Context Aggregation Module (CAM) to reduce the sensitivity to dropout noise. As shown in Fig.~\ref{fig:CAM}, CAM starts with a max-pooling with a relatively large kernel size. The max-pooling aggregates contextual information around a pixel with a much larger receptive field, and it is less sensitive to missing data within its receptive field. Also, max-pooling can be computed efficiently even with a large kernel size. Two cascaded convolution layers then follow the max-pooling layer with a ReLU activation in between. Following~\cite{hu2018squeeze}, we use the \textit{sigmoid} function to normalize the output of the module and use an element-wise multiplication to combine the output with the input. As shown in Fig.~\ref{fig:CAM_Comparison}, the proposed module is much less sensitive to dropout noise -- with the same corrupted input data, the error is significantly reduced.

In SqueezeSegV2, we insert CAM after the output of the first three modules (1 convolution layer and 2 FireModules), where the receptive fields of the filters are small. As can be seen in later experiments, CAM 1) improves the accuracy when trained on real data, and 2) reduces the domain gap while trained on synthetic data and testing on real data.

\subsection{Focal Loss}
\label{ssec:Focal} 
LiDAR point clouds have a very imbalanced distribution of point categories -- there are many more background points such as roads, sky, and trees, than there are foreground objects such as cars, pedestrians. This imbalanced distribution makes the model focus more on easy-to-classify background points that contribute no useful learning signals, with the foreground objects not being adequately addressed during training.

To address this problem, we replace the original cross entropy loss from SqueezeSeg~\cite{wu2017squeezeseg} with a focal loss~\cite{lin2018focal}. The focal loss modulates the loss contribution from different pixels and focuses on hard examples. For a given pixel label $t$, and the predicted probability of $p_t$, focal loss~\cite{lin2018focal} adds a modulating factor $(1-p_t)^{\gamma}$ to the cross-entropy loss. The focal loss for that pixel is thus
\begin{equation}
FL(p_t) =-(1-p_t)^{\gamma}\log{(p_t)}
\end{equation}

When a pixel is misclassified and $p_t$ is small, the
modulating factor is near 1, and the loss is unaffected. As
$p_t \rightarrow1$, the factor goes to 0, and the loss for well-classified
pixels is down-weighted. The focusing parameter $\gamma$ smoothly adjusts the rate such that well-classified examples are down-weighted.
When $\gamma = 0$, the Focal Loss is equivalent to the Cross-Entropy Loss. As $\gamma$ increases, the effect of the modulating factor is likewise increased. We choose $\gamma$ to be $2$ in our experiments.

\subsection{Other Improvements}
\textbf{LiDAR Mask}: Besides the original (x, y, z, intensity, depth) channels, we add one more channel -- a binary mask indicating if each pixel is missing or existing. As we can see from Table~\ref{tab:SqueezeSegV2}, the addition of the mask channel significantly improves segmentation accuracy for cyclists.

\textbf{Batch Normalization}: Unlike SqueezeSeg~\cite{wu2017squeezeseg}, we also add batch normalization (BN)~\cite{ioffe2015batch} after every convolution layer. The BN layer is designed to alleviate the issue of internal covariate shift -- a common problem for training a deep neural network. We observe an improvement in car segmentation after using BN layers in Table~\ref{tab:SqueezeSegV2}.

\section{Domain adaptation training}
\label{sec:Stronger}
\begin{figure}[!t]
\begin{center}
    \begin{subfigure}[b]{0.8\textwidth}
        \centering\includegraphics[width=\linewidth]{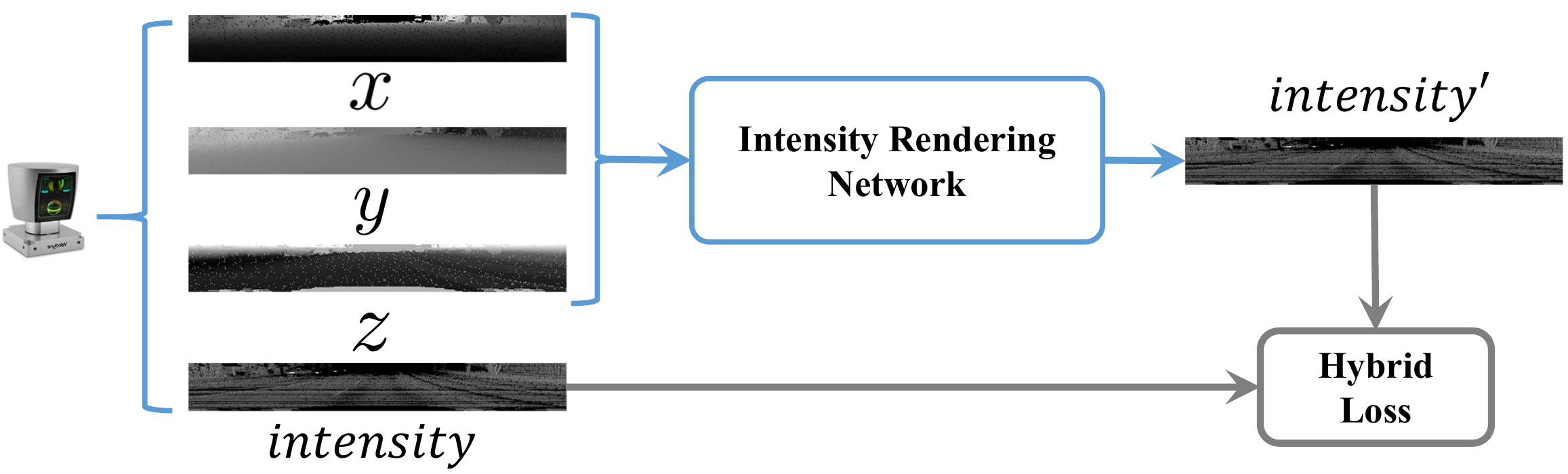}
        \caption{Pre-training: Learned Intensity Rendering}
        \label{fig:LIR}
    \end{subfigure} \hfill
    
    \begin{subfigure}[b]{0.8\textwidth}
        \centering\includegraphics[width=\linewidth]{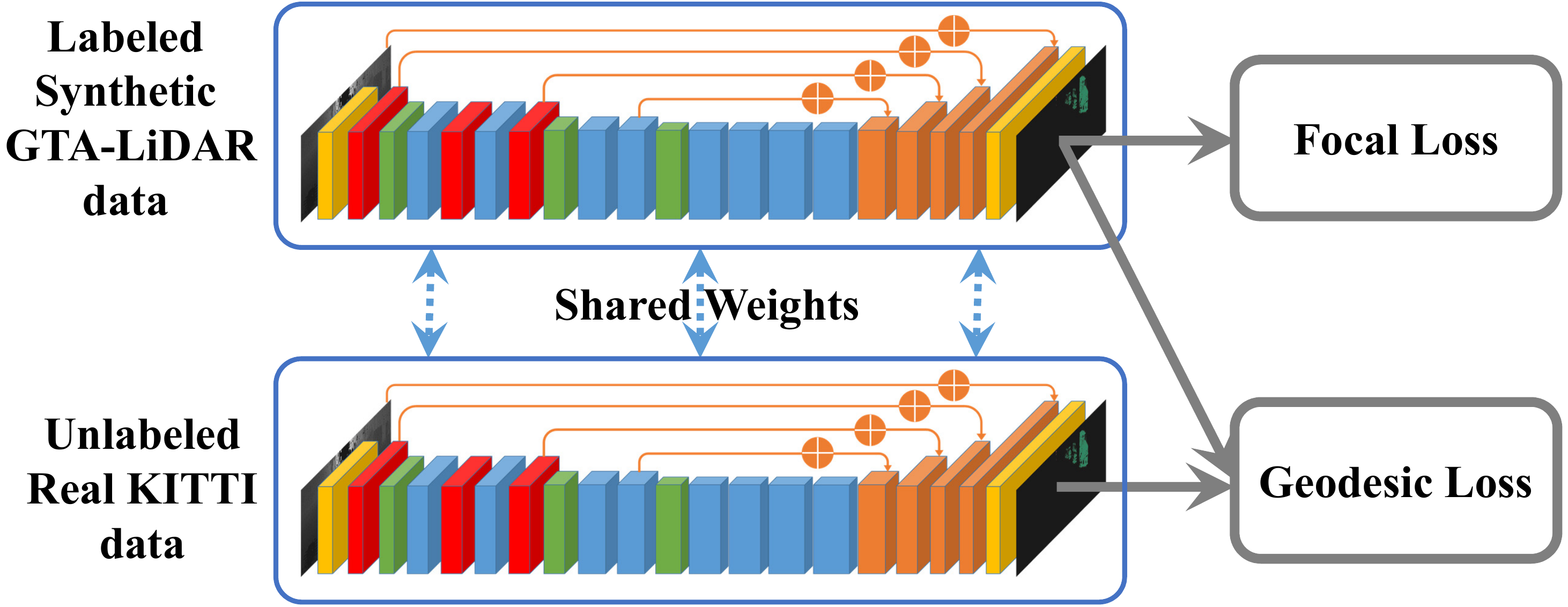}
        \caption{Training: Geodesic Correlation Alignment}
        \label{fig:GCA}
    \end{subfigure} \hfill
    
        \begin{subfigure}[b]{0.8\textwidth}
        \centering\includegraphics[width=\linewidth]{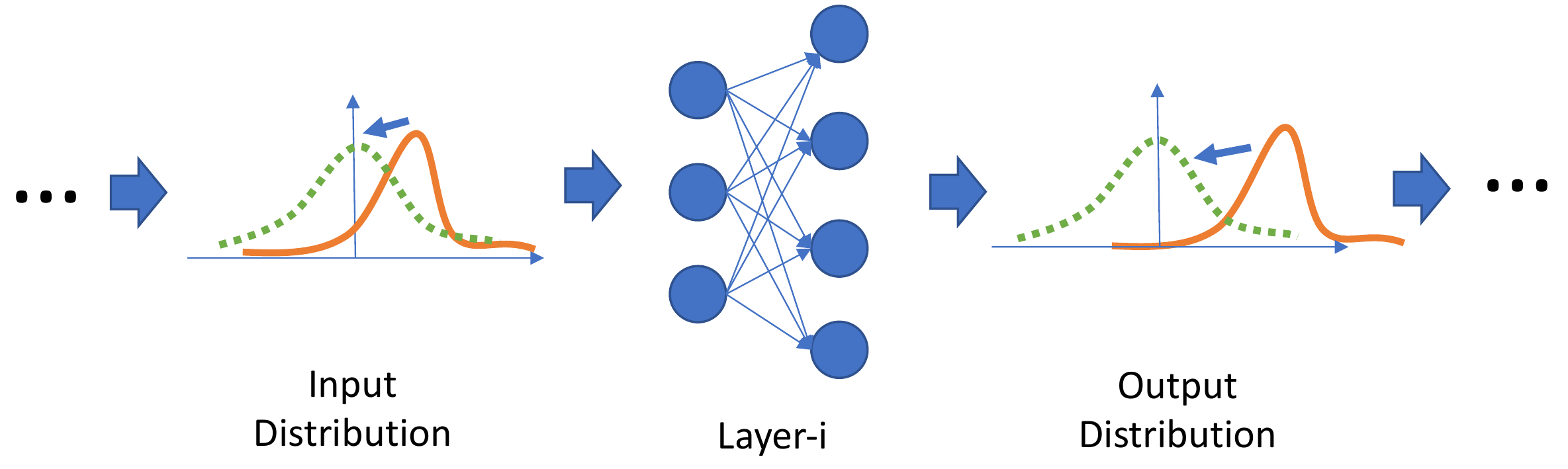}
        \caption{Post-training: Progressive Domain Calibration}
        \label{fig:PDC}
    \end{subfigure} \hfill
\caption[The domain adaptation framework for SqueezeSegV2.]{The framework of the proposed unsupervised domain adaptation method for road-object segmentation from the synthetic GTA-LiDAR dataset to the real-world KITTI dataset.}
\label{fig:DAFramework}
\end{center}
\end{figure}

In this section, we introduce our unsupervised domain adaptation pipeline that trains SqueezeSegV2 on synthetic data and improves its performance on real data. We construct a large-scale 3D LiDAR point cloud dataset, GTA-LiDAR, with 100,000 LiDAR scans simulated on GTA-V. To deal with the \emph{domain shift} problem, we employ three strategies: learned intensity rendering, geodesic correlation alignment, and progressive domain calibration, as shown in Fig.~\ref{fig:DAFramework}.

\subsection{The GTA-LiDAR Dataset}
\label{ssec:GTA-LiDAR}
We synthesize 100,000 LiDAR point clouds in GTA-V to train SqueezeSegV2. We use the framework in \cite{philip2018free} to generate depth semantic segmentation maps, and use the method in \cite{yue2018lidar} to do Image-LiDAR registration in GTA-V. Following \cite{yue2018lidar}, we collect 100,000 point cloud scans by deploying a virtual car to drive autonomously in the virtual world. GTA-V provides a wide variety of scenes, car types, traffic conditions, which ensures the diversity of our synthetic data. Each point in the synthetic point cloud contains one label, one distance, and $x, y, z$ coordinates. However, it does not contain intensity, which represents the magnitude of the reflected laser signal. Also, the synthetic data does not contain dropout noise as in the real data. Because of such distribution discrepancies, the model trained on synthetic data fails to transfer to real data.

\subsection{Learned Intensity Rendering}
\label{ssec:Intensity}
\begin{figure*}[!t]
\begin{center}
\centering \includegraphics[width=1.\linewidth]{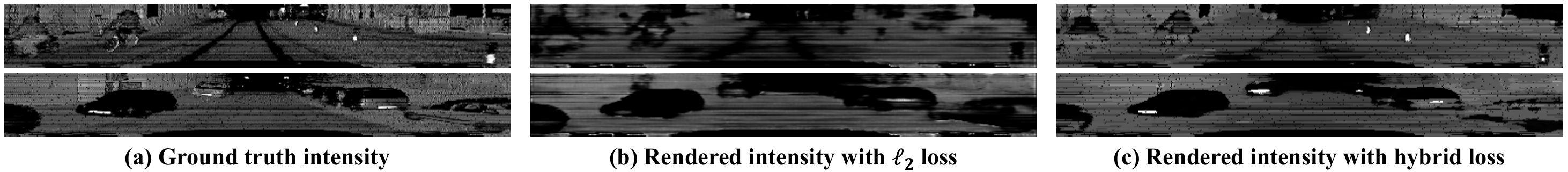}
\caption{Rendered v.s. ground truth intensity in the KITTI dataset.}
\label{fig:PredictedIntensity}
\end{center}
\end{figure*}

The synthetic data only contains $x, y, z, depth$ channels but does not have intensity. As shown in SqueezeSeg~\cite{wu2017squeezeseg}, intensity is an important signal. A visualization of the intensity map is provided in Figure \ref{fig:PredictedIntensity} (a). Intensity not only encodes the structure information of the environment, but also reflects geometry and surface features that are unique to certain objects such as cars. The absence of intensity can lead to severe accuracy loss. Rendering realistic intensity is a non-trivial task, since a multitude of factors that affect intensity, such as surface materials and LiDAR sensitivity, are generally unknown to us.

To solve this problem, we propose a method called \textit{learned intensity rendering}. The idea is to use a network to take the $x, y, z, depth$ channels of the point cloud as input, and predict the intensity. Such rendering network can be trained with unlabeled LiDAR data, which can be easily collected as long as a LiDAR sensor is available. As shown in Fig.~\ref{fig:LIR}, we train the rendering network in a self-supervision fashion, splitting the $x, y, z$ channels as input to the network and the intensity channel as the label. The structure of the rendering network is almost the same as SqueezeSeg, except that the CRF layer is removed. 

The intensity rendering can be seen as a regression problem, where the $\ell_2$ loss is a natural choice. However,  $\ell_2$ fails to capture the multi-modal distribution of the intensity -- given the same input of $x, y, z$, the intensity can differ. To model this property, we designed a hybrid loss function that involves both classification and regression. We divide the intensity into $n=10$ regions, with each region having a reference intensity value. The network first predicts which region the intensity belongs. Once the region is selected, the network further predicts a deviation from the reference intensity. This way, the categorical prediction can capture the multi-modal distribution of the intensity, and the deviation prediction leads to more accurate estimations. We train the rendering network on the KITTI \cite{geiger2012we} dataset with the hybrid loss function and measure its accuracy with mean squared error (MSE). Compared to $\ell_2$ loss, the converged MSE drops significantly by 3X from 0.033 to 0.011. A few rendered results using two different losses are shown in Fig.~\ref{fig:PredictedIntensity}. After training the rendering network, we feed synthetic GTA-LiDAR data into the network to render point-wise intensities. 

\subsection{Geodesic Correlation Alignment}
\label{ssec:LE}
After rendering intensity, we train SqueezeSegV2 on the synthetic data with focal loss. However, due to distribution discrepancies between synthetic data and real data, the trained model usually fails to generalize to real data.

To reduce this domain discrepancy, we adopt \textit{geodesic correlation alignment} during training. As shown in Fig.~\ref{fig:GCA}, at every step of training, we feed in one batch of synthetic data and one batch of real data to the network. We compute the focal loss on the synthetic batch, where labels are available. Meanwhile, we compute the geodesic distance \cite{morerio2018minimal} between the output distributions of two batches. The total loss now contains both the focal loss and the geodesic loss. Where the focal loss focuses on training the network to learn semantics from the point cloud, the geodesic loss penalizes discrepancies between batch statistics from two domains. Note that other distances, such as the Euclidean distance can also be used to align the domain statistics. However, we choose the geodesic distance over the Euclidean distance since it takes into account the manifold’s curvature. More details can be found in \cite{morerio2018minimal}.

We denote the synthetic input data as $X_{sim}$, synthetic labels as $Y_{sim}$ the real input data as $X_{real}$. Our loss function can be computed as
\begin{equation}
 FL(X_{sim}, Y_{sim})  + \lambda \cdot GL(X_{sim}, X_{real}), 
\end{equation}
where $FL$ denotes focal loss between the synthetic label and network prediction, $GL$ denotes the geodesic loss between batch statistics of synthetic and real data. $\lambda$ is a weight coefficient, and we set it to 10 in our experiment. Note that in this step, we only require unlabeled real data, which is much easier to obtain than annotated data as long as a LiDAR sensor is available.

\subsection{Progressive Domain Calibration}
\label{ssec:PBA}

\begin{algorithm}[!t]
{
\small
\KwIn{Unlabeled real data $\mathcal{X}$, model $\mathcal{M}$}
$\mathcal{X}^{(0)} \leftarrow  \mathcal{X}$ \\
\For{layer $l$ in the model $\mathcal{M}$}{
   $\mathcal{X}^{(l)} \leftarrow \mathcal{M}^{(l)}(\mathcal{X}^{(l-1)})$\\
   $\mu^{(l)} \leftarrow \mathbb{E} (\mathcal{X}^{(l)}$), $\sigma^{(l)} \leftarrow \sqrt{Var(\mathcal{X}^{(l)})}$  \\
   Update the BatchNorm parameters of $\mathcal{M}^{(l)}$ \\
   $\mathcal{X}^{(l)} \leftarrow (\mathcal{X}^{(l)} - \mu^{(l)}) / \sigma^{(l)}$  \\
}
\KwOut{Calibrated model $\mathcal{M}$}

}
\small\caption{Progressive Domain Calibration}
\label{Alg:MTSSRLearning}
\end{algorithm}

After training SqueezeSegV2 on synthetic data with geodesic correlation alignment, each layer of the network learns to recognize patterns from its input and extract higher-level features. However, due to the non-linear nature of the network, each layer can only work well if its input is constrained within a certain range. Taking the ReLU function as an example, if somehow its input distribution shifts below 0, the output of the ReLU becomes all zero. Otherwise, if the input shifts towards larger than 0, the ReLU becomes a linear function. For deep learning models with multiple layers, distribution discrepancies from the input data can lead to distribution shift at the output of each layer, which is accumulated or even amplified across the network and eventually leads to serious degradation of performance, as illustrated in Fig. \ref{fig:PDC}.  

To address this problem, we employ a post-training procedure called progressive domain calibration (PDC). The idea is to break the propagation of the distribution shift through each layer with progressive layer-wise calibration. For a network trained on synthetic data, we feed the real data into the network. Starting from the first layer, we compute its output statistics (mean and variance) under the given input, and then re-normalize the output's mean to be 0 and its standard deviation to be 1, as shown in Fig.~\ref{fig:PDC}. Meanwhile, we update the batch normalization parameters (mean and variance) of the layer with the new statistics. We progressively repeat this process for all layers of the network until the last layer. Similar to geodesic correlation alignment, this process only requires unlabeled real data, which is presumably abundant. This algorithm is summarized in Algorithm~\ref{Alg:MTSSRLearning}.
A similar idea was proposed in \cite{li2018adaptive}, but PDC is different since it performs calibration progressively, making sure that the calibrations of earlier layers do not impact those of later layers.

\section{Experiments}
\label{sec:Experiments}
In this section, we introduce the details of our experiments. 
We train and test SqueezeSegV2 on a converted KITTI \cite{geiger2012we} dataset as \cite{wu2017squeezeseg}. To verify the generalization ability, we further train SqueezeSegV2 on the synthetic GTA-LiDAR dataset and test it on the real world KITTI dataset.

\begin{figure*}[!t]
\begin{center}
\vspace{3pt}
\centering \includegraphics[width=\linewidth]{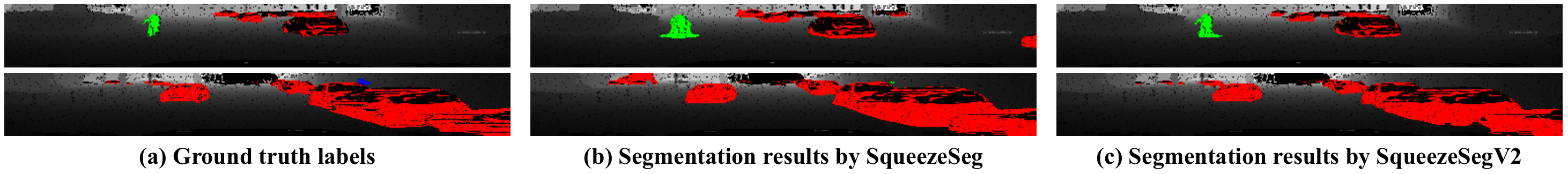}
\caption[Segmentation results of SqueezeSeg and SqueezeSegV2]{Segmentation result comparison between SqueezeSeg~\cite{wu2017squeezeseg} and our SqueezeSegV2 (red: car, green: cyclist). Note that in first row, SqueezeSegV2 produces much more accurate segmentation for the cyclist. In the second row, SqueezeSegV2 avoids a falsely detected car that is far away.}
\label{fig:SqueezeSegV1V2}
\end{center}
\end{figure*}

\begin{figure*}[!t]
\begin{center}
\centering \includegraphics[width=\linewidth]{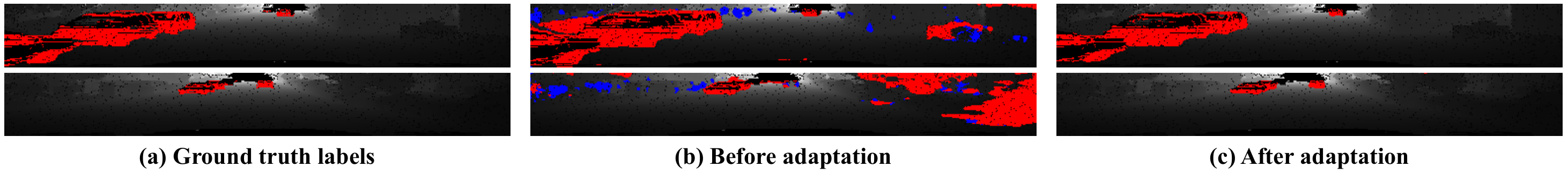}
\caption[Segmentation results before and after domain adaptation.]{Segmentation result comparison before and after domain adaptation (red: car, blue: pedestrian).}
\label{fig:DomainAdaptationResult}
\end{center}
\end{figure*}

\subsection{Experimental Settings}
\label{ssec:Settings}
We compare the proposed method with SqueezeSeg~\cite{wu2017squeezeseg}, the state-of-the-art model for semantic segmentation from 3D LiDAR point clouds. We use KITTI~\cite{geiger2012we} as the real-world dataset. KITTI provides images, LiDAR scans, and 3D bounding boxes organized in sequences. Following~\cite{wu2017squeezeseg}, we obtain point-wise labels from 3D bounding boxes. All points within a bounding box are considered part of the target object. In total, 10,848 samples with point-wise labels are collected. For SqueezeSegV2, the dataset is split into a training set with 8,057 samples and a testing set with 2,791 samples. 
For domain adaptation, we train the model on GTA-LiDAR and test it on KITTI for comparison.

Similar to~\cite{wu2017squeezeseg}, we evaluate our model's performance on class-level segmentation tasks by a point-wise comparison of the predicted results with ground-truth labels. We employ intersection-over-union (IoU) \cite{wu2018squeezeseg} as our evaluation metric, which is defined as $IoU_c = \frac{|\mathcal{P}_c \cap \mathcal{G}_c|}{|\mathcal{P}_c \cup \mathcal{G}_c|}$, where $\mathcal{P}_c$ and $\mathcal{G}_c$ respectively denote the predicted and ground-truth point sets that belong to class-$c$. $|\cdot|$ denotes the cardinality of a set.

\begin{table}[!t]
\begin{center}

\caption[Segmentation performance of SqueezeSegV2 and baselines]{Segmentation performance (IoU, \%) comparison between the proposed SqueezeSegV2 (+BN+M+FL+CAM) model and state-of-the-art baselines on the KITTI dataset. }
\begin{tabular}
{c | c c c c}
\hline
& Car & Pedestrian & Cyclist & Average\\
\hline
SqueezeSeg~\cite{wu2017squeezeseg} & 64.6 & 21.8  &  25.1  & 37.2 \\
+BN & 71.6  & 15.2  & 25.4 & 37.4 \\
+BN+M  & 70.0  & 17.1 & 32.3 & 39.8\\
+BN+M+FL   & 71.2 & 22.8  &  27.5 & 40.5 \\
+BN+M+FL+CAM & \textbf{73.2} & \textbf{27.8} & \textbf{33.6} & \textbf{44.9} \\
\hline
PointSeg \cite{wang2018pointseg} & 67.4 & 19.2 & 32.7 & 39.8 \\
\hline
\end{tabular}
\label{tab:SqueezeSegV2}
\end{center}
\textbf{+BN} denotes using batch normalization. \textbf{+M} denotes adding LiDAR mask as input. \textbf{+FL} denotes using focal loss. \textbf{+CAM} denotes using the CAM module.
\end{table}

\subsection{Improved Model Structure}
\label{ssec:Results_SqueezeSegV2}
The performance comparisons, measured in IoU, between the proposed SqueezeSegV2 model and baselines  are shown in Table~\ref{tab:SqueezeSegV2}. Some segmentation results are shown in Fig.~\ref{fig:SqueezeSegV1V2}.

From the results, we have the following observations. (1) both batch normalization and the mask channel can produce better segmentation results - batch normalization boosts segmentation of cars, whereas the mask channel boosts segmentation of cyclists. (2) Focal loss improves the segmentation of pedestrians and cyclists. The number of points corresponding to pedestrians and cyclists is low relative to a large number of background points. This class imbalance causes the network to focus less on the pedestrian and cyclist classes. Focal loss mitigates this problem by focusing the network on optimization of these two categories. (3) CAM significantly improves the performance of all the classes by reducing the network's sensitivity to dropout noise. 

\subsection{Domain Adaptation Pipeline}
\label{ssec:Results_SqueezeSegUDA}

\begin{table}[!t]
\begin{center}
\small
\caption[Segmentation performance of the domain adaptation pipeline and baselines.]{Segmentation performance (IoU, \%) of the proposed domain adaptation pipeline from GTA-LiDAR to the KITTI.}
\begin{tabular}
{c | c  c }
\hline
& \multicolumn{1}{c}{Car} & \multicolumn{1}{c}{Pedestrian}\\
\hline
SQSG trained on GTA \cite{wu2017squeezeseg}  &  29.0 & -  \\
SQSG trained on GTA-LiDAR & 30.0 & 2.1 \\
+LIR   &  42.0 & 16.7  \\
+LIR+GCA &  48.2 & 18.2\\
+LIR+GCA+PDC  &  50.3 & 18.6  \\
+LIR+GCA+PDC+CAM  &  \textbf{57.4}  & \textbf{23.5}  \\
\hline
SQSG trained on KITTI w/o intensity \cite{wu2017squeezeseg} & 57.1 & - \\
\hline
\end{tabular}
\label{tab:DA_Result}
\end{center}
 \textbf{SQSG} denotes SqueezeSeg. \textbf{+LIR} denotes using learned intensity rendering. \textbf{+GCA} denotes using geodesic correlation alignment. \textbf{+PDC} denotes using progressive domain calibration. \textbf{+CAM} denotes using the CAM module.
\end{table}

The performance comparisons, measured in IoU, between the proposed domain adaptation pipeline and baselines  are shown in Table~\ref{tab:DA_Result}. Some segmentation results are shown in Fig.~\ref{fig:DomainAdaptationResult}. From the results, we have the following observations. (1) Models trained on the source domain without any adaptation does not perform well. \emph{Domain discrepancy} lowers models' transferability from the source domain to the target domain. (2) All adaptation methods are effective, with the combined pipeline performing the best, demonstrating its effectiveness. (3) Adding the CAM to the network also significantly boosts the performance on the real data, supporting our hypothesis that dropout noise is a significant source of domain discrepancy. Therefore, improving the network to make it more robust to dropout noise can help reduce the domain gap. (4) Compared with \cite{wu2017squeezeseg} where a SqueezeSeg model is trained on the real KITTI dataset but without intensity, our SqueezeSegV2 model trained purely on synthetic data and unlabeled real data achieves a better accuracy, showing the effectiveness of our domain adaptation training pipeline. (5) Compared with our latest SqueezeSegV2 model trained on the real KITTI dataset, there is still an obvious performance gap. Adapting the segmentation model from synthetic LiDAR point clouds is still a challenging problem.

\section{Conclusion}
\label{sec:Conclusion}
In this chapter, we present SqueezeSegV2 and show two improvement over SqueezeSegV1. First, we improve the segmentation performance of SqueezeSegV1. We design a context aggregation module to mitigate the impact of dropout noise. Together with other improvements including focal loss, batch normalization and a LiDAR mask channel, SqueezeSegV2 sees accuracy improvements of 6.0\% to 8.6\% in various pixel categories over the original SqueezeSegV1. 
In SqueezeSegV1, we proposed the idea of using simulated data to train the model. However, experiments show that due to the severe domain shift, the model trained with simulated data fails to generalize to the real-world. To solve this problem, we propose a domain adaptation pipeline with three components: learned intensity rendering, geodesic correlation alignment, and progressive domain calibration. The proposed pipeline significantly improves the real-world accuracy of the model trained on synthetic data by 28.4\%, even out-performing a baseline model \cite{wu2017squeezeseg} trained on the real dataset. This is a significant improvement, since collecting and annotating real data is extremely expensive, especially for LiDAR data. Being able to leverage simulated data means we can bypass the data collection, therefore greatly improves the data efficiency of neural networks. Compared with the augmented SqueezeSegV2 model that is trained on real data, we still see a performance gap. However, this result demonstrates the potential for further research on this topic. 

%% file: chap7.tex
\chapter{Hardware Efficiency: Shift \& Synetgy}
\label{chap:shift}

In order to make a deep learning system run faster, consume less energy, achieve higher accuracy, we not only need to design compact deep neural networks, we must also optimize hardware processors that they run on. Ideally, this process should involve the co-design of both deep neural networks and hardware processors. In reality, however, we found that the two communities of model design and hardware design are divided. On the model design side, most previous efforts on mainly focus on optimizing hardware-agnostic metrics such as parameter size or FLOPs (or more strictly, the number of Multiply-and-Accumulate operations). However, those proxies do not always reflect actual efficiencies, such as latency and energy. Some neural networks achieve lower FLOPs but also become slower due to complicated network structures. Meanwhile, in the hardware design community, many improvements were solely focusing on the hardware side without leveraging the latest progress of neural network design. In this chapter, we discuss the following \textit{key question}:
\begin{quote}
    Can we co-design neural networks and hardware accelerators to further improve the efficiency of deep learning systems? 
\end{quote}
We will first discuss a strategy to drastically reduce the parameters and FLOPs of neural networks by simplifying spatial convolutions all the way to a novel and simple operator named ``Shift''. This not only significantly reduces parameter size and FLOPs of a network but also allows us to build neural networks with only 1x1 convolutions, which enables to build simplified and customized hardware accelerators to achieve significant speedup over previous state-of-the-art. 

\section{Introduction and related work}

\begin{figure}[ht]
\centering
\includegraphics[width=0.7\textwidth]{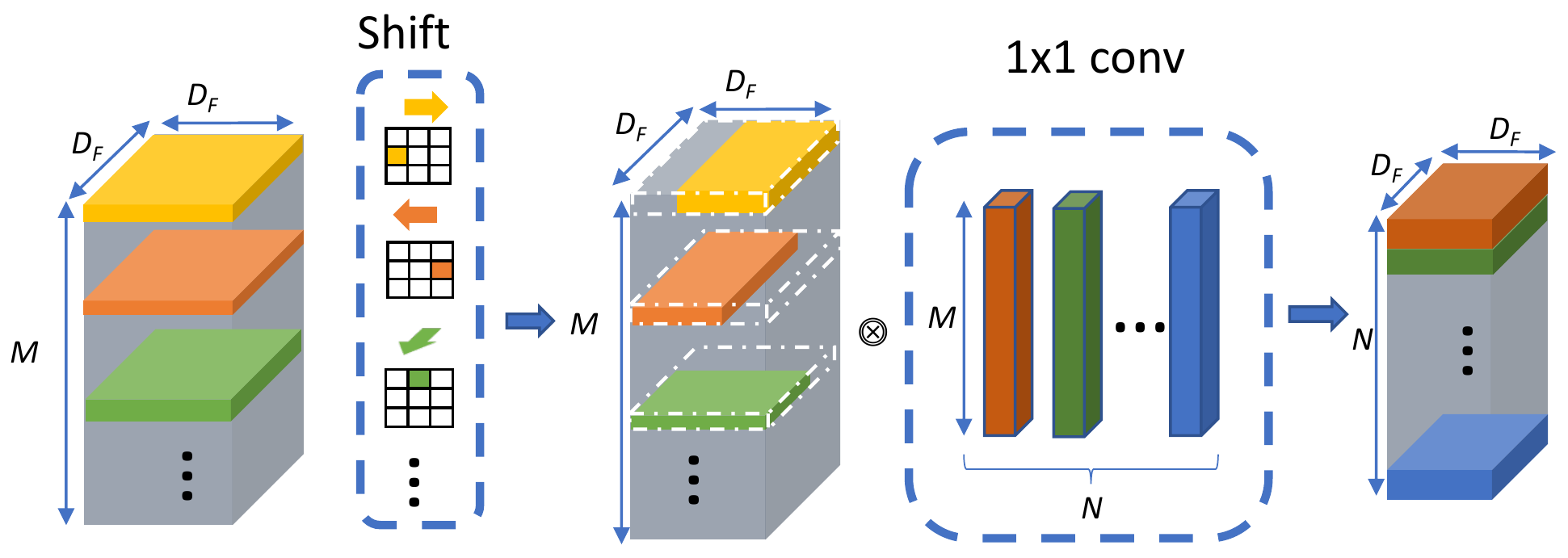}
   \caption[A shift operation followed by a 1x1 convolution.]{Illustration of a shift operation followed by a 1x1 convolution. The shift operation collect data spatially and the 1x1 convolution mixes information across channels.}
\label{fig:shift-conv}
\end{figure}

Convolutional neural networks (CNNs) rely on spatial convolutions with kernel sizes of 3x3 or larger to aggregate spatial information within an image. However, both the FLOPs and model size of spatial convolutions grow quadratically with respect to kernel sizes. In the VGG-16 model \cite{VGG}, 3x3 convolutions account for 15 million parameters, and the \textit{fc1} layer, effectively a 7x7 convolution, accounts for 102 million parameters and even more FLOPs. 

Several strategies have been adopted to reduce the size of spatial convolutions, therefore computations. ResNet\cite{resnet} employs a ``bottleneck module,'' placing two 1x1 convolutions before and after a 3x3 convolution, reducing its number of input and output channels. Despite this, 3x3 convolutional layers still account for  50\% of all parameters in ResNet models with bottleneck modules. SqueezeNet \cite{SqueezeNet} adopts a ``fire module,'' where the outputs of a 3x3 convolution and a 1x1 convolution are concatenated along the channel dimension. 
Recent networks such as ResNext~\cite{RexNext}, MobileNet~\cite{MobileNet}, and Xception~\cite{Xception} adopt group convolutions and depth-wise separable convolutions as alternatives to standard spatial convolutions. In theory, depth-wise convolutions require less computation. However, it is difficult to implement depth-wise convolutions efficiently in practice, as their arithmetic intensity (ratio of FLOPs to memory accesses) is too low to efficiently utilize the hardware.
Such a drawbacks are also mentioned in \cite{ShuffleNet, Xception}. ShuffleNet~\cite{ShuffleNet} integrates depth-wise convolutions, point-wise group convolutions, and channel-wise shuffling to further reduce parameters and complexity. ShuffleNetV2\cite{ma2018shufflenet} further develops this idea and proposes a more compact CNN, where less than 10\% of FLOPs are contributed by spatial convolutions. Another work, \cite{LocalBinaryCNN} inherits the idea of a separable convolution to freeze spatial convolutions and learn only point-wise convolutions. This does reduce the number of learnable parameters but falls short of saving FLOPs or model size. 

\begin{figure*}[ht!]
\centering
\includegraphics[width=1.\textwidth]{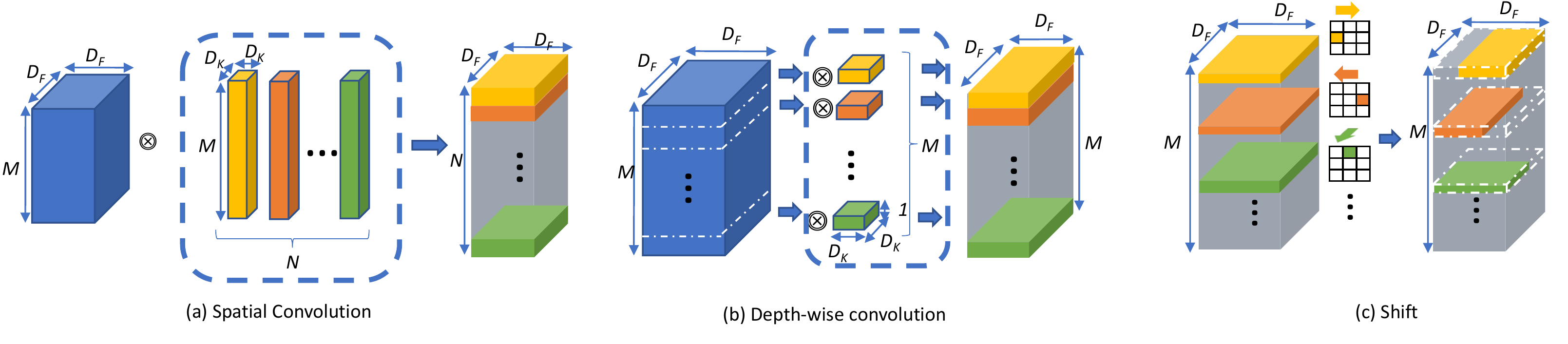}
   \caption[Spatial convolution vs. depth-wise convolution vs. shift. ]{Illustration of (a) spatial convolutions, (b) depth-wise convolutions and (c) shift. In (c), the 3x3 grids denote a shift matrix with a kernel size of 3. The lighted cell denotes a 1 at that position and white cells denote 0s.}
\label{fig:conv-depth-shift}
\end{figure*}

Our approach is a more radical one, which is to sidestep spatial convolutions entirely. In this chapter, we present the \textit{shift operation} (Figure~\ref{fig:shift-conv}) as an alternative to spatial convolutions. 
The shift operation moves each channel of its input tensor in a different spatial direction. A \textit{shift-based module} interleaves shift operations with point-wise convolutions, which further mixes spatial information across channels. 
Unlike spatial convolutions, the shift operation itself requires zero FLOPs and zero parameters. As opposed to depth-wise convolutions, shift operations can be easily and efficiently implemented. More importantly, with the shift operation, we can build a CNN with only 1x1 convolutions. This greatly simplifies the hardware design since it allows us to build a dedicated compute unit customized for 1x1 convolutions to achieve further speedup. 

Our approach is orthogonal to model compression~\cite{DeepCompression}, tensor factorization~\cite{TensorFact} and low-bit networks~\cite{XnorNet}. As a result, any of these techniques could be composed with our proposed method to further reduce model size.

We demonstrate the efficacy of the shift operator by building a family of network architectures called \textit{ShiftNet} and evaluating them on a wide range of tasks including image classification, face verification, and style transfer. Using significantly fewer parameters, ShiftNet attains competitive performance. Furthermore, we combine the idea of shift operator with ShuffleNetV2\cite{ma2018shufflenet}, and co-designed a hardware accelerator named \textit{Synetgy} which achieved significant speedup over the previous state-of-the-art. 

\section{The Shift module and network design}

We first review the standard spatial and depth-wise convolutions illustrated in Figure~\ref{fig:conv-depth-shift}. Consider the spatial convolution in Figure~\ref{fig:conv-depth-shift}(a), which takes a tensor $F \in \mathbb{R}^{D_F \times D_F \times M}$ as input. Let $D_F$ denote the height and width and $M$ denote the channel size. The kernel of a spatial convolution is a tensor $K\in \mathbb{R}^{D_K \times D_K \times M \times N}$, where $D_K$ denotes the kernel's spatial height and width, and $N$ is the number of filters. For simplicity, we assume the stride is 1 and that the input/output have identical spatial dimensions. Then, the spatial convolution outputs a tensor $G \in \mathbb{R}^{D_F\times D_F \times N}$, which can be computed as
\begin{equation}
    G_{k, l, n} = \sum_{i, j, m}K_{i, j, m, n} F_{k+\hat{i},l+\hat{j}, m},
\end{equation}
where $\hat{i} = i - \lfloor D_K/2\rfloor, \hat{j} = j - \lfloor D_K/2 \rfloor$ are the re-centered spatial indices; $k, l$ and $i, j$ index along spatial dimensions and $n, m$ index into channels.
The number of parameters required by a spatial convolution is $M\times N\times D_K^2$ and the computational cost is $M\times N\times D_K^2 \times D_F^2 $. As the kernel size $D_K$ increases, we see the number of parameters and computational cost grow quadratically. 

A popular variant of the spatial convolution is a depth-wise  convolution \cite{MobileNet,Xception}, which is usually followed by a point-wise convolution (1x1 convolution). Altogether, the module is called the depth-wise separable convolution.
A depth-wise convolution, as shown in Figure~\ref{fig:conv-depth-shift}(b), aggregates spatial information from a $D_K \times D_K$ patch within each channel, and can be described as 
\begin{equation}
    \hat{G}_{k, l, m} = \sum_{i, j}\hat{K}_{i, j, m} F_{k+\hat{i},l+\hat{j}, m},
\end{equation}
where $\hat{K} \in \mathbb{R}^{D_F\times D_F\times M}$ is the depth-wise convolution kernel. This convolution comprises $M\times D_K^2$ parameters and $M \times D_K^2 \times D_F^2$ FLOPs. As in standard spatial convolutions, the number of parameters and computational cost grow quadratically with respect to the kernel size $D_K$. Finally, point-wise convolutions mix information across channels, giving us the following output tensor 
\begin{equation}
    G_{k, l, n} = \sum_{m}P_{m, n} \hat{G}_{k,l, m},
    \label{eqn:point-wise}
\end{equation}
where $P\in\mathbb{R}^{M\times N}$ is the point-wise convolution kernel. 

In theory, depth-wise convolution requires less computation and fewer parameters. In practice, however, this means memory access dominates computation, thereby limiting the use of parallel hardware. 
For standard convolutions, the ratio between computation \textit{vs.} memory access is 
\begin{equation}
    \frac{M\times N \times D_F^2 \times D_K^2}{D_F^2\times (M+N) + D_K^2\times M \times N},
    \label{eqn:conv-AI}
\end{equation}
while for depth-wise convolutions, the ratio is 
\begin{equation}
    \frac{M\times D_F^2 \times D_K^2}{D_F^2\times 2M + D_K^2\times M}.
    \label{eqn:depth-AI}
\end{equation}
A lower ratio here means that more time is spent on memory accesses, which are several orders of magnitude slower and more energy-consuming than FLOPs. 
This drawback implies an I/O-bound device will be unable to achieve maximum computational efficiency.

\subsection{The Shift Operation}

The shift operation, as illustrated in Figure~\ref{fig:conv-depth-shift}(c), can be viewed as a special case of depth-wise convolutions. Specifically, it can be described logically as:
\begin{equation}
    \tilde{G}_{k, l, m} = \sum_{i, j}\tilde{K}_{i, j, m} F_{k+\hat{i},l+\hat{j}, m}.
    \label{eqn:shift}
\end{equation}
The kernel of the shift operation is a tensor $\tilde{K} \in \mathbb{R}^{D_F \times D_F\times M}$ such that 
\begin{equation}
    \tilde{K}_{i, j, m} = 
    \begin{cases}
        1, & \text{if } i=i_m \text{ and } j=j_m,  \\
        0, & \text{otherwise}.
    \end{cases}
\end{equation}
Here $i_m, j_m$ are channel-dependent indices that assign one of the values in $\tilde{K}_{:, :, m} \in \mathbb{R}^{D_K\times D_K}$ to be 1 and the rest to be 0. We call $\tilde{K}_{:, :, m}$ a shift matrix. 

For a shift operation with kernel size $D_K$, there exist $D_K^2$ possible shift matrices, each of them corresponding to a shift direction. If the channel size $M$ is no smaller than $D_K^2$, we can construct a shift matrix that allows each output position $(k, l)$ to access all values within a $D_K \times D_K$ window in the input. We can then apply another point-wise convolution per Eq.~(\ref{eqn:point-wise}) to exchange information across channels. 

Unlike spatial and depth-wise convolutions, the shift operation itself does not require parameters or floating-point operations (FLOPs). Instead, it is a series of memory operations that adjusts channels of the input tensor in certain directions. A more sophisticated implementation can fuse the shift operation with the following 1x1 convolution, where the 1x1 convolution directly fetches data from the shifted address in memory. With such an implementation, we can aggregate spatial information using shift operations, for free.

\subsection{Constructing Shift Kernels}

For a given kernel size $D_K$ and channel size $M$, there exists $D_K^2$ possible shift directions, making $(D_K^2)^M$ possible shift kernels. An exhaustive search over this state space for the optimal shift kernel is prohibitively expensive. 

To reduce the state space, we use a simple heuristic: divide the $M$ channels evenly into $D_K^2$ groups, where each group of $\lfloor M/D_K^2 \rfloor$ channels adopts one shift. We will refer to all channels with the same shift as a \textit{shift group}. The remaining channels are assigned to the ``center'' group and are not shifted.

However, finding the optimal permutation, \textit{i.e.}, how to map each channel-$m$ to a shift group, requires searching a combinatorially large search space. To address this issue, we introduce a modification to the shift operation that makes input and output invariant to channel order: We denote a shift operation with channel permutation $\pi$ as $\mathcal{K}_{\pi}(\cdot)$, so we can express Eq.~(\ref{eqn:shift}) as $\tilde{G} = \mathcal{K}_{\pi}(F)$. We permute the input and output of the shift operation as  
\begin{equation}
\tilde{G} = \mathcal{P}_{\pi_2}(\mathcal{K}_{\pi}(\mathcal{P}_{\pi_1}(F))) = (\mathcal{P}_{\pi_2}\circ \mathcal{K}_{\pi} \circ \mathcal{P}_{\pi_1}) (F),
\label{eqn:perm-shift-perm}
\end{equation}
where $\mathcal{P}_{\pi_i}$ are permutation operators and $\circ$ denotes operator composition. However, permutation operators are discrete and therefore difficult to optimize. As a result, we process the input $F$ to Eq.~(\ref{eqn:perm-shift-perm}) by a point-wise convolution $\mathcal{P}_1 (F)$. We repeat the process for the output $\tilde{G}$ using $\mathcal{P}_2 (\tilde{G})$. The final expression can be written as 
\begin{equation}
\begin{aligned}
    G & = (\mathcal{P}_2 \circ \mathcal{P}_{\pi_2} \circ \mathcal{K}_{\pi} \circ \mathcal{P}_{\pi_1} \circ \mathcal{P}_1 ) (F) \\
    & = ((\mathcal{P}_2 \circ \mathcal{P}_{\pi_2}) \circ \mathcal{K}_{\pi} \circ (\mathcal{P}_{\pi_1} \circ \mathcal{P}_1) ) (F) \\ 
    & = (\hat{\mathcal{P}}_2 \circ \mathcal{K}_{\pi} \circ \hat{\mathcal{P}}_1) (F), 
\end{aligned}
\label{eqn:SCS-operation}
\end{equation}
where the final step holds, as there exists a bijection from the set of all $\mathcal{P}_i$ to the set of all $\mathcal{\hat{P}}_i$, since the permutation operator $\mathcal{P}_{\pi_i}$ is bijective by construction. As a result, it suffices to learn $\hat{\mathcal{P}}_1$ and $\hat{\mathcal{P}}_2$ directly. Therefore, this augmented shift operation Eq.~(\ref{eqn:SCS-operation}) can be trained with stochastic gradient descent end-to-end, without regard for channel order. So long as the shift operation is sandwiched between two point-wise convolutions, different permutations of shifts are equivalent. Thus, we can choose an arbitrary permutation for the shift kernel, after fixing the number of channels for each shift direction.

\subsection{Shift-based Modules}

First, we define a \textit{module} to be a collection of layers that perform a single function, e.g.~ResNet's bottleneck \textit{module} or SqueezeNet's fire \textit{module}. Then, we define a \textit{group} to be a collection of repeated modules.

\begin{figure}[th]
\centering
\includegraphics[width=.45\textwidth]{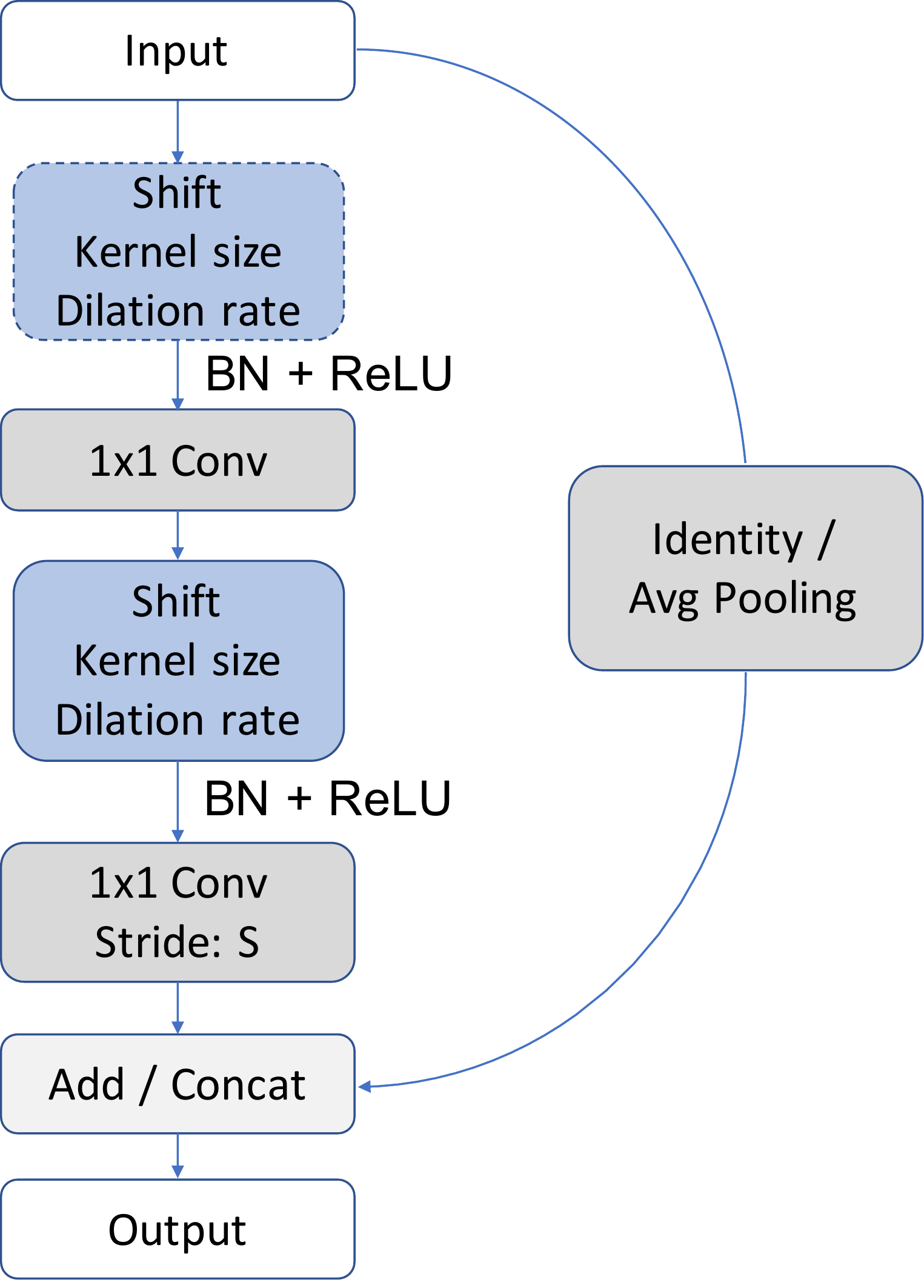}
   \caption[Shift-based modules.]{Illustration of the \textit{Conv-Shift-Conv} $CSC$ module and the \textit{Shift-Conv-Shift-Conv} $(SC^2)$ module.}
\label{fig:CSC}
\end{figure}

Based on the analysis in previous sections, we propose a module using shift operations as shown in Figure~\ref{fig:CSC}. The input tensor is first processed by point-wise convolutions. Then, we perform a shift operation to redistribute spatial information. Finally, we apply another set of point-wise convolutions to mix information across channels. Both sets of point-wise convolutions are preceded by batch normalization and a non-linear activation function (ReLU). Following ShuffleNet~\cite{ShuffleNet}, we use an additive residual connection when input and output are of the same shape and use average pooling with concatenation when we down-sample the input spatially and double the output channels. We refer to this as a \textit{Conv-Shift-Conv} or $CSC$ module. A variant of this module includes another shift operation before the first point-wise convolution; we refer to this as the \textit{Shift-Conv-Shift-Conv} or $SC^2$ module. This allows the designer to further increase the receptive field of the module. 

As with spatial convolutions, shift modules are parameterized by several factors that control their behavior. We use the kernel size of the shift operation to control the receptive field of the $CSC$ module. Akin to the dilated convolution, the ``dilated shift'' samples data at a spatial interval, which we define to be the \textit{dilation rate} $\mathcal{D}$. The \textit{stride} of the $CSC$ module is defined to be the stride of the second point-wise convolution so that spatial information is mixed in the shift operation before down-sampling. Similar to the bottleneck module used in ResNet, we use the ``expansion rate'' $\mathcal{E}$ to control the intermediate tensor's channel size. With bottleneck modules, 3x3 convolutions in the middle are expensive computationally, forcing small intermediate channel sizes. However, the shift operation allows kernel size $D_F$ adjustments without affecting parameter size and FLOPs. As a consequence, we can employ a shift module to allow larger intermediate channel sizes, where sufficient information can be gathered from nearby positions.

\section{Experiments for shift}

We first assess the shift module's ability to replace convolutional layers and then adjust hyperparameter $\mathcal{E}$ to observe tradeoffs between model accuracy, model size, and computation. 
We then construct a range of shift-based networks and investigate their performance for a number of different applications.

\subsection{Operation choice and hyperparameters}

Using ResNet, we juxtapose the use of 3x3 convolutional layers with the use of $CSC$ modules, by replacing all of ResNet's basic modules (two 3x3 convolutional layers) with $CSC$s to make ``ShiftResNet''. To compare with the depth-wise convolution, we replace shift in the $CSC$ module with a depthwise convolution to make \textit{Depthwise-ResNet}, or DWResNet. For ResNet, ShiftResNet, and DWResNet, we use two Tesla K80 GPUs with batch size 128 and a starting learning rate of 0.1, decaying by a factor of 10 after 32k and 48k iterations, as in \cite{resnet}. In these experiments, we use the CIFAR10 version of ResNet: a convolutional layer with 16 3x3 filters; 3 groups of basic modules with output channels 16, 32, 64; and a final fully-connected layer. A basic module contains two 3x3 convolutional layers followed by batchnorm and ReLU in parallel with a residual connection. With ShiftResNet, each group contains several $CSC$ modules. We use three ResNet models: in ResNet20, each group contains 3 basic modules. For ResNet56, each contains 9, and for ResNet110, each contains 18. By toggling the hyperparameter $\mathcal{E}$, the number of filters in the $CSC$ module's first set of 1x1 convolutions, we can reduce the number of parameters in ``ShiftResNet'' by nearly 3 times without any loss in accuracy, as shown in Table~\ref{tab:res-shift-vs-conv}. Table~\ref{tab:shift-cifar10} and Table~\ref{tab:shift-cifar100} summarize CIFAR10 and CIFAR100 results across all $\mathcal{E}$ and ResNet models.

We next compare different strategies for parameter reduction. We reduce ResNet's parameters to match that of ShiftResNet for some $\mathcal{E}$, denoted ResNet-$\mathcal{E}$ and ShiftResNet-$\mathcal{E}$, respectively. We use two separate approaches: 1) module-wise: decrease the number of filters in each module's first 3x3 convolutional layer; 2) net-wise: decrease every module's input and output channels by the same factor. As Table~\ref{tab:robust} shows, convolutional layers are less resilient to parameter reduction, with the shift module preserving accuracy 8\% better than both reduced ResNet models of the same size, on CIFAR100. In Table~\ref{tab:robust-imagenet}, we likewise find improved resilience on ImageNet as ShiftResNet achieve better accuracy with millions fewer parameters.

\begin{figure}[h]
\centering
\begin{tikzpicture}
\begin{axis}[
    title={Accuracy vs FLOPs tradeoff},
    xlabel={FLOPs (Millions)},
    ylabel={Accuracy (Top 1 CIFAR100)},
    ylabel style={yshift=-0.5em},
    xmin=0, xmax=300,
    ymin=50, ymax=80,
    xtick={0,50,100,150,200,250,300},
    ytick={50,60,70,80},
    legend pos=south east,
    ymajorgrids=true,
    grid style=dashed,
    width=240pt
]
 
\addplot+[
    purple,
    mark=triangle*,
    solid,
    mark options={solid}
    ]
    coordinates {
    (6,55.62)(17,62.32)(32,68.64)(48,69.82)};
\addplot+[
    green!85!black,
    mark=triangle*,
    solid,
    mark options={solid}
    ]
    coordinates {
    (16,65.21)(45,69.77)(89,72.13)(133,73.64)
    };
\addplot[mark=triangle*,teal!50!cyan]
    coordinates {
    (30,67.84)(87,71.83)(174,72.56)(260,74.10)
    };
\addplot+[
    purple,
    mark=square*,
    dashed,
    mark options={solid}
    ]
    coordinates {
    (4.8,52.40)(14,60.61)(27,64.27)(41,66.37)};
\addplot+[
    green!85!black,
    mark=square*,
    dashed,
    mark options={solid}
    ]
    coordinates {
    (14,56.78)(41,64.49)(83,67.45)(126,68.63)
    };
\addplot[mark=square*,
    teal!50!cyan,
    dashed]
    coordinates {
    (28,60.44)(82,66.16)(167,68.87)(253,70.14)
    };
    \legend{ShiftResNet20,ShiftResNet56,ShiftResNet110,ResNet20,ResNet56,ResNet110}
\end{axis}
\end{tikzpicture}
\caption[Accuracy vs. FLOPs of ShiftResNet and baselines.]{Tradeoff curves further to the top left are more efficient, with higher accuracy per FLOP. This figure shows ShiftResNet is more efficient than ResNet, in FLOPs.}
\label{fig:acc-v-flops-tradeoff}
\end{figure}
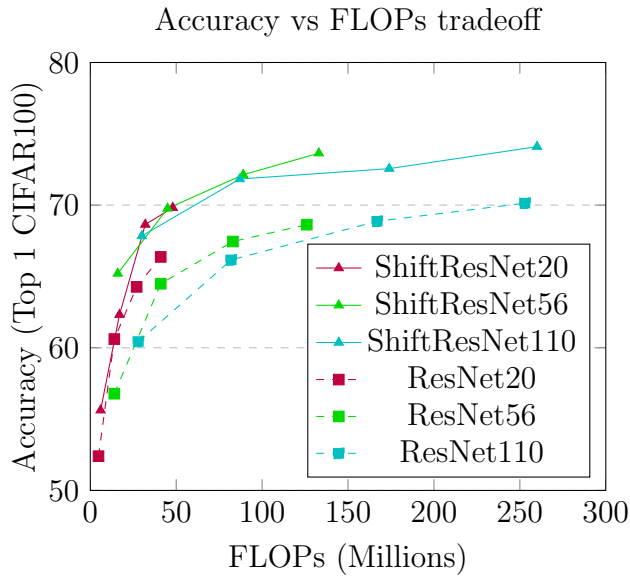

Table~\ref{tab:shrink-model-size} shows that ShiftResNet consistently outperforms ResNet, when both are constrained to use 1.5x fewer parameters. Figure~\ref{fig:acc-v-param-tradeoff} shows the tradeoff between CIFAR100 accuracy and number of parameters for the hyperparameter $\mathcal{E} \in \{1,3,6,9\}$ across both \{ResNet, ShiftResNet\} models using varying numbers of layers $\ell \in \{20,56,110\}$. Figure~\ref{fig:acc-v-flops-tradeoff} examines the same set of possible models and hyperparameters but between CIFAR100 accuracy and the number of FLOPs. Both figures show that ShiftResNet models provide superior trade-off between accuracy and parameters/FLOPs. 

\begin{table}[h]
\begin{center}
\caption{Parameters for shift vs convolution, with fixed accuracy on CIFAR-100.}
\label{tab:res-shift-vs-conv}
\begin{tabular}{c | c c c c}
\hline
Model & \thead{Top1 Acc} & \thead{FLOPs} & \thead{Params} \\
\hline
ShiftResNet56-3 & \textbf{69.77}\% & \textbf{44.9M} & \textbf{0.29M}\\
ResNet56 & 69.27\% & 126M  & 0.86M \\
\hline
\end{tabular}
\end{center}
\end{table}

\begin{table}
\begin{center}
\caption{Reduction resilience for shift vs convolution, with fixed parameters.}
\label{tab:robust}
\begin{tabular}{c | c c c c}
\hline
Model & \thead{CIFAR-100 Acc} & \thead{FLOPs} & \thead{Params} \\
\hline
ShiftResNet110-1 & \textbf{67.84}\% & 29M & \textbf{203K} \\
ResNet110-1 & 60.44\% & \textbf{28M} & 211K \\
\hline
\end{tabular}
\end{center}
\end{table}

\begin{table}
\begin{center}
\caption{Reduction resilience for shift vs convolution on ImageNet.}
\label{tab:robust-imagenet}
\begin{tabular}{c c c}
\hline
\thead{ {\normalsize ShiftResNet50} \\ Top1 / Top5 Acc} & \thead{{\normalsize ResNet50} \\ Top1 / Top5 Acc} & \thead{{\normalsize Parameters} \\ Shift50 / ResNet50} \\
\hline
\textbf{75.6} / \textbf{92.8} & 75.1 / 92.5 & 22M / 26M\\
\textbf{73.7} / \textbf{91.8} & 73.2 / 91.6 & 11M / 13M \\
\textbf{70.6} / 89.9 & 70.1 / 89.9 & 6.0M / 6.9M \\
\hline
\end{tabular}
\end{center}
\end{table}

\begin{table}
\begin{center}
\caption{Performance across ResNet 
models, with 1.5 fewer parameters}
\label{tab:shrink-model-size}
\begin{tabular}{c | c c c c}
\hline
No. Layers & \thead{ {\normalsize ShiftResNet-6} \\ CIFAR100 Top 1} & \thead{{\normalsize ResNet } \\ CIFAR100 Top 1} \\
\hline
20 & \textbf{68.64\%} & 66.25\%\\
56 & \textbf{72.13\%} & 69.27\%\\
110 & \textbf{72.56\%} & 72.11\%\\
\hline
\end{tabular}
\end{center}
\vspace{-0.2in}
\end{table}

\begin{figure}[h]
\centering
\begin{tikzpicture}
\begin{axis}[
    title={Accuracy vs parameters tradeoff},
    xlabel={Parameters (Millions)},
    ylabel={Accuracy (Top 1 CIFAR100)},
    ylabel style={yshift=-0.5em},
    xmin=0, xmax=2,
    ymin=50, ymax=80,
    xtick={0,0.5,1,1.5,2},
    ytick={50,60,70,80},
    legend pos=south east,
    ymajorgrids=true,
    grid style=dashed,
    width=240pt
]
 
\addplot+[
    purple,
    mark=triangle*,
    solid,
    mark options={solid}
    ]
    coordinates {
    (0.03,55.62)(0.1,62.32)(0.19,68.64)(0.28,69.82)};
\addplot+[
    green!85!black,
    mark=triangle*,
    solid,
    mark options={solid}
    ]
    coordinates {
    (0.1,65.21)(0.29,69.77)(0.58,72.13)(0.87,73.64)
    };
\addplot[
    mark=triangle*,
    teal!50!cyan
    ]
    coordinates {
    (0.2,67.84)(0.59,71.83)(1.18,72.56)(1.76,74.10)
    };
\addplot+[
    purple,
    mark=square*,
    dashed,
    mark options={solid}
    ]
    coordinates {
    (0.03,52.40)(0.1,60.61)(0.19,64.27)(0.28,66.37)};
\addplot+[
    green!85!black,
    mark=square*,
    dashed,
    mark options={solid}
    ]
    coordinates {
    (0.1,56.78)(0.29,64.49)(0.58,67.45)(0.87,68.63)
    };
\addplot[mark=square*,teal!50!cyan,dashed]
    coordinates {
    (0.2,60.44)(0.59,66.16)(1.18,68.87)(1.76,70.14)
    };
    \legend{ShiftResNet20,ShiftResNet56,ShiftResNet110,ResNet20,ResNet56,ResNet110}
\end{axis}
\end{tikzpicture}
\caption[Accuracy vs. parameter size of ShiftResNet and baselines.]{This figure shows that ShiftResNet family members are  significantly more efficient than their corresponding ResNet family members. Tradeoff curves further to the top left are more efficient, with higher accuracy per parameter. For ResNet, we take the larger of two accuracies between module-wise and net-wise reduction results.}
\label{fig:acc-v-param-tradeoff}
\end{figure}
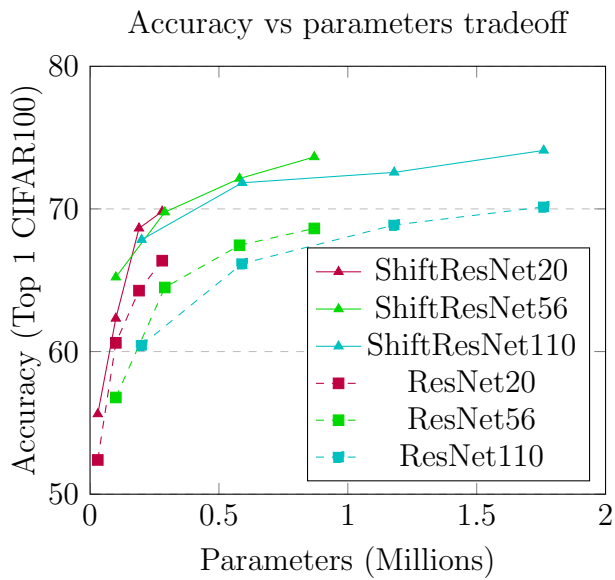

\begin{table*}
\centering
\caption{Shift operator analysis on CIFAR10}
\label{tab:shift-cifar10}
\begin{tabular}{c | c c c c c}
Model & $\mathcal{E}$ & \thead{ShiftResNet \\ Acc (\%)} & \thead{ ResNet \\ (Module) \\ Acc (\%)} & \thead{ Params \\ (million)} & \thead{ FLOPs \\ (million) } \\
\hline
20 & 1 & 86.66 & 85.54 & 0.03 & 6 \\
20 & 3 & 90.08 & 88.33 & 0.10 & 17 \\
20 & 6 & 90.59 & 90.09 & 0.19 & 32 \\
20 & 9 & 91.69 & 91.35 & 0.28 & 48 \\
\hline
56 & 1 & 89.71 & 87.46 & 0.10 & 16 \\
56 & 3 & 92.11 & 89.40 & 0.29 & 45 \\
56 & 6 & 92.69 & 89.89 & 0.58 & 89 \\
56 & 9 & 92.74 & 92.01 & 0.87 & 133 \\
\hline
110 & 1 & 90.34 & 76.82 & 0.20 & 29 \\
110 & 3 & 91.98 & 74.30 & 0.59 & 87 \\
110 & 6 & 93.17 & 79.02 & 1.18 & 174 \\
110 & 9 & 92.79 & 92.46 & 1.76 & 260 \\
\hline
\end{tabular}
\caption*{
ResNet (Module) denotes the module-level reduction on the original ResNet models.}
\end{table*}

\begin{table*}
\centering
\caption{Shift operator analysis on CIFAR100}
\label{tab:shift-cifar100}
\begin{tabular}{c | c c c c c c c}
Model & $\mathcal{E}$ & \thead{ShiftResNet \\ Acc (\%)} & \thead{ResNet \\ (Module) \\ Acc (\%)}  & \thead{ResNet \\ (Net) \\ Acc (\%)} & \thead{DWResNet \\ Acc (\%)} & \thead{ Params \\ (million)} & \thead{FLOPs \\ (million)} \\
\hline
20 & 1 & 55.63 & 52.40 & 49.58 & 61.32 &  0.03 & 6 \\
20 & 3 & 62.32 & 60.61 & 58.16 & 64.51 & 0.10 & 17 \\
20 & 6 & 68.64 & 64.27 & 63.22 & 65.38 & 0.19 & 32 \\
20 & 9 & 69.82 & 66.25 & 66.25 & 65.59 & 0.28 & 48 \\
\hline
56 & 1 & 65.21 & 56.78 & 56.62 & 65.30 & 0.10 & 16 \\
56 & 3 & 69.77 & 62.53 & 64.49 & 66.49 & 0.29 & 45 \\
56 & 6 & 72.13 & 61.99 & 67.45 & 67.46 & 0.58 & 89 \\
56 & 9 & 73.64 & 69.27 & 69.27 & 67.75 & 0.87 & 133 \\
\hline
110 & 1 & 67.84 & 39.90 & 60.44 & 65.80 & 0.20 & 29 \\
110 & 3 & 71.83 & 40.52 & 66.61 & 67.22 & 0.59 & 87 \\
110 & 6 & 72.56 & 40.23 & 68.87 & 68.11 & 1.18 & 174 \\
110 & 9 & 74.10 & 72.11 & 72.11 & 68.39 & 1.76 & 260 \\
\hline
\end{tabular}
\caption*{ResNet (Module) and ResNet (Net) denote module-level and net-level reduction on the original ResNet models. DWResNet denotes the result of Depthwise-ResNet. }
\end{table*}

\subsection{ShiftNet}

Even though ImageNet classification is not our primary goal, to further investigate the effectiveness of the shift operation, we use the proposed $CSC$ module shown in Figure~\ref{fig:CSC} to design a class of efficient models called ShiftNet and present its classification performance on standard benchmarks to compare with state-of-the-art small models. 

Since an external memory access consumes orders-of-magnitudes more energy than a single arithmetic operation \cite{pedram2016dark}\footnote{According to \cite{pedram2016dark}, under 45nm technology, an off-chipt DRAM access consumes 3556x more energy than an addition operation for 16-bit integers.}, our primary goal in designing ShiftNet is to optimize the number of parameters and thereby to reduce memory footprint.
In addition to the general desirability of energy efficiency, 
the main targets of ShiftNet are mobile and IOT applications, 
where memory footprint, even
more so than FLOPs, are a primary constraint.
In these application domains, small models can be packaged in mobile and IoT applications that are delivered within the 100-150MB limit for mobile over-the-air updates. In short, our design goal for ShiftNets is to attain competitive accuracy with fewer parameters.

The ShiftNet architecture is described in Table~\ref{tab:shiftnet-arch}. Since parameter size does not grow with shift kernel size, we use a larger kernel size of 5 in earlier modules. We adjust the expansion parameter $\mathcal{E}$ to scale the parameter size in each $CSC$ module. We refer to the architecture described in Table~\ref{tab:shiftnet-arch} as ShiftNet-A. We shrink the number of channels in all $CSC$ modules by 2 for ShiftNet-B. We then build a smaller and shallower network, with $\{1, 4, 4, 3\}$ CSC modules in groups $\{1, 2, 3, 4\}$ with channel sizes of $\{32, 64, 128, 256\}$, $\mathcal{E}=1$ and kernel size is 3 for all modules. We name this shallow model ShiftNet-C. 
We train the three ShiftNet variants on the ImageNet 2012 classification dataset~\cite{ILSVRC15} with 1.28 million images and evaluate on the validation set of 50K images. We adopt data augmentations suggested by ~\cite{Xception} and weight initializations suggested by~\cite{resnet}. We train our models for 90 epochs on 64 Intel KNL instances using Intel Caffe~\cite{Caffe} with a batch size of 2048, an initial learning rate of 0.8, and learning rate decay by $10$ every 30 epochs. 

In Table~\ref{tab:shiftnet_result}, we show classification accuracy and parameters size for ShiftNet and other state-of-the-art models. Especially, MobileNet is considered as a strong baseline for efficient models, but its training protocol is not clearly explained in~\cite{MobileNet}. Therefore, we trained MobileNet with the same training protocol as ShiftNet, and report both the accuracy reproduced by ourselves and the accuracy reported in~\cite{MobileNet}, with the reported accuracy in brackets. We compare ShiftNet-\{A, B, C\} with 3 groups of models with similar levels of accuracy. In the first group, ShiftNet-A is \textbf{34X} smaller than VGG-16, while the top-1 accuracy drop is only $1.4\%$. 1.0-MobileNet-224 has similar parameter size as ShiftNet-A. Under the same training protocol, MobileNet has worse accuracy than ShiftNet-A, while the reported accuracy is better. In the second group, ShiftNet-B's top-1, top-5 accuracy is $2.3\%$ and $0.7\%$ worse than its MobileNet counterpart, but it uses fewer parameters. We compare ShiftNet-C with SqueezeNet and AlexNet, and we can achieve better accuracy with $2/3$ the number of SqueezeNet's parameters, and \textbf{77X} smaller than AlexNet.

\begin{table}[]
\centering
\caption{ShiftNet architecture}
\label{tab:shiftnet-arch}
\begin{tabular}{c|c|c|c|c|c}
\hline
Group & \begin{tabular}[c]{@{}c@{}}Type/\\ Stride\end{tabular} & Kernel            & $\mathcal{E}$ & \begin{tabular}[c]{@{}c@{}}Output\\ Channel\end{tabular} & Repeat \\ \hline
- & Conv /s2      & 7$\times$7      &   -     & 32                                                       & 1      \\ \hline 
1 & $\mbox{CSC}$ / s2      & 5$\times$5      & 4      & 64                                                       & 1      \\ 
\newline & $\mbox{CSC}$ / s1      & 5$\times$5      & 4      &                                                        & 4      \\ \hline
2 & $\mbox{CSC}$ / s2      & 5$\times$5      & 4      & 128                                                      & 1      \\ 
\newline & $\mbox{CSC}$ / s1      & 5$\times$5      & 3      &                                                       & 5      \\ \hline 
3 & $\mbox{CSC}$ / s2      & 3$\times$3      & 3      & 256                                                      & 1      \\ 
\newline & $\mbox{CSC}$ / s1      & 3$\times$3      & 2      &                                                      & 6      \\ \hline 
4 & $\mbox{CSC}$ / s2      & 3$\times$3      & 2      & 512                                                      & 1      \\ 
\newline & $\mbox{CSC}$ / s1      & 3$\times$3      & 1      &                                                       & 2      \\ \hline 
- & Avg Pool      & 7$\times$7  &   -     & 512                                                      & 1      \\ \hline
- & FC            & - &     -     & 1k                                                       & 1      \\ \hline
\end{tabular}
\end{table}

\begin{table}[]
\centering
\caption{ShiftNet results on Imagenet}
\label{tab:shiftnet_result}
\begin{tabular}{c|cc}
\hline
Model  & \thead{Accuracy\\ Top-1 / Top-5} & \thead{Parameters \\ (Millions)} \\ \hline
VGG-16~\cite{VGG} & \textbf{71.5} / \textbf{90.1} & 138 \\
GoogleNet~\cite{googlenet} & 69.8 / - & 6.8 \\
ShiftResNet-0.25 (ours) & 70.6 / 89.9 & 6.0 \\
ShuffleNet-2$\times$~\cite{ShuffleNet} & 70.9 / - & 5.4 \\
1.0 MobileNet-224~\cite{MobileNet}* & 67.5 (70.6) / 86.6 & 4.2\\
Compact DNN~\cite{compactDNN} & 68.9 / 89.0 & 4.1 \\
ShiftNet-A (ours)  & 70.1 / 89.7 & \textbf{4.1} \\
\hline
0.5 MobileNet-224~\cite{MobileNet}* & \textbf{63.5} (63.7) / 84.3 & 1.3\\ 
ShiftNet-B (ours) & 61.2 / 83.6 & \textbf{1.1} \\
\hline
AlexNet~\cite{alexnet} & 57.2 / 80.3 & 60 \\
SqueezeNet~\cite{SqueezeNet} & 57.5 / 80.3 & 1.2 \\
ShiftNet-C (ours) & \textbf{58.8} / \textbf{82.0} & \textbf{0.78} \\
\hline 
\end{tabular}
    \caption*{* We list both our reproduced accuracy and reported accuracy from~\cite{MobileNet}. The reported accuracy is in brackets.}
\end{table}

\subsection{Face embedding}

We continue to investigate the shift operation for different applications. Face verification and recognition are becoming increasingly popular on mobile devices. Both functionalities rely on face embedding, which aims to learn a mapping from face images to a compact embedding in Euclidean space, where face similarity can be directly measured by embedding distances. Once the space has been generated, various face-learning tasks, such as facial recognition and verification, can be easily accomplished by standard machine learning methods with feature embedding. Mobile devices have limited computation resources, therefore creating small neural networks for face embedding is a necessary step for mobile deployment.

FaceNet~\cite{schroff2015facenet} is one state-of-the-art face embedding approach. The original FaceNet is based on Inception-Resnet-v1~\cite{szegedy2016rethinking}, which contains 28.5 million parameters, making it difficult to be deployed on mobile devices. We propose a new model ShiftFaceNet based on ShiftNet-C from the previous section, which only contains 0.78 million parameters. 

Following~\cite{parkhi2015deep}, we train FaceNet and ShiftFaceNet by combining the softmax loss with center loss~\cite{wen2016discriminative}. We evaluate the proposed method on three datasets for face verification: given a pair of face images, a distance threshold is selected to classify the two images belonging to the same or different entities. The LFW dataset~\cite{huang2007labeled} consists of 13,323 web photos with 6,000 face pairs.
The YTF dataset~\cite{wolf2011face} includes 3,425 with 5,000 video pairs for video-level face verification.
The MS-Celeb-1M dataset (MSC)~\cite{guo2016ms} comprises 8,456,240 images for 99,892 entities.
In our experiments, we randomly select 10,000 entities from MSC as our training set, to learn the embedding space. We test on 6,000 pairs from LFW, 5,000 pairs from YTF, and 100,000 pairs randomly generated from MSC, excluding the training set.
In the pre-processing step, we detect and align all faces using a multi-task CNN~\cite{zhang2016joint}.
Following~\cite{schroff2015facenet},
the similarity between two videos is computed as the average similarity of 100 random pairs of frames, one from each video. Results are shown in Table~\ref{tab:face_accuracy}. 
The original FaceNet contains 28.5 million parameters, while the ShiftFaceNet only contains 0.78 million parameters. With ShiftFaceNet, we are able to reduce the parameter size by \textbf{35X}, with at most $2\%$ drop of accuracy in the above three verification benchmarks. 

\begin{table}
\centering
\small
\caption[Face verification accuracy for ShiftFaceNet vs FaceNet]{Face verification accuracy for ShiftFaceNet vs FaceNet~\cite{schroff2015facenet}}
\vspace{-0.1in}
\label{tab:face_accuracy}
\begin{tabular}{c | c c | c c}
\hline
 & \multicolumn{2}{c|}{Accuracy$\pm$ STD (\%)} & \multicolumn{2}{c}{Area under curve (\%)} \\
\hline
 & FaceNet & ShiftFaceNet & FaceNet & ShiftFaceNet \\
LFW & 97.1$\pm$1.3 & 96.0$\pm$1.4 & 99.5 & 99.4\\
YTF & 92.0$\pm$1.1 & 90.1$\pm$0.9 & 97.3 & 96.1\\
MSC & 79.2$\pm$1.7 & 77.6$\pm$1.7 & 85.6 & 84.4\\
\hline
\end{tabular}
\end{table}

\subsection{Style transfer}

\newcommand{\STincludegraphics}[2][]{\includegraphics[width = 0.9in, height=0.9in,#1]{#2}}
{
\setlength\tabcolsep{2pt}
\begin{figure*}
\begin{tabular}{c | c c c | c c c}
\textsc{Content} & \textsc{Style} &  \textsc{Shift}&  \textsc{Original} & \textsc{Style} & \textsc{Shift} &  \textsc{Original} \\
\subfloat{\STincludegraphics[]{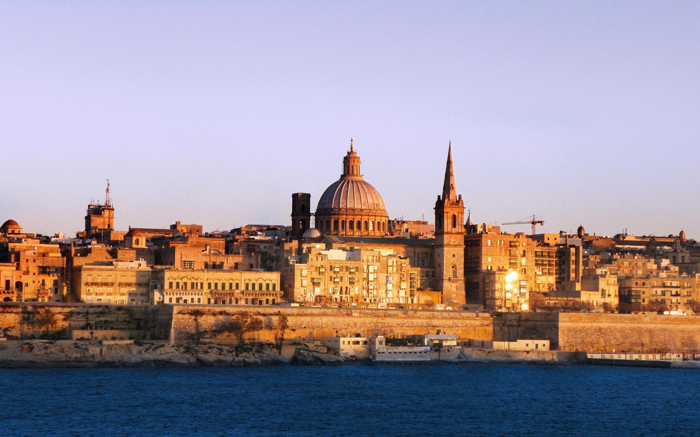}} &
 \multirow{1}{*}[4ex]{\subfloat{\STincludegraphics[]{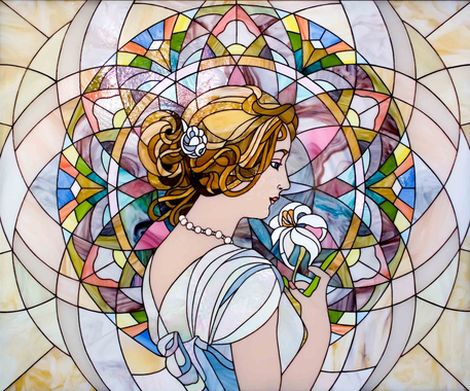}}} &
\subfloat{\STincludegraphics[]{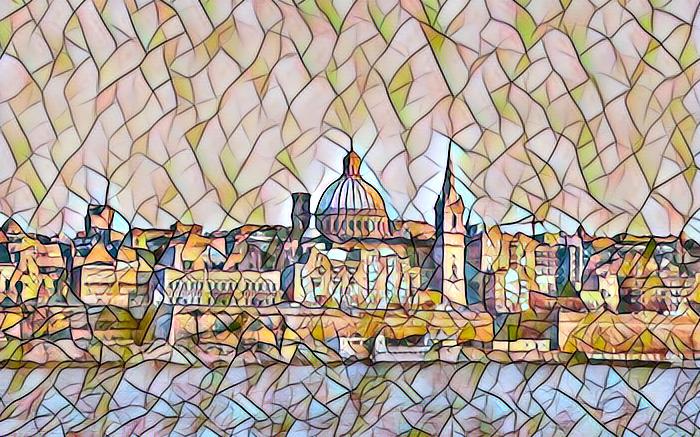}} &
\subfloat{\STincludegraphics[]{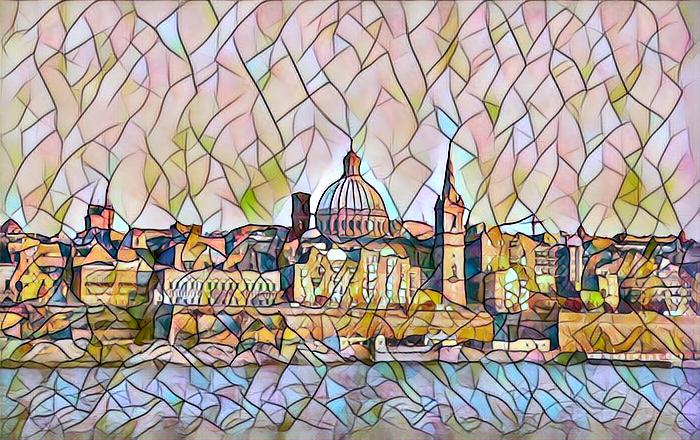}} &
\multirow{1}{*}[4ex]{\subfloat{\STincludegraphics[]{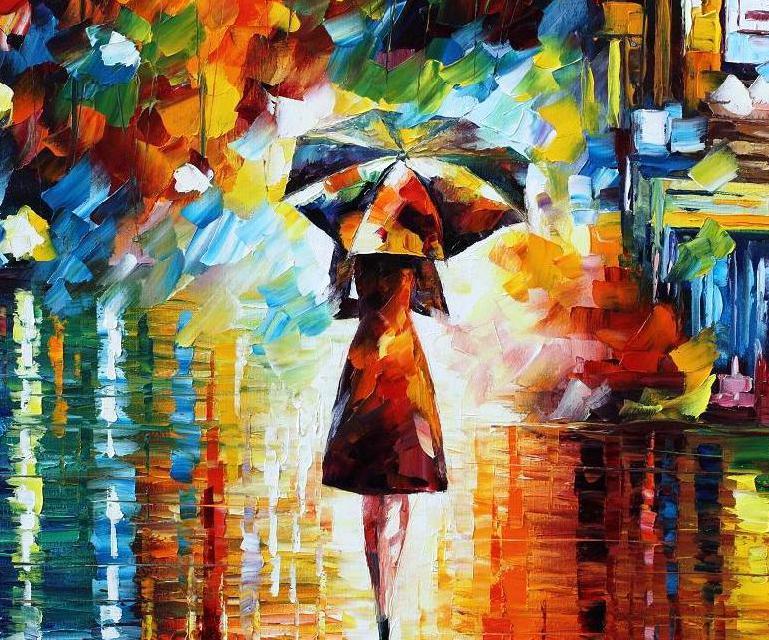}}} &
\subfloat{\STincludegraphics[]{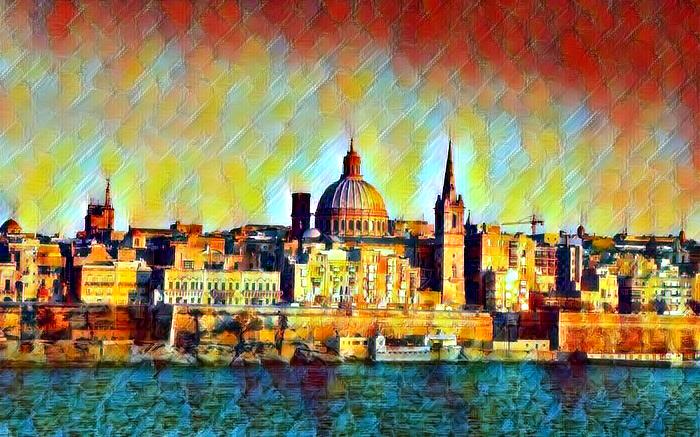}} &
\subfloat{\STincludegraphics[]{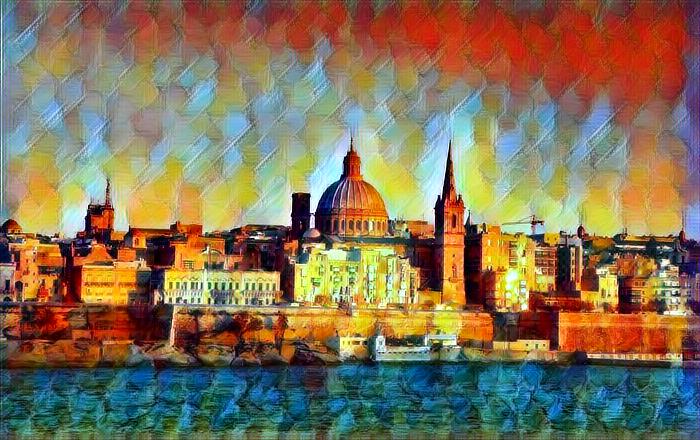}} \\
\subfloat{\STincludegraphics[]{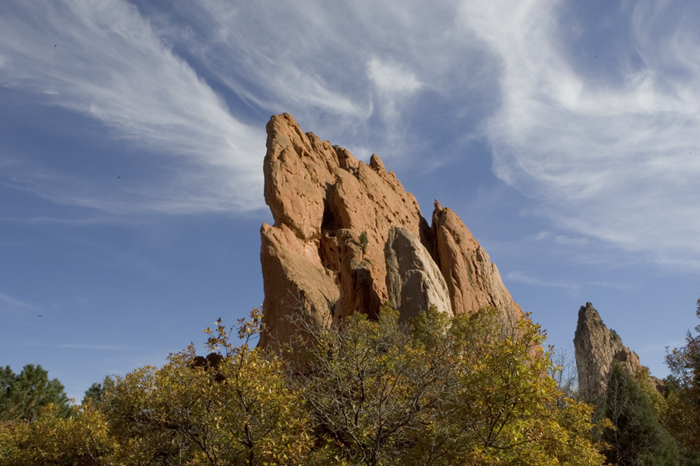}} & 
&
\subfloat{\STincludegraphics[]{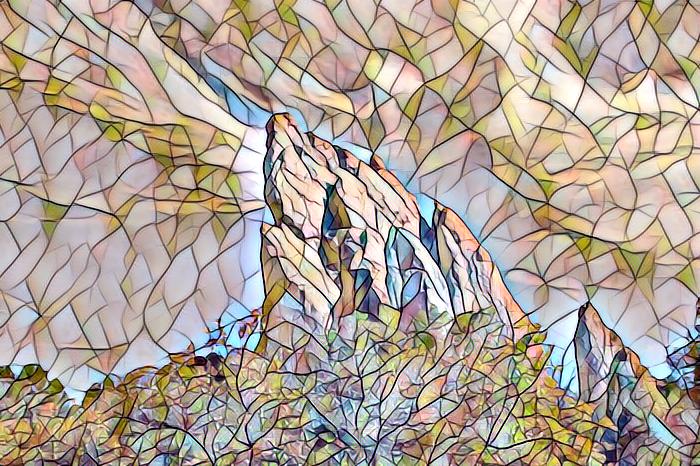}} &
\subfloat{\STincludegraphics[]{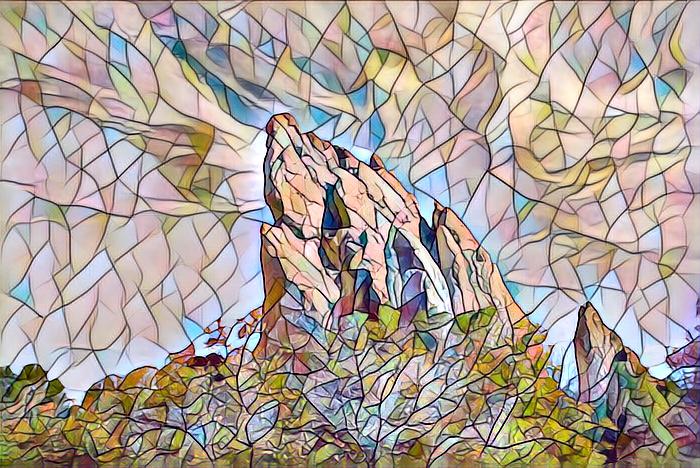}} &
&
\subfloat{\STincludegraphics[]{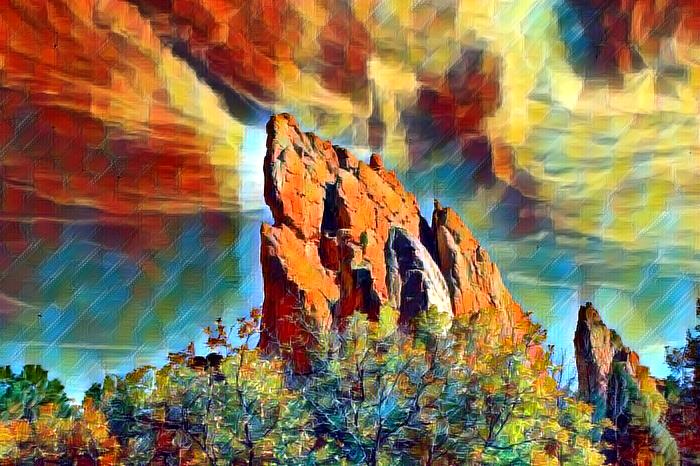}} & 
\subfloat{\STincludegraphics[]{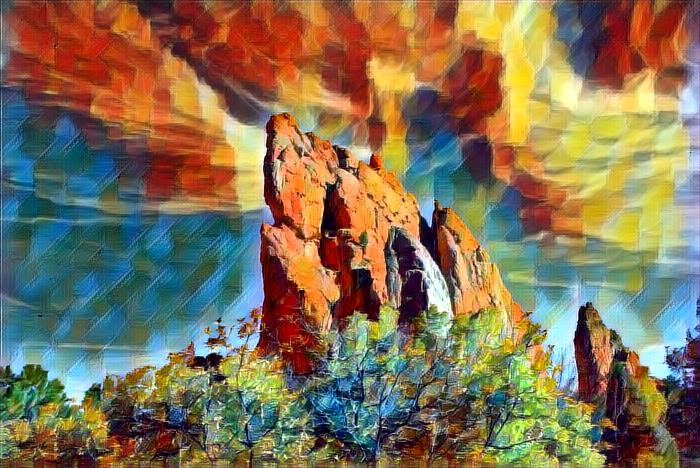}} 
\end{tabular}
\caption{Style transfer results using shift operators}
\label{fig:style}
\end{figure*}
}

Artistic style transfer is another popular application on mobile devices. It is an image transformation task where the goal is to combine the \textit{content} of one image with the \textit{style} of another. Although this is an ill-posed problem without definite quantitative metrics, a successful style transfer requires that networks capture both minute textures and holistic semantics, for content and style images.

Following \cite{gatys-style-transfer, johnson-style-transfer}, we use perceptual loss functions to train a style transformer. In our experiment, we use a VGG-16 network pretrained on ImageNet to generate the perceptual loss. The original network trained by Johnson \textit{et al.} \cite{johnson-style-transfer} consists of three downsampling convolutional layers, five residual modules, and three upsampling convolutional layers. All non-residual convolutions are followed by an instance normalization layer \cite{instance-norm}. In our experiments, we replace all but the first and last convolution layers with shifts followed by 1x1 convolutions. We train the network on the COCO \cite{coco} dataset using the previously reported hyperparameter settings $\lambda_s \in \{1e10, 5e10, 1e11\}, \lambda_c = 1e5$. By replacing convolutions with shifts, we achieve an overall \textbf{6X} reduction in the number of parameters from 1.9 million to 0.3 million, with minimal degradation in image quality. Examples of stylized images generated by original and shift based transformer networks can be found in Figure~\ref{fig:style}.

\section{DiracDeltaNet: Co-designing a hardware friendly CNN using shift}
The shift operator not only reduces the parameter size and FLOPs, more importantly, it eliminates the need for spatial convolution, which makes hardware design much easier. In this section, we discuss the design of DiracDeltaNet, a hardware friendly neural network enabled by the shift operator. 

An ideal CNN for embedded FPGA acceleration should satisfy the following aspects: 
1) The network should not contain too many parameters or FLOPs but should still maintain a competitive accuracy. 
2) The network structure should be hardware friendly such mapping to hardware is easy. 
3) The network's operation set should be simplified for efficient FPGA implementation. 
4) The network's weights and activations should be quantized to low-precision fixed-point numbers without much accuracy loss. 

\subsection{ShuffleNetV2}
ShuffleNetV2-1.0x \cite{ma2018shufflenet} is one of the state-of-the-art efficient models. It has a top-1 accuracy of 69.4\% on ImageNet (2\% lower than VGG16) but contains only 2.3M parameters (60x smaller than VGG16) and 146M FLOPs (109x smaller than VGG16).

\begin{figure}[!t]
\begin{center}
    \begin{subfigure}[b]{0.8\textwidth}
        \centering\includegraphics[width=0.7\linewidth]{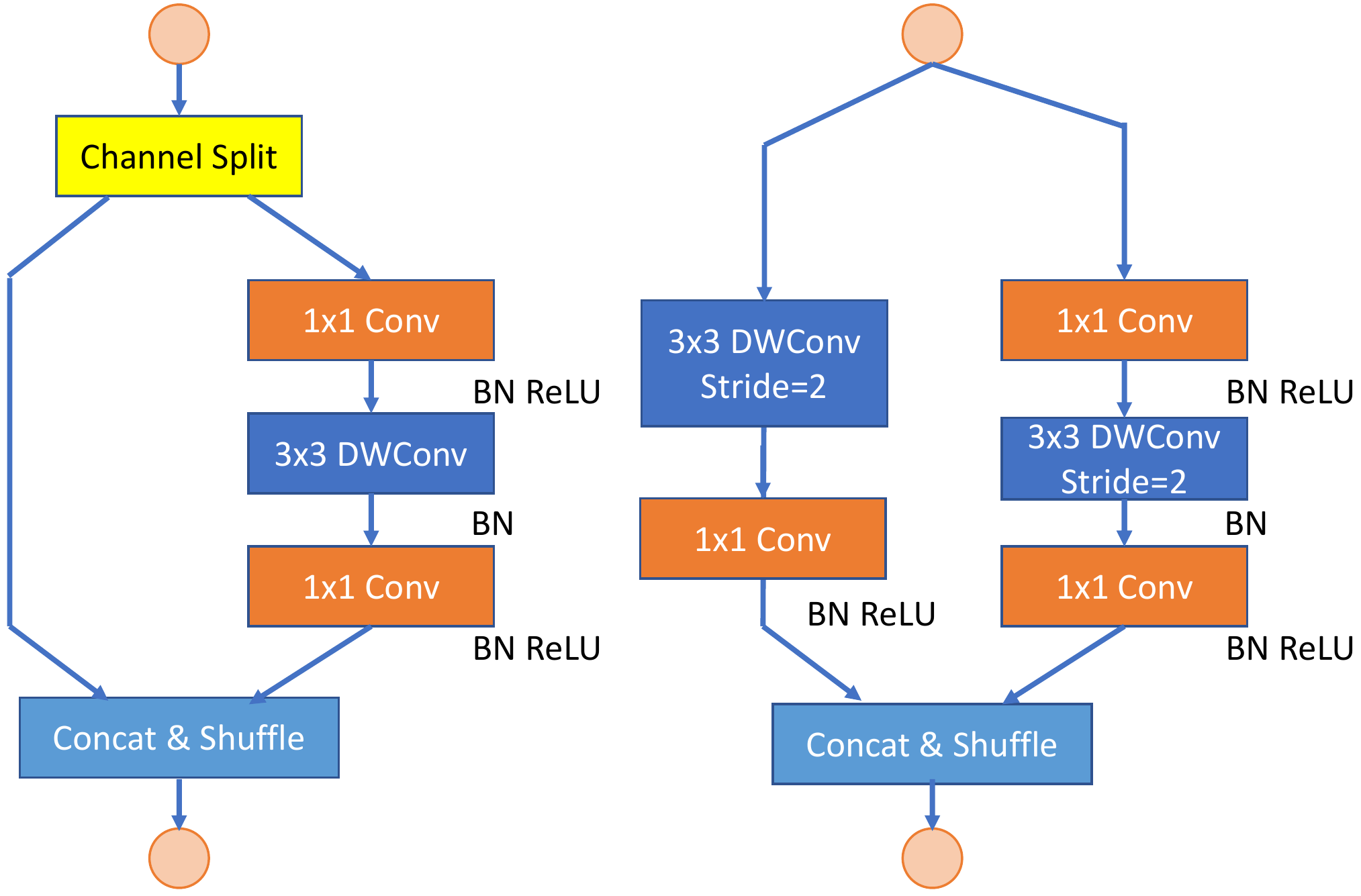}
        \caption{ShuffleNetV2 blocks \cite{ma2018shufflenet}. }
        \label{fig:shufflenet-blocks}
    \end{subfigure} \hfill
    \begin{subfigure}[b]{0.8\textwidth}
        \centering\includegraphics[width=0.7\linewidth]{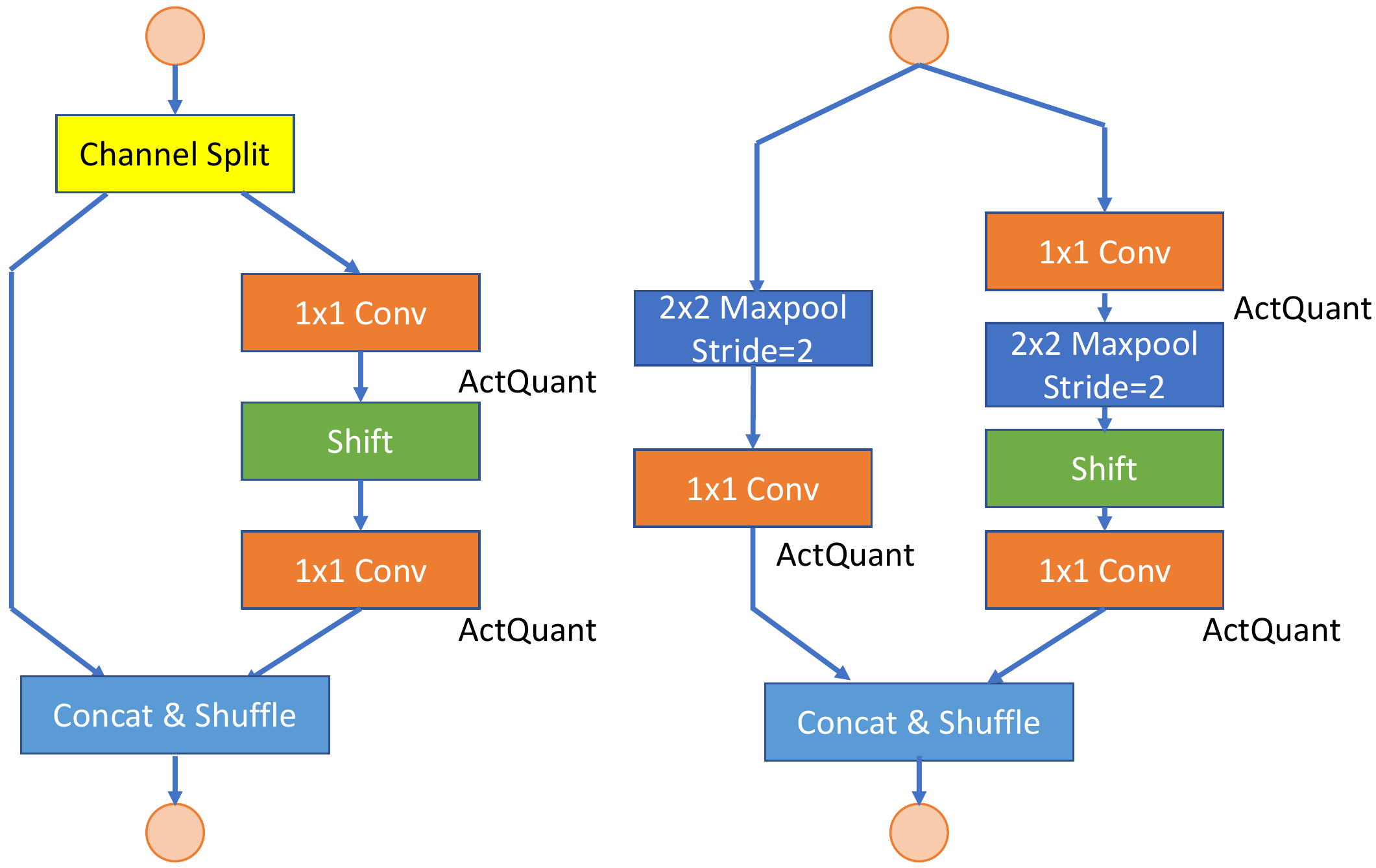}
        \caption{Our modified DiracDeltaNet blocks. We replace depth-wise convolutions with shift operations. In the downsampling blocks, we use stride-2 max-pooling and shift operations to replace stride-2 depth-wise convolutions. We also double the filter number of the 1st 1$\times$1 convolution on the non-skip branch in each module.}
        \label{fig:dirac-block}
    \end{subfigure} \hfill    
\caption{ShuffleNetV2 blocks vs. DiracDeltaNet blocks}
\label{fig:netblocks}
\end{center}
\end{figure}

The block-level structure of ShuffleNetV2 is illustrated in Fig. \ref{fig:shufflenet-blocks}. The input feature map of the block is first split into two parts along the channel dimension. The first branch of the network does nothing to the input data and directly feeds the input to the output. The second branch performs a series of 1$\times$1 convolutions, 3$\times$3 depth-wise convolutions and another 1$\times$1 convolution operations on the input. Outputs of two branches are then concatenated along the channel dimension. Channel shuffle \cite{zhang1707shufflenet} is then applied to exchange information between branches. In down-sampling blocks, depth-wise 3$\times$3 convolutions with a stride of 2 are applied to both branches of the block to reduce the spatial resolution. 1$\times$1 convolutions are used to double the channel size of input feature maps. These blocks are cascaded to build a deep ConvNet. We refer readers to \cite{ma2018shufflenet} for the macro-structure description of the ShuffleNetV2.

\begin{figure}[!t]
\begin{center}
\centering \includegraphics[width=0.6\linewidth]{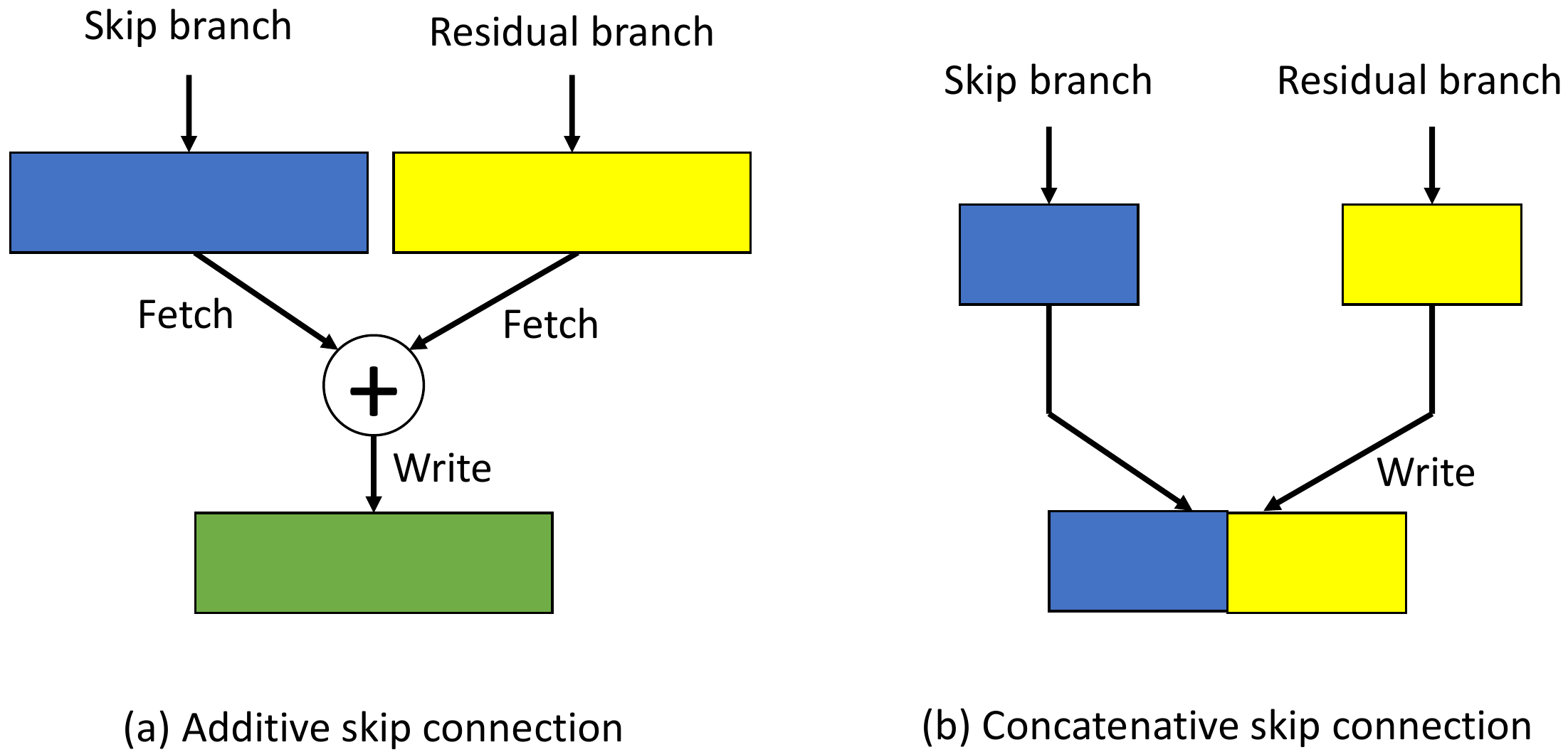}
\caption[Additive Skip Connections vs. Concatenative Skip Connections.]{Additive Skip Connections vs. Concatenative Skip Connections. Rectangles represent data tensors.}
\label{fig:add-vs-concat}
\end{center}
\end{figure}

We select ShuffleNetV2-1.0x not only because of its small model size and low FLOP count but also because it uses concatenative skip connections instead of additive skip connections. Additive skip connections, as illustrated in Fig. \ref{fig:add-vs-concat}(a), were first proposed in \cite{he2016deep}. It effectively alleviates the difficulty of training deep neural networks and therefore improves accuracy. It is widely used in many ConvNet designs. However, additive skip connections are not efficient on FPGAs. As illustrated in Fig. \ref{fig:add-vs-concat}(a), both the skip and the residual branches' data need to be fetched on-chip to conduct the addition. Though additions do not cost too much computation, the data movement is expensive. Concatenative skip connections, as illustrated in Fig. \ref{fig:add-vs-concat}(b), were first proposed in \cite{huang2017densely}. It has a similar positive impact on network training and accuracy. With concatenative skip connections, data from skip branch is already in off-chip DRAMs. So we can concatenate the two branches simply by writing the residual branch data next to the skip branch data. This avoids the extra memory access in additive skip connections and alleviates the memory bandwidth pressure.

\subsection{DiracDeltaNet}
\label{DiracDeltaNet}
Based on ShuffleNetV2, we build DiracDeltaNet through the following modifications: 1) we replace all the 3$\times$3 convolutions with shift and 1$\times$1 convolutions; 2) we reduce the kernel size of max-pooling from 3$\times$3 to 2$\times$2; 3) we modify the order of channel shuffle.

\begin{figure}[!t]
\begin{center}
\centering \includegraphics[width=0.9\linewidth]{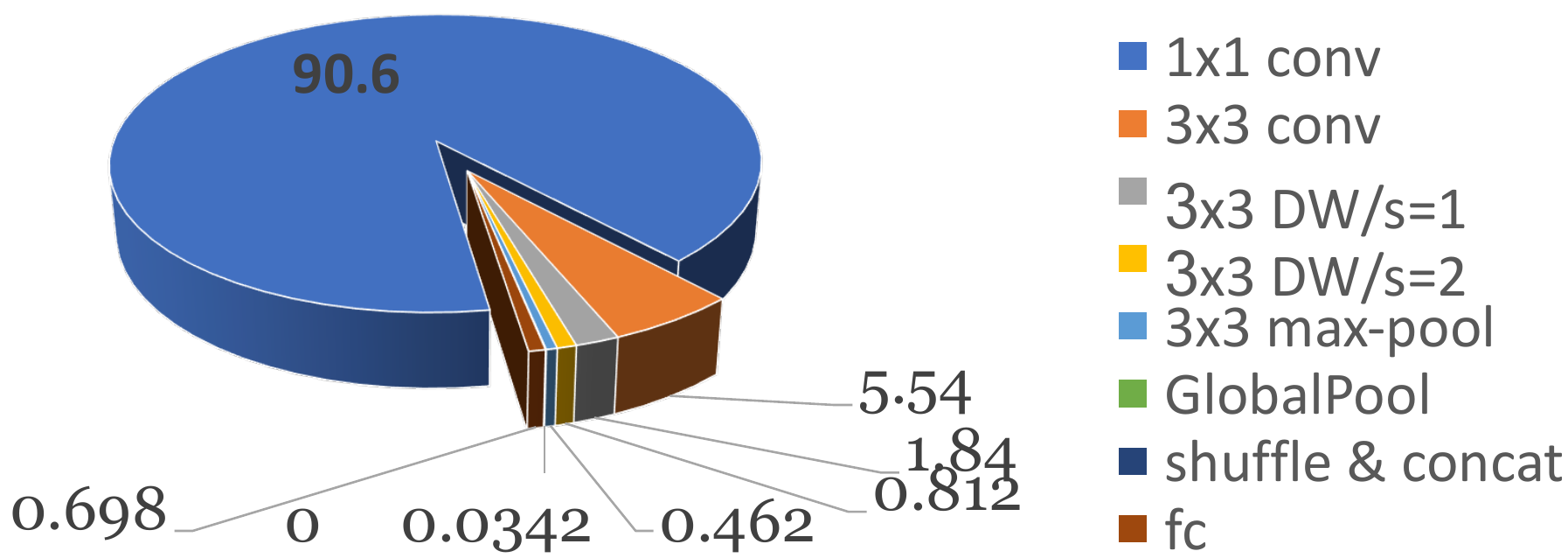}
\caption{Types of operators and their FLOPs in ShuffleNetV2.}
\label{fig:op-FLOPs}
\end{center}
\end{figure}

Fig. \ref{fig:op-FLOPs} shows different types of operators in ShuffleNetV2 and their FLOP contributions. Note that due to heavy adoption of depthwise convolutions, spatial convolutions, including regular 3x3 and depthwise 3x3 account for less than 10\% of the total FLOPs. However, this introduces a dilemma for hardware design. Since we need to support both spatial and 1x1 convolutions, we need to either build a general compute unit that supports both, or build dedicated compute unites for each of the operators. The first strategy sacrifices the performance for generality. The second strategy, however, wastes hardware resources on spatial convolutions that are not frequently used. To avoid this dilemma, we use the shift operator to get rid of all the spatial convolutions, which enables us to simplify the hardware and only build a compute unit for 1x1 convolutions.

For 3$\times$3 depth-wise convolutions in ShuffleNetV2, we directly replace them with shift operations, as shown in Fig. \ref{fig:dirac-block}. This direct replacement can lead to some accuracy loss. To mitigate this, we double the output filter number of the first 1$\times$1 convolution on the non-skip branch from Fig. \ref{fig:dirac-block}. Nominally, doubling the output channel size increases both FLOP count and parameter size by a factor of 2. However, this is mitigated by the performance improvement of the customization for 1$\times$1 convolutions. In the downsample block, we directly replace the stridden 3$\times$3 depth-wise convolutions with a stride-2 2$\times$2 max-pooling. Unlike \cite{wu2017shift}, our shift operation only uses 4 cardinal directions (up, down, left, right) in addition to the identity mapping (no-shift). This simplifies our hardware implementation of the shift operation without hurting accuracy. 

The first stage of ShuffleNetV2 consists of a 3$\times$3 convolution with a stride of 2 and filter number of 24. It is then followed by a 3$\times$3 max-pooling with a stride of 2. We replace these two layers to a module consisting of a series of 1$\times$1 convolution, 2$\times$2 max-pooling, and shift operations, as shown in Table \ref{tab:dirac-macro}. Compared with the original 3$\times$3 convolutions, our proposed module has more parameters (2144 vs 648) and FLOPs (30.5M vs 8.1M). But the implementation and execution cost of the proposed first stage is negligible compared to a 3$\times$3 convolution layer. After training the network, we find that this module gives near equal accuracy than the original 3$\times$3 convolution module. With our new module, we can eliminate the remaining 3$\times$3 convolutions from our network, enabling us to allocate more computational resources to 1$\times$1 convolutions and thereby increasing parallelism and throughput. 

In addition to replacing all 3$\times$3 convolutions, we also reduce the max-pooling kernel size from 3$\times$3 to 2$\times$2. By using the same pooling kernel size as the stride, we eliminate the need to buffer extra data on the pooling kernel boundaries, thereby achieving better efficiency. Our experiments also show that reducing the max-pooling kernel size does not impact accuracy.

\begin{figure}[!t]
\begin{center}
\centering 
\includegraphics[width=.7\linewidth]{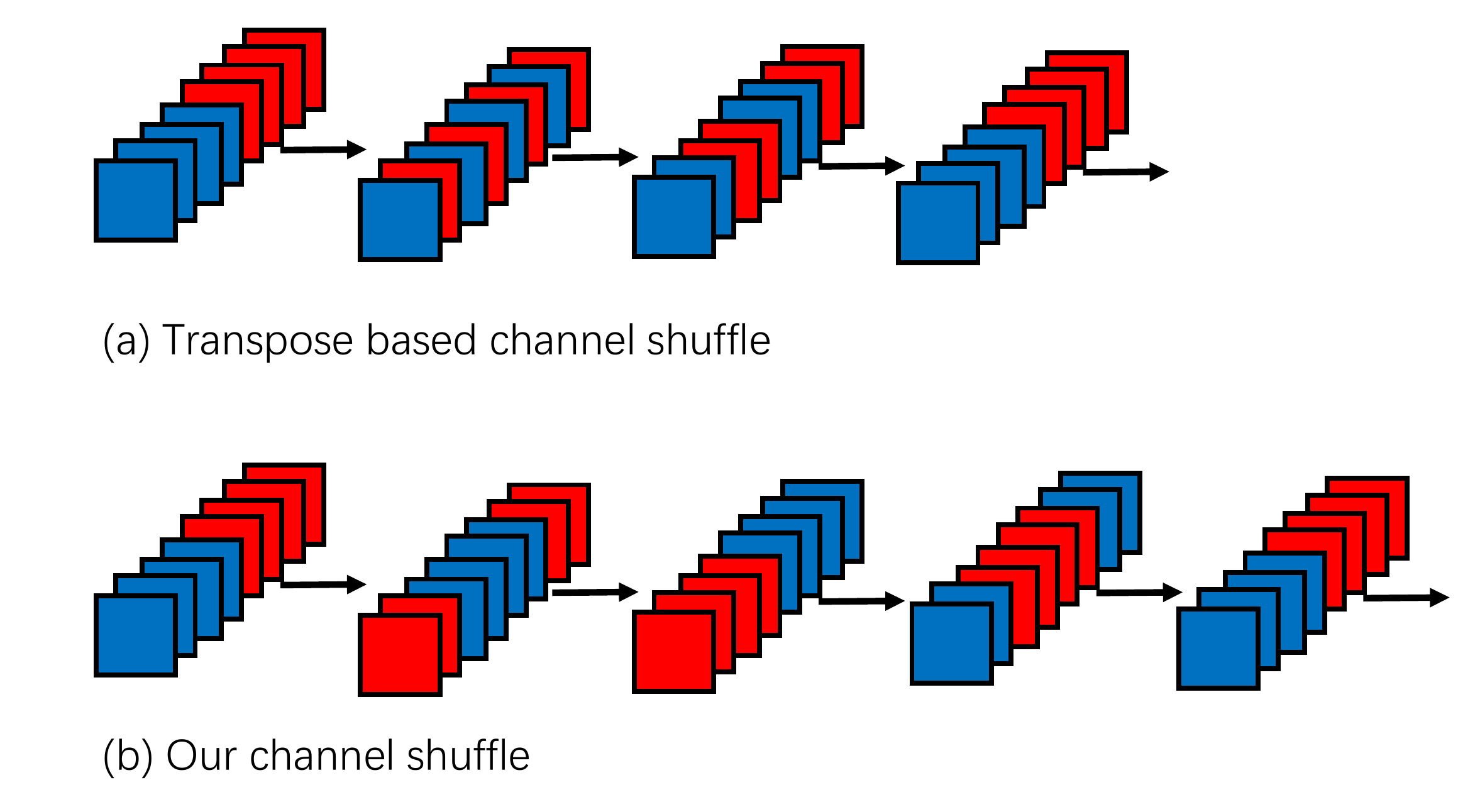}
\caption[Transpose based shuffle vs. proposed shuffle.]{Transpose based shuffle (ShuffleNetV2) vs. our HW efficient shuffle (DiracDeltaNet)}
\label{fig:shuffle}
\end{center}
\end{figure}

We also modify the channel shuffle's order to make it more hardware efficient. ShuffleNetV2 uses transpose operation to mix channels from two branches. This is illustrated in Fig. \ref{fig:shuffle}(a), where blue and red rectangles represent channels from different branches. The transpose based shuffling is not hardware friendly since it breaks the contiguous data layout. Performing channel shuffle in this manner will require multiple passes of memory read and write. We propose a more efficient channel shuffle showed in Fig. \ref{fig:shuffle}(b). We perform a circular shift to the feature map along the channel dimension. We can have the same number of channels exchanged between two branches while preserving the contiguity of the feature map and minimizing the memory accesses.

We name the modified ShuffleNetV2-1.0x model as DiracDeltaNet. The name comes from the fact that our network only contains 1$\times$1 convolutions. With a kernel size of 1, the kernel functions can be seen as discrete 2D Dirac Delta functions. DiracDeltaNet's macro-structure is summarized in Table \ref{tab:dirac-macro}. Stage 2,3,4 consist of chained DiracDeltaNet blocks depicted in Fig. \ref{fig:netblocks} with different feature map size, channel size and stride. We adopt the training recipe and hyperparameters described in \cite{ma2018shufflenet}. We train DiracDeltaNet for 90 epoch with linear learning rate decay, the initial learning rate of 0.5, 1024 batch size and 4e-5 weight decay. A comparison between ShuffleNetV2-1.0x and our DiracDeltaNet is summarized in Table \ref{tab:net-compare}.

\begin{table}[]
\centering
\caption{Macro-structure of DiracDeltaNet}
\begin{tabular}{c|ccccc}
\hline
Layer                                                                                     & \begin{tabular}[c]{@{}c@{}}Output\\ size\end{tabular}                   & \begin{tabular}[c]{@{}c@{}}Kernel\\ size\end{tabular}       & Stride                                                        & \#Repeat                                                      & \begin{tabular}[c]{@{}c@{}}Output \\ channel\end{tabular} \\ \hline
Image                                                                                     & 224                                                                     &                                                             &                                                               &                                                               & 3                                                         \\ \hline
\begin{tabular}[c]{@{}c@{}}Conv1\\ Maxpool\\ shift\\ Conv2\\ Maxpool\\ shift\end{tabular} & \begin{tabular}[c]{@{}c@{}}224\\ 112\\ 112\\ 112\\ 56\\ 56\end{tabular} & \begin{tabular}[c]{@{}c@{}}1\\ 2\\3\\1\\ 2\\ 3\end{tabular} & \begin{tabular}[c]{@{}c@{}}1\\ 2\\ 1\\ 1\\ 2\\ 1\end{tabular} & \begin{tabular}[c]{@{}c@{}}1\\ 1\\ 1\\ 1\\ 1\\ 1\end{tabular} & \begin{tabular}[c]{@{}c@{}}32\\ \\ \\ 64\end{tabular}     \\ \hline
Stage 2                                                                                    & \begin{tabular}[c]{@{}c@{}}28\\ 28\end{tabular}                         &                                                             & \begin{tabular}[c]{@{}c@{}}2\\ 1\end{tabular}                 & \begin{tabular}[c]{@{}c@{}}1\\ 3\end{tabular}                 & 128                                                       \\ \hline
Stage 3                                                                                   & \begin{tabular}[c]{@{}c@{}}14\\ 14\end{tabular}                         &                                                             & \begin{tabular}[c]{@{}c@{}}2\\ 1\end{tabular}                 & \begin{tabular}[c]{@{}c@{}}1\\ 7\end{tabular}                 & 256                                                       \\ \hline
Stage 4                                                                                   & \begin{tabular}[c]{@{}c@{}}7\\ 7\end{tabular}                           &                                                             & \begin{tabular}[c]{@{}c@{}}2\\ 1\end{tabular}                 & \begin{tabular}[c]{@{}c@{}}1\\ 3\end{tabular}                 & 512                                                       \\ \hline
Conv5   & 7    & 1   & 1     & 1    & 1024    \\ \hline
GlobalPool   & 1   & 7   &   &  1    & 1024 \\ \hline
FC     &   &    &   & 1   & 1000   \\ \hline
\end{tabular}
\label{tab:dirac-macro}
\vspace{-5pt}
\end{table}

\begin{table}[]
\centering
\caption{ShuffleNetV2-1.0x vs. DiracDeltaNet}
\begin{tabular}{c|cccc}
\hline
                & MACs & \#Params & Top-1 acc & Top-5 acc \\ \hline
ShuffleNetV2-1.0x    & 146M  &  2.3M   & 69.4\%       &   -        \\ \hline
DiracDeltaNet & 330M  &  3.3M    & 68.9\%     &   88.7\%     \\ \hline
\end{tabular}
\label{tab:net-compare}
\end{table}

\subsection{Quantization}
\label{Quantization}

To further reduce the cost of DiracDeltaNet, we apply quantization to convert floating-point weights and activations to low-precision integer values. For network weights, we follow DoReFa-Net \cite{zhou2016dorefa} to quantize full-precision weights as
\begin{equation}
w_k = 2Q_k(\frac{\tanh (w)}{2\text{max}(|\tanh (w)|)} + 0.5)-1.
\end{equation}
Here, $w$ denotes the latent full-precision weight of the convolution kernel. $Q_k(\cdot)$ is a function that quantizes its input in the range of $[0, 1]$ to its nearest neighbor in $\{\frac{i}{2^k-1}| i=0,\cdots 2^{k-1}\}$.  

We follow PACT \cite{choi2018pact} to quantize each layer's activation as
\begin{equation}
\begin{gathered}
{y^l} = PACT\left( {{x^l}} \right) = \frac{{\left| {{x^l}} \right| - \left| {{x^l} -  {{\alpha ^l}} } \right| + {{\alpha ^l}}}}{2},\\
{y^l} = {Q_k}\left( {{y^l}/ {{\alpha ^l}}} \right) \cdot {{\alpha ^l}} .
\end{gathered}
\end{equation}
$x^l$ is the activation of layer-$l$. $PACT(\cdot)$ is a function that clips the activation $x^l$ to the range between $[0, { {{\alpha ^l}} }]$. $\alpha^l$ is a layer-wise trainable upper bound, determined by the training of the network. $y^l$ is the clipped activation from layer-$l$ and it is further quantized to $y_k^l$, a k-bit activation tensor. Note that activations from the same layer share the same floating-point coefficient $\alpha^l$, but activations from different layers can have different coefficients. This is problematic for the concatenative skip connection, since if the coefficients $\alpha^l$ and $\alpha^{l-1}$ are different, we need to first cast $y_k^{l-1}$ and $y_k^{l}$ from fixed-point to floating-point, re-calculate a coefficient for the merged activation, and quantize it again to new fixed-point numbers. This process is very inefficient. 

In our experiment, we notice that most of the layers in the DiracDeltaNet have similar coefficients with values. Therefore, we rewrite equation (\ref{eqn:act_quant}) as 
\begin{equation}
\label{eqn:act_quant_new}
{y^l} = {Q_k}\left( {{y^l}/ {{\alpha ^l}}} \right) \cdot {{s} }.
\end{equation}
where $s$ is a coefficient shared by the entire network. This step ensures that activations from different layers of the network are quantized and normalized to the same scale of $[0, {{s}}]$. As a result, we can concatenate activations from different layers directly without extra computation. Moreover, by using the same coefficient of $s$ across the entire network, the convolution can be computed completely via fixed-point operations. The coefficient $s$ can be fixed before or leave it as trainable. A general rule is that we should let $s$ have similar values of $\alpha^l$ from different layers. Otherwise, if $s/\alpha^l$ is either too small or too large, it can cause gradient vanishing or exploding problems in training, which leads to a worse accuracy of the network.

In our network, we merge the PACT function and activation quantization into one module and name it ActQuant. The input to ActQuant is the output of 1$\times$1 convolutions. Since the input and weight of the convolution are both quantized into fixed-point integers, the output is also integers. Then, ActQuant is implemented as a look-up-table whose parameters are determined during training and fixed during inference.

\begin{figure}[!t]
\begin{center}
\centering 
\includegraphics[width=0.9\linewidth]{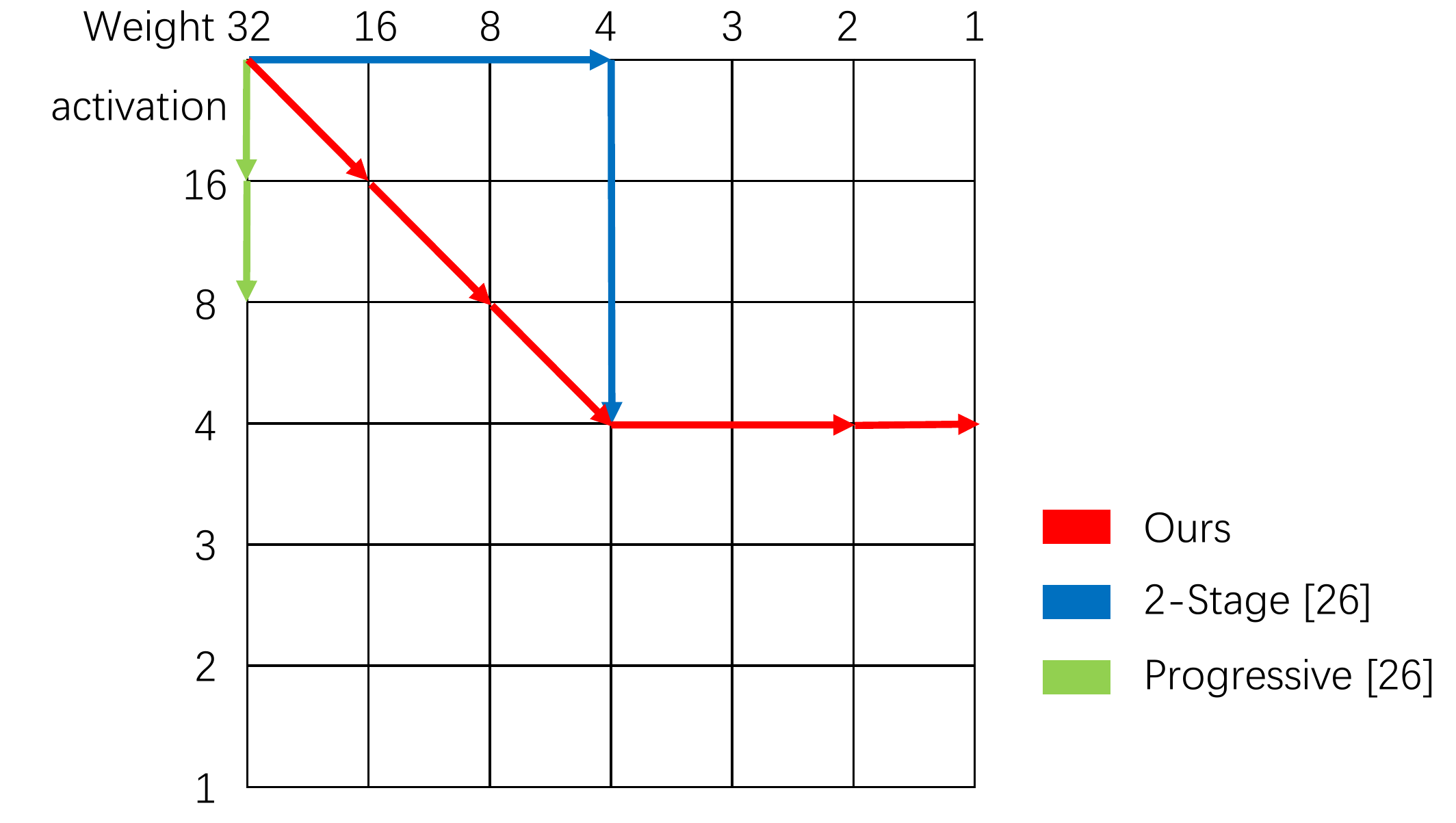}
\caption{Progressive quantization schedule.}
\label{fig:grid}
\end{center}
\end{figure}

We follow \cite{Zhuang2017progressive} to quantize the network progressively from full-precision to the desired low-precision numbers. The process is illustrated in Fig. \ref{fig:grid}, where x-axis denotes bit-width of weights and y-axis denotes the bit-width of activations. We start from the full-precision network, train the network to convergence, and follow a path to progressively reduce the precision for weights or activations. At each point, we fine-tune the network for 50 epochs with step learning rate decay. Formally, we denote each point in the grid as a quantization configuration ${\mathscr{C}_{w,a}}\left( N_{w} \right)$. Here $w$ represents the bitwidth of weight. $a$ is the bitwidth of activation. $N_w$ is the network containing the quantized parameters. The starting configuration would be the full precision network ${\mathscr{C}_{32,32}}\left( N_{32} \right)$. Starting from this configuration, one can either go down to quantize the activation or go right to reduce the bitwidth of weight. More aggressive steps can be taken diagonally or even across several grids. The two-stage and progressive optimization methods proposed in \cite{Zhuang2017progressive} can be represented as two paths in Fig. \ref{fig:grid}. 

In our work, we start from ${\mathscr{C}_{32,32}}\left( N_{32} \right)$. Then we use $N_{32}$ to initialize $N_{16}$ and obtain ${\mathscr{C}_{16,16}}\left( N_{16} \right)$. And we apply step lr decay fine-tuning onto $N_{16}$ to recover the accuracy loss due to the quantization. After several epochs of fine-tuning, we get the desired low-precision configuration ${\mathscr{C}_{16,16}}\left( N_{16}' \right)$ with no accuracy loss. Following the same procedures, we are able to first go diagonally in the quantization grid to ${\mathscr{C}_{4,4}}\left( N_{4} \right)$ with less than 1\% top-5 accuracy loss compared to its full precision counterpart. 

\begin{table}
\caption{Quantization result on DiracDeltaNet}
\centering
\begin{tabular}{c|c|c}
\hline
 & full   & w4a4  \\ 
\hline
Top-1 Acc & 68.9\% & 68.3\%  \\
Top-5 Acc & 88.7\% & 88.1\% \\
\hline
\end{tabular}
\label{tab:quantization_result}
\end{table}

We use a pre-trained ResNet50 label-refinery \cite{bagherinezhad2018label} to boost the accuracy of the quantized model. Even with such low-precision quantization, our quantized model still preserves a very competitive top-5 accuracy of 88.1\%. Most of the previous quantization works \cite{choi2018pact, Zhuang2017progressive, zhou2016dorefa} are only effective on large models such as VGG16, AlexNet or ResNet50. Our quantization result is summarized in Table \ref{tab:quantization_result}.

\section{Synetgy: co-designing a hardware accelerator for DiracDeltaNet}
\subsection{Synetgy}
We aggressively simplified ShuffleNetV2's operator set. Our modified network is composed of the following operators: $1\times1$ convolution, $2\times2$ max-pooling, shift, shuffle, and concatenation. 1x1 convolution is the only computational operator. 

This allows us to co-design a simplified hardware accelerator, Synetgy, and we tailor it to support the operators above and improve hardware efficiency. Since DiracDeltaNet does not contain spatial convolutions, we built a dedicated compute unit only for 1x1 convolution. The compute of the fully-connected layer can also be mapped onto our convolution unit. Shuffle operation is not fully supported on FPGA. CPU-based memory copy is needed to maintain the memory layout. And the remaining average-pooling layer is offloaded to the ARM processor on the SoC platform.

\begin{figure}[h]
\centering
\includegraphics[width=.9\linewidth]{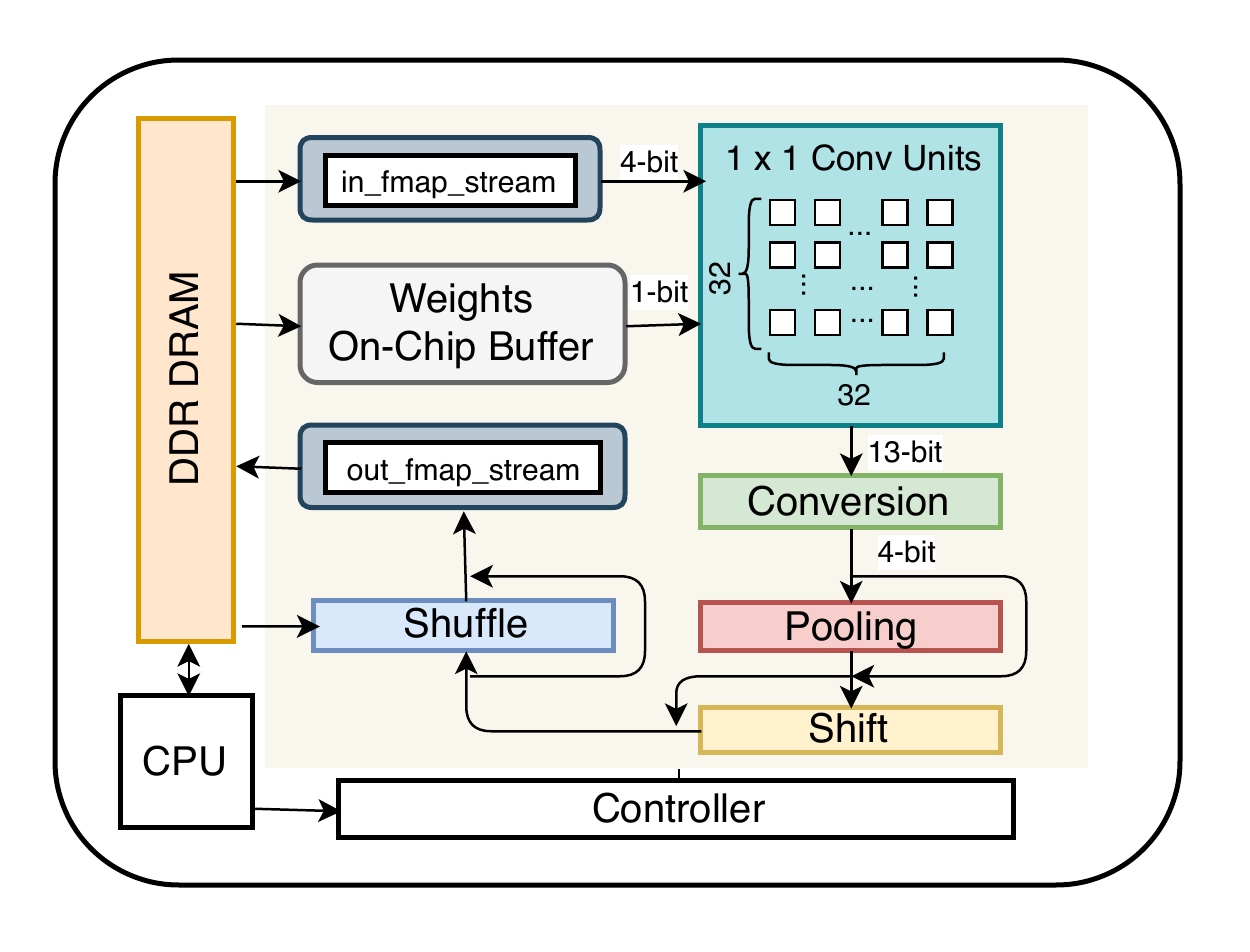}
\caption{Accelerator architecture of synetgy.}
\label{fig:accel_arch}
\end{figure}

Fig. \ref{fig:accel_arch} shows the overall accelerator architecture design. Our accelerator, highlighted in light yellow, can be invoked by the CPU for computing one $1\times1$ Conv-Pooling-Shift-Shuffle subgraph at a time. The CPU provides supplementary support to the accelerator. Due to its simplicity, the accelerator implementation only took two people one month to finish. For more details about the hardware design, we refer readers to \cite{yang2018synetgy}.

\subsection{Experimental Results}

We implement our accelerator, Synetgy, on an Ultra96 development board with Xilinx Zynq UltraScale+ MPSoC targeted at embedded applications. Our implementation runs at 250 MHz. Power measurements are obtained via a power monitor. We measured 5.3W with no workload running on the programming logic side and 5.5W max power on the Ultra96 power supply line when running our network.

\begin{table*}[ht]
\centering
\caption{Performance comparison of Synetgy and previous works}
\begin{tabular}{c|ccccccc}
\hline
\textbf{}         & \textbf{\cite{qiu2016going}} & \textbf{\cite{liang2018fp}} & \textbf{\cite{suda2016throughput}} & \textbf{\cite{guo2017software}} & \textbf{\cite{jiao2017accelerating}} & \textbf{ \cite{blott2018finnr}} & \textbf{Ours}   \\ \hline
Model & VGG-SVD & AlexNet & VGG16 & VGG16 & DoReFa & DoReFa & \thead{Dirac- \\ -DeltaNet} \\
Platform  & \thead{Zynq \\ XC7Z045}   & \thead{Stratix \\V}  & \thead{Stratix \\V}    & \thead{Zynq \\ 7Z020}   & \thead{ Zynq \\ 7Z020}   & \thead{Zynq \\ ZU3EG}  & \thead{ Zynq \\ ZU3EG }  \\ 
Speed (fps)  & 4.5  & \textbf{864.7}  & 3.8   & 5.7  & 106.0  & 200.0  & 66.3 \\ 
Top-1 Acc & 64.64   & 42.90  & 66.58  & 67.72  & 46.10  & 50.3  & \textbf{68.30}  \\ 
Top-5 Acc & 86.66  & 66.80  & 87.48  & 88.06   & 73.10  & N/A  & \textbf{88.12} \\ 
Precision & 16b  & 16b  & 8-16b  & 8b  & 2b  & 1-2b  & 4-4b \\ 
Power (W)  & 3.0 & 26.2 & 19.1  & 3.0  & 2.3 & 10.2  & 5.5 \\ \hline
\end{tabular}
\label{tab:time}
\end{table*}

\begin{table}[]
\centering
\caption{Frame rate with different batch sizes}
\begin{tabular}{|c|cccccc|}
\hline 
Batch Size  & 1 & 2 & 4 & 8 & 10 & 16 \\ \hline
Frame Rate (fps)  & 58.7  & 72.9 & 84.1 & 94.4 & 95.9 & 96.5 \\    
\hline
\end{tabular}
\label{tab:batchsize}
\end{table}

We compare our accelerator against previous work in Table~\ref{tab:time}. ConvNets for ImageNet classification are usually orders of magnitude more complex than CIFAR10 classification. Therefore, we only compare accelerators targeting ConvNets for ImageNet classification with reasonable accuracy. Our work focuses on achieving competitive accuracy while improving the actual inference speed in terms of frames per second. Our experiments show that we successfully achieve those two goals. From the table, we can make the following observations: 
1) Synetgy achieves the highest top-1 and top-5 accuracy on ImageNet. The only previous work that comes close to our accuracy is \cite{guo2017software}, but its frame rate is 11.6$\times$ slower than ours. 
2) Among the embedded accelerators whose top-1 accuracy is higher than 60\%, which is a loose constraint, our model achieves the fastest inference speed. 
3) Without the accuracy constraint, the speed of \cite{liang2018fp, jiao2017accelerating, blott2018finnr} can go as fast as 864.7 frames per second. But their accuracy is rather low.
4) The peak attainable throughput of our accelerator is 418 GOPs, which is close to the theoretical compute roofline.  
Our average throughput (47.09 GOPs) is currently limited by the low hardware utilization. The inefficiency is mainly from the software shuffle operations and the first convolution layer whose input dimension is 3, which is much less than the hardware tiling factor $IC$.
However, Synetgy still achieves competitive frame rate, demonstrating the efficacy of our co-design methodology. We see the opportunity of significant frame rate improvement through further algorithm-hardware co-design.

\section{Conclusion}
In this chapter, we discuss the hardware efficiency of deep neural networks.We show that we can achieve significant improvement of efficiency by of co-designing efficient neural networks and hardware processors. We started from designing efficient neural networks by replacing spatial convolutions. Spatial convolution was regarded as an irreplaceable part in convolution neural networks, since CNNs rely on spatial convolutions to aggregate spatial information. However, spatial convolutions are also computationally expensive and the cost grows quadratically with the kernel size. In this chaper, we show that spatial convolution is not the only operator to aggregate spatial information. We present the shift operator, a zero-FLOP, zero-parameter operator to achieve the same goal as spatial convolutions. It not only reduces parameters and FLOPs of a neural network significantly in a wide range of applications, more importantly, it greatly simplifies the operators involved in a CNN. This opens a new opportunity of for developing highly customized hardware accelerators optimized for 1x1 convolutions. Applying this philosophy into reality, we design a new network called DiracDeltaNet, and an embedded FPGA-based accelerator called Synetgy. Due to the operator simplification, hardware designers can devote all the resources on the board to 1x1 convolutions. Such optimized implementation leads to 11.2x speedup over the previous state-of-the-art while achieving higher accuracy. The idea of the shift operator is further improved by later works to achieve better efficiency \cite{chen2019all, he2019addressnet}, enable new software-hardware co-design \cite{kungmaestro}, and is extended to process temporal information \cite{lin2018temporal}. 

%% file: chap8.tex
\chapter{Design efficiency: Differentiable Neural Architecture Search}
\label{chap:dnas}

In this Chapter, we discuss the \textbf{design efficiency} of deep neural networks. The performance of a neural network is highly dependent on its model architecture. However, finding the optimal architecture is a difficult problem, since the design space for neural architectures is typically combinatorially large. The conventional approach for neural network design is to manually and iteratively modify neural architectures, which can take weeks or months since we typically need to explore hundreds or even thousands of variants and training each of them takes a long time. This motivates us to consider the following \textbf{key question}:
\begin{quote}
    Can we reduce the cost of designing neural networks by
    automatically exploring the design space?
\end{quote}
In this chapter, we discuss a new algorithm called Differentiable Neural Architecture Search (DNAS) for automatic and fast neural network design. When applied to two different problems, mixed-precision quantization and efficient ConvNet search, DNAS discovers neural architectures that surpass the previous state-of-the-art models designed manually and automatically, while the design process is hundreds of times faster than previous automatic design pipelines. 

\section{Introduction}
In many computer vision tasks, a better neural network architecture design usually leads to a significant accuracy improvement. In previous works accuracy improvement came at the cost of higher computational complexity, making it more challenging to deploy neural networks to mobile devices, where computing capacity is limited. Instead of solely focusing on accuracy, recent work also aims to optimize for efficiency, especially latency. However, designing efficient and accurate neural networks is a challenging task due to the following reasons.

\begin{figure}[!t]
\begin{center}
    \begin{subfigure}[b]{0.8\textwidth}
        \centering\includegraphics[width=\linewidth]{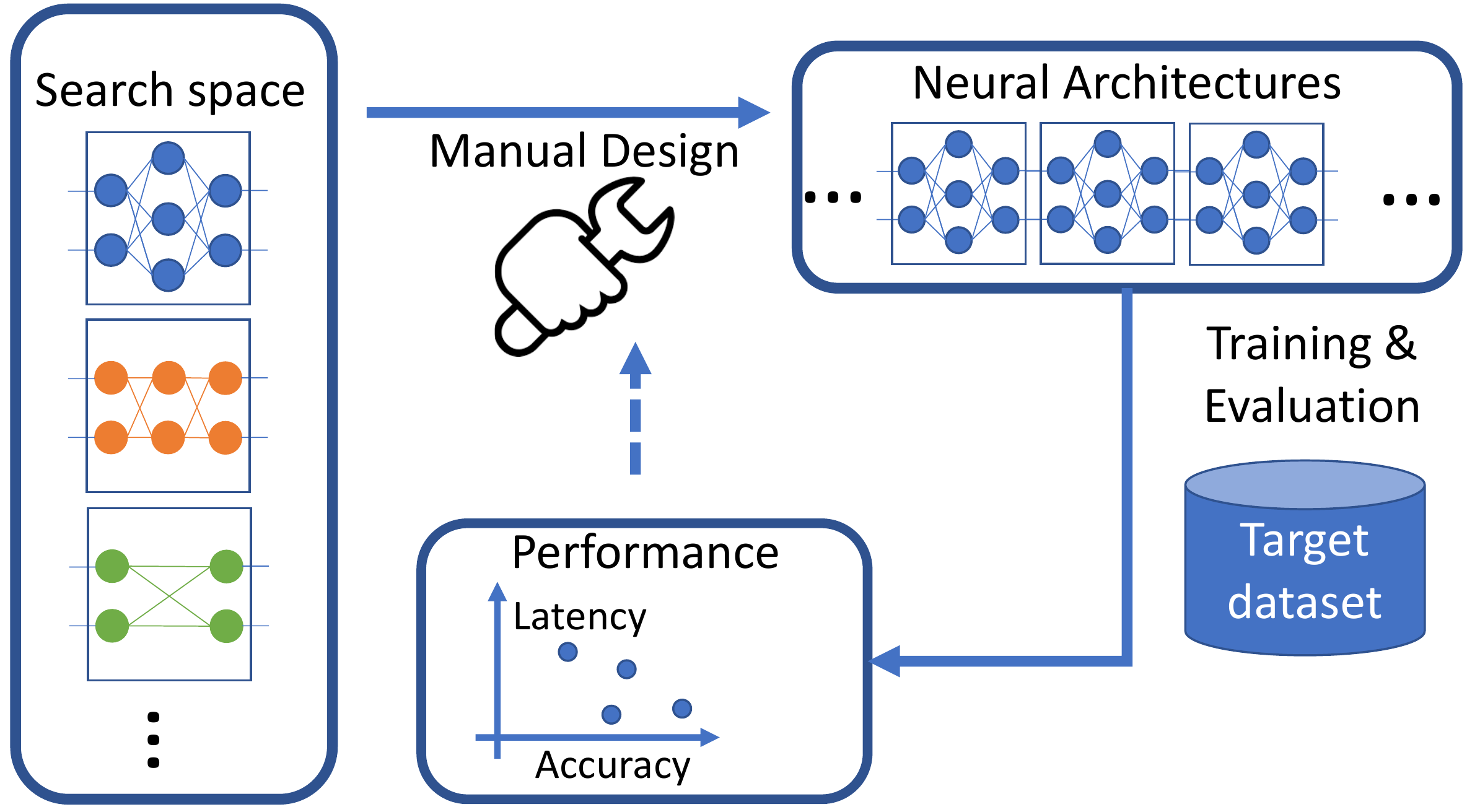}
        \caption{A typical flow of manual design.}
        \label{fig:manual_flow}
    \end{subfigure} \hfill
    \begin{subfigure}[b]{0.8\textwidth}
        \centering\includegraphics[width=\linewidth]{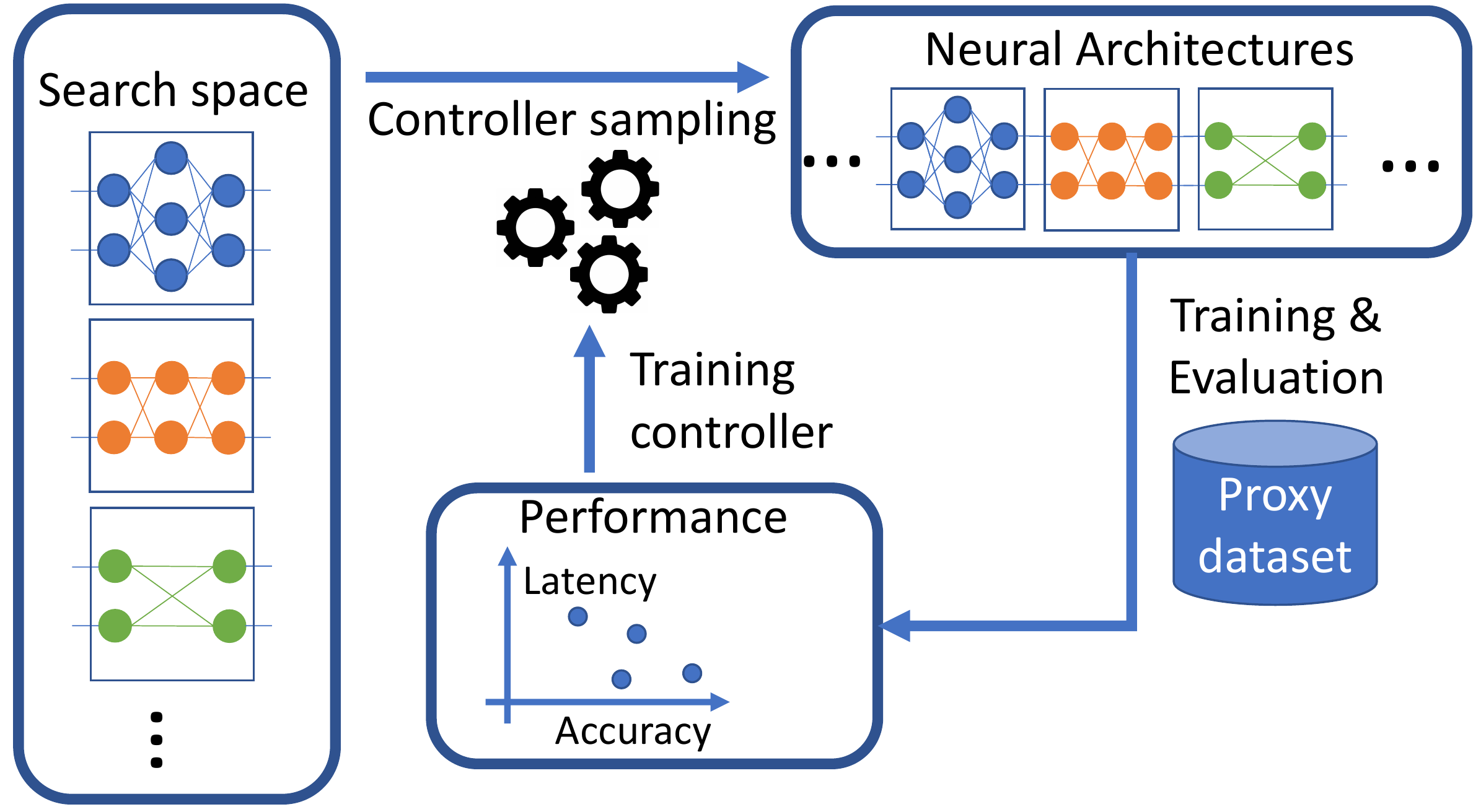}
        \caption{A typical flow of reinforcement learning based neural architecture search.}
        \label{fig:rl_nas_flow}
    \end{subfigure} \hfill
\caption[Manual design and reinforcemenet learning based NAS.]{Illustration of manual design and reinforcement learning based neural architecture search.}
\label{fig:baseline_design_flow}
\end{center}
\end{figure}

\textbf{Intractable design space}: The design space of a neural network is combinatorially large. Using VGG16 \cite{simonyan2014very} as a motivating example: VGG16 contains 16 layers. Assume for each layer of the network, we can choose a different kernel size from $\{1, 3, 5\}$ and a different filter number from $\{32, 64, 128, 256, 512\}$. Even with such simplified design choices and shallow layers, the design space contains $(3\times 5)^{16} \approx 6\times 10^{18}$ possible architectures. However, training a neural network is very time-consuming, typically taking days or even weeks. As a result, previous neural network design rarely explores the design space in a comprehensive way. A typical flow of manual design is illustrated in Figure \ref{fig:manual_flow}. Designers propose initial architectures and train them on the target dataset. Based on the performance, designers evolve the architectures accordingly. Limited by the time cost of training, the design flow has to stop after a few iterations, which is far too few to sufficiently explore the design space. 

Starting from \cite{zoph2016neural}, recent works adopt neural architecture search (NAS) to explore the design space automatically. Many previous works \cite{zoph2016neural,zoph2017learning, tan2018mnasnet} use reinforcement learning (RL) to guide the search and a typical flow is illustrated in Figure \ref{fig:rl_nas_flow}. A controller samples architectures from the search space to be trained. To reduce the training cost, sampled architectures are trained on a smaller proxy dataset such as CIFAR-10 or trained for fewer epochs on ImageNet. The performance of the trained networks is then used to train and improve the controller. Previous works \cite{zoph2016neural,zoph2017learning, tan2018mnasnet} have demonstrated the effectiveness of such methods in finding accurate and efficient networks. However, training each architecture is still time-consuming, and it usually takes thousands of architectures to train the controller. As a result, the computational cost of such methods is prohibitively high.

\textbf{Nontransferable optimality}: the optimality of neural architectures is conditioned on many factors such as input resolutions and target devices. Once these factors change, the optimal architecture is likely to be different. A common practice to reduce the FLOP count of a network is to shrink the input resolution. A smaller input resolution may require a smaller receptive field of the network and therefore shallower layers. On a different device, the same operator can have a different latency, so we need to adjust the architecture to achieve the best accuracy-efficiency trade-off. Ideally, we should design different architectures case-by-case. In practice, however, limited by the computational cost of previous manual and automated approaches, we can only realistically design one architecture and use it for all conditions.

\textbf{Inconsistent efficiency metrics}: Most of the efficiency metrics we care about are dependent on not only the architecture but also the hardware and software configurations on the target device. Such metrics include latency, power, and energy - in this work, we mainly focus on latency. To simplify the problem, most of the previous works adopt hardware-agnostic metrics such as FLOPs (more strictly, the number of multiply-add operations) to evaluate efficiency. However, a network with lower FLOP count is not necessarily faster. For example, NasNet-A \cite{zoph2017learning} has a similar FLOP count to MobileNetV1 \cite{howard2017mobilenets}, but its complicated and fragmented cell-level structure is not hardware friendly, so the actual latency is higher \cite{sandler2018mobilenetv2}. The inconsistency between hardware agnostic metrics and actual efficiency makes the design more difficult.

\begin{figure}[h]
\begin{center}
\includegraphics[width=1.\linewidth]{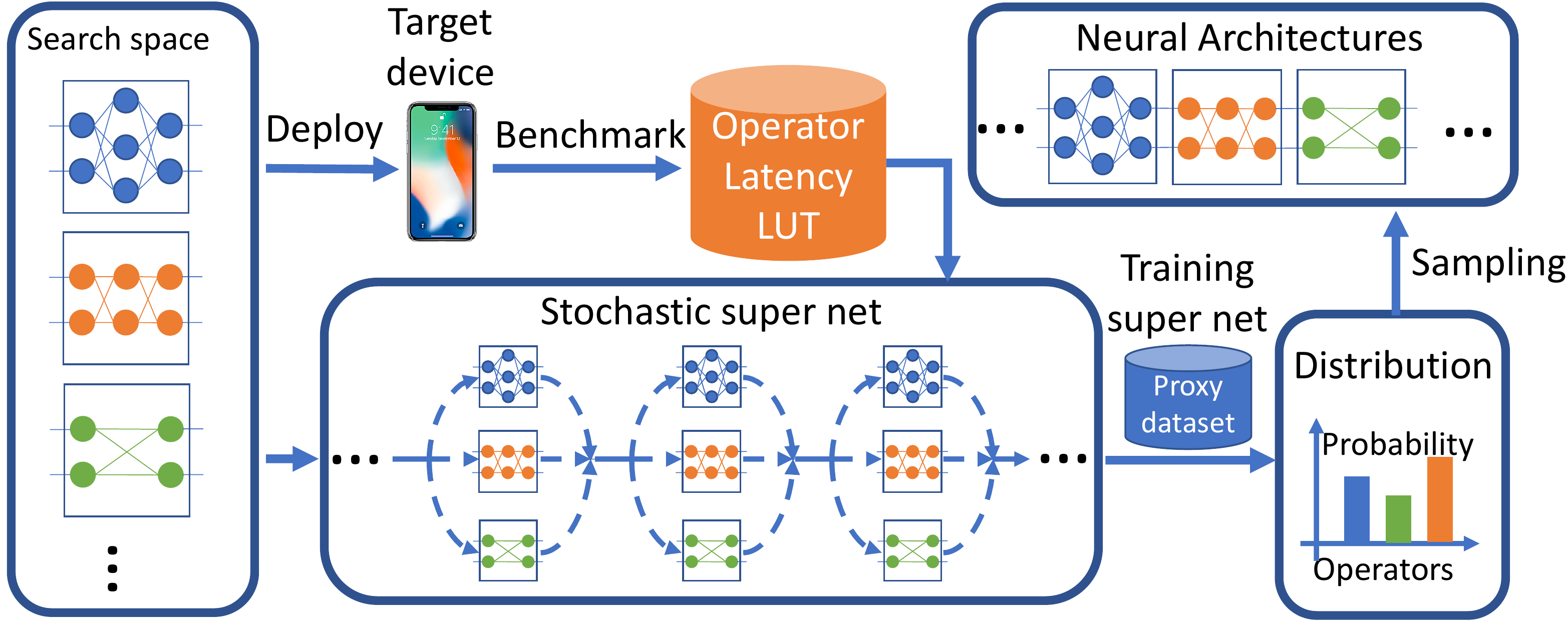}
\end{center}
\caption[DNAS for ConvNet design.]{Differentiable neural architecture search (DNAS) for ConvNet design. DNAS explores a layer-wise space where each layer of a ConvNet can choose a different block. The search space is represented by a stochastic super net. The search process trains the stochastic super net using SGD to optimize the architecture distribution. Optimal architectures are sampled from the trained distribution. The latency of each operator is measured on target devices and used to compute the loss for the super net.}
\label{fig:dnas_flow}
\end{figure}

To address the above problems, we propose to use differentiable neural architecture search (DNAS) to discover hardware-aware efficient ConvNets for computer vision problems. The flow of our algorithm is illustrated in Figure \ref{fig:dnas_flow}. DNAS allows us to explore a layer-wise search space where we can choose a different block for each layer of the network. Following \cite{veniat2017learning}, DNAS represents the search space by a super net whose operators execute stochastically. We relax the problem of finding the optimal architecture to find a distribution that yields the optimal architecture. By using the Gumbel Softmax technique \cite{jang2016categorical}, we can directly train the architecture distribution using gradient-based optimization such as SGD. The search process is extremely fast compared with previous reinforcement learning (RL) based methods. The loss used to train the stochastic super net consists of both the cross-entropy loss that leads to better accuracy and the latency loss that penalizes the network's latency on a target device. To estimate the latency of an architecture, we measure the latency of each operator in the search space and use a lookup table model to compute the overall latency by adding up the latency of each operator. Using this model allows us to quickly estimate the latency of architectures in this enormous search space. More importantly, it makes the latency differentiable with respect to layer-wise block choices.

We applied DNAS to solve two different problems. The first one is mixed-precision quantization. For a deep neural network with $N$ layers and $M$ candidate precisions in each layer, we want to find an optimal assignment of precisions to minimize the cost in terms of model size, memory footprint or computation, while keeping the accuracy. An exhaustive combinatorial search has exponential complexity ($\mathcal{O}(M^N)$). Using DNAS to search for layer-wise quantization strategies for ResNet on CIFAR10 and ImageNet, we surpass the state-of-the-art compression. Our quantized models with 21.1x smaller model size or 103.9x smaller computational cost can still outperform baseline quantized or even full precision models. 

In the second problem, we apply DNAS to search for efficient ConvNet architectures. We name the models discovered by DNAS as FBNets (\textit{\textbf{F}acebook-\textbf{B}erkeley-\textbf{Net}s}). FBNets surpass the state-of-the-art efficient ConvNets designed manually and automatically. FBNet-B achieves 74.1\% top-1 accuracy with 295M FLOPs and 23.1 ms latency on an Samsung S8 phone, 2.4x smaller and 1.5x faster than MobileNetV2-1.3. Being better than MnasNet, FBNet-B's search cost is 216 GPU-hours, 421x lower than the cost for MnasNet estimated based on \cite{tan2018mnasnet}. Such a low search cost enables us to re-design ConvNets case-by-case. For different resolution and channel scaling, FBNets achieve 1.5\% to 6.4\% absolute gain in top-1 accuracy compared with MobileNetV2 models. The smallest FBNet achieves 50.2\% accuracy and 2.9 ms latency (345 frames per second) with a batch size of 1 on Samsung S8. Using DNAS to search for device-specific ConvNet, an iPhone-X-optimized model achieves 1.4x speedup on an iPhone X compared with a Samsung-optimized model.

\section{Related work}
\textbf{Efficient ConvNet models}: Designing efficient ConvNet has attracted much research attention in recent years. SqueezeNet \cite{iandola2016squeezenet} is one of the early works focusing on reducing the parameter size of ConvNet models. It is originally designed for classification, but later extended to object detection \cite{wu2017squeezedet} and LiDAR point-cloud segmentation \cite{wu2018squeezeseg, wu2018squeezesegv2}. Following SqueezeNet, SqueezeNext \cite{gholami2018squeezenext} and ShiftNet \cite{wu2017shift} achieve further parameter size reduction. Recent works change the focus from parameter size to FLOPs. MobileNetV1 and MobileNetV2 \cite{howard2017mobilenets, sandler2018mobilenetv2} use depthwise convolutions to replace the more expensive spatial convolutions. ShuffleNet \cite{zhang1707shufflenet} uses group convolution and shuffle operations to reduce the FLOP count further. More recent works incorporate the fact that FLOP count does not always reflect the actual hardware efficiency. To improve actual latency, ShuffleNetV2 \cite{ma2018shufflenet} proposes a series of practical guidelines for efficient ConvNet design. Synetgy \cite{yang2018synetgy} combines ideas from ShuffleNetV2 and ShiftNet to co-design hardware friendly ConvNets and FPGA accelerators. 

\textbf{Network quantization} received a lot of research attention in recent years. Early works such as \cite{han2015deep, zhu2016trained, leng2017extremely} mainly focus on quantizing neural network weights while still using 32-bit activations. Quantizing weights can reduce the model size of the network and therefore reduce storage space and over-the-air communication cost. More recent works such as \cite{rastegari2016xnor,zhou2016dorefa,choi2018pact,jung2018joint, zhuang2018training} quantize both weights and activations to reduce the computational cost on CPUs and dedicated hardware accelerators. Most of the works use the same precision for all or most of the layers of a network. The problem of mixed-precision quantization is rarely explored. 

\textbf{Neural Architecture Search}: \cite{zoph2016neural, zoph2017learning} first propose to use reinforcement learning (RL) to search for neural architectures to achieve competitive accuracy with low FLOPs. Early NAS methods are computationally expensive. Recent works try to reduce the computational cost by weight sharing \cite{pham2018efficient} or using gradient-based optimization \cite{liu2018darts}. Early works on NAS \cite{zoph2017learning,pham2018efficient,liu2018darts} focus on the cell level architecture search, and the same cell structure is repeated in all layers of a network. However, such fragmented and complicated cell-level structures are not hardware friendly, and the actual efficiency is low. Most recently, \cite{tan2018mnasnet} explores a stage-level hierarchical search space, allowing different blocks for different stages of a network, while blocks inside a stage are still the same. Instead of focusing on FLOPs, \cite{tan2018mnasnet} aims to optimize the latency on target devices.  Besides searching for new architectures, works such as \cite{yang2018netadapt, he2018amc} focus on adapting existing models to improve efficiency. Such methods finetune neural architectures around a given design instead of searching from a larger design space, so the performance is limited. In comparison to prior works, our work searches a combinatorially large design space in a significantly faster speed and optimizes for actual latency for different given target devices. 

\section{Differentiable neural architecture search}
\subsection{Neural architecture search}
Formally, the problem of neural architecture search (NAS) can be formulated as
\begin{equation}
\label{eqn:nas}
\underset{a \in \mathcal{A}}{\text{min}} ~ \underset{\vw_a}{\text{min}}
~ \mathcal{L}(a, \vw_a)
\end{equation}
Here, $a$ denotes a neural architecture, $\mathcal{A}$ denotes the architecture space. $\vw_a$ denotes the weights of architecture $a$. $\mathcal{L}(\cdot, \cdot)$ represents the loss function on a target dataset given the architecture $a$ and its parameter $\vw_a$. The loss function is differentiable with respect to $\vw_a$, but not to $a$. As a consequence, the computational cost of solving the problem in (\ref{eqn:nas}) is prohibitively high. To solve the inner optimization problem requires training a neural network $a$ to convergence, which can take days. The outer problem has a discrete search space with combinatorial complexity. To solve the problem efficiently, the key is to avoid enumerating the search space and evaluating each candidate architecture one-by-one.

\subsection{Differentiable neural architecture search}
\begin{figure}[h]
\begin{center}
\includegraphics[width=\linewidth]{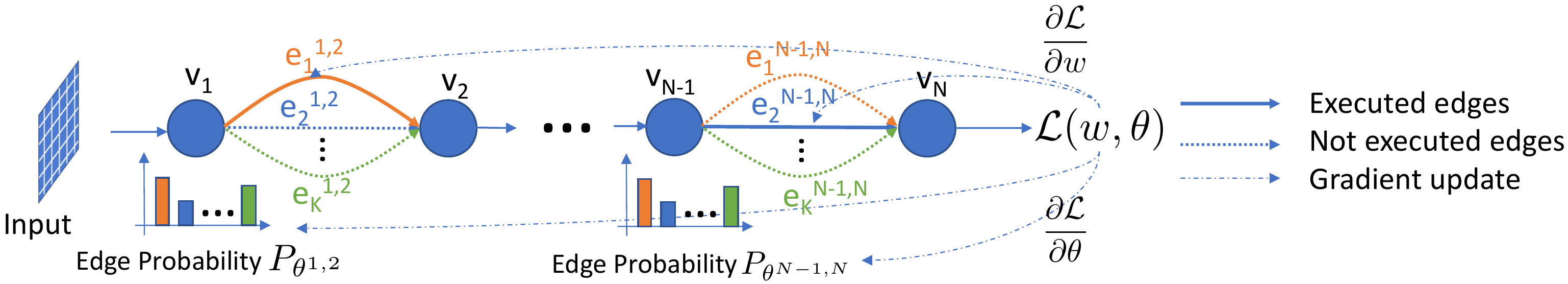}
\end{center}
\caption[Illustration of a stochastic super net.]{Illustration of a stochastic super net. Nodes represent data tensors and edges represent operators. Edges are executed stochastically following the distribution $P_\vtheta$. $\vtheta$ denotes the architecture parameter and $\vw$ denotes network weights. The stochastic super net is fully differentiable.}
\label{fig:supernet}
\end{figure}

We discuss the idea of differentiable neural architecture search (DNAS). The idea is illustrated in Fig. \ref{fig:supernet}. We start by constructing a super net to represent the architecture space $\mathcal{A}$. The super net is essentially a computational DAG (directed acyclic graph) that is denoted as $G=(V, E)$. Each node $v_i \in V$ of the super net represents a data tensor. Between two nodes $v_i$ and $v_j$, there can be $K^{ij}$ edges connecting them, indexed as $e_k^{ij}$. Each edge represents an operator parameterized by its trainable weight $w_k^{ij}$. The operator takes the data tensor at $v_i$ as its input and computes its output as $e_k^{ij}(v_i;w_k^{ij})$. To compute the data tensor at $v_j$, we sum the output of all incoming edges as
\begin{equation}
\label{eqn:super_node}
    v_j = \sum_{i, k} e_k^{ij}(v_i;w_k^{ij}).
\end{equation}
With this representation, any neural net architecture $a \in \mathcal{A}$ can be represented by a subgraph $G_a(V_a, E_a)$ with $V_a \subseteq V, E_a \subseteq E$. For simplicity, in a candidate architecture, we keep all the nodes of the graph, so $V_a = V$. And for a pair of nodes $v_i, v_j$ that are connected by $K^{ij}$ candidate edges, we only select one edge. Formally, in a candidate architecture $a$, we re-write Equation (\ref{eqn:super_node}) as 
\begin{equation}
\label{eqn:masked_node}
    v_j = \sum_{i, k} m_k^{ij} e_k^{ij}(v_i;w_k^{ij}),
\end{equation}
where $m_k^{ij}\in \{0, 1\}$ is an ``edge-mask'' and 
$\sum_k m_k^{ij} = 1$. Note that though the value of $m_k^{ij}$ is discrete, we can still compute the gradient to $m_k^{ij}$. Let $\vm$ be a vector that consists of $m_k^{ij}$ for all $e_k^{ij} \in E$.  For any architecture $a\in\mathcal{A}$, we can encode it using an ``edge-mask'' vector $\vm_a$. So we re-write the loss function in Equation (\ref{eqn:nas}) to an equivalent form as $\mathcal{L}(\vm_a, \vw_a)$. 

We next convert the super net to a stochastic super net whose edges are executed stochastically. For each edge $e_k^{ij}$, we let $\rM_k^{ij} \in \{0, 1\}$ be a random variable and we execute edge $e_k^{ij}$ when $\rM_k^{ij}$ is sampled to be 1. We assign each edge a parameter $\theta_k^{ij}$ such that the probability of executing $e_k^{ij}$ is
\begin{equation}
    P_{\vtheta^{ij}} (\rM_k^{ij}=1) = \text{softmax}(\theta_k^{ij} , \vtheta^{ij}) = \frac{\exp(\theta_k^{ij})}{\sum_{k=1}^{K^{ij}} \exp(\theta_k^{ij})}.
\end{equation}
The stochastic super net is now parameterized by $\vtheta$, a vector whose elements are $\theta_k^{ij}$ for all $e_k^{ij} \in E$. From the distribution $P_\vtheta$, we can sample a mask vector $\vm_a$ that corresponds to a candidate architecture $a \in \mathcal{A}$. We can further compute the expected loss of the stochastic super net as $\mathbb{E}_{\ra \sim P_\vtheta} \left[\mathcal{L}(\vm_a, \vw_a)\right]$. 
The expectation of the loss function is differentiable with respect to $\vw_a$, but not directly to $\vtheta$, since we cannot directly back-propagate the gradient to $\vtheta$ through the discrete random variable $\vm_a$. To estimate the gradient, we can use Straight-Through estimation \cite{bengio2013estimating} or REINFORCE \cite{williams1992simple}. Our final choice is to use the Gumbel Softmax technique \cite{jang2016categorical}, which will be explained in the next section. Now that the expectation of the loss function becomes fully differentiable, we re-write the problem in Equation (\ref{eqn:nas}) as 
\begin{equation}
\label{eqn:stochastic_nas}
\underset{\vtheta }{\text{min}} ~ \underset{\vw_a}{\text{min}}
~ \mathbb{E}_{\ra \sim P_\vtheta} \left[\mathcal{L}(\vm_a, \vw_a)\right]
\end{equation}
The combinatorial optimization problem of solving for the optimal architecture $a \in \mathcal{A}$ is relaxed to solving for the optimal architecture-distribution parameter $\vtheta$ that minimizes the expected loss. Once we obtain the optimal $\vtheta$, we acquire the optimal architecture by sampling from $P_\vtheta$.

\subsection{DNAS with Gumbel Softmax}
We use stochastic gradient descent (SGD) to solve Equation (\ref{eqn:stochastic_nas}). The optimization process is also denoted as training the stochastic super net. We compute the Monte Carlo estimation of the gradient
\begin{equation}
\label{eqn:gradient}
\nabla_{\vtheta, \vw_a} \mathbb{E}_{\ra \sim P_\vtheta}  \left[\mathcal{L}(\vm_a, \vw_a) \right] \approx \frac{1}{B} \sum_{i=1}^B \nabla_{\vtheta, \vw_{a_i}}\mathcal{L}(\vm_{a_i}, \vw_{a_i}),
\end{equation}
where $a_i$ is an architecture sampled from distribution $P_\vtheta$ and $B$ is the batch size. Equation (\ref{eqn:gradient}) provides an unbiased estimation of the gradient, but it has high variance, since the size of the architecture space is orders of magnitude larger than any feasible batch size $B$. Such high variance for gradient estimation makes it difficult for SGD to converge.  

To address this issue, we use Gumbel Softmax proposed by \cite{jang2016categorical, maddison2016concrete} to control the edge selection. For a node pair $(v_i, v_j)$, instead of applying a ``hard'' sampling and execute only one edge, we use Gumbel Softmax to apply a ``soft'' sampling. We compute $\rM_k^{ij}$ as
\begin{equation}
    \label{eqn:gumbel_softmax}
    \rM_k^{ij} = \text{GumbelSoftmax} (\theta_k^{ij} | \vtheta^{ij}) = \frac{\exp((\theta_k^{ij} + \rg_k^{ij}) / \tau)}{\sum_k \exp((\theta_k^{ij} + \rg_k^{ij}) / \tau) },  ~ g_k^{ij} \sim \text{Gumbel(0, 1)}.
\end{equation}
$\rg_k^{ij}$ is a random variable drawn from the Gumbel distribution. Note that under this relaxation, $\rM_k^{ij}$ becomes a continuous random variable. It is directly differentiable with respect to $\theta_k^{ij}$ and we don't need to pass gradient through the random variable $\rg_k^{ij}$. Therefore, the gradient of the loss function with respect to $\vtheta$ can be computed as
\begin{equation}
    \nabla_{\vtheta} \mathbb{E}_{\ra \sim P_\vtheta}  \left[\mathcal{L}(\vm_a, \vw_a) \right] = \mathbb{E}_{\rvg \sim \text{Gumbel}(0, 1)} \left[
    \frac{\partial \mathcal{L}(\vm_a, \vw_a)}{\partial \rvm_a} \cdot \frac{\partial \rvm_a(\vtheta, \rvg)}{\partial \vtheta}
    \right].
\end{equation}
A temperature coefficient $\tau$ is used to control the behavior of the Gumbel Softmax. As $\tau \rightarrow \infty$, $\vm^{ij}$ becomes a continuous random variable following a uniform distribution. Therefore, in Equation (\ref{eqn:masked_node}), all edges are executed and their outputs are averaged. The gradient estimation in Equation (\ref{eqn:gradient}) are biased but the variance is low, which is favorable during the initial stage of the training. As $\tau \rightarrow 0$, $\vm^{ij}$ gradually becomes a discrete random variable following the categorical distribution of $P_{\vtheta^{ij}}$. When computing Equation (\ref{eqn:masked_node}), only one edge is sampled to be executed. The gradient estimation then becomes unbiased but the variance is high. This is favorable towards the end of the training. In our experiment, we use an exponential decaying schedule to anneal the temperature as 
\begin{equation}
\label{eqn:t_schedule}
    \tau = T_0 \exp(- \eta\times epoch),
\end{equation}
where $T_0$ is the initial temperature when training begins. We decay the temperature exponentially after every epoch. Using the Gumbel Softmax trick effectively stabilizes the super net training.

In some sense, our work is in the middle ground of two previous works: ENAS by \cite{pham2018efficient} and DARTS by \cite{liu2018darts}. ENAS samples child networks from the super net to be trained independently while DARTS trains the entire super net together without decoupling child networks from the super net. By using Gumbel Softmax with an annealing temperature, our DNAS pipeline behaves more like DARTS at the beginning of the search and behaves more like ENAS at the end.

\subsection{The DNAS pipeline}
Based on the analysis above, we propose a differentiable neural architecture search pipeline, summarized in Algorithm \ref{alg:dnas}. We first construct a stochastic super net $G$ with architecture parameter $\vtheta$ and weight $\vw$. We train $G$ with respect to $\vw$ and $\vtheta$ separately and alternately. Training the weight $\vw$ optimizes all candidate edges (operators). However, different edges can have different impact on the overall performance. Therefore, we train the architecture parameter $\vtheta$, to increase the probability to sample those edges with better performance, and to suppress those with worse performance. To ensure generalization, we split the dataset for architecture search into $\mathcal{X}_\vw$, which is used specifically to train $\vw$, and $\mathcal{X}_\vtheta$, which is used to train $\vtheta$. The idea is illustrated in Fig. \ref{fig:supernet}.

In each epoch, we anneal the temperature $\tau$ for gumbel softmax with the schedule in Equation (\ref{eqn:t_schedule}). To ensure $\vw$ is sufficiently trained before updating $\vtheta$, we postpone the training of $\vtheta$ for $N_{warmup}$ epochs. At the end of the training, we draw samples $a \sim P_\vtheta$. These sampled architectures are then trained on the training dataset $\mathcal{X}_{train}$ and evaluated on the test set $\mathcal{X}_{test}$.

The formulation of DNAS is general. The stochastic super net can represent any network architectures, as long as it can be represented by a DAG. Edges of the graph can represent any types of operators, as long as their input and output nodes have the same data shape. Such flexibility allows us to apply DNAS to search for different neural architectures for different problems, as we will show in the following sections.

\begin{algorithm}[H]
\SetAlgoLined
\KwIn{Stochastic super net $G = (V, E)$ with parameter $\vtheta$ and $\vw$, 
 searching dataset $\mathcal{X}_\vw$ and $\mathcal{X}_\vtheta$, training dataset $\mathcal{X}_{train}$, test dataset $\mathcal{X}_{test}$;}
 $Q_A \leftarrow \emptyset$ \; 
 \For{$epoch =0,\cdots N$}{
    $\tau \leftarrow T_0 \exp(-\eta \times epoch)$\;
    Train $G$ with respect to $\vw$ for one epoch\;
    \If{$epoch > N_{warmup}$}{
        Train $G$ with respect to $\vtheta$ for one epoch\;
    }
 } 
Sample architectures $a \sim P_\vtheta$; Push $a$ to $Q_A$\; 
 \For{$a \in Q_A$}{
    Train $a$ on $\mathcal{X}_{train}$ to convergence\;
    Test $a$ on $\mathcal{X}_{test}$\;
 }
\KwOut{Trained architectures $Q_A$.}
\caption{The DNAS pipeline.}
\label{alg:dnas}
\end{algorithm}

\section{DNAS for mixed-precision quantization}

Conventionally, 32-bit (full-precision) floating point numbers are used to represent weights and activations of neural nets. For resource constrained applications, low precision numbers can be used to represent neural networks to reduce the computational cost effectively. This technique is called quantization. Most of the existing quantization methods often represent all weights and activations using the same precision (bit-width). However, if we are allowed to compress different layers of a network to different precisions, can we further compress the network? 

Mixed-precision computation is widely supported by hardware platforms such as CPUs, FPGAs, and dedicated accelerators. To leverage this, we need to study how should we decide the precision for each layer such that we can maintain the accuracy of the network while minimizing the cost in terms of model size or computation. This is the problem of mixed-precision quantization. For an $N$-layer network where each layer can choose from $M$ candidate precisions, exhaustive search yields $\mathcal{O}(M^N)$ complexity.

\begin{figure}[h]
\begin{center}
\includegraphics[width=0.9\linewidth]{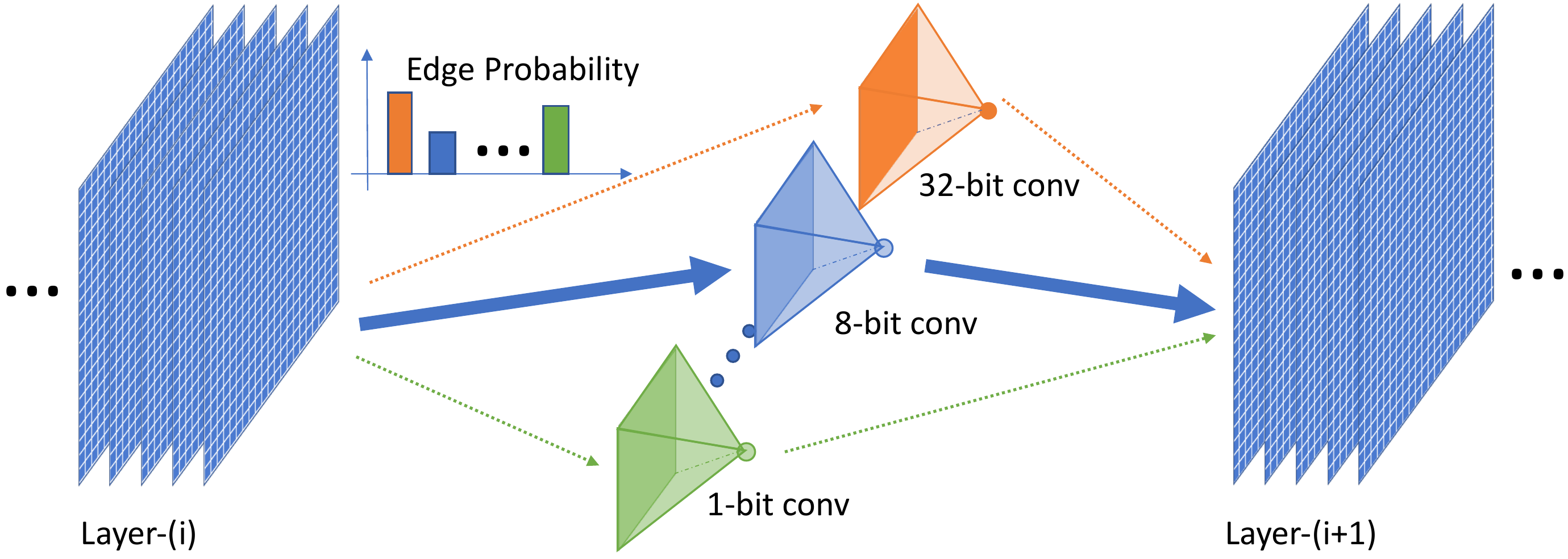}
\end{center}
\caption[One layer of a super net for mixed-precision quantization of a ConvNet.]{One layer of a super net for mixed-precision quantization of a ConvNet. Nodes in the super net represent feature maps, edges represent convolution operators with different bit-widths.}
\label{fig:mixed_quant}
\end{figure}

We use the DNAS framework to solve the mixed-precision quantization problem. For a ConvNet, we first construct a super net that has the same ``macro-structure'' (number of layers, number of filters each layer, etc.) as the given network. As shown in Fig. \ref{fig:mixed_quant}. Each node $v_i$ in the super net corresponds to the output tensor (feature map) of layer-$i$. Each candidate edge $e_k^{i,i+1}$ represents a convolution operator whose weights or activation are quantized to a lower precision.  

\subsection{Quantization method}
To quantize the network, we adopt conventional strategies, DoReFa-Net \cite{zhou2016dorefa} to quantize weights, and PACT \cite{choi2018pact} to quantize activations. 

\begin{equation}
\label{eqn:weight_quant}
w_k = 2Q_k(\frac{\tanh (w)}{2\text{max}(|\tanh (w)|)} + 0.5).
\end{equation}
$w$ denotes the latent full-precision weight of a network. $Q_k(\cdot)$ denotes a $k$-bit quantization function that quantizes a continuous value $w \in [0, 1]$ to its nearest neighbor in $\{\frac{i}{2^k - 1} | i = 0, \cdots, 2^k - 1\}$. To quantize activations, we follow \cite{choi2018pact} to use a bounded activation function followed by a quantization function as
\begin{equation}
\label{eqn:act_quant}
\begin{gathered}
y = PACT(x) = 0.5(|x| - |x-\alpha| + \alpha),\\
y_k = Q_k(y/\alpha) \cdot \alpha.
\end{gathered}
\end{equation}
Here, $x$ is the full precision activation, $y_k$ is the quantized activation. $PACT(\cdot)$ is a function that bounds the output between $[0, \alpha]$. $\alpha$ is a learnable upper bound of the activation function.

\subsection{Loss function}

In order to encourage using lower-precision weights and activations, we define the loss function as 
\begin{equation}
\label{eqn:mixed_qunt_loss}
    \mathcal{L}(a, \vw_a) = \text{CrossEntropy}(a) \times \mathcal{C}(Cost(a)).
\end{equation}
$Cost(a)$ denotes the cost of a candidate architecture and $\mathcal{C}(\cdot)$ is a weighting function to balance the cross entropy term and the cost term. To compress the model size, we define the cost as 
\begin{equation}
\label{eqn:cost_param}
Cost(a) = \sum_{e_k^{ij} \in E} m_k^{ij}\times \text{\#PARAM}(e_k^{ij}) \times \text{weight-bit}(e_k^{ij}),
\end{equation}
where $\text{\#PARAM}(\cdot)$ denotes the number of parameters of a convolution operator and $\text{weight-bit}(\cdot)$ denotes the bit-width of the weight. $m_k^{ij}$ is the edge selection mask described in Equation (\ref{eqn:masked_node}). Alternatively, to reduce the computational cost by jointly compressing both weights and activations, we use the cost function
\begin{equation}
\label{eqn:cost_flop}
Cost(a) = \sum_{e_k^{ij}\in E} m_k^{ij}\times \text{\#FLOP}(e_k^{ij}) \times \text{weight-bit}(e_k^{ij}) \times \text{act-bit}(e_k^{ij}),
\end{equation}
where $\text{\#FLOP}(\cdot)$ denotes the number of floating point operations of the convolution operator, $\text{weight-bit}(\cdot)$ denotes the bit-width of the weight and $\text{act-bit}(\cdot)$ denotes the bit-width of the activation. Note that in a candidate architecture, $m_k^{ij}$ have binary values $\{0, 1\}$. In the super net, we allow $m_k^{ij}$ to be continuous so we can compute the expected cost of the super net.. 

To balance the cost term with the cross entropy term in Equation (\ref{eqn:mixed_qunt_loss}), we define
\begin{equation}
\label{eqn:cost_weighting}
    \mathcal{C}(Cost(a)) = \beta (\log(Cost(a)))^\gamma.
\end{equation}
where $\beta$ is a coefficient to adjust the initial value of $\mathcal{C}(Cost(a))$ to be around 1. $\gamma$ is a coefficient to adjust the relative importance of the cost term vs. the cross-entropy term. A larger $\gamma$ leads to a stronger cost term in the loss function, which favors efficiency over accuracy.

\section{Mixed-precision quantization experiments}

\subsection{CIFAR10 experiments}

In the first experiment, we focus on quantizing ResNet20, ResNet56, and ResNet110 \cite{he2016deep} on CIFAR10 \cite{krizhevsky2009learning} dataset. We start by focusing on reducing model size, since smaller models require less storage and communication cost, which is important for mobile and embedded devices. We only perform quantization on weights and use full-precision activations. We conduct mixed-precision search at the block level -- all layers in one block use the same precision. Following the convention, we do not quantize the first or the last layer. We construct a super net whose macro architecture is exactly the same as our target network. For each block, we can choose a precision from $\{0, 1, 2, 3, 4, 8, 32\}$. If the precision is 0, we simply skip this block so the input and output are identical. If the precision is 32, we use the full-precision floating point weights. For all other precisions with $k$-bit, we quantize weights to $k$-bit fixed-point numbers. 

Our experiment result is summarized in Table \ref{tab:cifar_model_size}. For each quantized model, we report its accuracy and model size compression rate compared with 32-bit full precision models. The model size is computed by Equation (\ref{eqn:cost_param}). Among all the models we searched, we report the one with the highest test accuracy (top) and the one with the highest compression rate (bottom). We compare our method with \cite{zhu2016trained}, where they use 2-bit (ternary) weights for all the layers of the network, except the first convolution and the last fully connect layer. From the table, we have the following observations: 1) All of our most accurate models out-perform their full-precision counterparts by up to 0.37\% while still achieves 11.6 - 12.5X model size reduction. 2) Our most efficient models can achieve 16.6 - 20.3X model size compression with accuracy drop less than 0.39\%. 3) Compared with \cite{zhu2016trained}, our model achieves up to 1.59\% better accuracy. This is partially due to our improved training recipe as our full-precision model's accuracy is also higher. But it still demonstrates that our models with searched mixed-precision assignment can very well preserve the accuracy. 

\begin{table}[]
\begin{tabular}{cc|ccc|cc}
\hline
                                                &      & \multicolumn{3}{c|}{DNAS (ours)}       & \multicolumn{2}{c}{TTQ \cite{zhu2016trained}} \\ \cline{3-7} 
                                                &      & Full  & Most Accurate & Most Efficient & Full               & 2bit                       \\ \hline
\multicolumn{1}{c|}{\multirow{2}{*}{ResNet20}}  & Acc  & 92.35 & 92.72 (+0.37) & 92.00 (-0.35)  & 91.77              & 91.13 (-0.64)              \\
\multicolumn{1}{c|}{}                           & Comp & 1.0   & 11.6          & 16.6           & 1.0                & 16.0                       \\ \hline
\multicolumn{1}{c|}{\multirow{2}{*}{ResNet56}}  & Acc  & 94.42 & 94.57 (+0.15) & 94.12 (-0.30)  & 93.20              & 93.56 (+0.36)              \\
\multicolumn{1}{c|}{}                           & Comp & 1.0   & 14.6          & 18.93          & 1.0                & 16.0                       \\ \hline
\multicolumn{1}{c|}{\multirow{2}{*}{ResNet110}} & Acc  & 94.78 & 95.07 (+0.29) & 94.39 (-0.39)  & -                  & -                          \\
\multicolumn{1}{c|}{}                           & Comp & 1.0   & 12.5          & 20.3           & -                  & -                          \\ \hline
\end{tabular}
\caption[Mixed-Precision Quantization for ResNet on CIFAR10 dataset.]{Mixed-Precision Quantization for ResNet on CIFAR10 dataset. We report results on ResNet\{20, 56, 110\}. In the table, we abbreviate accuracy as ``Acc'' and compression as ``Comp''.}
\label{tab:cifar_model_size}
\end{table}

\begin{figure}[h]
\begin{center}
\includegraphics[width=0.8\linewidth]{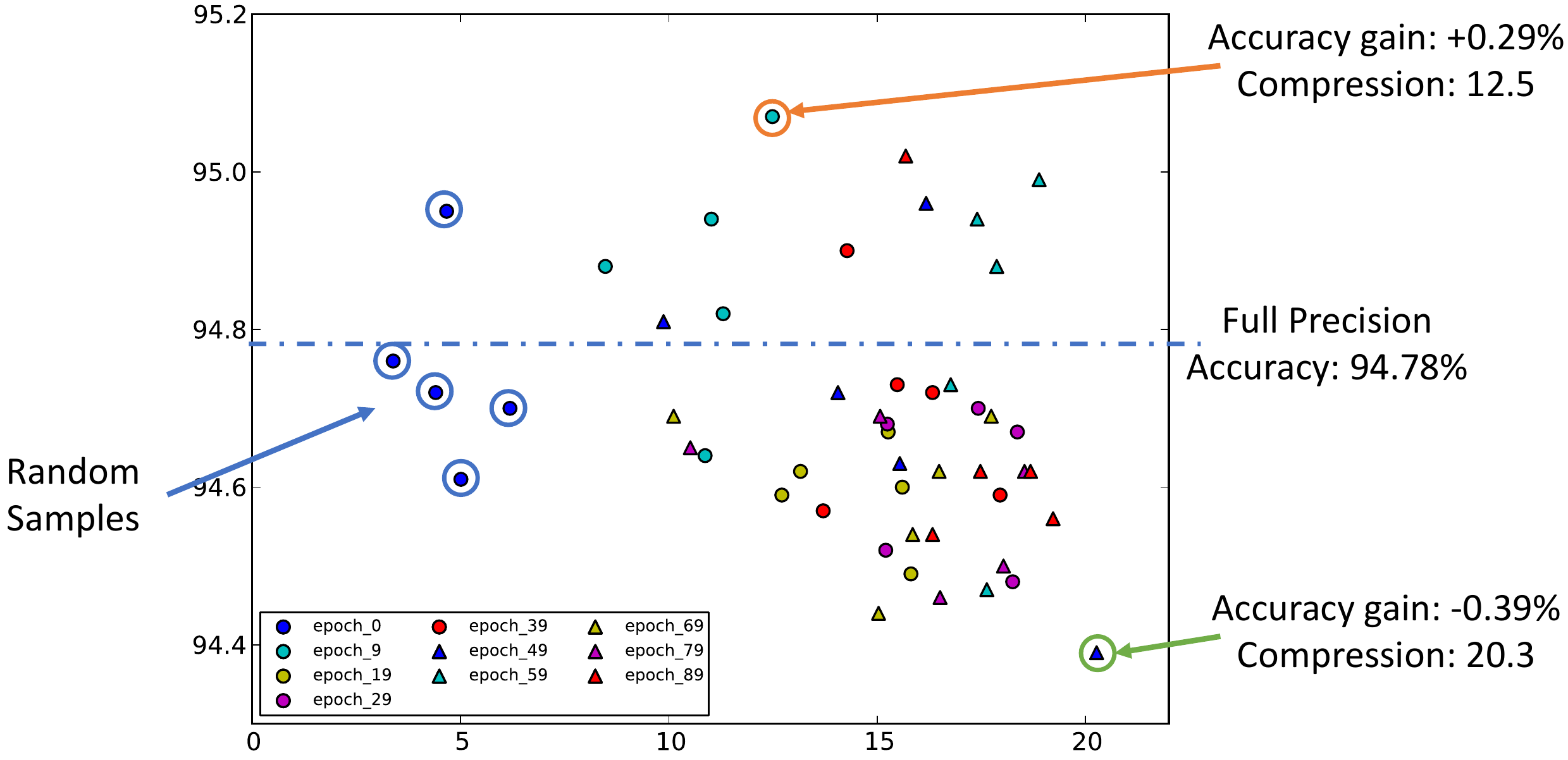}
\end{center}
\caption[Visualization of all searched architectures for ResNet110 on CIFAR10 dataset.]{Visualization of all searched architectures for ResNet110 on CIFAR10 dataset. x-axis is the compression rate of each model. y-axis is the accuracy.}
\label{fig:acc-comp}
\end{figure}

Table \ref{tab:layer-precision} compares the precision assignment for the most accurate and the most efficient models for ResNet20. Note that for the most efficient model, it directly skips the $3$rd block in group-1. In Fig. \ref{fig:acc-comp}, we plot the accuracy vs. compression rate of searched architectures of ResNet110. We observe that models with random precision assignment (from epoch 0) have significantly worse compression while searched precision assignments generally have higher compression rate and accuracy.

\begin{table}[]
\centering
\begin{tabular}{c|ccc|ccc|ccc}
\hline
               & g1b1 & g1b2 & g1b3 & g2b1 & g2b2 & g2b3 & g3b1 & g3b2 & g3b3 \\ \hline
Most Accurate  & 4    & 4    & 3    & 3    & 3    & 4    & 4    & 3    & 1    \\
Most Efficient & 2    & 3    & 0    & 2    & 4    & 2    & 3    & 2    & 1    \\ \hline
\end{tabular}
\caption[Layer-wise preicions of searched models.]{Layer-wise bit-widths for the most accurate vs. the most efficient architecture of ResNet20.}
\label{tab:layer-precision}
\end{table}

\subsection{ImageNet experiments}
We quantize ResNet18 and ResNet34 on the ImageNet ILSVRC2012 \cite{deng2009imagenet} dataset. In contrast with the original ResNet \cite{he2016deep}, we use the ``ReLU-only preactivation'' ResNet from \cite{he2016identity}. Similar to the CIFAR10 experiments, we conduct mixed-precision search at the block level. We do not quanitze the first and the last layer. 

We conduct two sets of experiments. In the first set, we aim at compressing the model size, so we only quantize weights and use the cost function from Equation (\ref{eqn:cost_param}). Each block contains convolution operators with weights quantized to $\{1, 2, 4, 8, 32\}$-bit. In the second set, we aim at compressing computational cost. So we quantize both weights and activations and use the cost function from Equation (\ref{eqn:cost_flop}). Each block in the super net contains convolution operators with weights and activations quantized to $\{(1,4), (2, 4), (3, 3), (4, 4), (8, 8), (32, 32)\}$-bit. The first number in the tuple denotes the weight precision and the second denotes the activation precision. The DNAS search is very efficient, taking less than 5 hours on 8 V100 GPUs to finish the search on ResNet18. 

\begin{table}[]
\begin{tabular}{cc|ccc|ccc}
\hline
                                               &      & \multicolumn{3}{c|}{DNAS (ours)}        &        & TTQ           & ADMM          \\ \cline{3-8} 
                                               &      & Full  & MA              & ME            & Full   & 2bit          & 3bit          \\ \hline
\multicolumn{1}{c|}{\multirow{2}{*}{ResNet18}} & Acc  & 71.03 & 71.21 (+0.18  ) & 69.58 (-1.45) & 69.6   & 66.6 (-3.0)   & 68.0 (-1.6)   \\
\multicolumn{1}{c|}{}                          & Comp & 1.0   & 11.2            & 21.1          & 1.0    & 16.0          & 10.7          \\ \hline
\multicolumn{1}{c|}{\multirow{2}{*}{ResNet34}} & Acc  & 74.12 & 74.61 (+0.49)   & 73.37 (-0.75) & \multicolumn{3}{c}{\multirow{2}{*}{-}} \\
\multicolumn{1}{c|}{}                          & Comp & 1.0   & 10.6            & 19.0          & \multicolumn{3}{c}{}                   \\ \hline
\end{tabular}
\caption[Mixed-precision quantization for model size compression.]{Mixed-precision quantization for ResNet on ImageNet for model size compression. 
In the table, we abbreviate accuracy as ``Acc'' and compression as ``Comp''.  ``MA'' denotes the most accurate model from architecture search and ``ME'' denotes the most efficient model.}
\label{tab:imagenet_model_size}
\end{table}

Our model size compression experiment is reported in Table \ref{tab:imagenet_model_size}. We report two searched results for each model. ``MA'' denotes the searched architecture with the highest accuracy, and ``ME'' denotes the most efficient. We compare our results with TTQ \cite{zhu2016trained} and ADMM \cite{leng2017extremely}. TTQ uses ternary weights (stored by 2 bits) to quantize a network. For ADMM, we cite the result with $\{-4, 4\}$ configuration where weights can have 7 values and are stored by 3 bits. We report the accuracy and model size compression rate of each model. From Table \ref{tab:imagenet_model_size}, we have the following observations: 1) All of our most accurate models out-perform full-precision models by up to 0.5\% while achieving 10.6-11.2X reduction of model size. 2) Our most efficient models can achieve 19.0 to 21.1X reduction of model size, still preserving competitive accuracy. 3) Compared with previous works, even our less accurate model has almost the same accuracy as the full-precision model with 21.1X smaller model size. This is partially because we use label-refinery \cite{bagherinezhad2018label} to effectively boost the accuracy of quantized models. But it still demonstrate that our searched models can very well preserve the accuracy, despite its high compression rate. 

\begin{table}[]
\begin{tabular}{cc|ccc|cccc}
\hline
                                               &              & \multicolumn{3}{c|}{DNAS (ours)}                                             & PACT   & DoReFA   & QIP    & GroupNet  \\ \cline{3-9} 
                                               &              & arch-1                    & arch-2                    & arch-3                     & W4A4   & W4A4     & W4A4   & W1A2G5    \\ \hline
\multicolumn{1}{c|}{\multirow{4}{*}{ResNet18}} & Acc          & 71.01                     & 70.64                     & 68.65                      & 69.2   & 68.1     & 69.3   & 67.6      \\
\multicolumn{1}{c|}{}                          & Full Acc     & 71.03                     & 71.03                     & 71.03                      & 70.2   & 70.2     & 69.2   & 69.7      \\
\multicolumn{1}{c|}{}                          & Acc $\Delta$ & -0.02                     & -0.39                     & -2.38                      & -1.0   & -2.1     & +0.1   & -2.1      \\
\multicolumn{1}{c|}{}                          & Comp         & 33.2                      & 62.9                      & 103.5                      & 64     & 64       & 64     & 102.4     \\ \hline
\multicolumn{1}{c|}{\multirow{4}{*}{ResNet34}} & Acc          & 74.21                     & \multicolumn{1}{l}{73.98} & 73.23                      & \multicolumn{4}{c}{\multirow{4}{*}{-}} \\
\multicolumn{1}{c|}{}                          & Full Acc     & \multicolumn{1}{l}{74.12} & \multicolumn{1}{l}{74.12} & \multicolumn{1}{l|}{74.12} & \multicolumn{4}{c}{}                   \\
\multicolumn{1}{c|}{}                          & Acc $\Delta$ & \multicolumn{1}{l}{+0.09} & \multicolumn{1}{l}{-0.14} & \multicolumn{1}{l|}{-0.89} & \multicolumn{4}{c}{}                   \\
\multicolumn{1}{c|}{}                          & Comp         & 40.8                      & \multicolumn{1}{l}{59.0}  & 87.4                       & \multicolumn{4}{c}{}                   \\ \hline
\end{tabular}
\caption[Mixed-precision quantization for computational cost reduction.]{Mixed-precision quantization for ResNet on ImageNet for computational cost compression. We abbreviate accuracy as ``Acc'' and compression rate as ``Comp''.  ``arch-\{1, 2, 3\}'' are three searched architectures ranked by accuracy. 
}
\label{tab:imagenet_flop}
\end{table}

Our experiment on computational cost compression is reported in Table \ref{tab:imagenet_flop}. We report three searched architectures for each model. We report the accuracy and the compression rate of the computational cost of each architecture. We compute the computational cost of each model using Equation (\ref{eqn:cost_flop}). We compare our results with PACT \cite{choi2018pact}, DoReFA \cite{zhou2016dorefa}, QIP \cite{jung2018joint}, and GroupNet \cite{zhuang2018training}. The first three use 4-bit weights and activations. We compute their compression rate as $(32 / 4) \times (32 / 4) = 64$. GroupNet uses binary weights and 2-bit activations, but its blocks contain 5 parallel branches. We compute its compression rate as $(32 / 1) \times (32 / 2) / 5 \approx 102.4$ The DoReFA result is cited from \cite{choi2018pact}. From table \ref{tab:imagenet_flop}, we have the following observations: 1) Our most accurate architectures (arch-1) have almost the same accuracy (-0.02\% or +0.09\%) as the full-precision models with compression rates of 33.2x and 40.8X. 2) Comparing arch-2 with PACT, DoReFa, and QIP, we have a similar compression rate (62.9 vs 64), but the accuracy is 0.71-1.91\% higher. 3) Comparing arch-3 with GroupNet, we have slightly higher compression rate (103.5 vs. 102.4), but 1.05\% higher accuracy.

\section{DNAS for efficient ConvNet search}
For the second application, we apply DNAS to search for new efficient ConvNet architectures for target hardware devices. DNAS allows us to define search space with high flexibility. However, to optimize the actual latency of a network, it is important to choose a design space that is intrinsically hardware friendly. We will discuss this in the search space section. In addition, to optimize the actual latency rather than the theoretical efficiency, we need to re-design the loss function to reflect the actual latency. We will discuss this in the loss function section. 

\subsection{The Search space}
\label{sec:search_space}
Most of the previous works \cite{zoph2016neural,zoph2017learning,pham2018efficient,liu2017progressive,liu2018darts} search for cell level architectures. The same cell level structure is used in all the layers across the network. However, many searched cell structures are very complicated and fragmented and are therefore slow when deployed to mobile CPUs \cite{sandler2018mobilenetv2,ma2018shufflenet}. Besides, at different layers, the same cell structure can have a different impact on the accuracy and latency of the overall network. As shown in \cite{tan2018mnasnet} and in our experiments, allowing different layers to choose different blocks leads to better accuracy and efficiency. 

In this work, we construct a layer-wise search space with a fixed macro-architecture, and each layer can choose a different block. Formally, for each layer of the super net, we have
\begin{equation}
\label{eqn:fb-mask}
    x_{l+1} = \sum_i m_{l, i} \cdot b_{l, i}(x_{l}),
\end{equation}
where $b_{l, i}(\cdot)$ is the $i$-th candidate block at layer $l$. $m_{l,i}$ is a random variable in $\{0, 1\}$ and is evaluated to 1 if block $b_{l, i}$ is sampled. $x_{l+1}$ and $x_{l}$ are feature maps of layer-$l$ and $l+1$. This is an instantiation of Equation (\ref{eqn:masked_node}). Similarly, we can use relax $m_{l,i}$ to a continuous random variable controlled by Gumbel Softmax function as
\begin{equation}
\label{eqn:fb-gumbel_softmax}
\begin{aligned}
    m_{l, i} & = \text{GumbelSoftmax}(\theta_{l, i}|\bm{\theta_{l}}) \\
             & = \frac{\exp[(\theta_{l,i} + g_{l,i})/\tau]}{\sum_i \exp[(\theta_{l,i} + g_{l,i})/\tau]}, 
\end{aligned}
\end{equation}
where $g_{l,i} \sim \text{Gumbel(0, 1)}$ is a random noise following the Gumbel distribution, $\theta_{l,i}$ is a trainable variable that controls the sampling probability of block-$i$ at layer-$l$.

The macro-architecture is described in Table \ref{tab:macro-space}. The macro architecture defines the number of layers and the input/output dimensions of each layer. The first and the last three layers of the network have fixed operators. For the rest of the layers, their block type needs to be searched. The filter numbers for each layer are hand-picked empirically. We use relatively small channel sizes for early layers, since the input resolution at early layers is large, and the computational cost (FLOP count) is quadratic to input size. 

\begin{table}[h]
\centering
\begin{tabular}{c|c|c|c|c}
\hline
Input shape             & Block       & f         & n        & s \\ \hline
$224^2 \times 3$        & 3x3 conv    & 16        & 1        & 2      \\
$112^2 \times 16$       & TBS         & 16        & 1        & 1      \\
$112^2 \times 16$       & TBS         & 24        & 4        & 2      \\
$56^2 \times 24$        & TBS         & 32        & 4        & 2      \\
$28^2 \times 32$        & TBS         & 64        & 4        & 2      \\
$14^2 \times 64$        & TBS         & 112       & 4        & 1      \\
$14^2 \times 112$       & TBS         & 184       & 4        & 2      \\
$7^2 \times 184$        & TBS         & 352       & 1        & 1      \\
$7^2 \times 352$        & 1x1 conv    & 1504 (1984)      & 1        & 1      \\
$7^2 \times 1504~(1984)$       & 7x7 avgpool & -         & 1        & 1      \\
$1504$                  & fc          & 1000      & 1        & -      \\ \hline
\end{tabular}
\caption[Macro-architecture of the search space. ]{Macro-architecture of the search space. Column-``Block'' denotes the block type. ``TBS'' means layer type needs to be searched. Column-$f$ denotes the filter number of a block. Column-$n$ denotes the number of blocks. Column-$s$ denotes the stride of the first block in a stage.  The filter size of the last 1x1 conv is 1504 for FBNet-A and 1984 for FBNet-\{B, C\}.}
\label{tab:macro-space}
\end{table}

\begin{figure}[!t]
\begin{center}
\centering \includegraphics[width=0.5\linewidth]{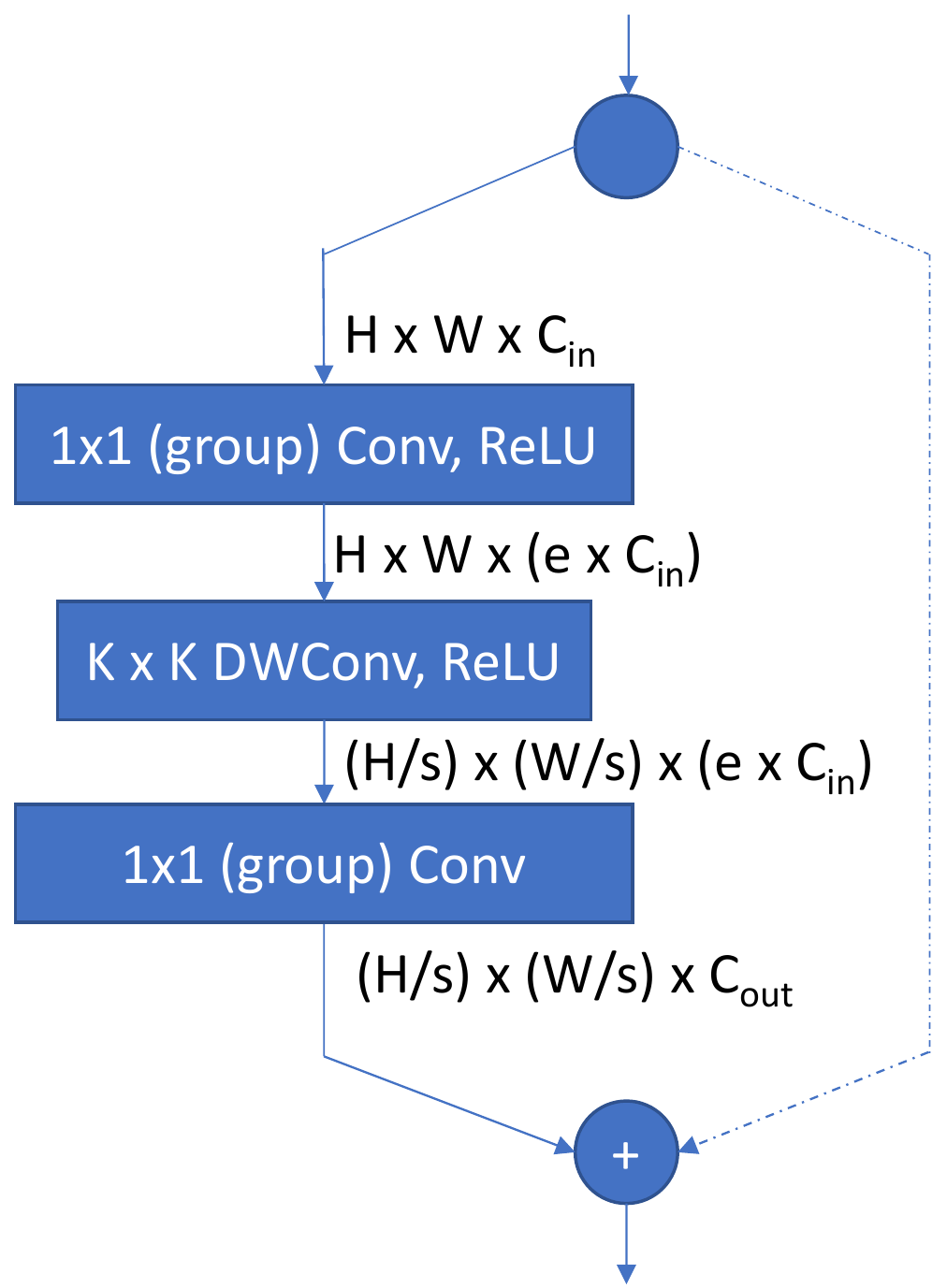}
\caption[The block structure of the micro-architecture search space.]{The block structure of the micro-architecture search space. Each candidate block in the search space can choose a different expansion rate, kernel size, and number of groups for group convolution.}
\label{fig:block}
\end{center}
\end{figure}

Each searchable layer in the network can choose a different block from the layer-wise search space. The block structure is inspired by MobileNetV2 \cite{sandler2018mobilenetv2} and ShiftNet \cite{wu2017shift}, and is illustrated in Figure \ref{fig:block}. It contains a point-wise (1x1) convolution, a K-by-K depthwise convolution where K denotes the kernel size, and another 1x1 convolution. ``ReLU'' activation functions follow the first 1x1 convolution and the depthwise convolution, but there are no activation functions following the last 1x1 convolution. If the output dimension stays the same as the input dimension, we use a skip connection to add the input to the output. Following \cite{sandler2018mobilenetv2, wu2017shift}, we use a hyperparameter, the expansion ratio $e$, to control the block. It determines how much do we expand the output channel size of the first 1x1 convolution compared with its input channel size. Following \cite{tan2018mnasnet}, we also allow choosing a kernel size of 3 or 5 for the depthwise convolution. In addition, we can choose to use group convolution for the first and the last 1x1 convolution to reduce the computation complexity. When we use group convolution, we follow \cite{zhang1707shufflenet} to add a channel shuffle operation to mix the information between channel groups. 

In our experiments, our layer-wise search space contains 9 candidate blocks, with their configurations listed in Table \ref{tab:micro-space}. Note we also have a block called ``skip'', which directly feed the input feature map to the output without actual computations. This candidate block essentially allows us to reduce the depth of the network. 

In summary, our overall search space contains 22 layers. The layer number is chosen empirically. Each layer can choose from 9 candidate blocks from Table \ref{tab:micro-space}, so it contains $9^{22} \approx 10^{21}$ possible architectures. 

Finding the optimal layer-wise block assignment from such enormous search space is non-trivial.

\begin{table}[]
\centering
\begin{tabular}{c|c|c|c}
\hline
Block type & expansion      & Kernel      & Group \\ \hline
k3\_e1     & 1              & 3           & 1     \\
k3\_e1\_g2 & 1              & 3           & 2     \\
k3\_e3     & 3              & 3           & 1     \\
k3\_e6     & 6              & 3           & 1     \\
k5\_e1     & 1              & 5           & 1     \\
k5\_e1\_g2 & 1              & 5           & 2     \\
k5\_e3     & 3              & 5           & 1     \\
k5\_e6     & 6              & 5           & 1     \\ 
skip       & -              & -           & -     \\ \hline
\end{tabular}
\caption{Configurations of candidate blocks in the search space.}
\label{tab:micro-space}
\end{table}

\subsection{Latency-aware loss function}
\label{sec:loss_function}
To optimize for actual latency, we re-design the loss function from (\ref{eqn:nas}) to reflect not only the accuracy of an architecture but also the latency of it on the target hardware. Following Equation (\ref{eqn:mixed_qunt_loss}), the loss function is defined as:
\begin{equation}
\label{eqn:loss}
    \mathcal{L}(a, w_a) = \text{ CE}(a, w_a) \cdot \alpha \log(\text{LAT}(a))^\beta.
\end{equation}
The first term $\text{CE}(a, w_a)$ denotes the cross-entropy loss of architecture $a$ with parameter $w_a$. The second term $\text{LAT}(a)$ denotes the latency of the architecture on the target hardware measured in micro-second. The coefficient $\alpha$ controls the overall magnitude of the loss function. The exponent coefficient $\beta$ modulates the magnitude of the latency term. 

The cross-entropy term can be easily computed. However, the latency term is more difficult, since we need to measure the actual runtime of an architecture on a target device. To cover the entire search space, we need to measure about $10^{21}$ architectures, which is an impossible task. 

To solve this problem, we use a latency lookup table model to estimate the overall latency of a network based on the runtime of each operator. More formally, we assume
\begin{equation}
\label{eqn:latency}
    \text{LAT}(a) = \sum_l \text{LAT} (b_l^{(a)}),
\end{equation}
where $b_l^{(a)}$ denotes the block at layer-$l$ from architecture $a$. This assumes that on the target processor, the runtime of each operator is independent of other operators. The assumption is valid for many mobile CPUs and DSPs, where operators are computed sequentially one by one. In case that inter-layer optimization can be applied, for example, two layers can be merged into one, the lookup table model can still be applied after such optimization. Using the lookup table model, by benchmarking the latency of a few hundred unique operators used in the search space, we can easily estimate the actual runtime of the $10^{21}$ architectures in the entire search space. 

As explained in Equation (\ref{eqn:stochastic_nas}), the architecture search problem is equivalent to using SGD to optimize the loss function with respect to parameter $\theta_{l,i}$. It is clear that the cross-entropy term from the loss function (\ref{eqn:loss}) is differentiable with respect to the mask $m_{l,i}$ and therefore $\theta_{l,i}$. For the latency term, since we use the lookup table based model for efficiency estimation, Equation (\ref{eqn:latency}) can be written as
\begin{equation}
\label{eqn:lut-latency}
    \text{LAT}(a) = \sum_l \sum_i m_{l,i} \cdot \text{LAT} (b_{l,i}).
\end{equation}
The latency of each operator $\text{LAT} (b_{l,i})$ is a constant coefficient, so the overall latency of architecture-$a$ is differentiable with respect to the mask $m_{l,i}$, therefore $\theta_{l,i}$. Now it is obvious that the loss function (\ref{eqn:loss}) is fully differentiable with respect to both weights $w_a$ and the architecture distribution parameter $\bm{\theta}$. This allows us to use SGD to efficiently solve problem (\ref{eqn:nas}). 

\section{Efficient ConvNet search experiments}
\subsection{ImageNet classification}
To demonstrate the efficacy of our proposed method, we use DNAS to search for ConvNet models on ImageNet 2012 classification dataset \cite{deng2009imagenet}, and we name the discovered models FBNets. We aim to discover models with high accuracy and low latency on target devices. In our first experiment, we target Samsung Galaxy S8 with a Qualcomm Snapdragon 835 platform. The model is deployed with Caffe2 with int8 inference engine for mobile devices.

Before the search starts, we first build a latency lookup table on the target device. Next, we train a stochastic super net. We set the input resolution of the network to 224-by-224. To reduce the training time, we randomly choose 100 classes from the original 1000 classes to train the stochastic super net. We train the stochastic super net for 90 epochs. In each epoch, we first train the operator weights $w_a$ and then the architecture probability parameter $\bm{\theta}$. $w_a$ is trained on 80\% of ImageNet training set using SGD with momentum. 
The architecture distribution parameter $\bm{\theta}$ is trained on the rest 20\% of ImageNet training set with Adam optimizer \cite{kingma2014adam}.
To control the temperature of the Gumbel Softmax from Equation (\ref{eqn:gumbel_softmax}), we use an exponentially decaying temperature. After the search finishes, we sample several architectures from the trained distribution $P_{\bm{\theta}}$, and train them from scratch. Our architecture search framework is implemented in pytorch \cite{paszke2017automatic} and searched models are trained in Caffe2. More training details will be provided in the supplementary materials. 

Our experimental results are summarized in Table \ref{tab:imagenet}. We compare our searched models with state-of-the-art efficient models both designed automatically and manually. The primary metrics we care about are top-1 accuracy on the ImageNet validation set and the latency. If the latency is not available, we use FLOP as the secondary efficiency metric. For baseline models, we directly cite the parameter size, FLOP count, and top-1 accuracy from the original paper. Since our network is deployed with caffe2 with a highly efficient in8 implementation, we have an unfair latency advantage against other baselines. Therefore, we implement the baseline models ourselves and measure their latency under the same environment for a fair comparison. For automatically designed models, we also compare the search method, search space, and search cost. 

\begin{table*}[h]
\centering
\begin{tabular}{c|ccc|cccc}
\hline
Model                                         & \begin{tabular}[c]{@{}c@{}}Search \\ method\end{tabular} & \begin{tabular}[c]{@{}c@{}}Search \\ space\end{tabular} & \begin{tabular}[c]{@{}c@{}}Search cost\\ (GPU hours \\/ relative)\end{tabular} & Params & FLOPs & \begin{tabular}[c]{@{}c@{}}CPU \\ Latency \\ (ms) \end{tabular} & \begin{tabular}[c]{@{}c@{}}Top-1 \\ acc \\ (\%)\end{tabular} \\ \hline
1.0-MNV2 \cite{sandler2018mobilenetv2} & manual                                                   & -                                                       & -                                                               & 3.4M         & 300M    & 21.7                                                & 72.0                                                      \\
1.5-SNV2 \cite{ma2018shufflenet}      & manual                                                   & -                                                       & -                                                               & 3.5M         & 299M    & 22.0                                                & 72.6                                                      \\
MnasNet-65 \cite{ma2018shufflenet}            & RL                                                       & stage                                              & 91K$^*$ / 421x                                     & 3.6M         & 270M    & -                                                      & 73.0                                                     \\ 
DARTS  \cite{liu2018darts}                    & gradient                                                 & cell                                                    & 288 / 1.33x                                                    & 4.9M         & 595M    & -                                                      & \textbf{73.1}                                                      \\ 
FBNet-A (ours)                              & gradient                                                 & layer                                              & 216 / 1.0x                                                    & 4.3M         & \textbf{249M}    & \textbf{19.8}                                                & 73.0                                                      \\ \hline
1.3-MNV2 \cite{sandler2018mobilenetv2} & manual                                                   & -                                                       & -                                                               & 5.3M         & 509M    & 33.8                                                 & \textbf{74.4}                                                      \\
MnasNet \cite{tan2018mnasnet}                 & RL                                                       & stage                                              & 91K$^*$ / 421x                                      & 4.2M         & 317M    & 23.7                                                & 74.0                                                      \\
NASNet-A  \cite{zoph2017learning}             & RL                                                       & cell                                                    & 48K / 222x                                       & 5.3M         & 564M    & -                                                      & 74.0                                                      \\
PNASNet \cite{liu2017progressive}             & SMBO                                                     & cell                                                    & 6K$^{\dag}$ / 27.8x                                      & 5.1M         & 588M    & -                                                      & 74.2                                                      \\ 
FBNet-B (ours)                              & gradient                                                 & layer                                              & 216 / 1.0x                                                     & 4.5M         & \textbf{295M}    & \textbf{23.1}                                                & 74.1                                                      \\ \hline
1.4-MNV2 \cite{sandler2018mobilenetv2} & manual                                                   & -                                                       & -                                                              & 6.9M         & 585M    & 37.4                                                   & 74.7                                                      \\
2.0-SNV2 \cite{ma2018shufflenet}      & manual                                                   & -                                                       & -                                                              & 7.4M         & 591M    & 33.3                                                   & \textbf{74.9}                                                      \\
MnasNet-92 \cite{tan2018mnasnet}              & RL                                                       & stage                                              & 91K$^*$ / 421x                                         & 4.4M         & 388M    & -                                                      & 74.8                                                     \\
FBNet-C (ours)                              & gradient                                                 & layer                                              & 216 / 1.0x                                                       & 5.5M         & \textbf{375M}    & \textbf{28.1}                                                   & \textbf{74.9}                                                     \\ \hline
\end{tabular}
\caption[ImageNet classification performance of FBNets and baselines.]{ImageNet classification performance compared with baselines. MNV2 denotes MobileNetV2. SNV2 denotes ShuffleNetV2. For baseline models, we directly cite the parameter size, FLOP count and top-1 accuracy on the ImageNet validation set from their original papers. For CPU latency, we deploy and benchmark the models on the same Samsung Galaxy S8 phone with Caffe2 int8 implementation. The details of MnasNet-\{64, 92\} are not disclosed from \cite{tan2018mnasnet} so we cannot measure the latency. *The search cost for MnasNet is estimated according to the description in \cite{tan2018mnasnet}. $\dag$ The search cost is estimated based on the claim from \cite{liu2017progressive} that PNAS \cite{liu2017progressive} is 8x lower than NAS\cite{zoph2017learning}. $\ddag$ The inference engine is faster than other models. } 
\label{tab:imagenet}
\end{table*}

Table \ref{tab:imagenet} divides the models into three categories according to their accuracy level. In the first group, FBNet-A achieves 73.0\% accuracy, better than 1.0-MobileNetV2 (+1.0\%), 1.5-ShuffleNet V2 (+0.4\%), and CondenseNet (+2\%), and are on par with DARTS and MnasNet-65. Regarding latency, FBNet-A is 1.9 ms (relative 9.6\%), 2.2 ms (relative 11\%), and 8.6 ms (relative 43\%) better than the MobileNetV2, ShuffleNetV2, and CondenseNet counterparts. Although we did not optimize for FLOP count directly, FBNet-A's FLOP count is only 249M, 50M smaller (relative 20\%) than MobileNetV2 and ShuffleNetV2, 20M (relative 8\%) smaller than MnasNet, and 2.4X smaller than DARTS. In the second group, FBNet-B achieves comparable accuracy with 1.3-MobileNetV2, but the latency is 1.46x lower, and the FLOP count is 1.73x smaller, even smaller than 1.0-MobileNetV2 and 1.5-ShuffleNet V2. Compared with MnasNet, FBNet-B's accuracy is 0.1\% higher, latency is 0.6ms lower, and FLOP count is 22M (relative 7\%) smaller. We do not have the latency of NASNet-A and PNASNet, but the accuracy is comparable, and the FLOP count is 1.9x and 2.0x smaller. In the third group, FBNet-C achieves 74.9\% accuracy, same as 2.0-ShuffleNetV2 and better than all others. The latency is 28.1 ms, 1.33x and 1.19x faster than MobileNet and ShuffleNet. The FLOP count is 1.56x, 1.58x, and 1.03x smaller than MobileNet, ShuffleNet, and MnasNet-92.

Among all the automatically searched models, FBNet's performance is much stronger than DARTS, PNAS, and NAS, and better than MnasNet. However, the search cost is orders of magnitude lower. MnasNet \cite{tan2018mnasnet} does not disclose the exact search cost (in terms of GPU-hours). However, it mentions that the controller samples 8,000 models during the search and each model is trained for five epochs. According to our experiments, training of MNasNet for one epoch takes 17 minutes using 8 GPUs. So the estimated cost for training 8,000 models for 5 epochs is about $17/60\times 5 \times 8 \times 8,000 \approx 91\times 10^3$ GPU hours. In comparison, the FBNet search takes 8 GPUs for only 27 hours, so the computational cost is only 216 GPU hours, or 421x faster than MnasNet, 222x faster than NAS, 27.8x faster than PNAS, and 1.33x faster than DARTS.

\begin{figure}[h]
\begin{center}
\includegraphics[width=1.\linewidth]{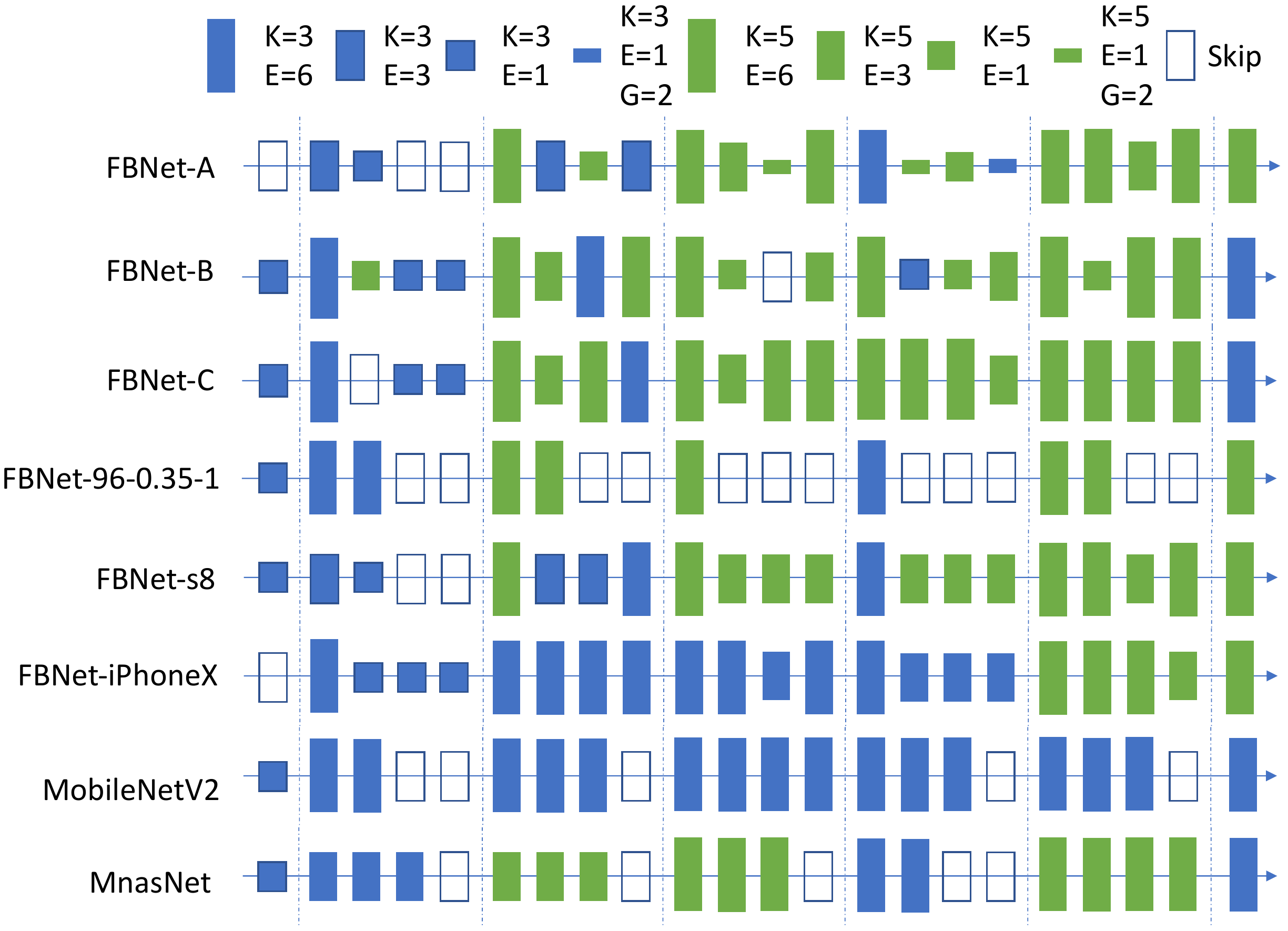}
\end{center}
\caption[Visualization of searched architectures.]{Visualization of searched architectures. We use colored boxes to denote blocks for each layer. We use different colors to denote the kernel size of the depthwise convolution, blue for kernel size of 3, green for kernel size of 5, and empty for skipping. We use height to denote the expansion rate of the block: 6, 3, 1, and 1 with group-2 convolution.}
\label{fig:arch_viz}
\end{figure}

We visualize some of our searched FBNets, MobileNetV2, and MnasNet in Figure \ref{fig:arch_viz}.

\subsection{Different resolution and channel size scaling}
A common technique to reduce the computational cost of a ConvNet is to reduce the input resolution or channel size without changing the ConvNet structure. This approach is likely to be sub-optimal. We hypothesize that with a different input resolution and channel size scaling, the optimal ConvNet structure will be different. To test this, we use DNAS to search for several different combinations of input resolution and channel size scaling. Thanks to the superior efficiency of DNAS, we can finish the search very quickly. The result is summarized in Table \ref{tab:scaling}. Compared with MobileNetV2 under the same input size and channel size scaling, our searched models achieve 1.5\% to 6.4\% better accuracy with similar latency. Especially the FBNet-96-0.35-1 model achieves 50.2\% (+4.7\%) accuracy and 2.9 ms latency (345 frames per second) on a Samsung Galaxy S8.

\begin{table*}[]
\centering
\begin{tabular}{c|c|cccc}
\hline
\begin{tabular}[c]{@{}c@{}}Input \&\\ Channel  \\ Scaling\end{tabular} & Model                  & Params & FLOPs & \begin{tabular}[c]{@{}c@{}}CPU \\ Latency\end{tabular} & Top-1 acc (\%) \\ \hline
\multirow{3}{*}{(224, 0.35)}                                            & MNV2-224-0.35   & 1.7M         & 59M     & 9.3 ms      & 60.3           \\
                                                                        & MNas-scale-224-0.35 & 1.9M         & 76M     & 10.7 ms     & 62.4 (+2.1)    \\
                                                                        & FB-224-0.35          & 2.0M         & 72M     & 10.7 ms     & 65.3 (+5.0)    \\ \hline
\multirow{3}{*}{(192, 0.50)}                                            & MNV2            & 2.0M        & 71M     & 8.4 ms      & 63.9           \\
                                                                        & Mnas-search-192-0.5 & -            & -       & -           & 65.6 (+1.7)    \\
                                                                        & FB-192-0.5 (ours)    & 2.6M        & 73M     & 9.9 ms      & 65.9 (+2.0)    \\ \hline
\multirow{3}{*}{(128, 1.0)}                                             & MNV2            & 3.5M         & 99M     & 8.4 ms      & 65.3           \\
                                                                        & Mnas-scale-128-1.0  & 4.2M         & 103M    & 9.2 ms      & 67.3 (+2.0)    \\
                                                                        & FB-128-1.0 (ours)    & 4.2M        & 92M     & 9.0 ms      & 67.0 (+1.7)    \\ \hline
\multirow{2}{*}{(128, 0.50)}                                            & MNV2            & 2.0M        & 32M     & 4.8 ms      & 57.7           \\
                                                                        & FB-128-0.5 (ours)    & 2.4M        & 32M   & 5.1 ms      & 60.0 (+2.3)    \\ \hline
\multirow{3}{*}{(96, 0.35)}                                             & MNV2            & 1.7M        & 11M     & 3.8 ms      & 45.5           \\
                                                                        & FB-96-0.35-1 (ours)  & 1.8M        & 12.9M   & 2.9 ms      & 50.2 (+4.7)    \\
                                                                        & FB-96-0.35-2 (ours)  & 1.9M        & 13.7M   & 3.6 ms      & 51.9 (+6.4)    \\ \hline
\end{tabular}
\caption[FBNets searched for different input resolution and channel scaling.]{FBNets searched for different input resolution and channel scaling. ``MNV2'' is for MobileNetV2. ``FB'' is for FBNet. ``Mnas'' is for MnasNet. Mnas-scale is the MnasNet model with input and channel size scaling. Mnas-search-192-0.5 is a model searched with an input size of 192 and channel scaling of 0.5. Details of it are not disclosed in \cite{tan2018mnasnet}, so we only cite the accuracy.}
\label{tab:scaling}
\end{table*}

\begin{table*}[h]
\centering
\begin{tabular}{c|ccccc}
\hline
Model & \#Parameters & \#FLOPs & \begin{tabular}[c]{@{}c@{}}Latency on \\ iPhone X\end{tabular} & \begin{tabular}[c]{@{}c@{}}Latency on\\ Samsung S8\end{tabular} & Top-1 acc (\%) \\ \hline
FBNet-iPhoneX      & 4.47M        & 322M    & \textbf{19.84 ms}                                                      & 23.33 ms                                                        & 73.20         \\
FBNet-S8           & 4.43M        & 293M    & 27.53 ms                                                               & \textbf{22.12 ms}                                               & 73.27         \\ \hline
\end{tabular}
\caption{FBNets searched for different devices.}
\label{tab:device}
\end{table*}

We visualize the architecture of FBNet-96-0.35-1 in Figure \ref{fig:arch_viz}, we can see that many layers are skipped, and the network is much shallower than FBNet-\{A, B, C\}, whose input size is 224. We conjecture that this is because with smaller input size, the receptive field needed to parse the image also becomes smaller, so having more layers will not effectively increase the accuracy.

\subsection{Different Target Devices}
In previous ConvNet design practices, the same ConvNet model is deployed to many different devices. However, this is sub-optimal since different computing platforms and software implementation can have different characteristics. To validate this, we conduct search targeting two mobile devices: Samsung Galaxy S8 with Qualcomm Snapdragon 835 platforms, and iPhone X with A11 Bionic processors. We use the same architecture search space, but different latency lookup tables collected from two target devices. All the architecture search and training protocols are the same. After we searched and trained two models, we deploy them to both Samsung Galaxy S8 and iPhone X to benchmark the overall latency. The result is summarized in Table. \ref{tab:device}. 

As we can see, the two models reach similar accuracy ($73.20\%$ vs. $73.27\%$). FBNet-iphoneX model's latency is 19.84 ms on its target device, but when deployed to a Samsung S8, its latency increases to 23.33 ms. On the other hand, FBNet-S8 reaches a latency of 22.12 ms on a Samsung S8, but when deployed to an iPhone X, the latency hikes to 27.53 ms, 7.69 ms (relatively 39\%) higher than FBNet-iPhone X. 
This demonstrates the necessity of re-designing ConvNets for different target devices. 

\begin{figure}[h]
\begin{center}
\includegraphics[width=.95\linewidth]{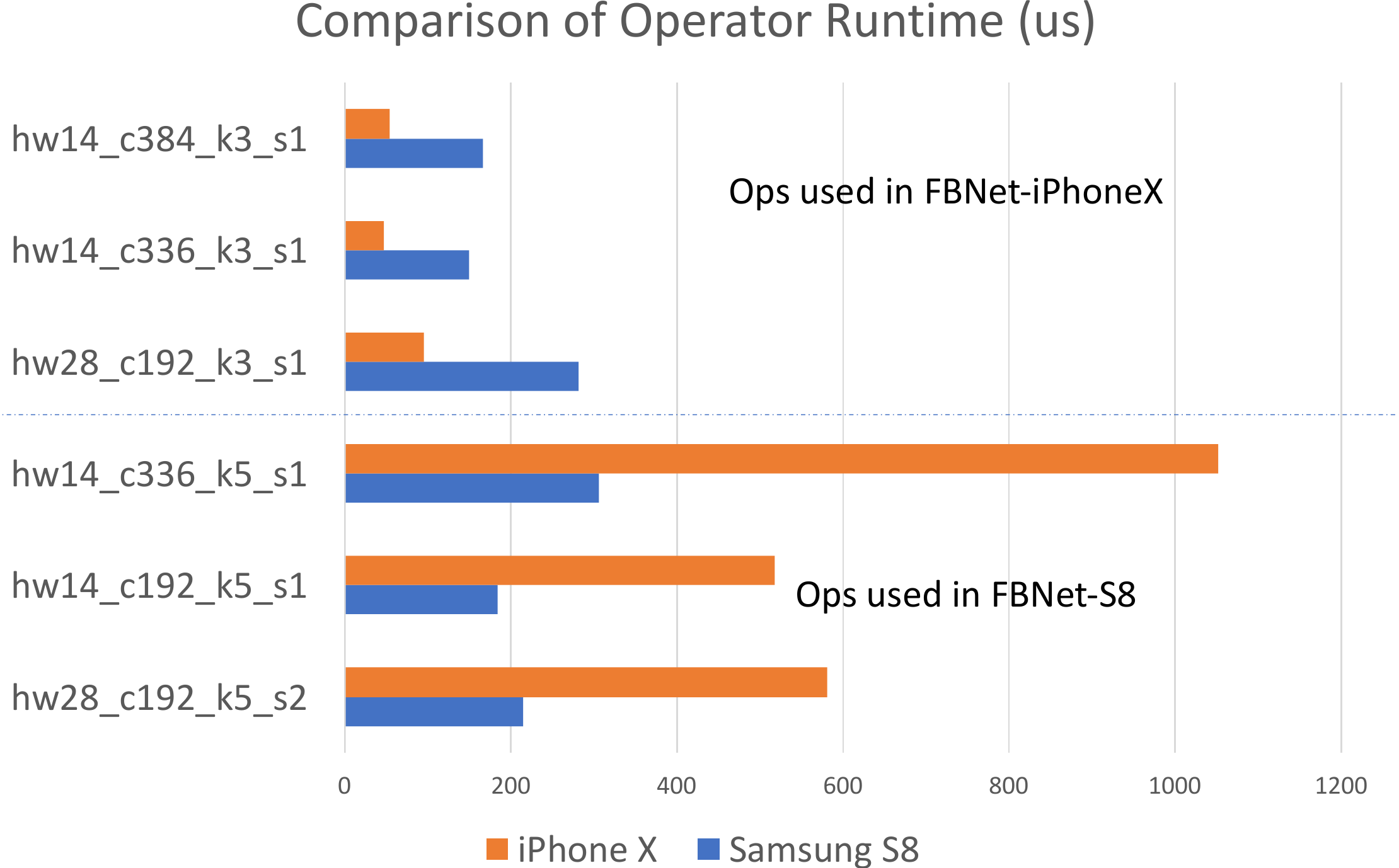}
\end{center}
\caption[Comparison of operator runtime on two devices.]{Comparison of operator runtime on two devices. Runtime is in micro-second (us). Orange bar denotes the runtime on iPhone X and blue bar denotes the runtime on Samsung S8. The upper three operators are faster on iPhone X, therefore they are automatically used in FBNet-iPhoneX. The lower three operators are faster on Samsung S8, and they are also automatically used in FBNet-S8.}
\label{fig:runtime_comparision}
\end{figure}

Two models are visualized in Figure \ref{fig:arch_viz}. Note that FBNet-S8 uses many blocks with 5x5 depthwise convolution while FBNet-iPhoneX only uses them in the last two stages. We examine the depthwise convolution operators used in the two models and compare their runtime on both devices. As shown in Figure \ref{fig:runtime_comparision}, the upper three operators are faster on iPhone X, therefore they are automatically used in FBNet-iPhoneX. The lower three operators are significantly faster on Samsung S8, and they are also automatically used in FBNet-S8. Notice the drastic runtime differences of the lower three operators on two target devices. It explains why the Samsung-S8-optimized model performs poorly on an iPhone X. This shows DNAS can automatically optimize the operator adoptions and generate different ConvNets optimized for different devices. 

\section{Conclusion}
In this chapter, we discuss the design efficiency of deep neural networks. Designing optimal deep neural networks for given applications and hardware accelerators is a difficult task, due to the challenges of huge design space, conditional optimality, and inaccurate efficiency metrics. To solve these problems, we present DNAS, a differentiable neural architecture search framework. It is a general neural architecture search algorithm that can be applied to different problems. It relaxed the combinatorial optimization problem to one that can be solved using stochastic gradient descent, and achieved two-orders of magnitudes of speedup. 

We applied DNAS to two problems. For mixed-precision quantization, we use DNAS to search for layer-wise precision assignment for ResNet on CIFAR10 and ImageNet. Our quantized models with 21.1x smaller model size or 103.9x smaller computational cost can still outperform baseline quantized or even full precision models. 

For hardware-efficient ConvNet search, we use DNAS to discover FBNets, a family of models that surpass state-of-the-art models, both manually and automatically designed: FBNet-B achieves 74.1\% top-1 accuracy with 295M FLOPs and 23.1 ms latency, 2.4x smaller and 1.5x faster than MobileNetV2-1.3 with the same accuracy. It also achieves better accuracy and lower latency than MnasNet, the state-of-the-art efficient model designed automatically; we estimate the search cost of DNAS is 420x smaller. Such efficiency allows us to conduct searches for different input resolutions and channel scaling. Discovered models achieve 1.5\% to 6.4\% accuracy gains. The smallest FBNet achieves 50.2\% accuracy with a latency of 2.9 ms (345 frames/sec) with batch size 1. Over the Samsung-optimized FBNet, the improved FBNet achieves 1.4x speed up on an iPhone X, showing DNAS is able to adapt to different target devices automatically. 

After the publication of DNAS, the search for smaller neural networks continued. For example, \cite{guo2019single} further developed the idea of the stochastic super net and combined it with evolution-based search algorithms. \cite{howard2019searching} combined manual design with architecture search and finetuning to find more efficient neural architectures. \cite{chen2019detnas} applied neural architecture search methods to solve object detection problems and conquered the limitation of two-stage training pipeline. For future research, we hope to see both design space innovation and search algorithm improvement. We also expect to see wider adoption of neural architecture search for more applications. 

%% file: conclusion.tex
\chapter{Conclusions}
\section{Review}
The success of deep neural networks is attributable to three factors: increased compute capacity, more complex models, and more data. However, in order to adopt deep neural networks to solve more practical problems, especially edge-based applications, we need to conquer the challenges of limited compute, limited model complexity, and limited data. In this thesis, we discussed our work on improving the efficiency of deep neural networks at four different levels. 

\textbf{Model efficiency}: It was believed that neural networks with higher computational complexity (FLOPs or parameter size) could achieve higher performance. However, the increased complexity makes it difficult to run neural networks on edge devices where compute capacity is limited. To address this problem, we focus on improving the model efficiency of neural networks by designing compact models that reach the same level of performance (accuracy) with significantly lower complexity (FLOPs and parameter size) and higher real-world efficiency (speed and energy). 

In Chapter \ref{chap:metrics}, we discussed efficiency metrics of deep neural networks. We discussed the background of computer architectures and compute characteristics of neural networks and introduced hardware-agnostic metrics (MACs, parameter size, activation size, and arithmetic intensity) and actual metrics (latency, throughput, power, and energy) for measuring the efficiency (complexity) of neural networks. 

In Chapter \ref{chap:sqdt}, we discussed the design of SqueezeDet, a convolutional neural network model for image-based object detection. SqueezeDet was designed to satisfy the need for autonomous driving to achieve high accuracy, fast inference speed, small model size, and low energy. By carefully designing the detection pipeline, network structure, and training protocol, we are able to train SqueezeDet to match the accuracy of our previous Faster-RCNN baselines while achieving 35.2x energy reduction, 30.4x model size reduction, and 19.7x inference speedup. 

In Chapter \ref{chap:sqsg}, we discussed the design of SqueezeSeg for LiDAR-based semantic segmentation. LiDAR is an essential sensor for autonomous vehicles, especially level 4 \& 5 urban taxis. It can sense accurate distance measurements from obstacles, which can be used for downstream planning and control. To parse semantic information from LiDAR point clouds, such as locating cars, pedestrians, and cyclists, traditional methods rely on hand-crafted features and multi-stage detection pipelines, which are slow and inaccurate. Instead, we present a convolution neural network-based model called SqueezeSeg. The network uses SqueezeNet as the backbone. It accepts a 2D LiDAR image transformed from a 3D point cloud, and predicts a point-wise label map, which is used to segment objects of interest. Thanks to the simplified pipeline and compact model design, SqueezeSeg is extremely fast, with the fastest version reaching 115 frames per second, while delivering good accuracy. 

\textbf{Data efficiency}: Deep neural networks require a massive amount of data to train, and it is observed \cite{sun2017revisiting} that the performance of deep neural networks improves logarithmically with the volume of the dataset. However, in many applications, collecting and annotating large datasets is a challenging task. This is especially true for LiDAR-based detection. In this thesis, we discuss two strategies to improve the data efficiency of deep neural networks and demonstrate them on the task of LiDAR segmentation. The first is to build advanced tools to make the annotation easier. The second is to leverage simulated data to train neural networks and transfer the model to the real world, thereby bypassing the need to collect and annotate real data. 

In Chapter \ref{chap:latte}, we discuss LATTE, a semi-automated LiDAR annotation tool. Compared with images, LiDAR point clouds are significantly more challenging to annotate. LiDAR sensors have much lower resolution compared with cameras, making it difficult even for a human to recognize objects. Moreover, annotating 3D objects requires more complex operations. Furthermore, most of the LiDAR point clouds are collected in sequences, and consecutive frames are highly correlated. In order to solve these problems, we built LATTE with three features: 1) With sensor fusion, we use image-based perception algorithms to detect objects and transform the prediction to a 3D point cloud. 2) We built a one-click annotation feature that simplifies the 3D bounding box annotation to simply one click on each target object. 3) We use tracking to transfer labels from one frame to subsequent frames automatically. With these three features, the LiDAR annotation process becomes 6.2x faster than the baseline without these features.  

In Chapter \ref{chap:sqsgv2}, we discuss our sim2real strategy to leverage synthetic data to train SqueezeSeg. Using GTA-V, we are able to synthesize a large amount of labeled LiDAR point cloud for training neural networks. However, due to the domain shift problem, a model trained on the simulation failed to transfer to the real world. We analyzed the source of domain shift, and discovered that 1) by improving the model to make it less sensitive to dropout noise in the LiDAR data and 2) by adopting a three-stage domain adaptation training pipeline, we are able to significantly boost the performance of SqueezeSeg trained on synthetic data from 30\% to 57.4\%. This result even out-performs an older version of SqueezeSeg trained on the real data. 

\textbf{Hardware efficiency}: We discuss the neural network model and hardware co-design in Chapter \ref{chap:shift}. In practical applications, merely reducing the hardware-agnostic complexity of neural networks, including FLOPs and parameter size, is not enough, since the essential metrics that people care about are speed and energy. To optimize for actual efficiency, we need to consider not only software design but also hardware design. In reality, however, we observed a gap between the neural network design community and the hardware design community. Most neural network designs only care about reducing FLOPs and parameter size, while the complicated structure of the model makes it challenging to map to hardware.
On the other hand, most of the hardware design did not leverage the latest progress of efficient neural networks. To close this gap, we first presented shift, a zero-FLOP, zero-parameter operator that replaces spatial convolutions. In a wide range of computer vision applications, using the shift operator not only significantly reduces the theoretical model complexity but more importantly, it simplifies the operator set of neural networks and allows us to build a ConvNet with only 1x1 convolutions. This, in turn, simplifies hardware design and allows us to build a customized compute unit optimized for 1x1 convolutions. Using the shift operator, we designed DiracDeltaNet and co-design an embedded FPGA based hardware accelerator named Synetgy. Compared with the previous state-of-the-art, Synetgy achieves 11.2x faster inference speed.

\textbf{Design efficiency}: In Chapter \ref{chap:dnas}, we discuss the design efficiency of neural networks. For a given application and an underlying hardware processor, it is a very challenging task to find the optimal neural network architecture. The challenges result from the facts that 1) neural networks typically have combinatorially large design spaces that are intractable. 2) The optimality of a neural network depends on many factors, including the hardware processors on which it will be run, but in reality, people can only afford to design one and fit to all conditions. 3) While people care about latency and energy, they are not always aligned with FLOPs and parameter size, and optimizing directly for latency and energy is very difficult. To address these problems, we present an automated neural architecture search algorithm named DNAS (Differentiable Neural Architecture Search). DNAS converts the combinatorial optimization problem of neural architecture search into a relaxed version such that we can use gradient-based methods to solve it. As a result, it is significantly faster than previous neural architecture search methods that rely on training neural architectures one by one. By designing a latency-aware loss function, we are able to directly optimize the latency of a network on a target hardware device. We applied DNAS to solve two problems. For mixed-precision quantization, we were able to compress a ResNet model trained on ImageNet by 21.1x without accuracy loss, surpassing the previous state-of-the-art. We also applied DNAS to search for models that would run on given mobile CPUs. The searched models named FBNets achieved better accuracy than previous searched and manually designed models, while the search cost is 421x lower than previous state-of-the-art NAS methods. 

\section{Impact of our work}
This thesis is a compilation of my research that has been previously published in eight papers \cite{SqueezeDet,wu2018squeezeseg,wu2018squeezesegv2,wang2019latte,wu2017shift,yang2018synetgy,wu2018mixed,wu2019fbnet} since I started working in this area in 2016. Despite the fact that we released these publications not long ago, they have begun to make impact on the efficient deep learning community. In this section, we summarize some of the impact of those works. 

We first released SqueezeDet\cite{wu2017squeezedet} in December 2016 and officially published the paper in June 2017. In less than three years, it has received 128 citations so far. Together with the paper, we also open-sourced the code to train, evaluate, and run SqueezeDet models on KITTI \cite{KITTI} and new datasets. Since its release, SqueezeDet has become a baseline for subsequent embedded object detectors \cite{li2018tiny}. The codebase of SqueezeDet was used by the research and industry community to develop new models for other perception tasks, such as SqueezeSeg \cite{wu2018squeezeseg,wu2018squeezesegv2}. It was used to study formal verification techniques for deep neural networks \cite{fremont2019scenic}. On the other hand, since SqueezeDet was mainly tuned for autonomous driving while contemporary object detectors such as YOLO \cite{YOLO} and SSD \cite{liu2016ssd} were applied to more general applications, SqueezeDet attracted less attention and did not
 became the mainstream method for subsequent object detection research.
 
 SqueezeSeg \cite{wu2018squeezeseg, wu2018squeezesegv2} was one of the first solution to apply deep neural networks to solve LiDAR-based perception problems. It proposed a novel problem formulation and established a new pipeline including data collection, training, and evaluation. Since its initial release in September 2017 and publication at ICRA in May 2018, it has received 59 citations. Many subsequent works followed this problem formulation and continued to improve the segmentation performance \cite{zhang2018liseg, wang2019ldls}. The idea proposed in SqueezeSeg of using simulated data to train neural networks was further developed by us in SqueezeSegV2  \cite{wu2018squeezesegv2} (published at ICRA 2019) and \cite{yue2018lidar} (published at ICMR 2018). In addition, SqueezeSeg \cite{wu2018squeezeseg} presented an idea for efficiently annotating LiDAR point clouds for pointwise labels. This idea is realized and extended in our later work LATTE \cite{wang2019latte}, which was accepted for publication at ITSC 2019.

The shift operator \cite{wu2017shift} was first published on arxiv in December 2017 and was officially published at CVPR in June 2018 as a spotlight oral paper. Following our paper, \cite{he2019addressnet} proposed a more efficient implementation of the shift operator and tested its effectiveness on GPUs. The operator is further improved by \cite{chen2019all} and is extended to process videos in the temporal domain by \cite{lin2018temporal}. Moreover, the shift operator is used as a critical component for neural network and hardware co-design for further efficiency optimization in \cite{yang2018synetgy,kungmaestro}.

DNAS \cite{wu2018mixed,wu2019fbnet} was first published on arxiv in December 2018 and was officially published at CVPR in June 2019 as an oral paper. By the time of this thesis' publication, in less than one year, this work has attracted much research attention and has received at least 30 citations. It has become the new state-of-the-art baseline for other manually and automatically designed neural networks \cite{howard2019searching, stamoulis2019single, liu2019metapruning}. Subsequent neural architecture search research continued to improve its search efficiency \cite{guo2019single, stamoulis2019single}, and extended the search to object detection problems \cite{chen2019detnas}. The mixed-precision application of DNAS has also inspired other researchers to solve this problem \cite{dong2019hawq} using other techniques.

\section{Future work}
At the end of this thesis, we discuss potential future extensions to research discussed in this thesis. 

\subsection{What is the limit of efficient neural networks?}
One of the fundamental questions for an efficient neural network is, what is the smallest network that can achieve a particular accuracy? SqueezeNet \cite{SqueezeNet} was a pioneer in efficient neural networks and it ``easily'' achieved a 50x parameter size reduction over AlexNet. This work made people realize that there exists redundancy in neural networks and inspired people to continue to shrink it as much as possible. From SqueezeNet to ShiftNet and DiractDeltaNet, the redundancy of spatial convolutions has been completely eliminated since the two networks only contain 1x1 convolutions. To further reduce the 1x1 convolutions, ShuffleNet\cite{ShuffleNet} and ShuffleNetV2\cite{ma2018shufflenet} introduced point-wise group convolution and shuffle operations. After that, NAS was used to push the limit of the design space to find the optimal neural architecture. However, recent work also observed a slowing trend of FLOPs and parameter size reduction. So the question is, are we close to the limit? Are there other redundancies we can explore?

We believe that there is still redundancy in current neural networks, especially ConvNets. A ConvNet is, by design, translation equivariant. However, ConvNets do not possess the property of scaling or rotation equivariance. As a result, to deal with feature scaling and rotation, ConvNets need to memorize the features in their convolution filters. By carefully re-designing the convolution filters, we may be able to reduce the redundancy further and make the network more stable towards variations. 

More fundamentally, can we address this problem using more theoretical approaches? Can we formally define concepts such as ``redundancy'' and ``capacity'' for neural networks? Can we adopt ideas such as Vapnik–Chervonenkis dimension \cite{abu1989vapnik} and Kolmogorov complexity \cite{schmidhuber1997discovering} to analyze neural networks and guide us to design efficient neural networks? 

\subsection{Automated co-design of neural networks \& hardware}
In this thesis, we discussed automated neural architecture search: DNAS and model-hardware co-design: Shift \& Synetgy. However, a natural question is: can we automate the co-design of neural networks and hardware processors. Currently, the DNAS framework considers the operator costs on the target hardware to be fixed. However, if given the flexibility of tuning hardware design parameters, such as dataflow design, or the number of MAC units, can we efficiently find the optimal neural network architecture and hardware configurations at the same time? In such a significantly larger search space, can we find significantly better solutions? What will be the best algorithm to explore this hybrid search space?

\subsection{On-device training and neural network adaptation}
Today's AI development pipeline separates the training and deployment of neural networks. The deployed neural networks are expected to perform well in all conditions. However, due to ubiquitous domain shifts, neural network models can quickly fail if they are not adapted properly to the new domain. In order to solve this problem, one solution is to break the division of training and deployment, and allow neural networks to be continuously trained with new data. To achieve this, we need to 1) develop a new software and hardware stack that supports on-device training, 2) develop better optimization techniques to drastically reduce the FLOP and memory cost of neural network training and 3) develop new algorithms, such as few-shot learning, meta learning, continual learning, and self-supervised learning to allow adapting and training neural networks with significantly less new data.